\documentclass[Journal, InsideFigs, letterpaper, NoLineNumbers]{ascelike-new}
\usepackage[utf8]{inputenc}
\usepackage[T1]{fontenc}
\usepackage{lmodern}
\usepackage{graphicx}
\usepackage{xcolor}
\usepackage[figurename=Fig.,labelfont=bf,labelsep=period]{caption}
\usepackage{subcaption}
\usepackage{amsmath}
\usepackage{newtxtext,newtxmath}
\usepackage[colorlinks=true,citecolor=red,linkcolor=black]{hyperref}
\usepackage{tabularx}
\usepackage{adjustbox}
\usepackage[authoryear]{natbib}
\usepackage{multirow}
\usepackage{pdflscape}
\usepackage{fancyhdr}
\usepackage{everypage}
\usepackage{lipsum}
\usepackage{setspace}

\newcommand{\Lpagenumber}{\ifdim\textwidth=\linewidth\else\bgroup
  \dimendef\margin=0 
  \ifodd\value{page}\margin=\oddsidemargin
  \else\margin=\evensidemargin
  \fi
  \raisebox{\dimexpr -\topmargin-\headheight-\headsep-0.5\linewidth}[0pt][0pt]{%
    \rlap{\hspace{\dimexpr \margin+\textheight+\footskip}%
    \llap{\rotatebox{90}{\thepage}}}}%
\egroup\fi}

\AddEverypageHook{\Lpagenumber}

\begin{document}

\title{Segment Any Crack: Deep Semantic Segmentation Adaptation for Crack Detection}

\author[1]{Ghodsiyeh Rostami, MSc}
\author[2]{Po-Han Chen, Ph.D., A.M.ASCE}
\author[3]{Mahdi S. Hosseini, Ph.D.}

\begingroup
\singlespacing 
\affil[1]{PhD Student, Department of Building, Civil, and Environmental Engineering, Concordia University, Montreal, Canada, H3G 1M8. Email: rose.rostami@concordia.ca}
\affil[2]{Professor, Department of Building, Civil, and Environmental Engineering, Concordia University, Montreal, Canada, H3G 1M8. Email: pohan.chen@concordia.ca (Corresponding author) }
\affil[3]{Assistant Professor, Department of Computer Science and Software Engineering, Concordia University, Montreal, Canada, H3G 1M8. Email: mahdi.hosseini@concordia.ca}

\maketitle
\endgroup

\begin{abstract}
Image-based crack detection algorithms are increasingly in demand in infrastructure monitoring, as early detection of cracks is of paramount importance for timely maintenance planning. While deep learning has significantly advanced crack detection algorithms, existing models often require extensive labeled datasets and high computational costs for fine-tuning, limiting their adaptability across diverse conditions. 
This study introduces an efficient selective fine-tuning strategy, focusing on tuning normalization components, to enhance the adaptability of segmentation models for crack detection. The proposed method is applied to the Segment Anything Model (SAM) and five well-established segmentation models. Experimental results demonstrate that selective fine-tuning of only normalization parameters outperforms full fine-tuning and other common fine-tuning techniques in both performance and computational efficiency, while improving generalization. The proposed approach yields a SAM-based model, Segment Any Crack (SAC), achieving a 61.22\% F1-score and 44.13\% IoU on the OmniCrack30k benchmark dataset, along with the highest performance across three zero-shot datasets and the lowest standard deviation. The results highlight the effectiveness of the adaptation approach in improving segmentation accuracy while significantly reducing computational overhead. 
\end{abstract}

\section{Introduction}
The aging and deterioration of constructed facilities, including bridges, roads, and buildings, during their operational lifecycle pose severe risks to the infrastructure's safety and integrity \citep{xiangCracksegmentationAlgorithmFusing2023,liAutomaticPavementCrack2022}. Specifically, cracks are the most common defect, developing from the surface of the structure to the deeper parts, leading to structure failure \citep{munawarImageBasedCrackDetection2021}. Therefore, regular inspections are crucial for early detection of cracks to ensure safety and enable in-time and cost-effective maintenance measures \citep{liAutomaticPavementCrack2022}. However, the prevalent manual inspection methods require specialized equipment and expert personnel, making the process labor-intensive, costly, and prone to safety risks \citep{halderRobotsInspectionMonitoring2023}. 
Over the past decades, automated visual inspection methods have been proposed, leveraging robotic platforms such as unmanned aerial vehicles and advanced imaging systems to enhance the efficiency and safety of inspections \citep{agnisarmanSurveyAutomationenabledHumanintheloop2019,pengUAVbasedMachineVision2021,zhangFullyAutomatedUnmanned2022}. 

With the rise of deep learning, crack recognition algorithms have been studied for detecting cracks from natural images and estimating their severity through geometric measurements, such as width and length \citep{dingCrackDetectionQuantification2023,liFastAccurateRoad2023}. To this end, semantic segmentation has become the mainstream approach for crack detection, enabling precise pixel-level localization and facilitating further quantification analyses \citep{guoSurfaceDefectDetection2024}. However, training deep segmentation models for crack detection remains challenging due to the limited availability of large-scale annotated crack datasets, making transfer learning a common strategy to achieve well-trained models \citep{iraniparastSurfaceConcreteCracks2023}. However, when adapting a pre-trained model from a source dataset to a specific domain such as crack detection, the issue of domain shift arises, where differences in data distribution can lead to ineffective training and degradation of the model's performance \citep{lianScalingShiftingYour2022}.

Moreover, as the complexity of deep learning models grows increasingly to capture more comprehensive feature representations, traditional transfer learning approaches, such as full tuning all the model's parameters or fine-tuning the decoder component, become computationally expensive and less efficient \citep{zhangParameterEfficientFineTuningFoundation2025}. Specifically, the emergence of vision foundation models, pre-trained on large-scale generic datasets, has introduced a paradigm shift, offering rich and transferable feature representations \citep{bommasaniOpportunitiesRisksFoundation2022}. Various parameter-efficient tuning strategies have been explored in deep learning literature; however, their application in crack detection remains underexplored. For instance, \citet{geFinetuningVisionFoundation2024} investigated the adaptation of vision foundation models for crack segmentation, specifically using the Segment Anything Model (SAM). While their study demonstrated the potential of fine-tuning approaches in achieving high-performance crack detection models, an open question remains: Can knowledge transfer be made more efficient and effective to develop a more generalized and robust crack detection model? Furthermore, what is the most optimal adaptation strategy for crack segmentation?

To address these questions, this study systematically explores different transfer learning techniques to identify the most efficient and effective adaptation method for crack detection. Specifically, inspired by the findings of \citet{zhaoTuningLayerNormAttention2023a}, we hypothesize that in transfer learning, since the source and domain-specific datasets have different data distributions, tuning only the normalization parameters of a pre-trained model can mitigate this domain shift and yield high-performing segmentation models. To validate this hypothesis, we conduct a comprehensive analysis of the effectiveness of normalization layers in adapting segmentation models for crack detection. Our study focuses on a SAM-based model, along with five well-established segmentation models: CrackFormer, DeepCrack, SegFormer, U-Net, and DeepLabv3+.

\section{Related Work}
Over the past decades, deep learning-based crack detection models have been extensively studied, adopting various architectures such as Convolutional Neural Networks (CNNs), Vision Transformers (ViTs), and, more recently, Vision Mamba. These advancements aim to improve not only detection accuracy and robustness but also computational efficiency, enabling faster and more reliable crack inspection in real-world applications. For instance, \citet{renImagebasedConcreteCrack2020} developed CrackSegNet, a deep FCN leveraging dilated convolution, pyramid pooling, and skip connections to enhance crack segmentation accuracy. Trained on 919 tunnel images (512×512 pixels), it outperformed U-Net and a conventional image processing method in F1-score and mIoU. Similarly, \citet{linDeepCrackATEffectiveCrack2023} developed DeepCrackAT, an encoder-decoder model with dilated convolution and attention mechanisms, trained on 1440 masonry and 2736 concrete/wall crack images (224×224 pixels). Despite achieving 76.67\% F1-score and 64.44\% mIoU, it had limitations in segmenting cracks with varying widths. 
To improve fine-grained crack detection, \citet{liuCrackFormerTransformerNetwork2021} proposed CrackFormer, a Transformer-based SegNet variant integrating self-attention and scaling-attention modules for enhanced contextual learning. It excelled in multi-scale pavement crack analysis with higher performance and fewer FLOPs. Moreover, \citet{zhouMixSegNetNovelCrack2024} introduced MixSegNet, a CNN-Transformer hybrid based on U-Net architecture, to capture fine details and long-range dependencies. It outperformed state-of-the-art models, including LRASPP, FCN, DeepLabV3, U-Net, AttuNet and Swin-UNet, on the Cracks-APCGAN dataset with 91.5\% F1-score and 84.8\% IoU. More recently, \citet{hanEnhancingPixellevelCrack2024} proposed MambaCrackNet, a hybrid Mamba-CNN model reducing fine crack misdetections. While outperforming conventional models, its dual feature-processing paths increased complexity, slowing inference. Additionally, Mamba’s constraints on input size and spatial dimensions may limit adaptability in crack detection applications.
  
However, models with limited capacity, trained on small-scale datasets, cannot effectively generalize to new environments and diverse scenes. To address this limitation, researchers have investigated the adaptation of large-scale pre-trained models for crack detection using transfer learning techniques.
For instance, \citet{egodawelaDeepLearningApproach2024} adapted SAM for segmenting crack pixels in UAV-captured images. Their dataset comprised 150 high-resolution images and 360,000 image frames extracted from a 20-minute video of indoor building surfaces captured by two collaborative drones. They fine-tuned SAM’s mask decoder, which has 4 million parameters, on this dataset. The finetuned SAM achieved IoU and dice score of 0.851 and 0.728 on the dataset. However, the majority of the crack images had relatively simple backgrounds, which may not reflect the complexity of real-world scenarios. Also, the finetuned SAM generates multiple segmentation masks per image and requires an operator to manually select the best mask, which is a tedious and time-consuming task, hindering the full automation of crack detection.
Therefore, there still remains the question of the optimal adaptation strategy for crack detection. This requires further investigation into efficient fine-tuning techniques that can balance performance, computational cost, and generalization of the model across diverse crack environments.

\section{Methodology}
The study consists of four main experiments: \textbf{1)} Hyperparameter tuning to achieve the optimal training hyperparameters such as loss function, learning rate, and batch size, \textbf{2)} Fine-Tuning ablation study on the Segment Anything Model (SAM) to evaluate and compare different fine-tuning approaches and selecting the most efficient one in terms of the minimum number of trainable parameters and computation efficiency, \textbf{3)} cross-model verification, by implementing the best fine-tuning strategy across different segmentation models assess its transferability, and \textbf{4)} zero-shot performance analysis to evaluate the performance of the fine-tuned models on three new datasets to assess the generalization capability of the fine-tuning strategy.

The methodology section is organized as follows: Section \ref{subsec:fine-tuning_ablation} describes different adaptation approaches investigated in this study using the SAM model, Section \ref{subsec:datasets} introduces the datasets used for training and zero-shot performance evaluation, Section \ref{subsec:hyperparameter_tuning} describes the experiments setup, the evaluation metrics, and hyperparameter tuning experiments, and Section \ref{subsec:cross-model} elaborates on the segmentation models used in this study.

\subsection{Fine-Tuning Methods} \label{subsec:fine-tuning_ablation}
Adapting deep segmentation models, specifically large-scale models such as vision foundation models, to domain-specific tasks poses risks and challenges, such as overfitting, due to the extensive parameterization and computational cost. Full tuning, which involves updating all model parameters, is often impractical for such models, as it demands high memory consumption and increased training time.
Instead, parameter-efficient and selective fine-tuning strategies have been widely explored as efficient alternatives that balance adaptation capability, parameter efficiency, and computational feasibility.

To systematically evaluate the impact of different fine-tuning strategies on crack segmentation, an ablation study was conducted using the Segment Anything Model (SAM) as the foundation model. The study focused on assessing the trade-offs between segmentation accuracy, parameter efficiency, and computational cost. Four fine-tuning approaches were investigated:

\textbf{(1) Normalization Parameters Fine-Tuning}: Normalization components play a critical role in stabilizing and accelerating the training of deep neural networks by mitigating internal covariate shifts \citep{ioffeBatchNormalizationAccelerating2015a}. Different normalization techniques have been proposed in the literature, including layer normalization \citep{baLayerNormalization2016a}, batch normalization \citep{ioffeBatchNormalizationAccelerating2015a}, and group normalization \cite{wuGroupNormalization2018}, each of which operates over different dimensions such as layers, batches, or groups of output activations. The normalization operation is formally defined as follows: Each pre-activation value is normalized by:

\begin{equation} \label{eq:batch nom1}
\hat{x} = \frac{x - \mu_{\text{batch}}}{\sigma_{\text{batch}} + \epsilon} \;.
\end{equation}

Where x represents the input activation, $\mu_{\text{batch}}$ and $\sigma_{\text{batch}}$ denote the mean and standard deviation across the mini-batch, and $\epsilon$ is a small constant for numerical stability. The normalized activation is then rescaled and shifted using trainable affine parameters:

\begin{equation} \label{eq:batch nom1}
y=\gamma\hat{x}+\beta \;.
\end{equation}
$\gamma$ and $\beta$ are learnable per-feature scale and shift parameters, respectively. These transformations allow the model to retain its representational capacity while benefiting from improved training stability and convergence efficiency.

\begin{figure}[h]
    \centering
    \includegraphics[width=4in]{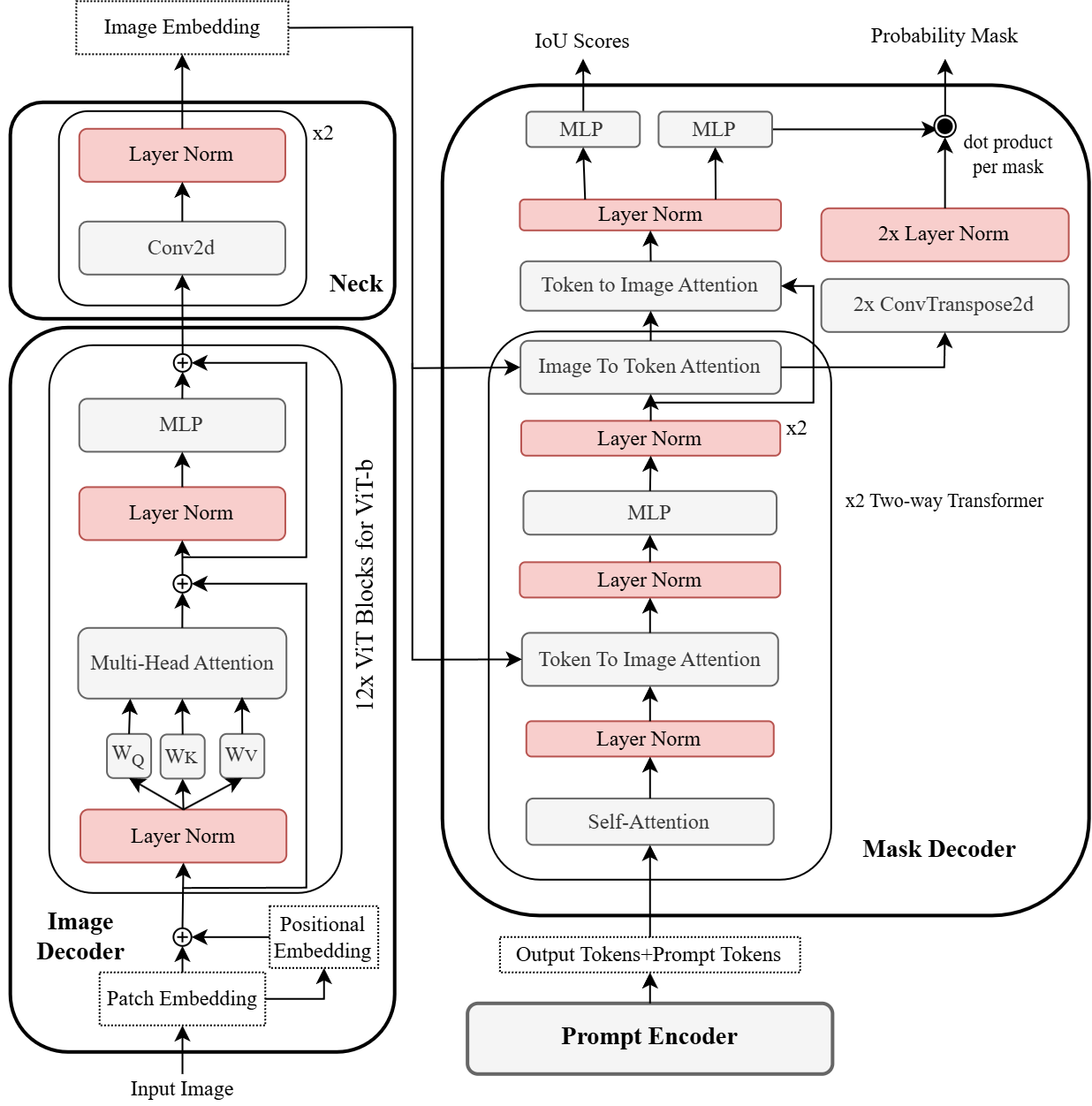}
    \caption{The proposed layer normalization fine-tuning method illustrated on the Segment Anything Model (SAM). Trainable modules are colored in \textcolor{red}{red} and frozen modules in \textcolor{gray}{grey}.}
    \label{fig:layernorm}
\end{figure}

In transfer learning, the main challenge is the differences in data distributions between the pre-training and target domains. Therefore, this study hypothesizes that only tuning the normalization parameters can help mitigate these discrepancies and improve model adaptation \citep{frankleTrainingBatchNormOnly2020}. To this end, all normalization parameters in SAM, which are layer normalization, were fine-tuned as the main adaptation strategy to refine feature scaling and distribution alignments as illustrated in Figure \ref{fig:layernorm}.

\textbf{(2) Encoder Selective Parameter-Efficient Fine-Tuning (PEFT):} 
In deep segmentation models, the early layers of the encoder mainly learn low-level features such as edges and textures, while the deeper layers capture high-level semantic representations. Given SAM's rich representations, an ablation study was conducted to investigate the impact of fine-tuning the later layers of the SAM’s encoder using Parameter-Efficient Fine-Tuning (PEFT) methods, specifically Low-Rank Adaptation (LoRA), to further enhance SAM's high-level semantics. Specifically, LoRA was applied to fine-tune the attention layers within different Vision Transformer (ViT) blocks of SAM. Also, the last linear layers of the multi-layer perceptron (MLP) modules in the later blocks were also fine-tuned using LoRA to refine the model’s feature representations for crack segmentation. 

\textbf{(3) Decoder Fine-Tuning:} 
Decoder fine-tuning is a widely adopted strategy in transfer learning, where only the decoder is fine-tuned while the encoder remains frozen. The underlying assumption is that the encoder’s learned feature representations are sufficiently generalizable to downstream tasks, while the decoder requires adaptation to effectively interpret these features for the domain-specific task, such as crack detection. This strategy is computationally efficient, as it requires updating only a limited number of parameters depending on the decoder's size.

\textbf{(4) CrackSAM Method}: The fine-tuning approach proposed by \citep{geFinetuningVisionFoundation2024} was also implemented for comparative evaluation. The method involves fully tuning SAM using a combination of two PEFT methods, LoRA and Adapter, for the encoder. Additionally, both the prompt encoder and mask decoder are fine-tuned to further enhance SAM’s adaptability for crack segmentation. It should be noted that no prompts are added in this approach.

\subsection{Datasets} \label{subsec:datasets}
\begin{figure}[h]
\captionsetup[subfigure]{font=footnotesize}
    \centering
    \begin{subfigure}[b]{0.13\textwidth}
        \adjustbox{trim=10 10 10 10,clip,width=2cm,height=2cm}{\includegraphics{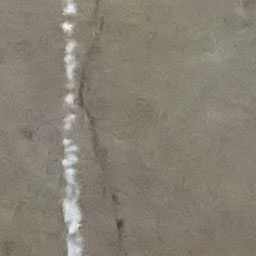}}
        \caption*{BCL}
    \end{subfigure}
    \begin{subfigure}[b]{0.13\textwidth}
        \adjustbox{trim=10 10 10 10,clip,width=2cm,height=2cm}{\includegraphics{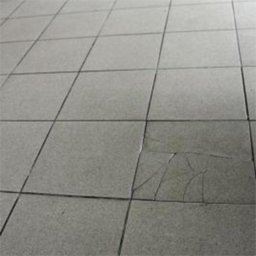}}
        \caption*{Ceramic}
    \end{subfigure}
    \begin{subfigure}[b]{0.13\textwidth}
        \adjustbox{trim=10 10 10 10,clip,width=2cm,height=2cm}{\includegraphics{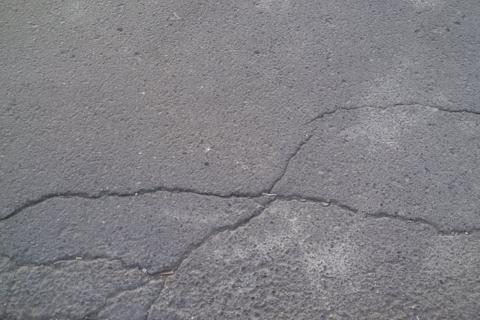}}
        \caption*{CFD}
    \end{subfigure}
    \begin{subfigure}[b]{0.13\textwidth}
        \adjustbox{trim=10 10 10 10,clip,width=2cm,height=2cm}{\includegraphics{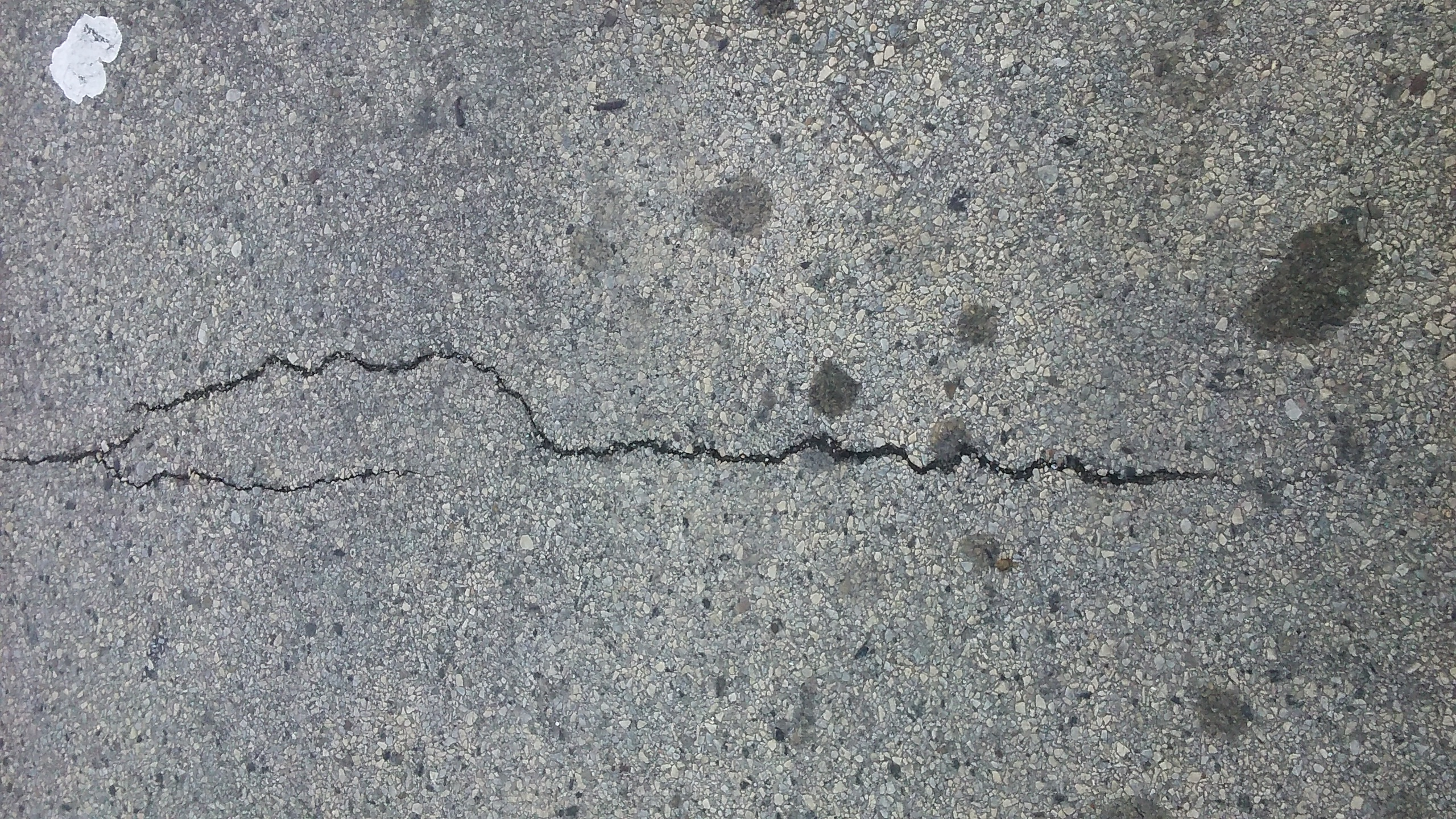}}
        \caption*{Crack500}
    \end{subfigure}
    \begin{subfigure}[b]{0.13\textwidth}
        \adjustbox{trim=10 10 10 10,clip,width=2cm,height=2cm}{\includegraphics{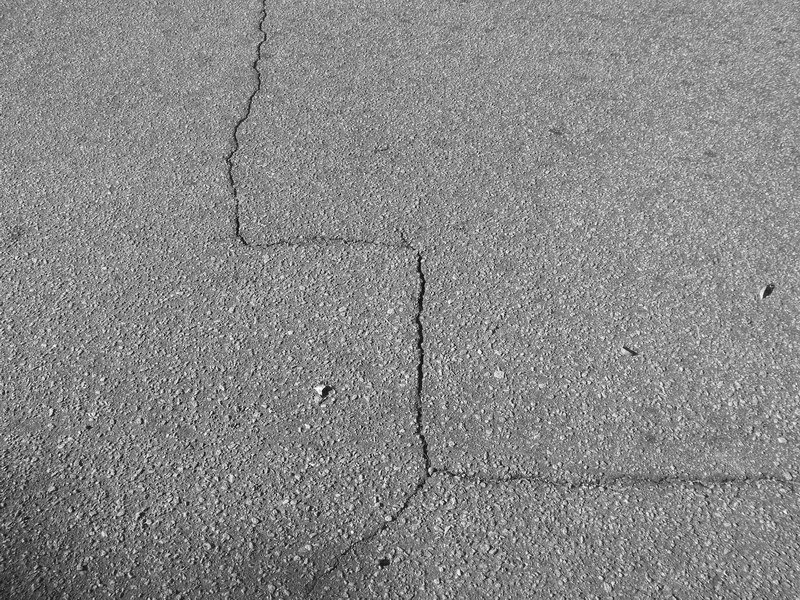}}
        \caption*{CrackTree260}
    \end{subfigure}

    \begin{subfigure}[b]{0.13\textwidth}
        \adjustbox{trim=10 10 10 10,clip,width=2cm,height=2cm}{\includegraphics{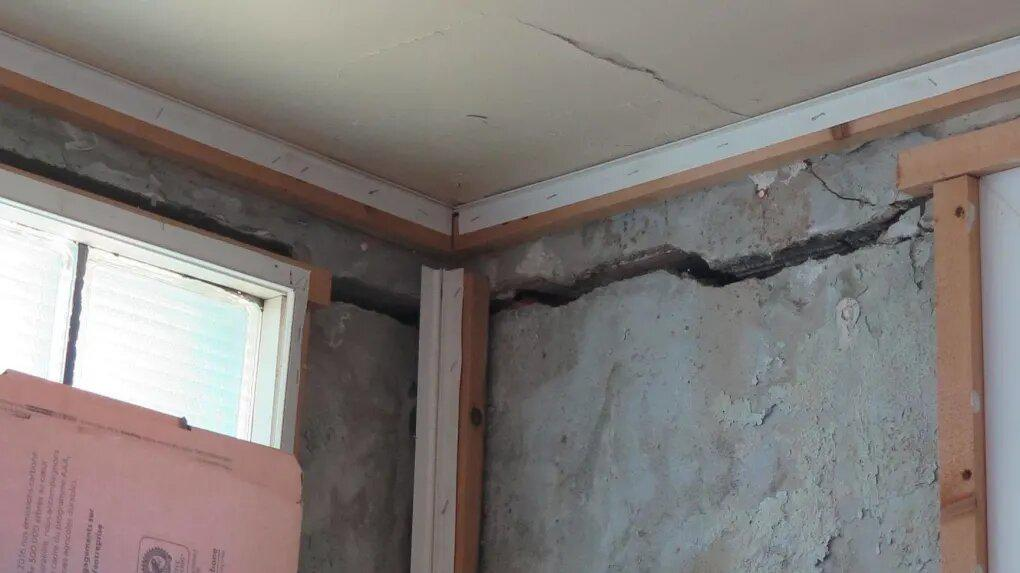}}
        \caption*{CrSpEE}
    \end{subfigure}
    \begin{subfigure}[b]{0.13\textwidth}
        \adjustbox{trim=10 10 10 10,clip,width=2cm,height=2cm}{\includegraphics{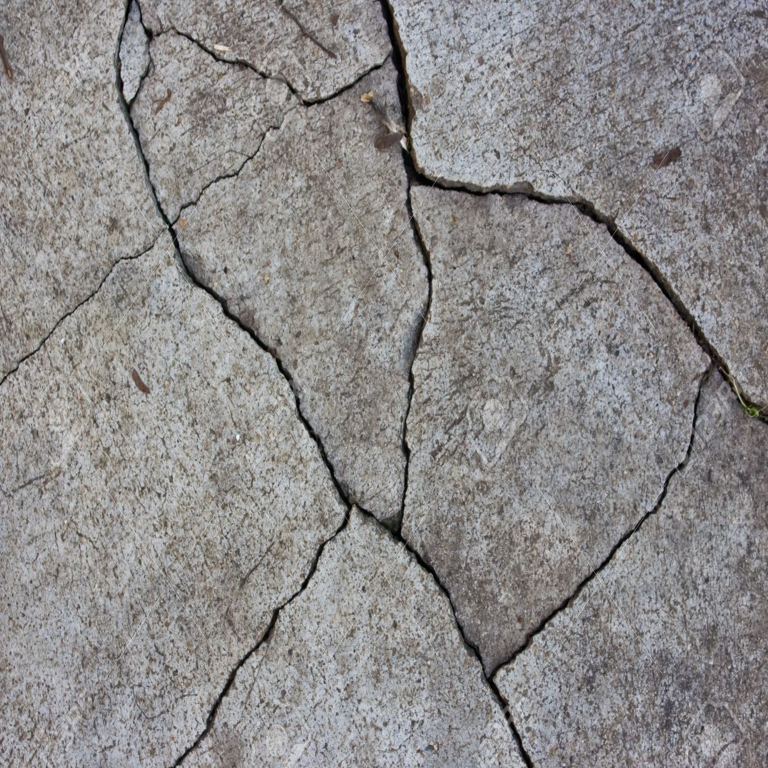}}
        \caption*{CSSC}
    \end{subfigure}
    \begin{subfigure}[b]{0.13\textwidth}
        \adjustbox{trim=10 10 10 10,clip,width=2cm,height=2cm}{\includegraphics{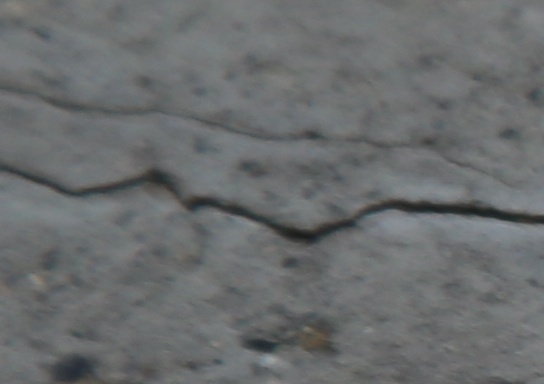}}
        \caption*{DeepCrack}
    \end{subfigure}
    \begin{subfigure}[b]{0.13\textwidth}
        \adjustbox{trim=10 10 10 10,clip,width=2cm,height=2cm}{\includegraphics{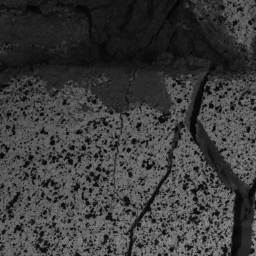}}
        \caption*{DIC SVS}
    \end{subfigure}
    \begin{subfigure}[b]{0.13\textwidth}
        \adjustbox{trim=10 10 10 10,clip,width=2cm,height=2cm}{\includegraphics{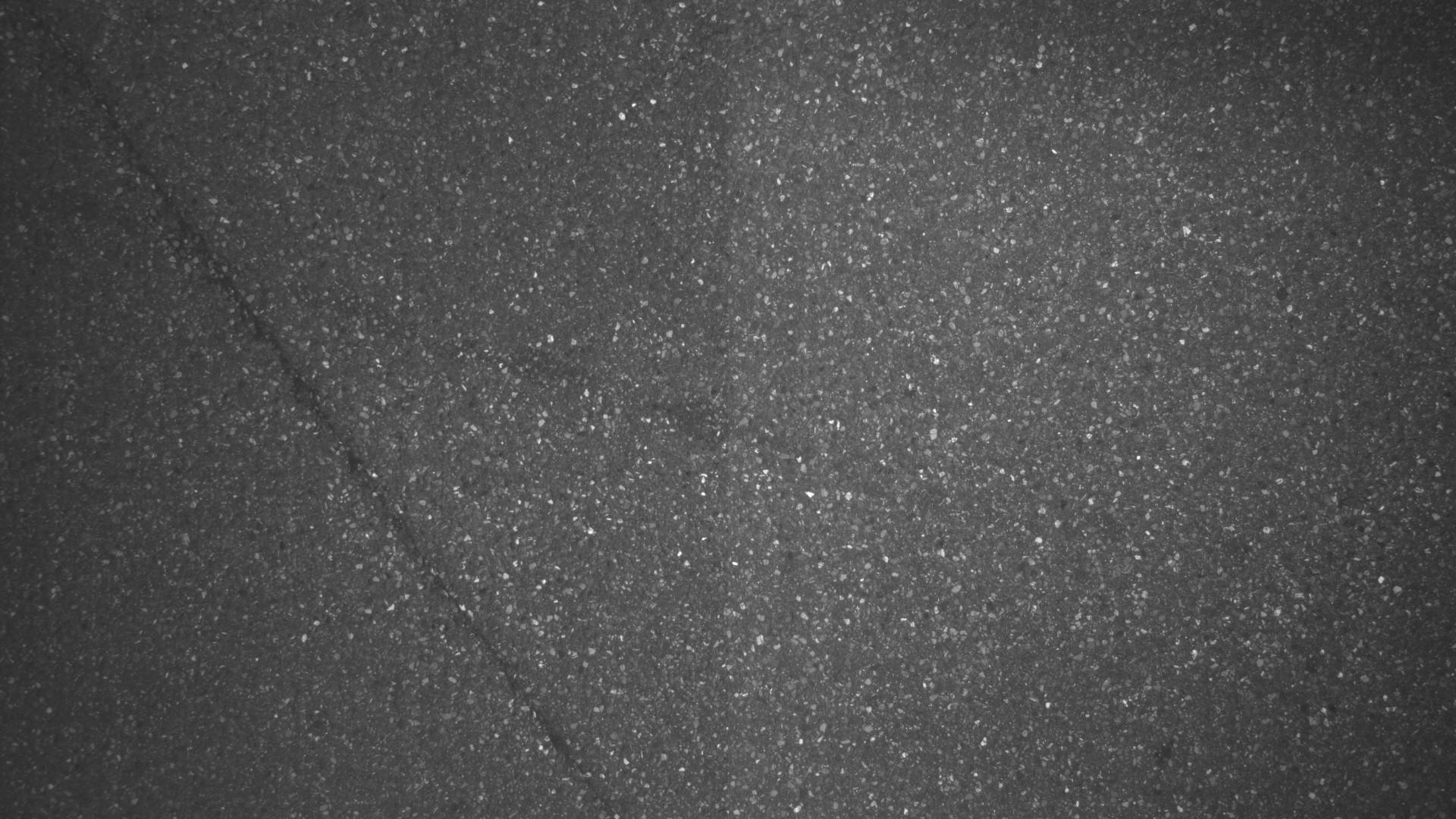}}
        \caption*{GAPS384}
    \end{subfigure}

    \begin{subfigure}[b]{0.13\textwidth}
        \adjustbox{trim=10 10 10 10,clip,width=2cm,height=2cm}{\includegraphics{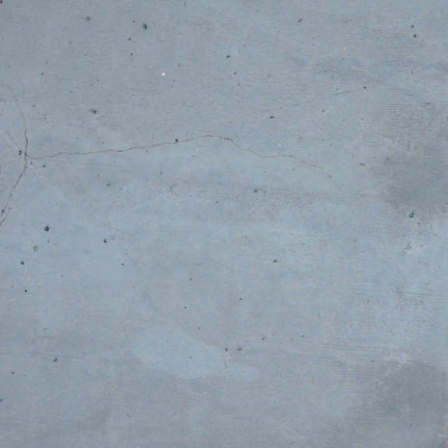}}
        \caption*{Khanh. Eugen}
    \end{subfigure}
    \begin{subfigure}[b]{0.13\textwidth}
        \adjustbox{trim=10 10 10 10,clip,width=2cm,height=2cm}{\includegraphics{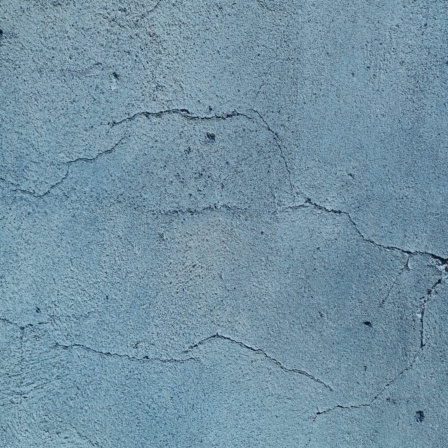}}
        \caption*{Khanh. Rissb.}
    \end{subfigure}
    \begin{subfigure}[b]{0.13\textwidth}
        \adjustbox{trim=10 10 10 10,clip,width=2cm,height=2cm}{\includegraphics{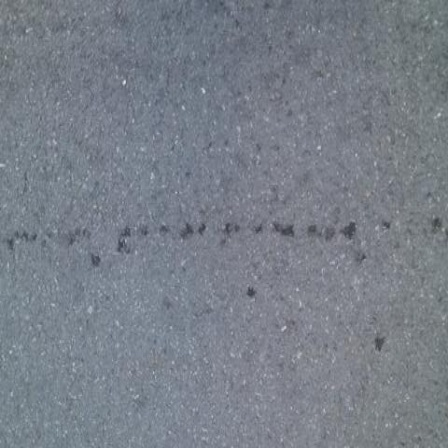}}
        \caption*{Khanh. Sylvie}
    \end{subfigure}
    \begin{subfigure}[b]{0.13\textwidth}
        \adjustbox{trim=10 10 10 10,clip,width=2cm,height=2cm}{\includegraphics{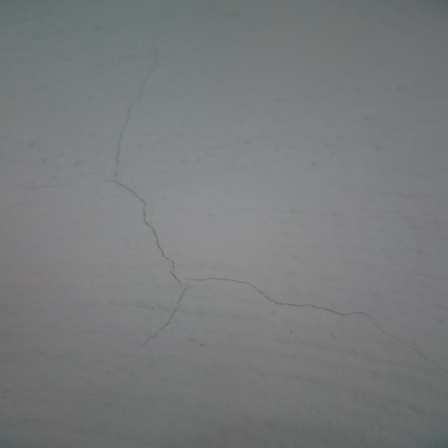}}
        \caption*{Khanh. Volker}
    \end{subfigure}
    \begin{subfigure}[b]{0.13\textwidth}
        \adjustbox{trim=10 10 10 10,clip,width=2cm,height=2cm}{\includegraphics{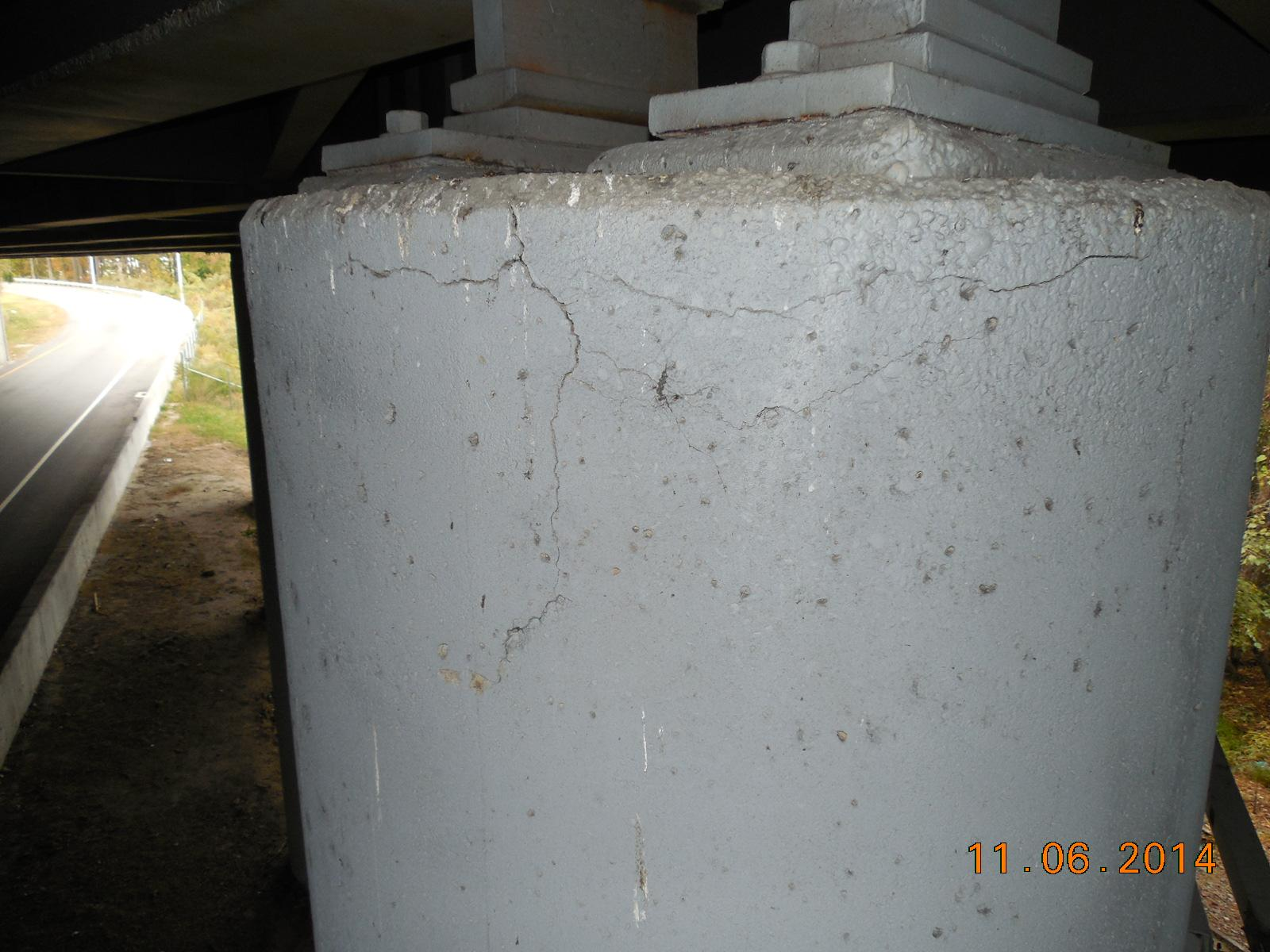}}
        \caption*{LCW}
    \end{subfigure}

    \begin{subfigure}[b]{0.13\textwidth}
        \adjustbox{trim=10 10 10 10,clip,width=2cm,height=2cm}{\includegraphics{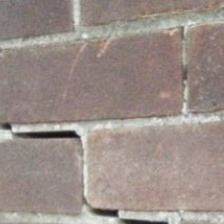}}
        \caption*{Masonry}
    \end{subfigure}
    \begin{subfigure}[b]{0.13\textwidth}
        \adjustbox{trim=10 10 10 10,clip,width=2cm,height=2cm}{\includegraphics{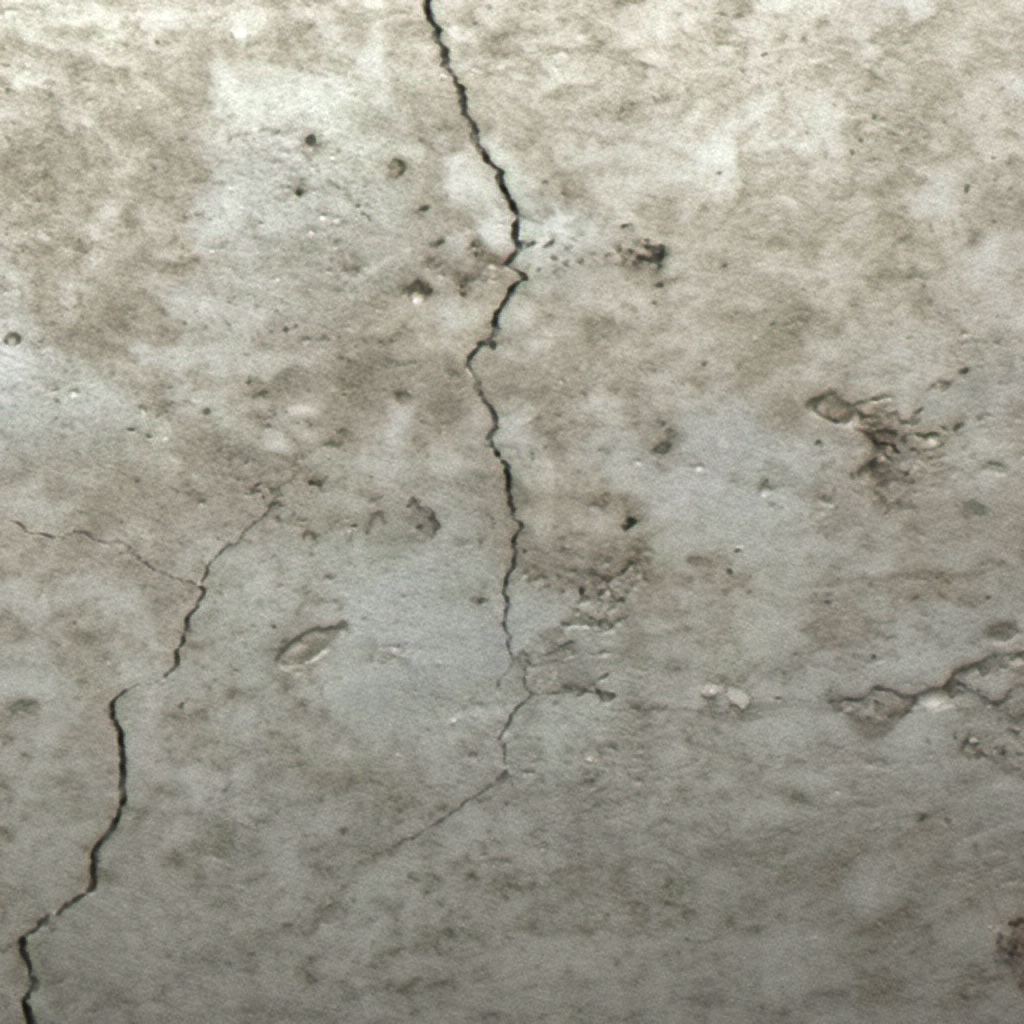}}
        \caption*{S2DS}
    \end{subfigure}
    \begin{subfigure}[b]{0.13\textwidth}
        \adjustbox{trim=10 10 10 10,clip,width=2cm,height=2cm}{\includegraphics{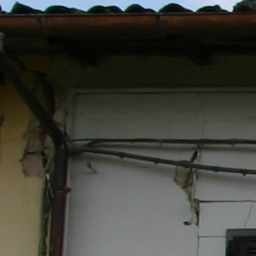}}
        \caption*{TopoDS}
    \end{subfigure}
    \begin{subfigure}[b]{0.13\textwidth}
        \adjustbox{trim=10 10 10 10,clip,width=2cm,height=2cm}{\includegraphics{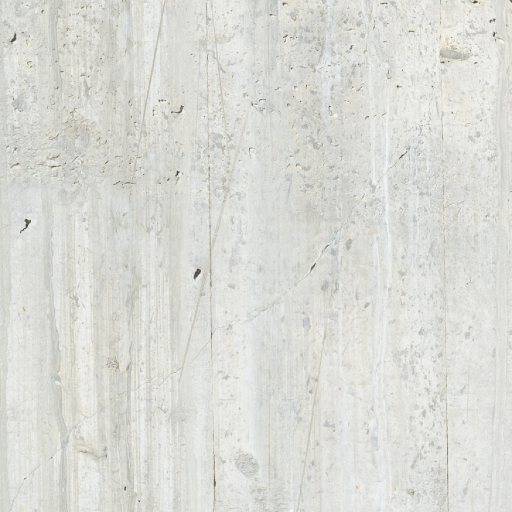}}
        \caption*{UAV75}
    \end{subfigure}
    \caption{Samples of the OmniCrack30k dataset. Note that Khanh. and Rissb. denote Khanh11k and Rissbilder, respectively.}
    \label{fig:omnicrack_samples}
\end{figure}

The dataset of the study is the state-of-the-art crack image segmentation dataset, OmniCrack30k \citep{benzOmniCrack30kBenchmarkCrack2024a}. The dataset is a comprehensive collection of 30k samples from over 20 sub-datasets comprising 9 billion pixels in total.  
The dataset includes a diverse range of crack images and corresponding binary masks across multiple materials including concrete, asphalt, ceramic, masonry, and steel as well as various surface types such as walls and roads. Additionally, it contains images captured under different fields of view, lighting conditions, and background textures, making it a highly challenging benchmark that reflects real-world scene variability. Sample images are illustrated in Figure \ref{fig:omnicrack_samples}. Image resolutions span in the $[81, 4608] \times [116, 4608]$ range. The dataset was partitioned into training, validation, and test sets, with detailed statistics provided in Table \ref{table:omnicrack split}. During both the training and testing phases, all images and their corresponding masks were resized to $256\times256$.

\begin{figure}[h]
\captionsetup[subfigure]{font=footnotesize}
    \centering
    \begin{subfigure}[b]{0.1\textwidth} 
        \subcaption*{\rotatebox{90}{Road420}} 
    \end{subfigure}
    \begin{subfigure}[b]{0.13\textwidth}
        \adjustbox{trim=30 30 30 30,clip,width=2.0cm,height=2cm}{\includegraphics{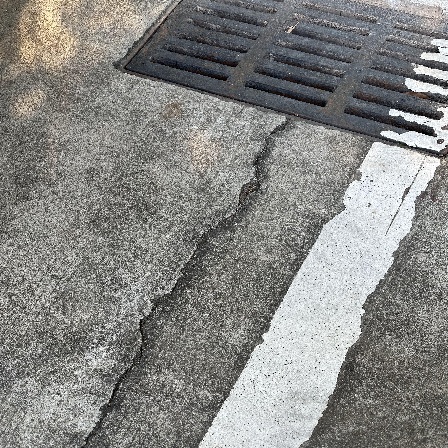}}
    \end{subfigure}
    \begin{subfigure}[b]{0.13\textwidth}
        \adjustbox{trim=30 30 30 30,clip,width=2cm,height=2cm}{\includegraphics{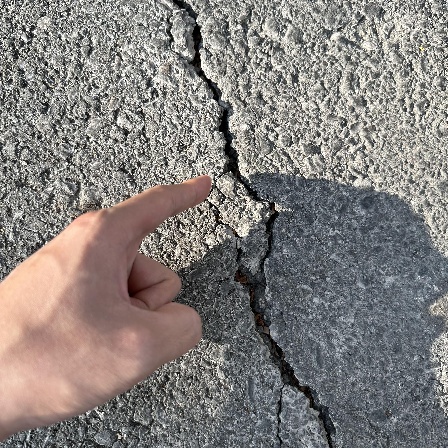}}
    \end{subfigure}
    \begin{subfigure}[b]{0.13\textwidth}
        \adjustbox{trim=30 30 30 30,clip,width=2cm,height=2cm}{\includegraphics{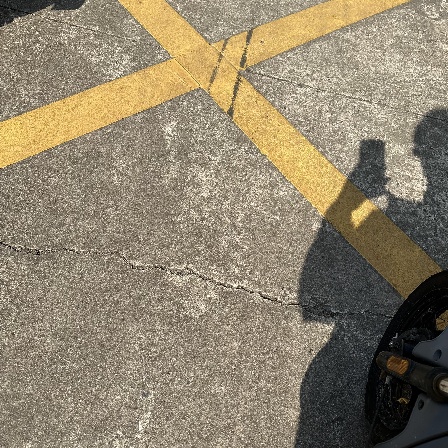}}
    \end{subfigure}
    \par
    \vspace{0.1cm} 
    \begin{subfigure}[b]{0.1\textwidth} 
        \subcaption*{\rotatebox{90}{Facade390}} 
    \end{subfigure}
    \begin{subfigure}[b]{0.13\textwidth}
        \adjustbox{trim=30 30 30 30,clip,width=2cm,height=2cm}{\includegraphics{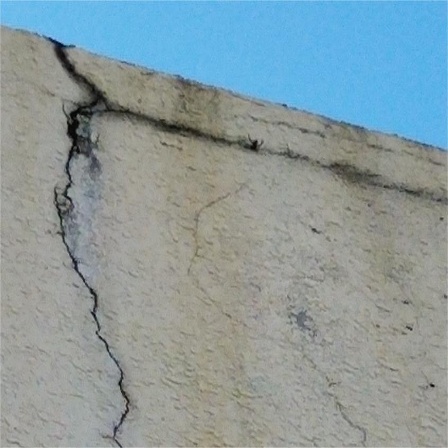}}
    \end{subfigure}
    \begin{subfigure}[b]{0.13\textwidth}
        \adjustbox{trim=30 30 30 30,clip,width=2cm,height=2cm}{\includegraphics{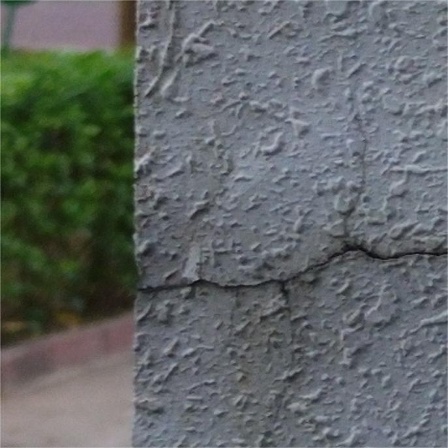}}
    \end{subfigure}
    \begin{subfigure}[b]{0.13\textwidth}
        \adjustbox{trim=30 30 30 30,clip,width=2cm,height=2cm}{\includegraphics{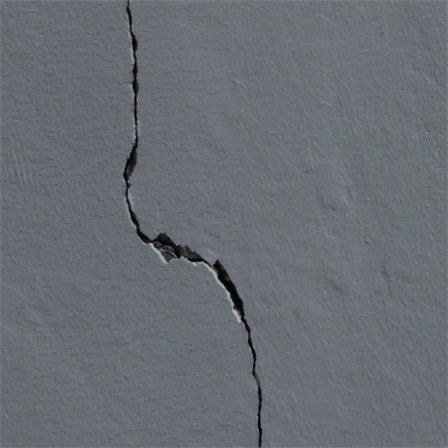}}
    \end{subfigure}
    \par
    \vspace{0.1cm} 
    \begin{subfigure}[b]{0.1\textwidth} 
        \subcaption*{\rotatebox{90}{Concrete3k}} 
    \end{subfigure}
    \begin{subfigure}[b]{0.13\textwidth}
        \adjustbox{trim=30 30 30 30,clip,width=2cm,height=2cm}{\includegraphics{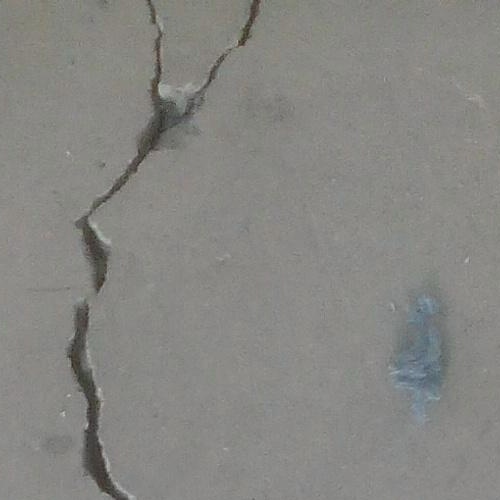}}
    \end{subfigure}
    \begin{subfigure}[b]{0.13\textwidth}
        \adjustbox{trim=30 30 30 30,clip,width=2cm,height=2cm}{\includegraphics{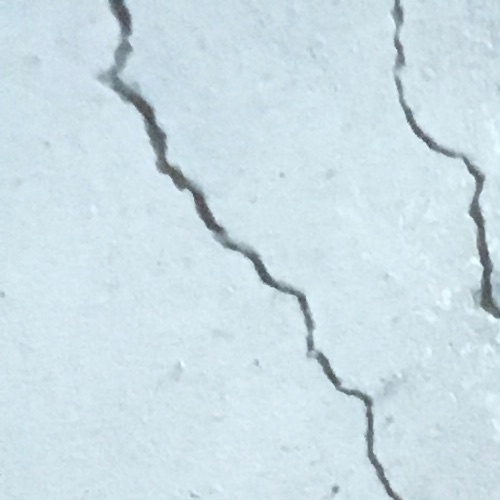}}
    \end{subfigure}
    \begin{subfigure}[b]{0.13\textwidth}
        \adjustbox{trim=30 30 30 30,clip,width=2cm,height=2cm}{\includegraphics{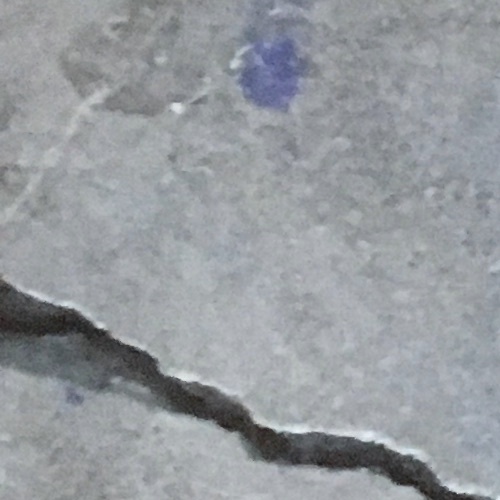}}
    \end{subfigure}
    \caption{Samples from the datasets used for zero-shot evaluations. The first, second, and third rows, correspond to the Road420, Facade390, and Concrete3k datasets, respectively.}
    \label{fig:zero_shot_samples}
\end{figure}

To further assess the robustness of the segmentation models, three out-of-distribution (OOD) datasets were used for zero-shot evaluation: Road420 and Facade390, introduced by \citep{geFinetuningVisionFoundation2024}, as well as Concrete3k collected by \citep{liRealtimeHighresolutionNeural2023, wangAutomaticConcreteCrack2022}. Sample images from each dataset are depicted in Figure \ref{fig:zero_shot_samples}.

\begin{table}[h]
\small
\renewcommand{\arraystretch}{1.25}
    \centering
    \caption{Overview of the crack datasets used in this study. OmniCrack30k is used for training and testing models, while Road420, Facade390, and Concrete3k are for zero-shot evaluations.}
    \label{table:omnicrack split}
    \small
    \begin{tabular}{l c c c c}
        \hline\hline 
        \multicolumn{1}{c}{} & 
        \multicolumn{1}{c}{Train} & 
        \multicolumn{1}{c}{Validation} & 
        \multicolumn{1}{c}{Test} &
        \multicolumn{1}{c}{Image Size}\\
        \hline
        OmniCrack30k & 22,158 & 13,277 & 4,582 & $[81, 4608] \times [116, 4608]$ \\
        Road420 & - & - & 420 & $448\times 448$ \\
        Facade390 & - & - & 390 & $448\times 448$\\
        Concrete3k & - & - & 3,000 & $500\times 500$\\
        \hline\hline
    \end{tabular}
    \normalsize
\end{table}

\subsection{Experimental Setup and Hyperparameter Tuning} \label{subsec:hyperparameter_tuning}
To evaluate the performance of the segmentation models, precision, recall, and F1-score  were used as the metric, which are computed based on True Positive (TP), True Negative (TN), False Positive (FP), and False Negative (FN) (Eq.\ref{eq:precision}-\ref{eq:f1-score}).  Additionally, to quantify the similarity between predicted masks and ground truth annotations, Intersection over Union (IoU) was used as the metric for this measurement (Eq. \ref{eq:iou}).

\begin{equation} \label{eq:precision}
Precision (Pre)=\frac{TP}{TP\ +\ FP} \;.
\end{equation}

\begin{equation} \label{eq:recall}
Recall (Re)=\frac{TP}{TP\ +\ FN} \;.
\end{equation}

\begin{equation} \label{eq:f1-score}
F1-Score=\frac{2\times Pr\ \times Re}{Pr\ +\ Re} \;.
\end{equation}

\begin{equation} \label{eq:iou}
IoU=\frac{\left|p\cap g\right|}{|p\cup g|} \;.
\end{equation}
In Eq.\ref{eq:iou}, $p$ and $g$ represent the predictions and ground truth labels, respectively.

In semantic segmentation, the standard loss function is the cross-entropy loss (Eq. \ref{eq:bce loss}).  However, since crack segmentation is an imbalanced problem, using cross-entropy loss, which treats samples from both classes (crack and non-crack) equally, may lead to a model bias toward the background. To mitigate this issue, a specific loss function should be considered to ensure balanced learning. According to the previous studies (\citep{geFinetuningVisionFoundation2024}, Dice loss (Eq. \ref{eq:dice loss}) can be effective in such imbalanced training problems \citep{milletariVNetFullyConvolutional2016b}, therefore, both Dice loss and a hybrid loss function combining cross-entropy and Dice losses were evaluated as hyperparameters to determine the most effective objective function for crack segmentation. Furthermore, in all experiments, AdamW was used as the optimizer with a weight decay of 0.00005 and cosine learning rate scheduler.

To determine the optimal learning rate and batch size, as well as the loss function, a systematic hyperparameter tuning was conducted on the SAM model using the first adaptation strategy, fine-tuning the normalization parameters. 
Three main configurations were defined as specified in Table \ref{table:hyperparameter tuning settings}. In configurations 1 and 3, a hybrid loss function was employed, incorporating a Lagrangian multiplier ($\lambda$) to balance the contribution of each loss component. The value of ($\lambda$) was treated as a hyperparameter to be optimized. To explore the hyperparameter space efficiently, random search was used, sampling 20\% and 5\% of the possible hyperparameter configurations, considering a range of values for learning rate and batch size. Each randomly selected experiment was conducted for four epochs, and the optimal hyperparameters were chosen based on the highest validation F1-score. All experiments were implemented using the PyTorch framework.

\begin{equation} \label{eq:bce loss}
L_{BCE}=-\sum_{i=1}^{N}\left[g_ilog\left(p_i\right)+\left(1-g_i\right)log\left(1-p_i\right)\right] \;.
\end{equation}

\begin{equation} \label{eq:dice loss}
L_{Dice}=1-\frac{2\sum_{i}^{N}{p_ig_i}}{\sum_{i}^{N}{p_i^2+}\sum_{i}^{N}g_i^2} \;.
\end{equation}

In Eq.\ref{eq:bce loss} and Eq.\ref{eq:dice loss}, $N$ is the number of pixels, $p_i$ and $g_i$ represent the predictions and ground truth labels, respectively.

\begin{table}[h]
\centering
\caption{Details of the three hyperparameter tuning configurations for the SAM-ViT-b model, optimized using the layer normalization fine-tuning method. In the first two configurations, 20\% of the hyperparameter combinations were tested by the random search, while the last configuration involved testing 5\% of the combinations. The table provides values for loss functions, $\lambda$, learning rates, and batch sizes used in each configuration.}
\label{table:hyperparameter_tuning_settings}
\small
\renewcommand{\arraystretch}{1.25}
\begin{tabular}{p{0.3cm} p{3.4cm} p{3.3cm} p{5cm} l}
    \hline\hline
    \multicolumn{1}{l}{Config.} &
    \multicolumn{1}{c}{Loss} & 
    \multicolumn{1}{c}{$\lambda$} & 
    \multicolumn{1}{c}{Learning Rate} &
    \multicolumn{1}{c}{Batch\textsuperscript{$\ast$}} \\
    \hline
    1 & $\lambda L_{BCE}+(1-\lambda)L_{Dice}$ & 
    $\begin{array}{l} [0.01, 0.05, 0.1, \\ 0.15, 0.2, 0.25, 0.3] \end{array}$ & 
    $\begin{array}{c} [0.0001:0.0001:0.0005, \\ 0.001, 0.002] \end{array}$ & $[2, 4, 8]$\\[5pt]
    
    2 & $\lambda L_{BCE}+(1-\lambda)L_{Dice}$ & 
    $\begin{array}{c} [0] \end{array}$ & 
    $\begin{array}{c} [0.0001:0.0001:0.0005, \\ 0.001, 0.002] \end{array}$ & $[2, 4, 8]$\\[5pt]
    
    3 & $L_{BCE} + \lambda L_{Dice}$ & 
    $\begin{array}{l} [0.5:0.05:0.95] \end{array}$ & 
    $\begin{array}{c} [0.0005:0.0002: 0.0014] \end{array}$ & $[2, 4, 8]$ \\
    \hline
    \multicolumn{2}{l}{\footnotesize\textsuperscript{$\ast$}Batch sizes.}\\
    \hline\hline
\end{tabular}
\end{table}

\subsection{Baseline Semantic Segmentation Models} \label{subsec:cross-model}
 To validate the effectiveness of the proposed fine-tuning approach, a comprehensive comparison was conducted using three crack segmentation models and four general-purpose segmentation models. The crack segmentation models include DeepCrack \citep{zouDeepCrackLearningHierarchical2019} (represented as DeepCrackZ), DeepCrack \citep{liuDeepCrackDeepHierarchical2019} (represented as DeepCrackL), and CrackFormer \citep{liuCrackFormerTransformerNetwork2021}. DeepCrackZ and DeepCrackL are based on convolutional neural networks (CNNs), and CrackFormer is a transformer-based architecture to capture long-range dependencies in crack structures.
 
 The four widely used segmentation models include: a) SegFormer \citep{xieSegFormerSimpleEfficient2021}, a transformer-based model designed for efficient and scalable segmentation, b) U-Net \citep{ronnebergerUNetConvolutionalNetworks2015}, a CNN-based U-shaped architecture with skip connections, initially developed for biomedical image segmentation, c) DeepLabv3+ \citep{chenRethinkingAtrousConvolution2017}, evaluated with two backbone architectures, ResNet50 and ResNet101, incorporating atrous convolution for multi-scale feature extraction.
 
 The specifications of the segmentation methods, along with SAM's, are provided in Table \ref{table:baseline_methods}. The pretrained weights for each model refer to those obtained from pretraining on their respective datasets. All segmentation methods were evaluated under both full fine-tuning and the proposed fine-tuning approach, enabling an in-depth analysis of their adaptation to crack segmentation task. 

\clearpage
\begin{landscape}
    \thispagestyle{empty}  
    \begin{table}[h]
        \centering
        \caption{Specifications of the baseline semantic segmentation methods considered in this study.}
        \label{table:baseline_methods}
        \small
        \renewcommand{\arraystretch}{1.25}
        \begin{tabular}{l l l p{1cm} | l l | l l l | c}
            \hline\hline
            \multicolumn{1}{c}{Method} & 
            \multicolumn{1}{c}{Architecture} &
            \multicolumn{1}{c}{Backbone} &
            \multicolumn{1}{c|}{\parbox[t]{1.5cm}{Norm \\ Layer}} &
            \multicolumn{1}{c}{\# Params} & 
            \multicolumn{1}{c|}{\parbox[t]{1.5cm}{\% Norm \\ Params}} &
            \multicolumn{1}{c}{\parbox[t]{1.5cm}{FLOPs\textsuperscript{$\ast$} \\ (G)}} & 
            \parbox[t]{2.5cm}{Inference Time \\ (s/image)} &
            \multicolumn{1}{c|}{\parbox[t]{2cm}{Pre-Trained \\ Dataset}} &
            \multicolumn{1}{c}{Tuning Method}\\
            \hline
            \hline
            DeepCrackZ & CNN & VGG16 & - & 31M & 0 & 137.46 &  0.198 & CrackTree260 & Full \\
            DeepCrackL & CNN & VGG16 & BN & 15M & 0.057 & 20.09 & 0.201 & DeepCrack & Full/Proposed \\
            CrackFormer & Transformer & \parbox[t]{2.5cm}{VGG16 \\ + Self-Attention} & GN & 5M & 0.797 & 22.71 &  0.106 & CrackTree260 & Full/Proposed\\ [5pt]
            \noalign{\vspace{0.2cm}} 
            \hline
            SegFormer & Transformer & MiT-B0 & BN \& LN & 3.7M & 0.207 & 15.73 & 0.147 & ImageNet1K & Full/Proposed\\
            U-Net & CNN & FCN & BN & 32.5M & 0.170 & 38.35 & 0.156 & ImageNet1K & Full/Proposed\\
            DeepLabv3+ & CNN & ResNet50 &  BN & 42M & 0.136 & 43.39 &  0.189 & COCO2017 & Full/Proposed\\
            DeepLav3+ & CNN & ResNet101 & BN & 61M & 0.180 & 62.84 &  0.227 & COCO2017 & Full/Proposed\\
            SAM & Transformer & ViT-Base & LN & 90M & 0.046 & 34.58 & 0.176 & SA-1B & Proposed\\
            \hline
            \multicolumn{10}{l}{\textsuperscript{$\ast$}Per image size of $256\times 256$.} \\
            \multicolumn{10}{l}{BN: Batch Normalization, GN: Group Normalization, LN: Layer Normalization} \\
            \hline\hline
        \end{tabular}%
        \normalsize
    \end{table}
\end{landscape}
\clearpage

\section{Results \& Discussion}
\subsection{Hyperparameter Tuning}
In the first hyperparameter tuning experiment, according to \citep{geFinetuningVisionFoundation2024}, a combination of binary cross entropy (BCE) and Dice losses was examined as the objective function. The learning rate was sampled from a range of $[0.0001:0.0001:0.0009, 0.001, 0.002]$, and the batch size was chosen from [2, 4, 8]. A random 20\% subset of all possible configurations was evaluated, resulting in 30 combinations, each trained for 4 epochs. 
The results, given in Table \ref{table:hyperparameter tuning results}, indicated that while training with batch size of 2 and 8 yielded a more stable training, the best-performing configuration was achieved with a Dice loss weight of $\lambda=0.2$, a learning rate of 0.0001, and a batch size of 4, yeilding a validation F1-score of 0.6646. The test performance of the model trained with these hyperparameters achieved an F1-score of 57.38\%, with an IoU of 40.24\%. 

Analysis of the results suggested that decreasing $\lambda$ and increasing the weight on Dice loss improves F1-score performance, highlighting the effectiveness of Dice loss in handling class imbalance. Based on this observation, two additional loss functions were considered for further experimentation: 1) Pure Dice Loss by setting $\lambda=0$ in $\lambda L_{BCE}+(1-\lambda)L_{Dice}$, 2) Weighted Hybrid Loss: A reformulated combined objective function, expressed as $L_{BCE} + \lambda L_{Dice}$, where a higher weight was assigned to Dice loss.

In the second tuning phase, experiments were conducted using Dice loss with the same learning rate range and batch size values. The best configuration included a learning rate of 0.0008, a batch size of 8 with a validation F1-score of 0.5977 (Table \ref{table:hyperparameter tuning results}). The test performance of the model trained with these hyperparameters yielded an F1-score of 58.28\% and an Intersection over Union (IoU) of 41.13\%. 

Furthermore, a final tuning experiment was conducted to optimize the hybrid loss function with an increased weight on Dice loss, where the Dice loss weight $\lambda$ was sampled from the range [0.5:0.05:0.95]. The learning rate was chosen from [0.0005:0.0002:0.0014], and batch size options remained [2, 4, 8]. This search covered 5\% of the possible combinations, resulting in 15 evaluations. The best configuration was identified with $\lambda=0.65$, a learning rate of 0.0005, and a batch size of 2, which showed a more stable training in comparison to batch size of 4 or 8 (Table \ref{table:hyperparameter tuning results}). This setting yielded the highest validation F1-score of 0.6696, and the test performance of the model fine-tuned according to this setting was 61.22\% F1-score and 44.13\% IoU. Therefore, this setting was chosen as the best configuration for the rest of the training experiments in the study.

The results demonstrated that a hybrid objective function with a stronger weight on Dice loss is most effective for crack segmentation, as it enhances both boundary detection and shape consistency. 

\begin{table}[h]
\centering
\caption{Results of the hyperparameter tuning experiments. The optimal hyperparameters are provided along with the model's test performance trained with each configuration. The best hyperparameters are highlighted in bold.}
\label{table:hyperparameter tuning results}
\small
\renewcommand{\arraystretch}{1.25}
\begin{tabular}{c c c c c c c}
    \hline\hline
    \multicolumn{1}{c}{Config.} & 
    \multicolumn{1}{c}{Loss} & 
    \multicolumn{1}{c}{$\lambda$} & 
    \multicolumn{1}{c}{Learning Rate} &
    \multicolumn{1}{c}{Batch} & 
    \multicolumn{1}{c}{F1-Score (\%)} &
    \multicolumn{1}{c}{IoU (\%)} \\
    \hline\hline
    1 & $\lambda L_{BCE}+(1-\lambda)L_{Dice}$ & 0.2 & 0.0001 & 4 & 57.38 & 40.24  \\
     2 & $L_{Dice}$ & - & 0.0008 & 8 & 58.28 & 41.13 \\
     \textbf{3} & $\mathbf{L_{BCE} + \lambda L_{Dice}}$ & \textbf{0.65} & \textbf{0.0005}  & \textbf{2} & \textbf{61.22} & \textbf{44.13} \\
    \hline
\end{tabular}%
\normalsize
\end{table}

\subsection{Fine-Tuning Ablation Study}
\begin{figure}[h] 
    \centering
    \includegraphics[width=3.0in]{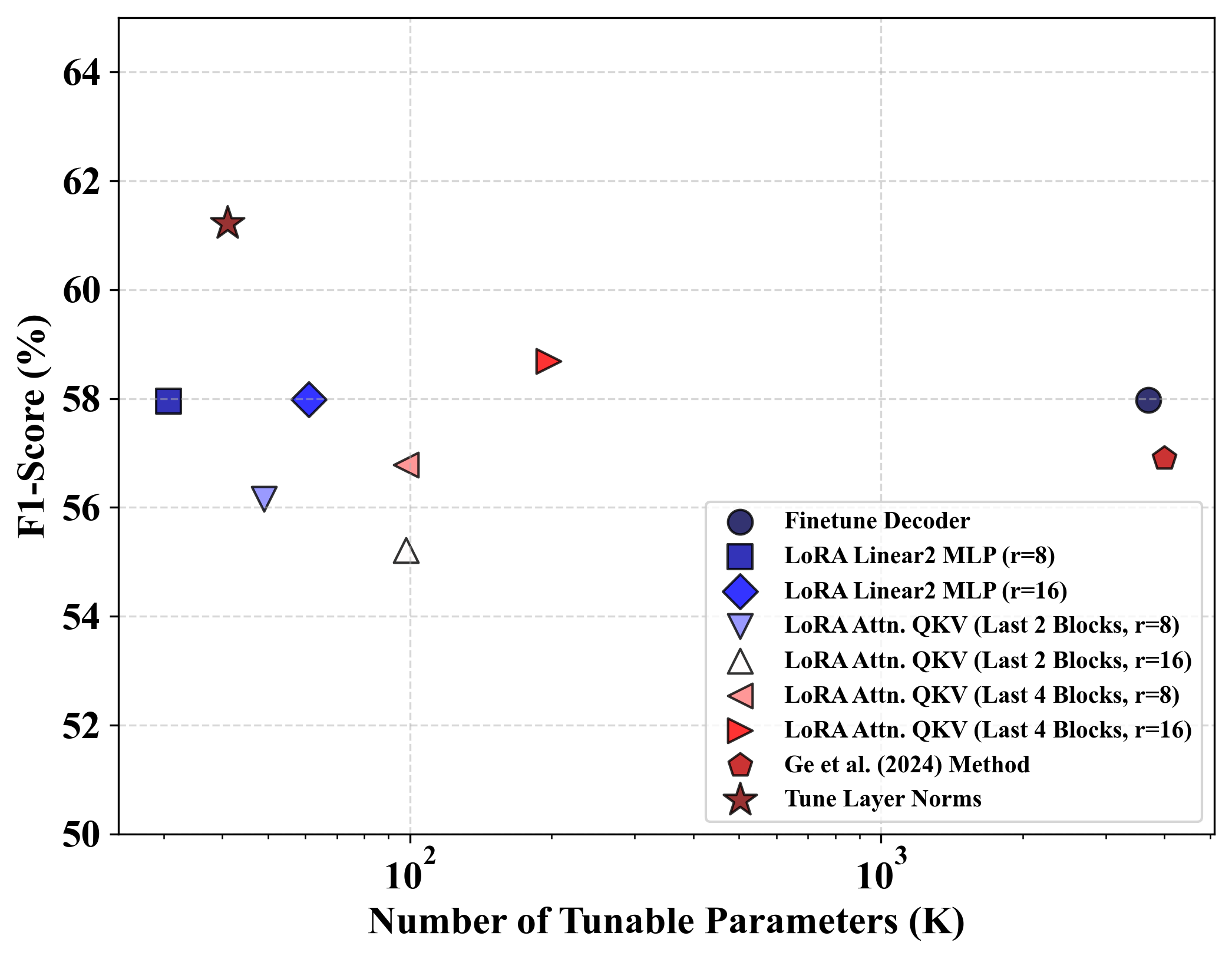}
    \caption{Ablation results of adapting the Segment Anything Model (SAM) using different fine-tuning methods. The proposed normalization fine-tuning method achieves the highest performance among other methods with a relatively lower number of trainable parameters.}
    \label{fig:finetuning_ablation}
\end{figure}

\begin{figure}[h]
    \captionsetup[subfigure]{labelformat=empty, font=footnotesize, justification=centering, position=top} 
    \centering

    \begin{minipage}{0.9\textwidth}
        \scriptsize
        \hspace*{1.6cm}
        \begin{tabular}{@{}c@{\hspace{7mm}}c@{\hspace{4mm}}c@{\hspace{5mm}}c@{\hspace{8mm}}c@{\hspace{7mm}}c@{}}
            & Input Image & GroundTruth & Probability & Overlay & Binary Mask
        \end{tabular}
    \end{minipage}

    \vspace{0.1cm} 

    \begin{minipage}{0.9\textwidth}
        \centering
        \begin{subfigure}[b]{0.1\textwidth}
            \subcaption*{\rotatebox{90}{\scriptsize Untuned}} 
        \end{subfigure}
        \begin{subfigure}[b]{0.8\textwidth}
            \adjustbox{trim=10 10 10 10,clip,max width=0.75\textwidth}{\includegraphics{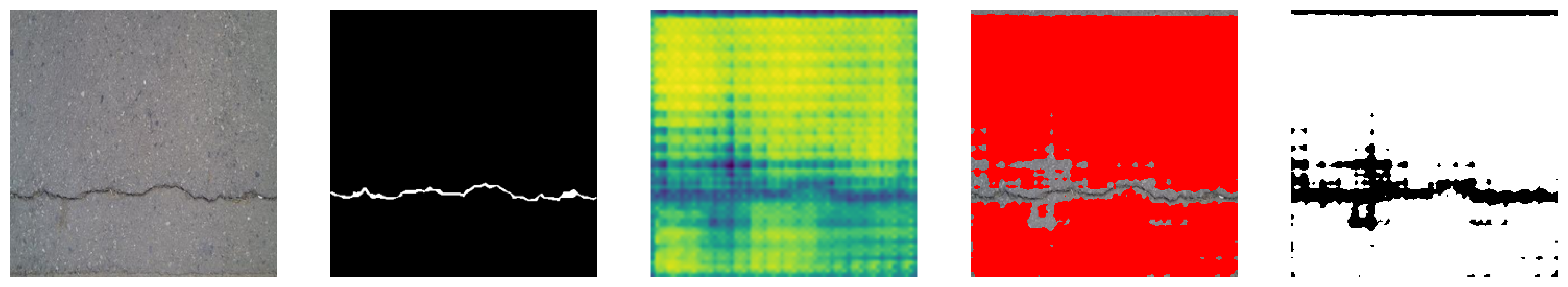}}
        \end{subfigure}

        \vspace{0.1cm}

        \begin{subfigure}[b]{0.1\textwidth} 
            \subcaption*{\rotatebox{90}{\scriptsize Fine-Tuned}} 
        \end{subfigure}
        \begin{subfigure}[b]{0.8\textwidth}
            \adjustbox{trim=10 10 10 10,clip,max width=0.75\textwidth}{\includegraphics{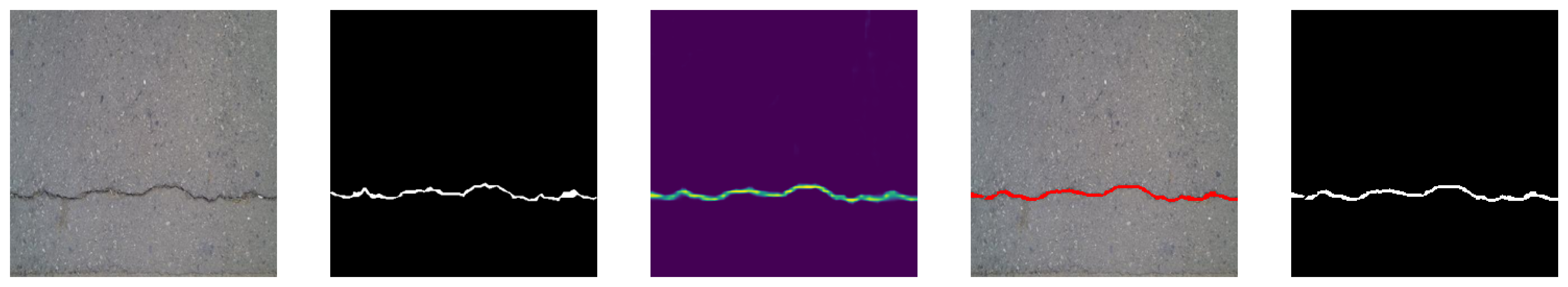}}
        \end{subfigure}
    \end{minipage}

    \vspace{1cm} 

    \begin{minipage}{0.9\textwidth}
        \centering
        \begin{subfigure}[b]{0.1\textwidth} 
            \subcaption*{\rotatebox{90}{\scriptsize Untuned}} 
        \end{subfigure}
        \begin{subfigure}[b]{0.8\textwidth}
            \adjustbox{trim=10 10 10 10,clip,max width=0.75\textwidth}{\includegraphics{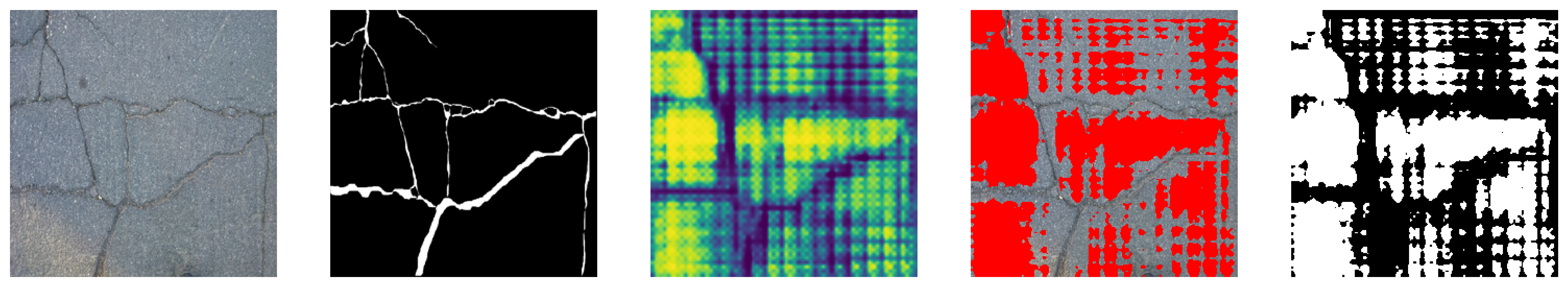}}
        \end{subfigure}

        \vspace{0.05cm}

        \begin{subfigure}[b]{0.1\textwidth} 
            \subcaption*{\rotatebox{90}{\scriptsize Fine-Tuned}} 
        \end{subfigure}
        \begin{subfigure}[b]{0.8\textwidth}
            \adjustbox{trim=10 10 10 10,clip,max width=0.75\textwidth}{\includegraphics{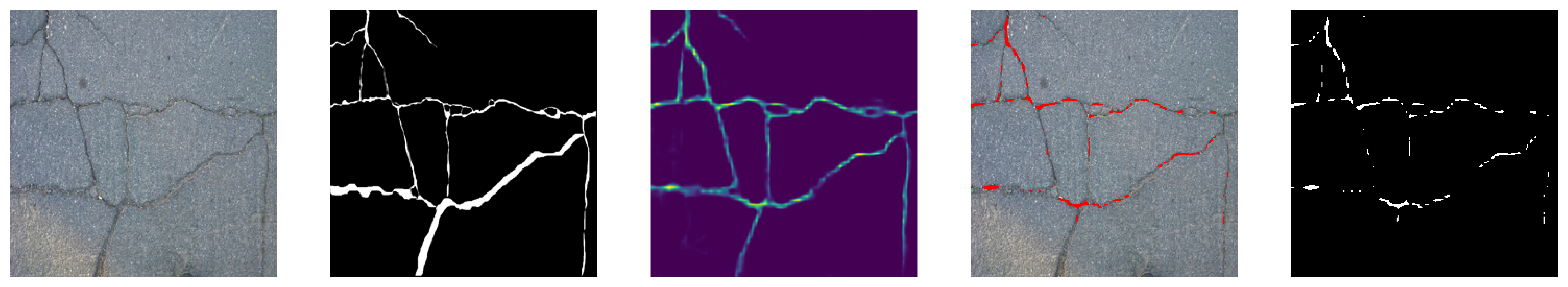}}
        \end{subfigure}
    \end{minipage}

    \vspace{1cm}

    \begin{minipage}{0.9\textwidth}
        \centering
        \begin{subfigure}[b]{0.1\textwidth} 
            \subcaption*{\rotatebox{90}{\scriptsize Untuned}} 
        \end{subfigure}
        \begin{subfigure}[b]{0.8\textwidth}
            \adjustbox{trim=10 10 10 10,clip,max width=0.75\textwidth}{\includegraphics{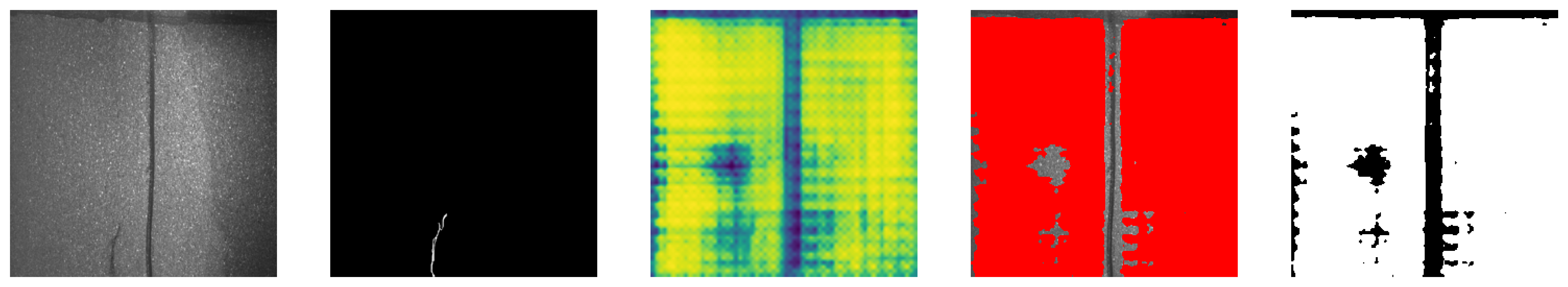}}
        \end{subfigure}

        \vspace{0.05cm}

        \begin{subfigure}[b]{0.1\textwidth} 
            \subcaption*{\rotatebox{90}{\scriptsize Fine-Tuned}} 
        \end{subfigure}
        \begin{subfigure}[b]{0.8\textwidth}
            \adjustbox{trim=10 10 10 10,clip,max width=0.75\textwidth}{\includegraphics{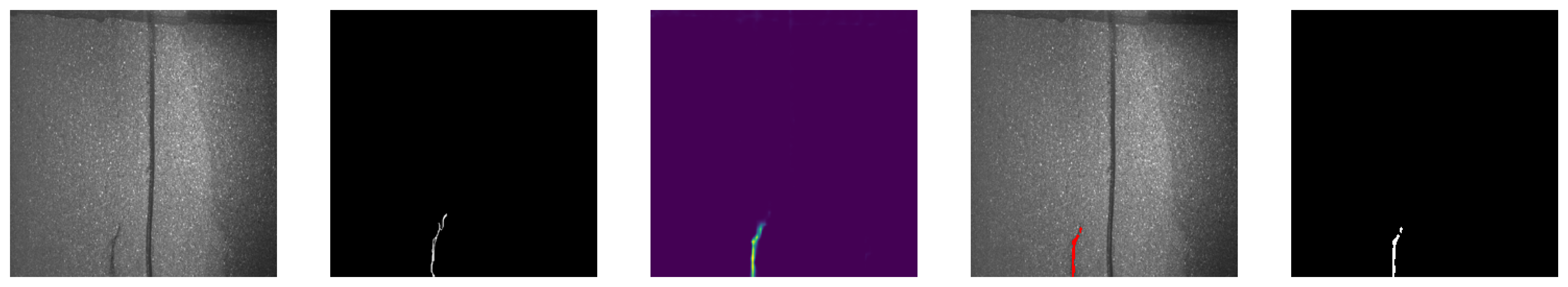}}
        \end{subfigure}
    \end{minipage}

    \caption{Comparison of the untuned and fine-tuned Segment Anything Model (SAM) segmentation results on three sample input images. In each case, the first row illustrates the prediction mask of the unturned SAM and the second row demonstrates the prediction output of the fine-tuned model.}
    \label{fig:sac_sam_qualitative}
\end{figure}

The fine-tuning ablation results are presented in Table \ref{table:sam fine-tuning ablation} and Figure \ref{fig:finetuning_ablation}, which illustrate the impact of different fine-tuning strategies on the segmentation performance of SAM on the OmniCrack30K dataset. The results indicate that tuning only the Layer Normalization (LN) parameters in both the encoder and decoder achieved the highest F1-score (61.22\%) and IoU (44.13\%) on OmniCrack30k test set, surpassing all other fine-tuning approaches while updating only 41K additional parameters (0.0458\% of the total model parameters). Sample qualitative results of the untuned and fine-tuned model are demonstrated in \ref{fig:sac_sam_qualitative}, which shows the effectiveness of the proposed tuning method in adapting large-scale foundation models to domain-specific tasks. The fine-tuned model is referred to as the Segment Any Crack (SAC) model.

Compared to full decoder fine-tuning, which requires 3.7M parameters, and the PEFT-based LoRA methods, which vary from 30K to 197K parameters, tuning the normalization layers provides a significantly more efficient trade-off between performance and parameter efficiency. This approach even outperforms the method proposed by \citet{geFinetuningVisionFoundation2024}, which employs a combination of LoRA-based encoder tuning and decoder fine-tuning with 4M parameters, further validating the effectiveness of selective normalization adaptation.

While achieving superior accuracy, the training time per iteration (12.3 min/it) for normalization tuning is higher than that of decoder-only tuning (7.9 min/it) but remains lower than that of the \citet{geFinetuningVisionFoundation2024}'s method (14.8 min/it). This suggests that while normalization tuning introduces a modest computational overhead, it remains a practical and scalable fine-tuning strategy for real-world applications. Overall, the results highlight the potential of parameter-efficient fine-tuning techniques, where strategically updating only normalization layers can yield substantial performance gains while keeping additional trainable parameters minimal.

\begin{table}[h]
\centering
\caption{Fine-tuning ablative results with SAM model. }
\label{table:sam fine-tuning ablation}
\small
\renewcommand{\arraystretch}{1.5}
\begin{tabular}{p{6.7cm} p{1.2cm} p{1.2cm} p{1cm}  p{1cm} p{1.4cm}}
\hline\hline
\multicolumn{1}{c}{Tuning Method} & 
\multicolumn{1}{c}{\# Params} & 
\multicolumn{1}{c}{\% Params} & 
\multicolumn{1}{c}{\parbox[t]{1cm}{F1\\ (\%)}} &
\multicolumn{1}{c}{\parbox[t]{1cm}{IoU\\ (\%)}} &
\multicolumn{1}{c}{\parbox[t]{1.4cm}{Training \\ Time \\(min/it)}}\\
\hline
No Finetuning & 0 & 0 &  13 & 17 & - \\
Finetune Decoder & 3.7M & 4.174 &  57.97 & 40.83 & 7.9 \\
\parbox[t]{8cm}{PEFT Encoder, LoRA -\\  2nd Linear of MLP, last ViT block, $r^{\:\ast}=8$} & 30.7K & 0.0339 &  57.95 & 40.81 & 9.9 \\
\parbox[t]{8cm}{PEFT Encoder, LoRA - \\  2nd Linear of MLP last ViT block, $r=16$} & 61K & 0.0679 &  57.99 & 40.85 & 9.9 \\
\parbox[t]{8cm}{PEFT Encoder, LoRA - \\  
Attn. QKV, last 2 ViT blocks, $r=8$} & 49K & 0.0543 &  56.16 & 39.05 & 8.3 \\

\parbox[t]{8cm}{PEFT Encoder, LoRA - \\  Attn. QKV, last 2 ViT blocks, $r=16$}  & 98K & 0.109 & 55.22 & 38.14 & 8.3\\

\parbox[t]{8cm}{PEFT Encoder, LoRA -\\  Attn. QKV, last 4 ViT blocks, $r=8$}& 98K& 0.109 & 56.78 & 39.66 & 8.3\\

\parbox[t]{8cm}{PEFT Encoder, LoRA - \\  Attn. QKV, last 4 ViT blocks, $r=16$} & 197K & 0.217 & 58.69 & 41.55 & 8.3\\
\citet{geFinetuningVisionFoundation2024} Method$^{\ast\ast}$ & 4M & 4.507 & 56.9 & 39.79 & 14.8\\
\textbf{Tune Layer Norms} & \textbf{41K} & \textbf{0.0458} & \textbf{61.22} & \textbf{44.13} & \textbf{12.3}\\
\hline
\multicolumn{3}{l}{\footnotesize $\ast$ $r$ represents the rank of the LoRA matrices} \\
\multicolumn{6}{l}{\footnotesize $\ast\ast$ PEFT Encoder with LoRA ($r=8$), Finetune Decoder and Prompt Encoder, No input prompt} \\
\hline\hline
\end{tabular}%
\normalsize
\end{table}

\subsection{Performance Evaluations}
\begin{figure}[h]
    \centering
    \begin{subfigure}{0.39\textwidth}
        \centering
        \caption*{(a)}
        \includegraphics[width=\linewidth]{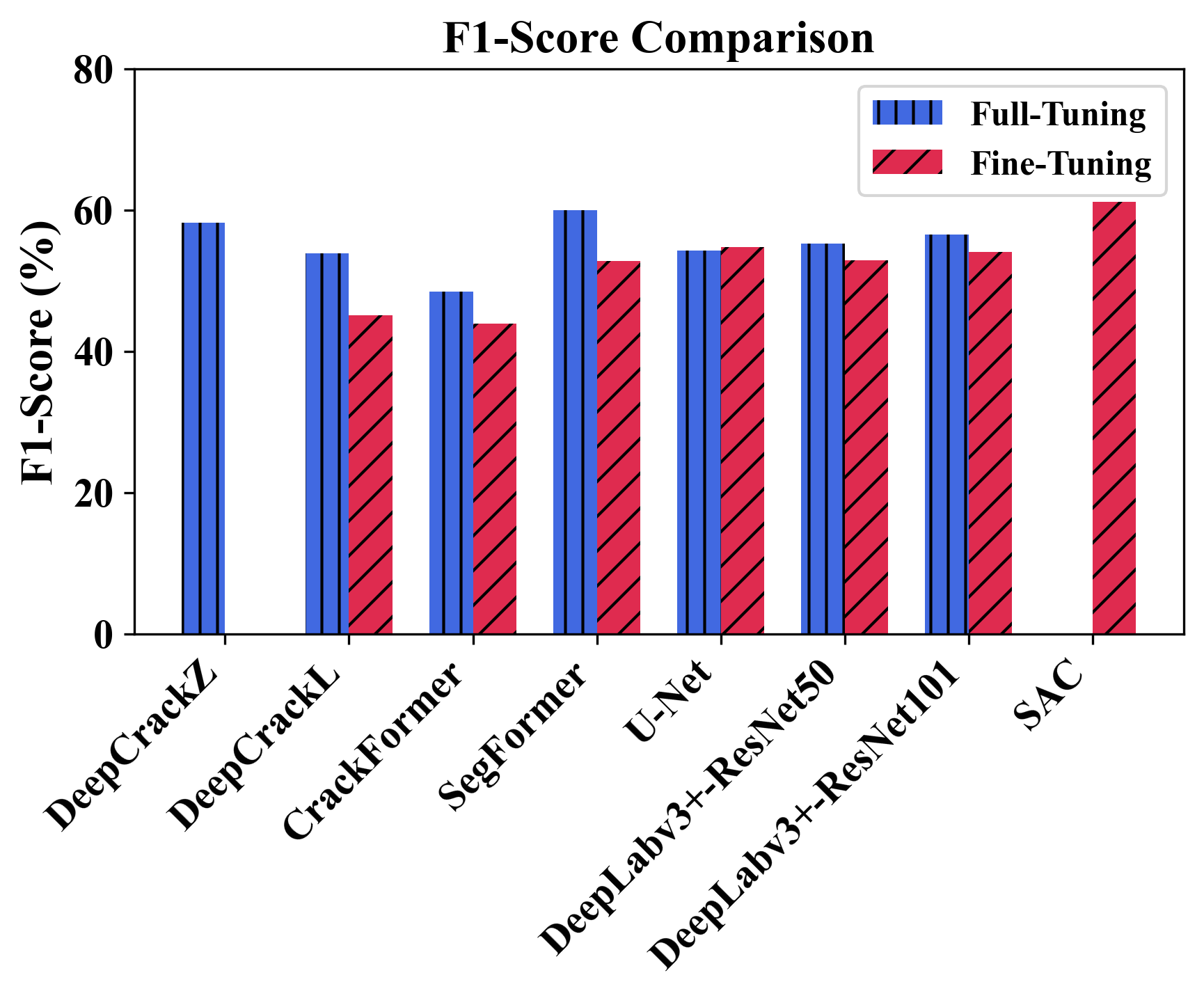}
    \end{subfigure}
    \begin{subfigure}{0.39\textwidth}
        \centering
        \caption*{(b)}
        \includegraphics[width=\linewidth]{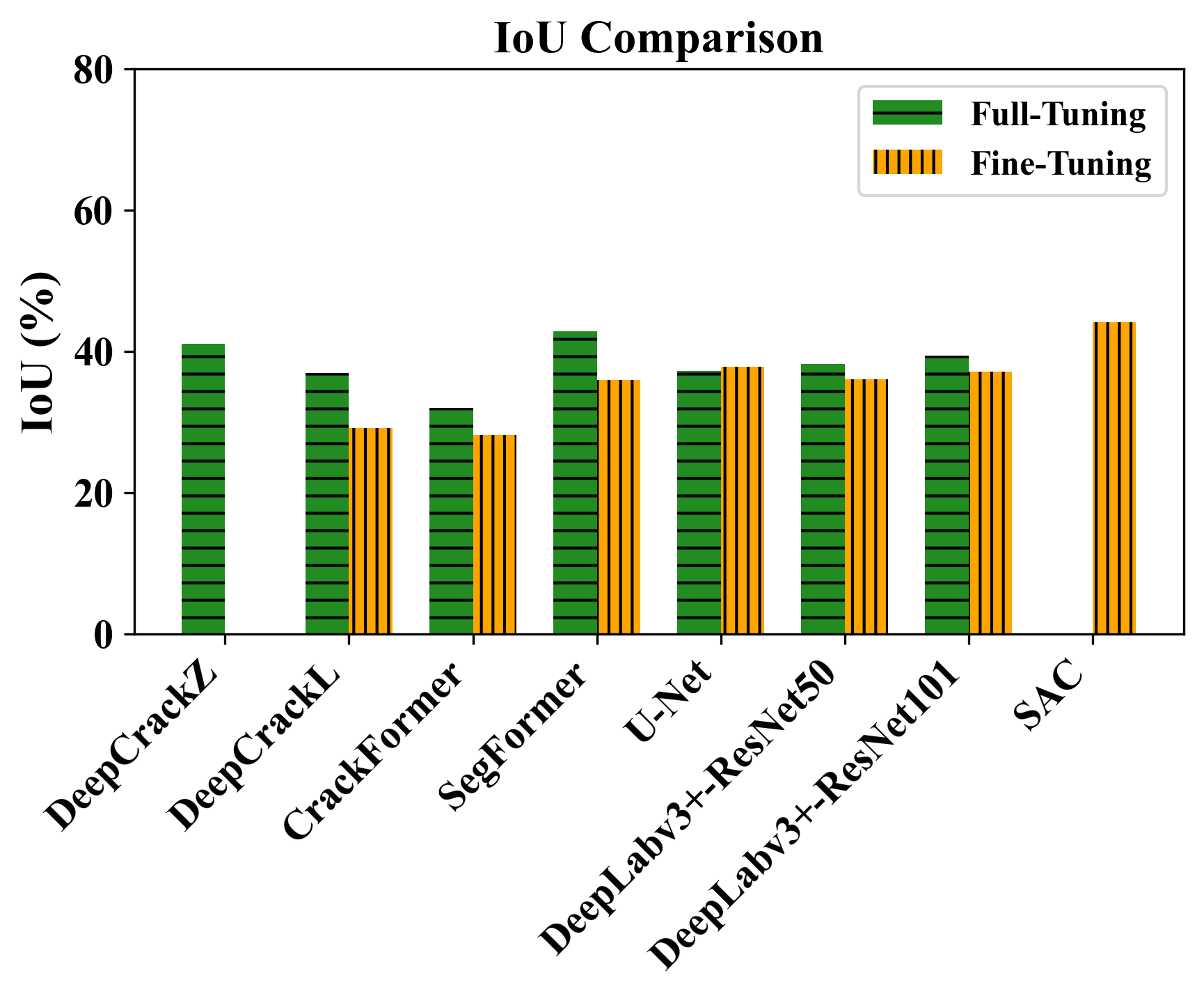}
        
    \end{subfigure}

    \begin{subfigure}{0.39\textwidth}
        \centering
        \caption*{(c)}
        \includegraphics[width=\linewidth]{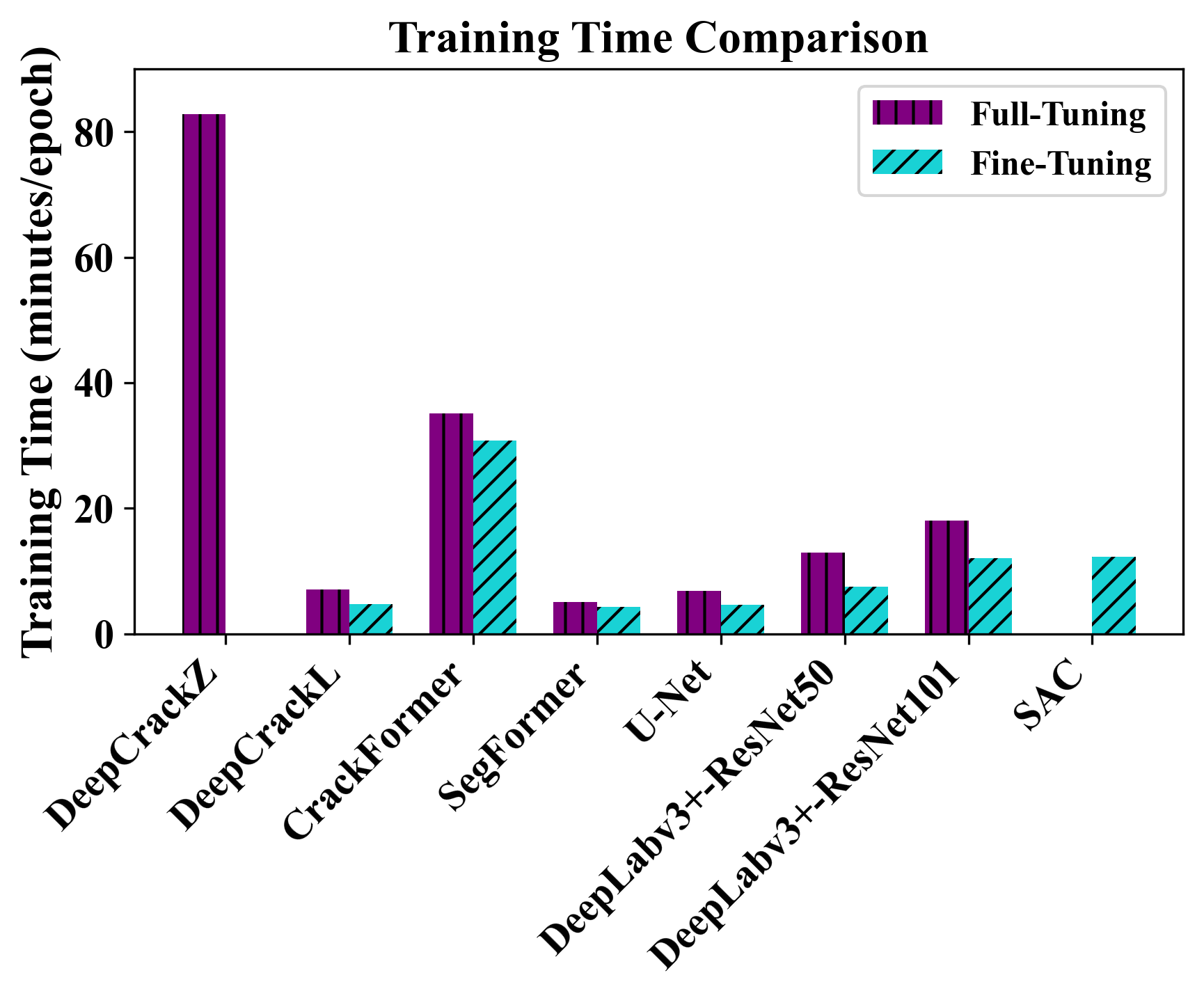}
        
    \end{subfigure}

    \caption{Comparison of the fully and fine-tuned segmentation models on OmniCrack30k test set. (a) and (b) illustrate the competitive performance gained by the proposed normalization tuning method compared to the full tuning method. (c) demonstrates the reduced training time per epoch using the proposed fine-tuning method.}
    \label{fig:performance_comparison}
\end{figure}

To further validate the effectiveness of the proposed fine-tuning approach, cross-model experiments were conducted by adapting the baseline segmentation methods using both full and fine-tuning methods. The performance results of each approach are presented in Tables \ref{table:fulltuning results} and \ref{table:fine-tuning results}.
In full-tuning, SegFormer achieved the best performance, with an F1-score of 59.98\% and an IoU of 42.85\%on OmniCrack30k test set, while maintaining a relatively low number of tunable parameters (3.7M) and the shortest training time per iteration (5.1 min/it). However, comparing the levels of performance in full and fine-tuning methods, as presented in Figure \ref{fig:performance_comparison}, it can be observed that fine-tuning only the normalization parameters yields competitive performance comparable to full tuning with significantly fewer trainable parameters. Also, this approach reduces the overall training time across all segmentation models as the number of tuning parameters drops. Furthermore, the fine-tuned SAM-based model (SAC) demonstrated superior performance, achieving the highest F1-score of 61.22\% and an IoU of 44.13\%, thereby outperforming other segmentation models in this study. These results highlight the potential of the proposed fine-tuning strategy in improving segmentation performance while optimizing computational efficiency.

\begin{figure}[h]
\captionsetup[subfigure]{labelformat=empty, font=footnotesize, justification=centering, position=top} 
    \centering
    \begin{subfigure}[b]{0.09\textwidth} 
        \subcaption*{AEL} 
    \end{subfigure}
    \begin{subfigure}[b]{0.1\textwidth}
        \subcaption*{Input Image} 
        \adjustbox{trim=10 10 10 10,clip,width=1.6cm,height=1.6cm}{\includegraphics{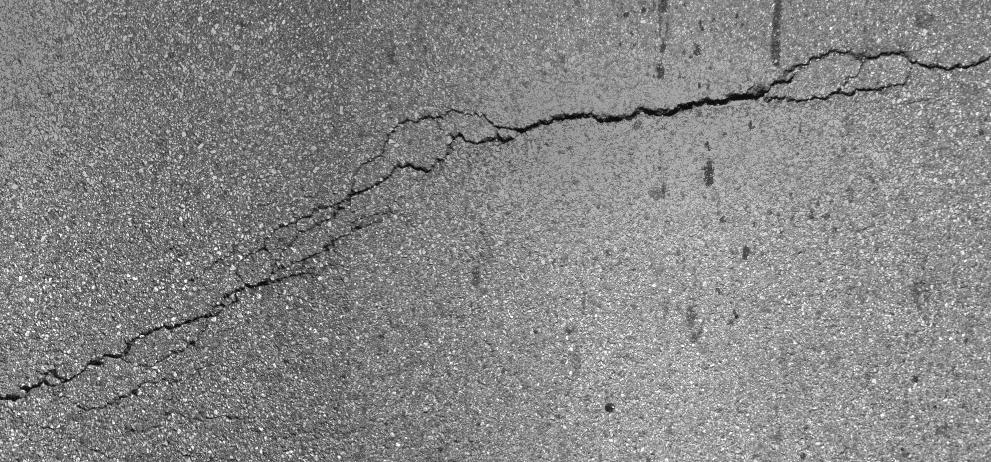}}
    \end{subfigure}
    \begin{subfigure}[b]{0.1\textwidth}
        \subcaption{GroundTruth} 
        \adjustbox{trim=10 10 10 10,clip,width=1.6cm,height=1.6cm}{\includegraphics{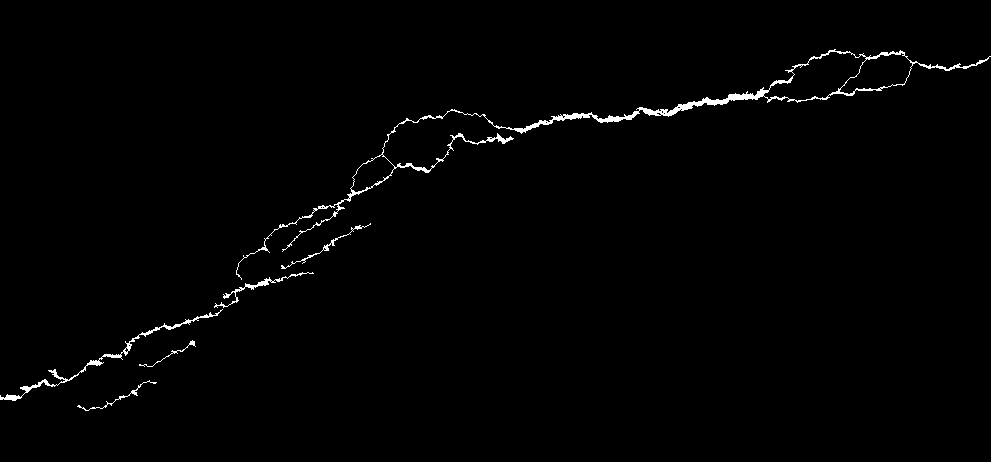}}
    \end{subfigure}
    \begin{subfigure}[b]{0.1\textwidth}
        \subcaption{SAC} 
       \adjustbox{trim=10 10 10 10,clip,width=1.6cm,height=1.6cm}{\includegraphics{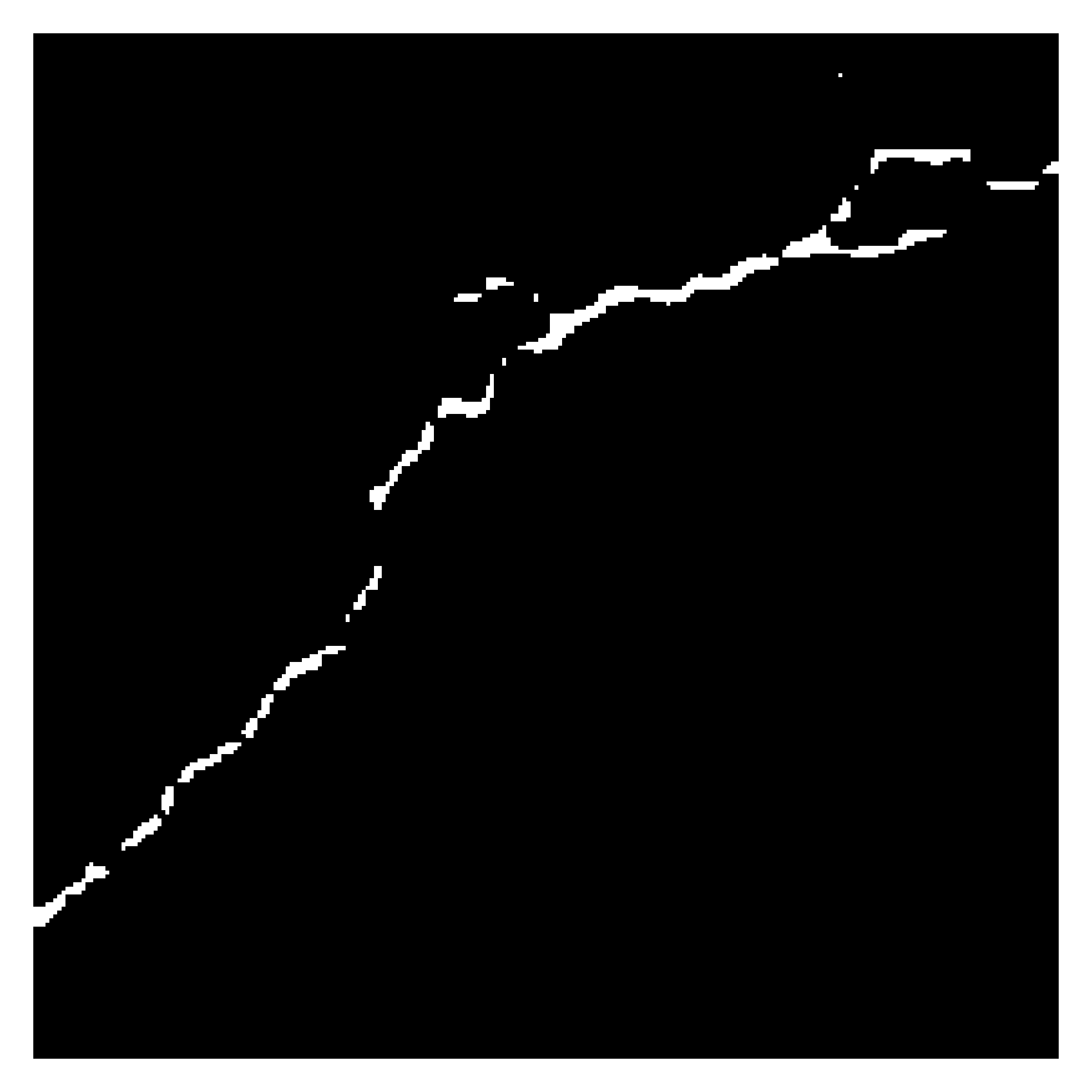}}
    \end{subfigure}
    \begin{subfigure}[b]{0.1\textwidth}
        \subcaption{DeepCrackL} 
        \adjustbox{trim=10 10 10 10,clip,width=1.6cm,height=1.6cm}{\includegraphics{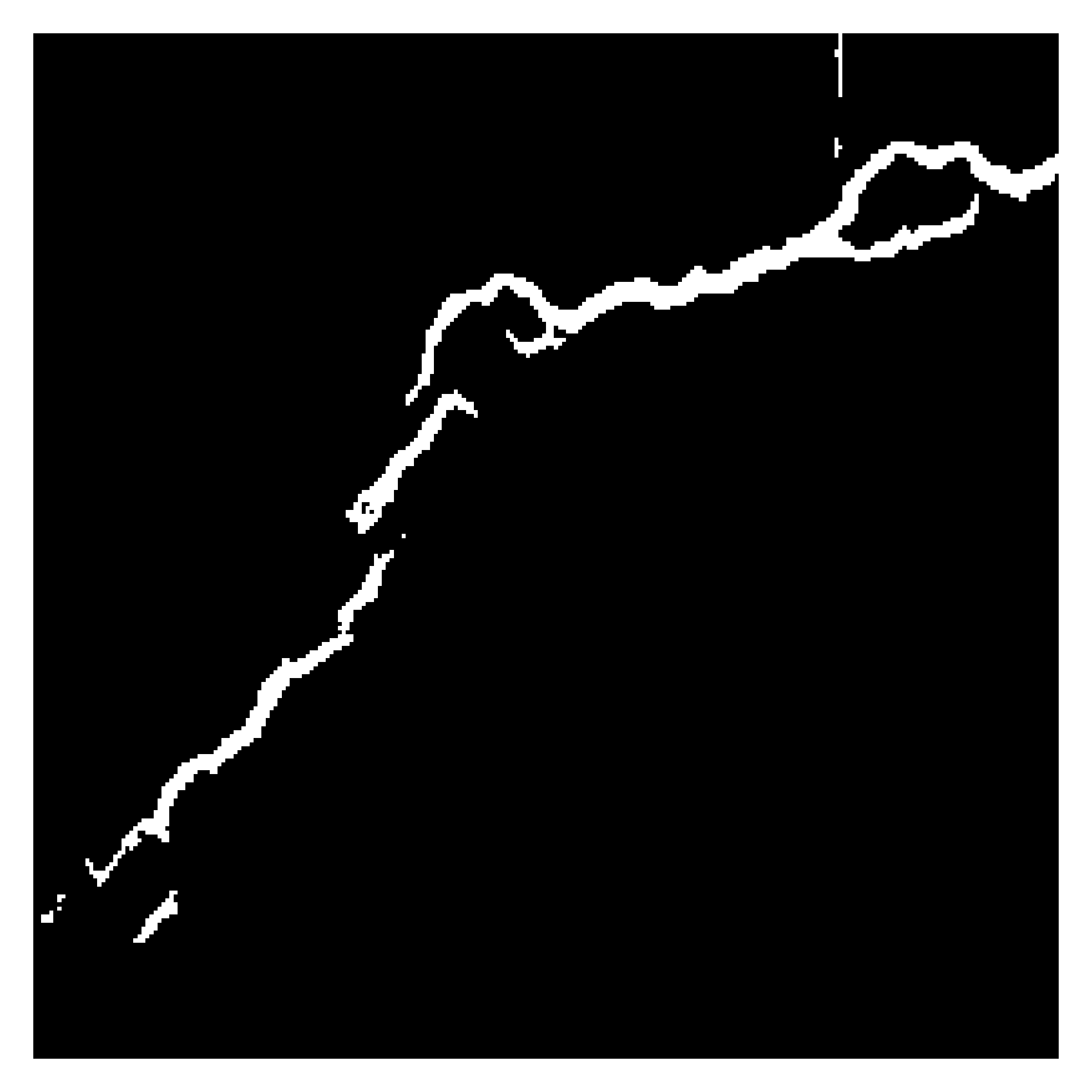}}
    \end{subfigure}
    \begin{subfigure}[b]{0.1\textwidth}
        \subcaption{CrackFormer} 
        \adjustbox{trim=10 10 10 10,clip,width=1.6cm,height=1.6cm}{\includegraphics{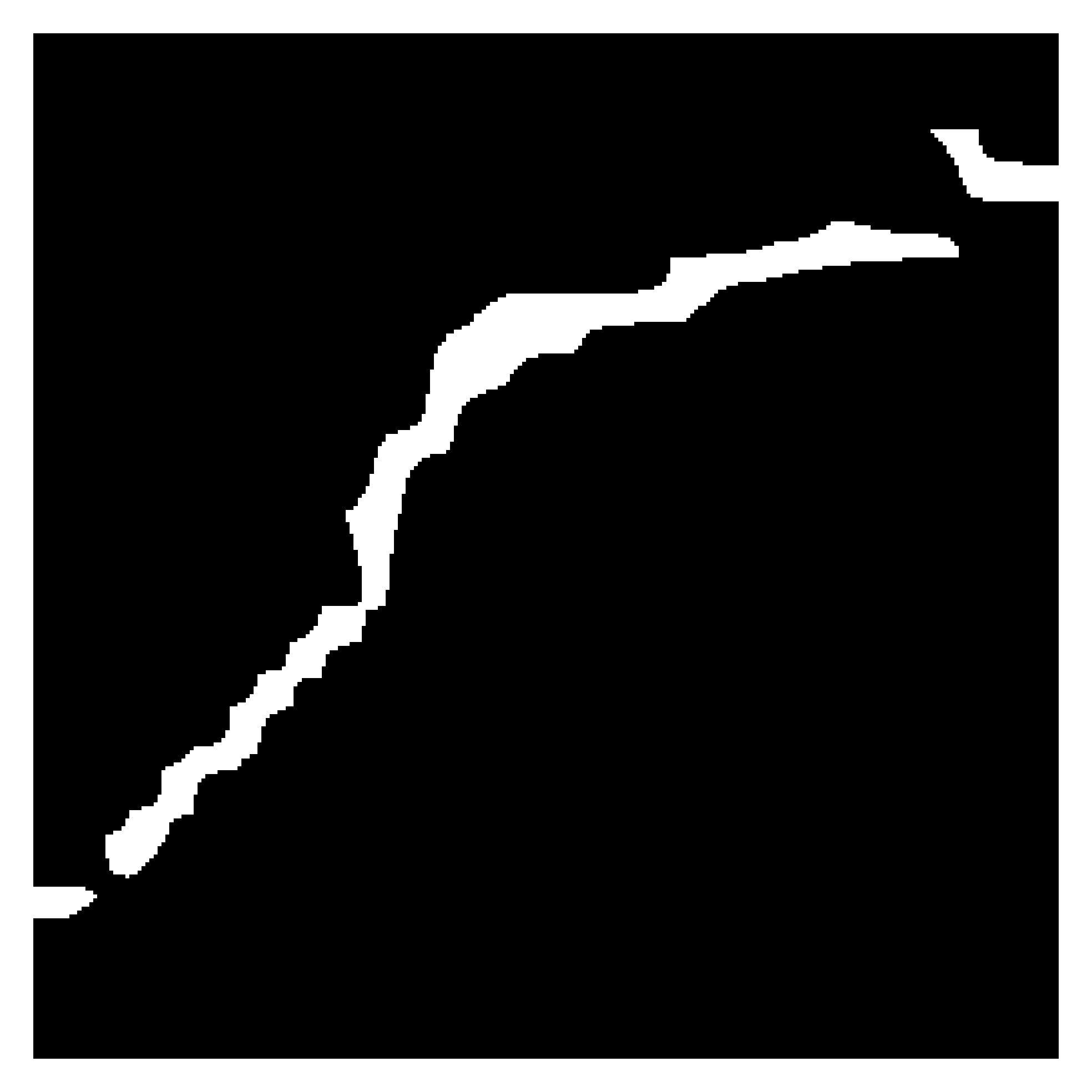}}
    \end{subfigure}
    \begin{subfigure}[b]{0.1\textwidth}
        \subcaption{SegFormer}
       \adjustbox{trim=10 10 10 10,clip,width=1.6cm,height=1.6cm}{\includegraphics{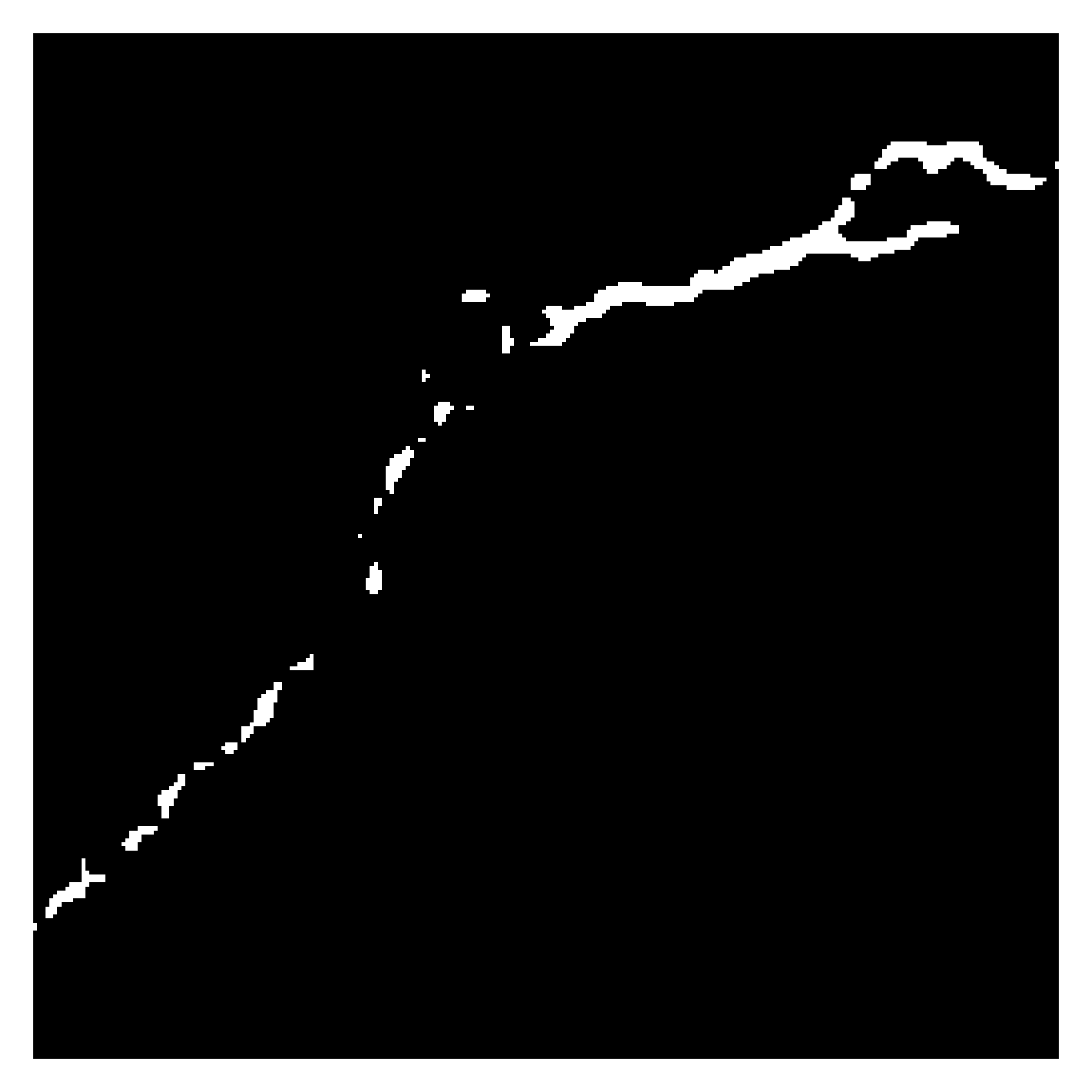}}
    \end{subfigure}
    \begin{subfigure}[b]{0.1\textwidth}
        \subcaption{U-Net} 
        \adjustbox{trim=10 10 10 10,clip,width=1.6cm,height=1.6cm}{\includegraphics{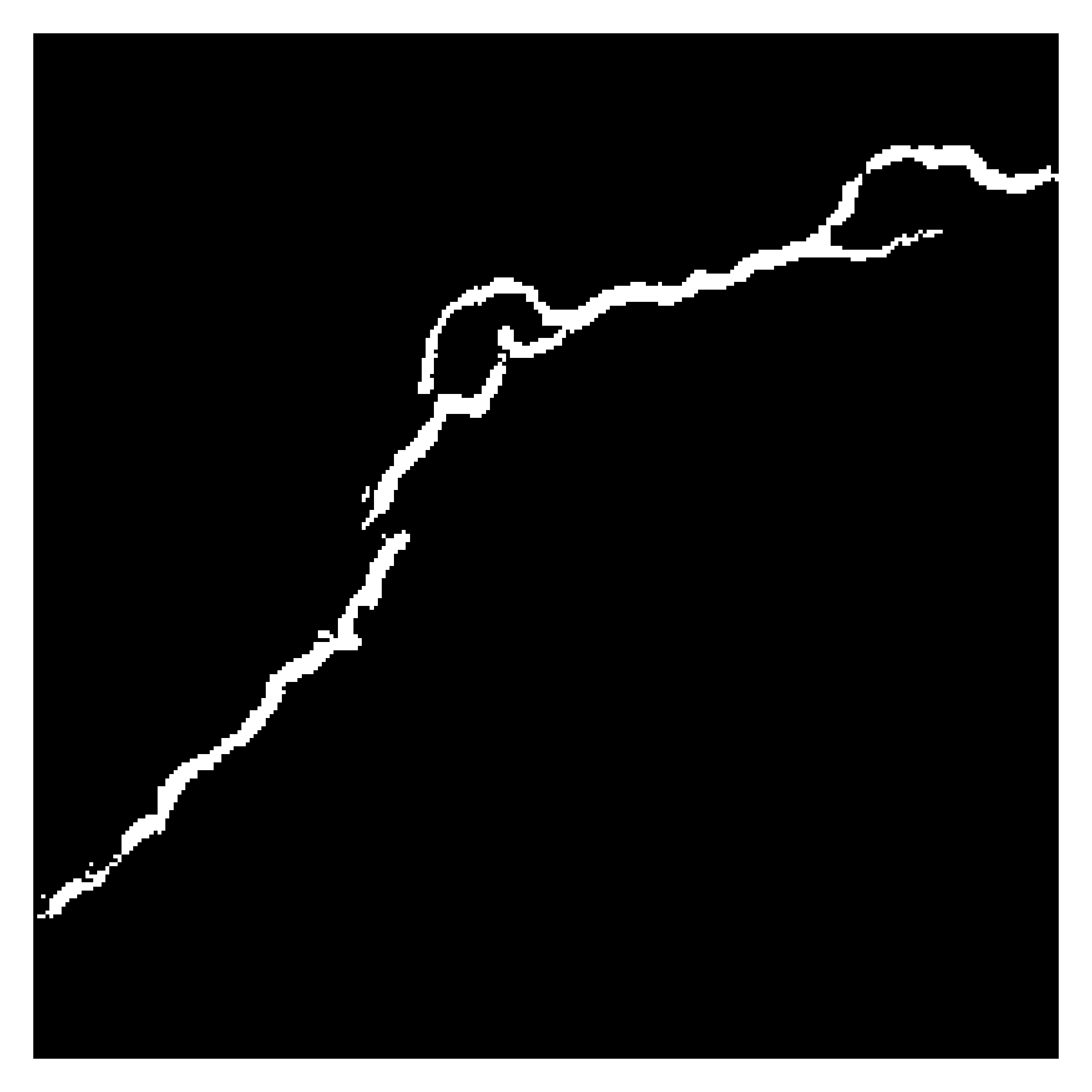}}
    \end{subfigure}
    \begin{subfigure}[b]{0.1\textwidth}
        \subcaption{DeepLabV3+} 
        \adjustbox{trim=10 10 10 10,clip,width=1.6cm,height=1.6cm}{\includegraphics{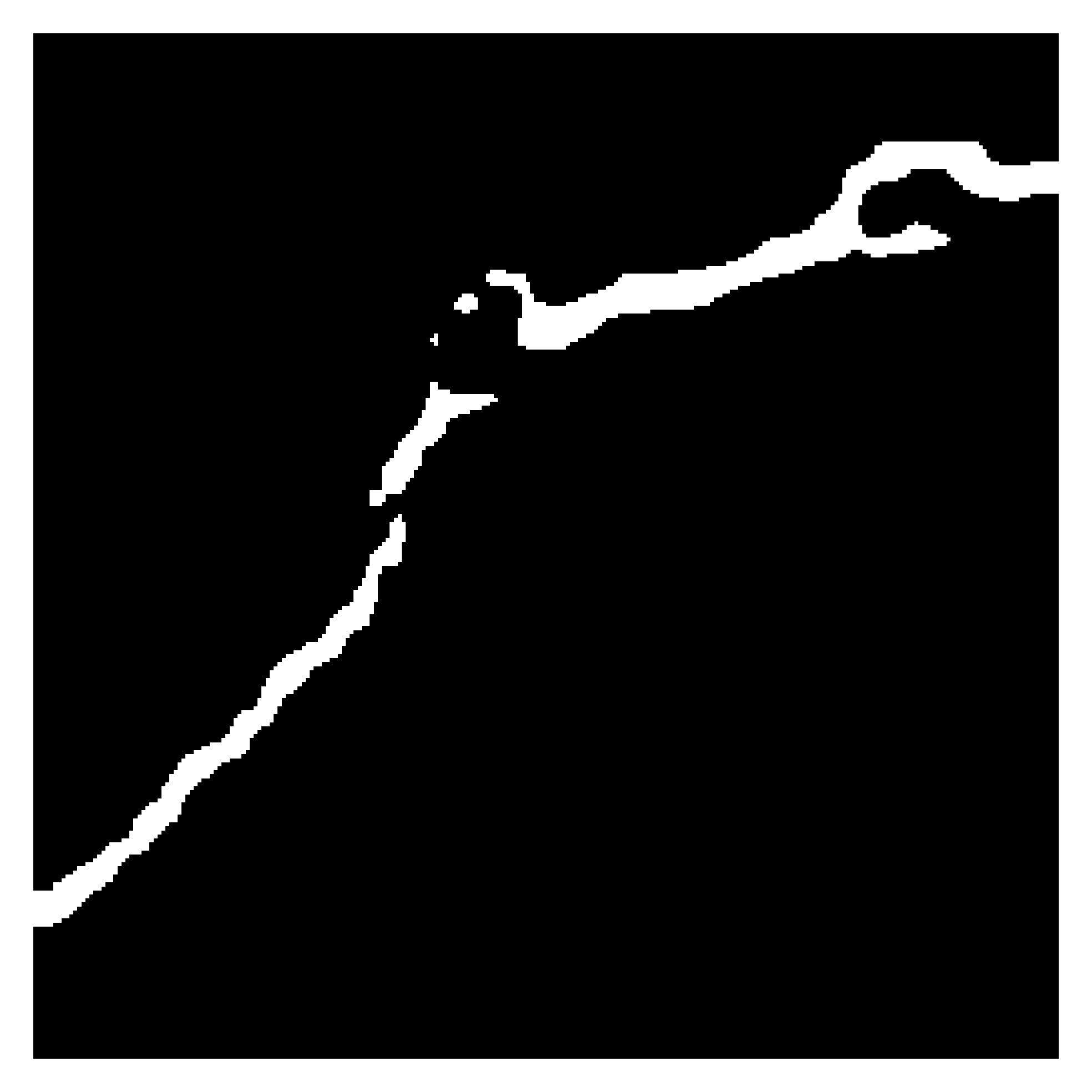}}
    \end{subfigure}
    
    \vspace{0.2cm} 

    \begin{subfigure}[b]{0.09\textwidth} 
        \subcaption*{BCL} 
    \end{subfigure}
    \begin{subfigure}[b]{0.1\textwidth}
        \adjustbox{trim=10 10 10 10,clip,width=1.6cm,height=1.6cm}{\includegraphics{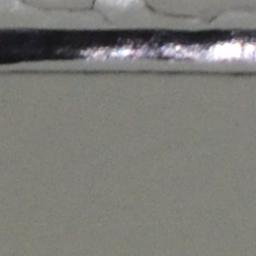}}
    \end{subfigure}
    \begin{subfigure}[b]{0.1\textwidth}
        \adjustbox{trim=10 10 10 10,clip,width=1.6cm,height=1.6cm}{\includegraphics{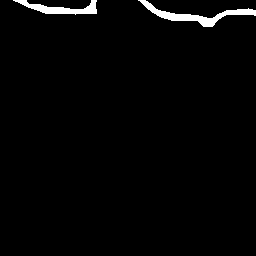}}
    \end{subfigure}
    \begin{subfigure}[b]{0.1\textwidth}
       \adjustbox{trim=10 10 10 10,clip,width=1.6cm,height=1.6cm}{\includegraphics{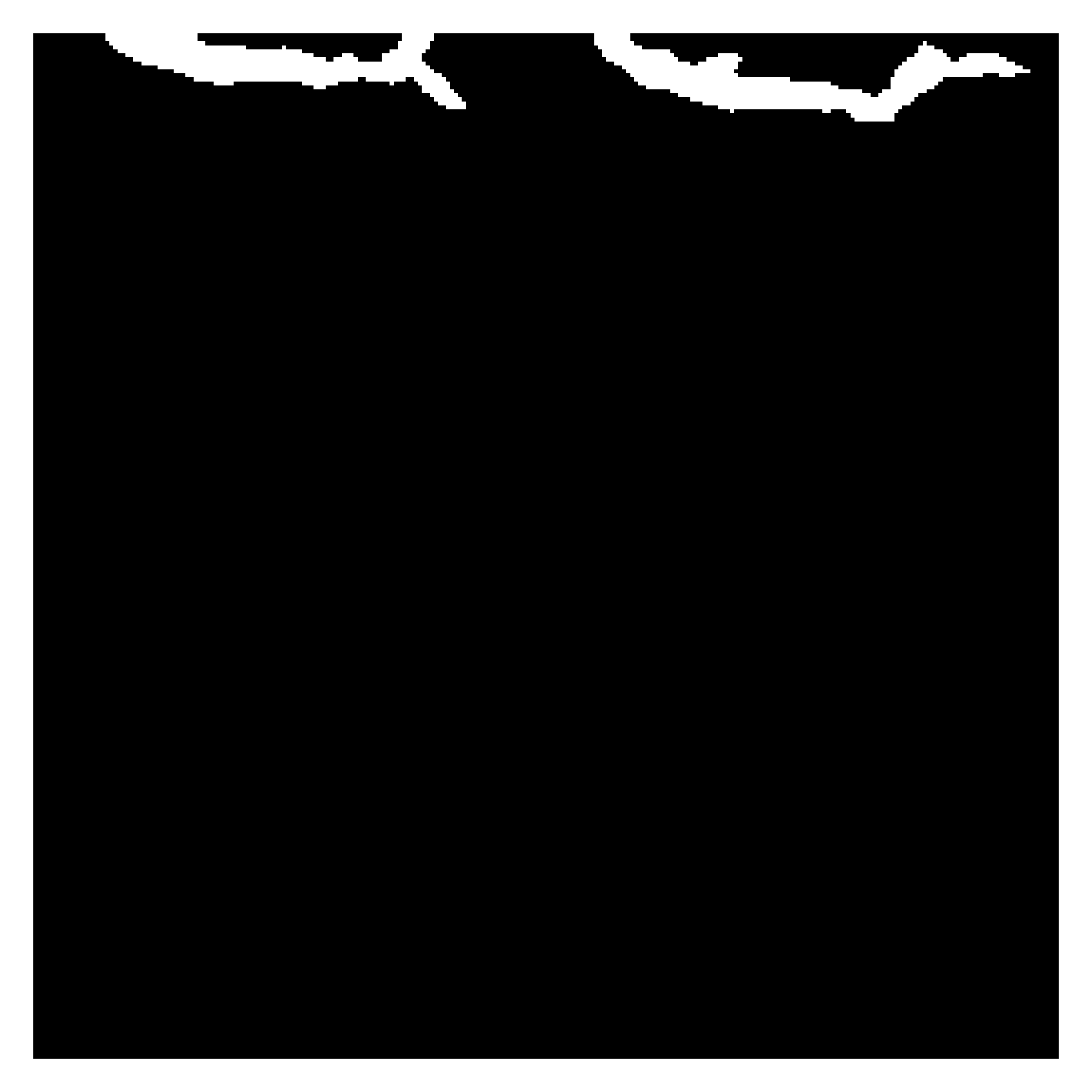}}
    \end{subfigure}
    \begin{subfigure}[b]{0.1\textwidth}
        \adjustbox{trim=10 10 10 10,clip,width=1.6cm,height=1.6cm}{\includegraphics{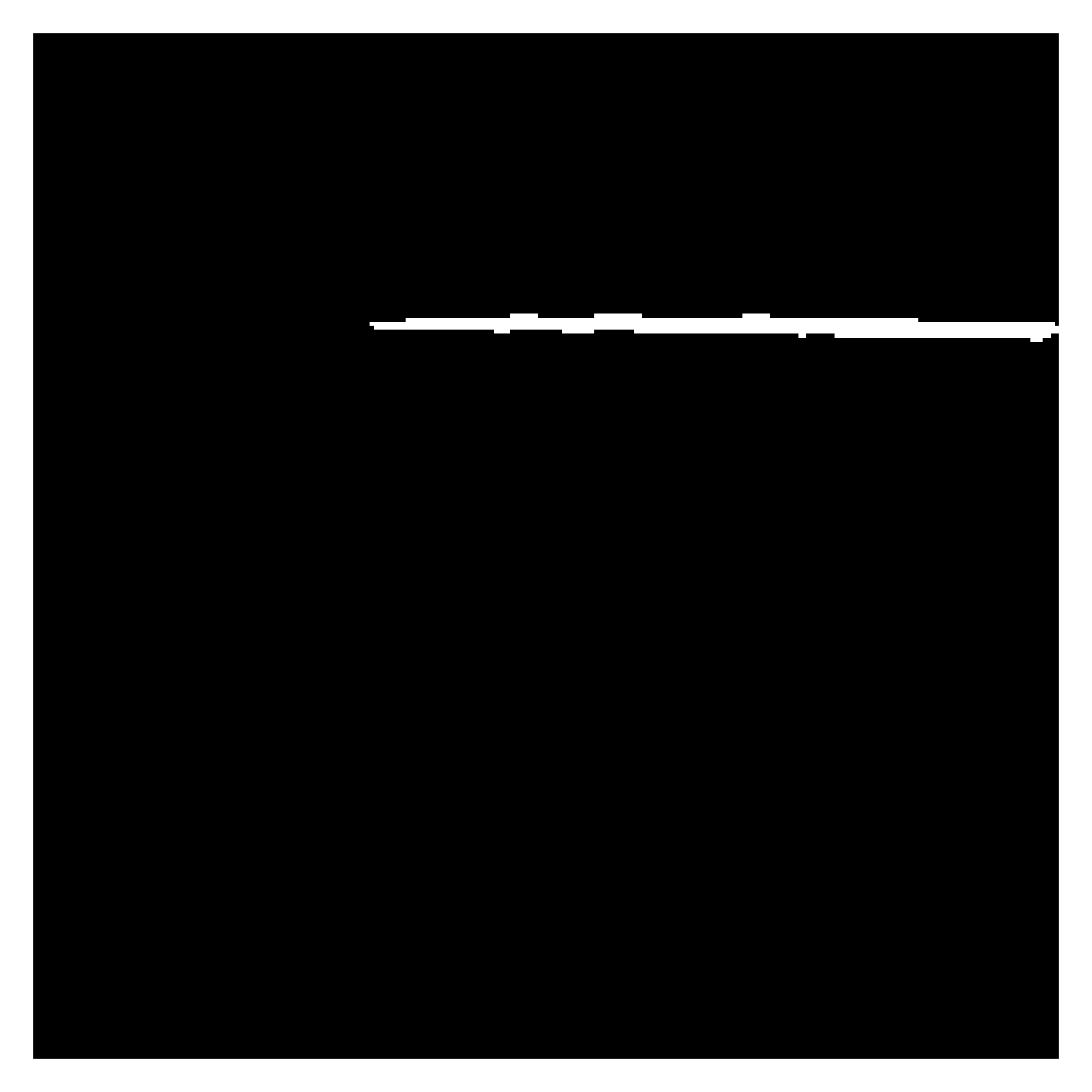}}
    \end{subfigure}
    \begin{subfigure}[b]{0.1\textwidth}
        \adjustbox{trim=10 10 10 10,clip,width=1.6cm,height=1.6cm}{\includegraphics{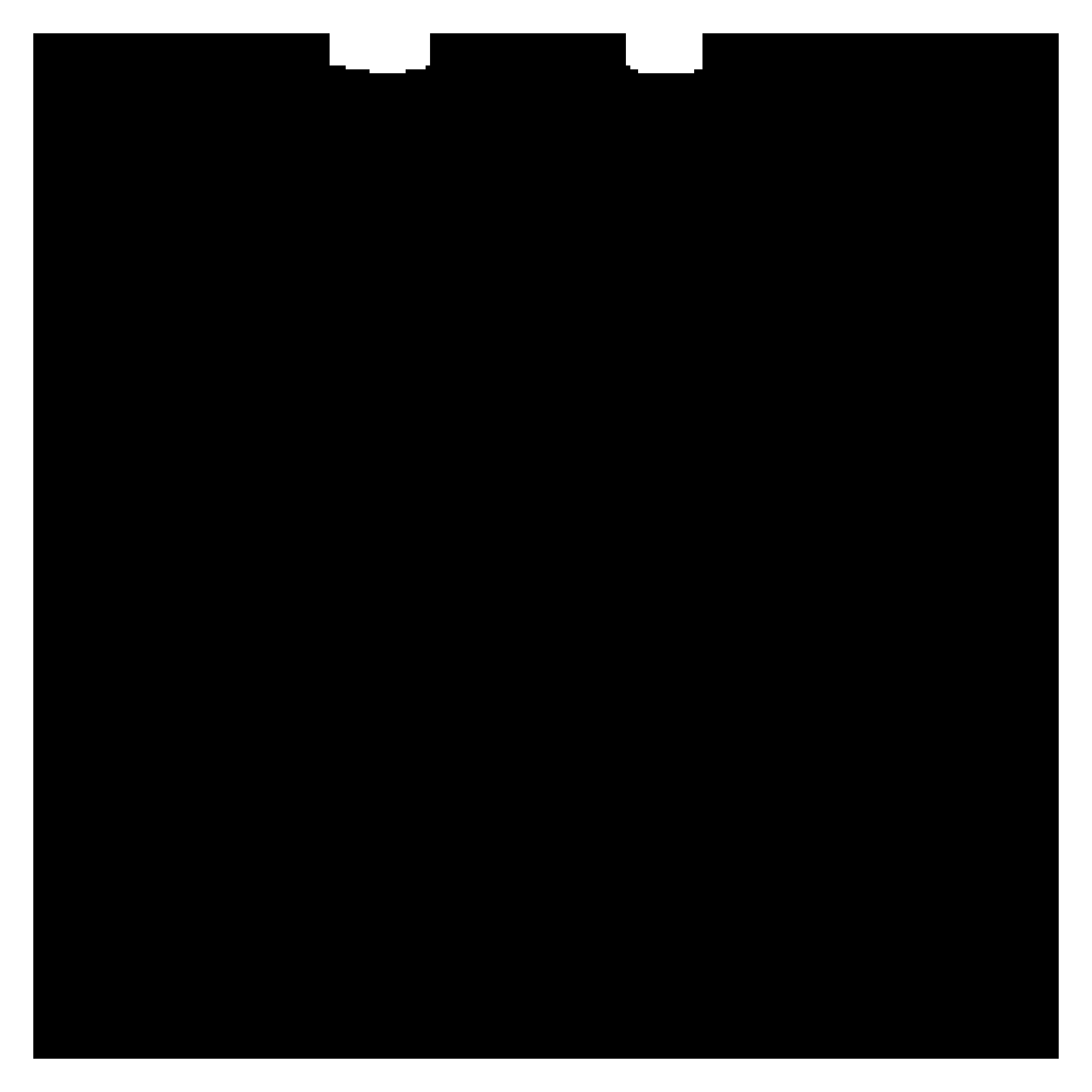}}
    \end{subfigure}
    \begin{subfigure}[b]{0.1\textwidth}
       \adjustbox{trim=10 10 10 10,clip,width=1.6cm,height=1.6cm}{\includegraphics{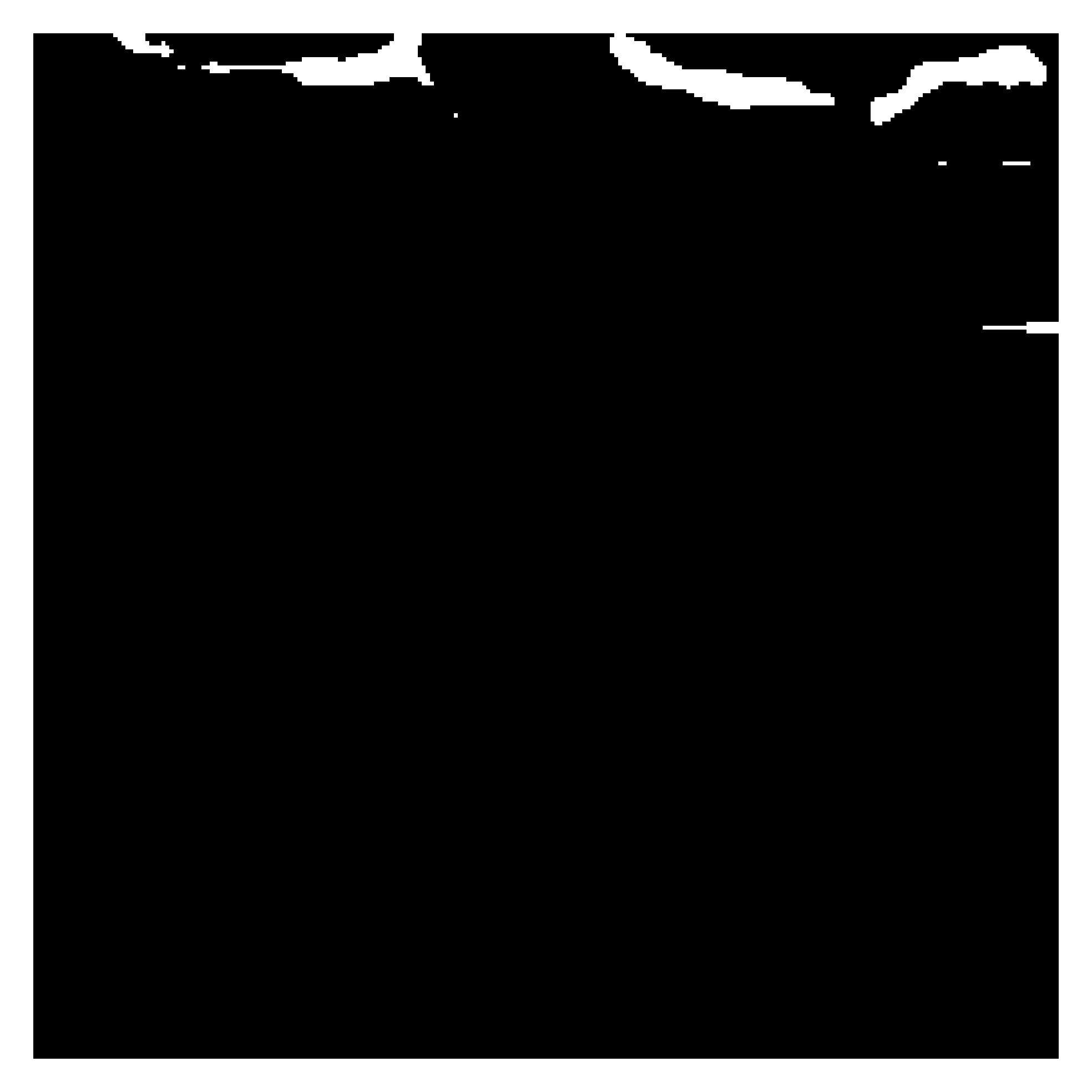}}
    \end{subfigure}
    \begin{subfigure}[b]{0.1\textwidth}
        \adjustbox{trim=10 10 10 10,clip,width=1.6cm,height=1.6cm}{\includegraphics{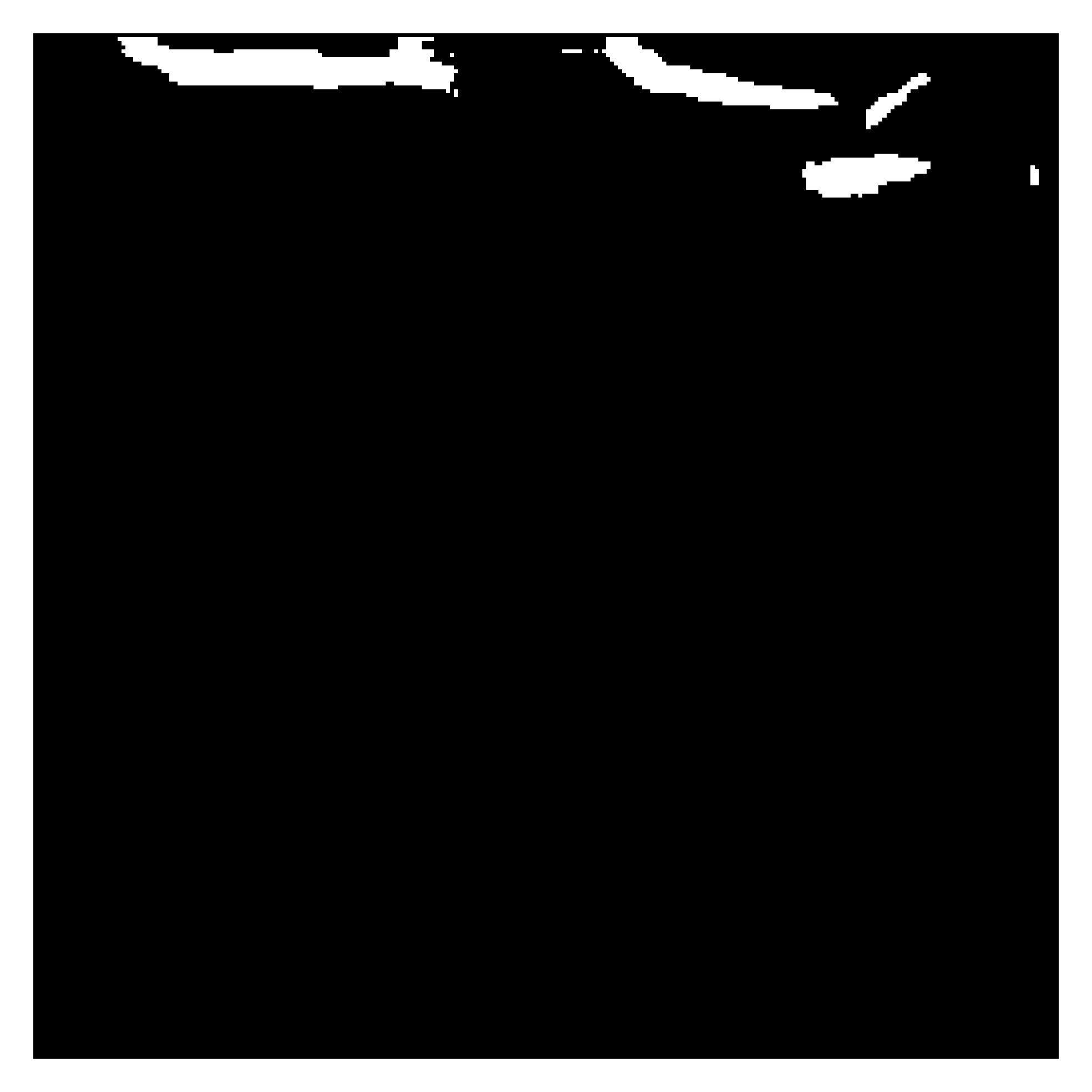}}
    \end{subfigure}
    \begin{subfigure}[b]{0.1\textwidth}
        \adjustbox{trim=10 10 10 10,clip,width=1.6cm,height=1.6cm}{\includegraphics{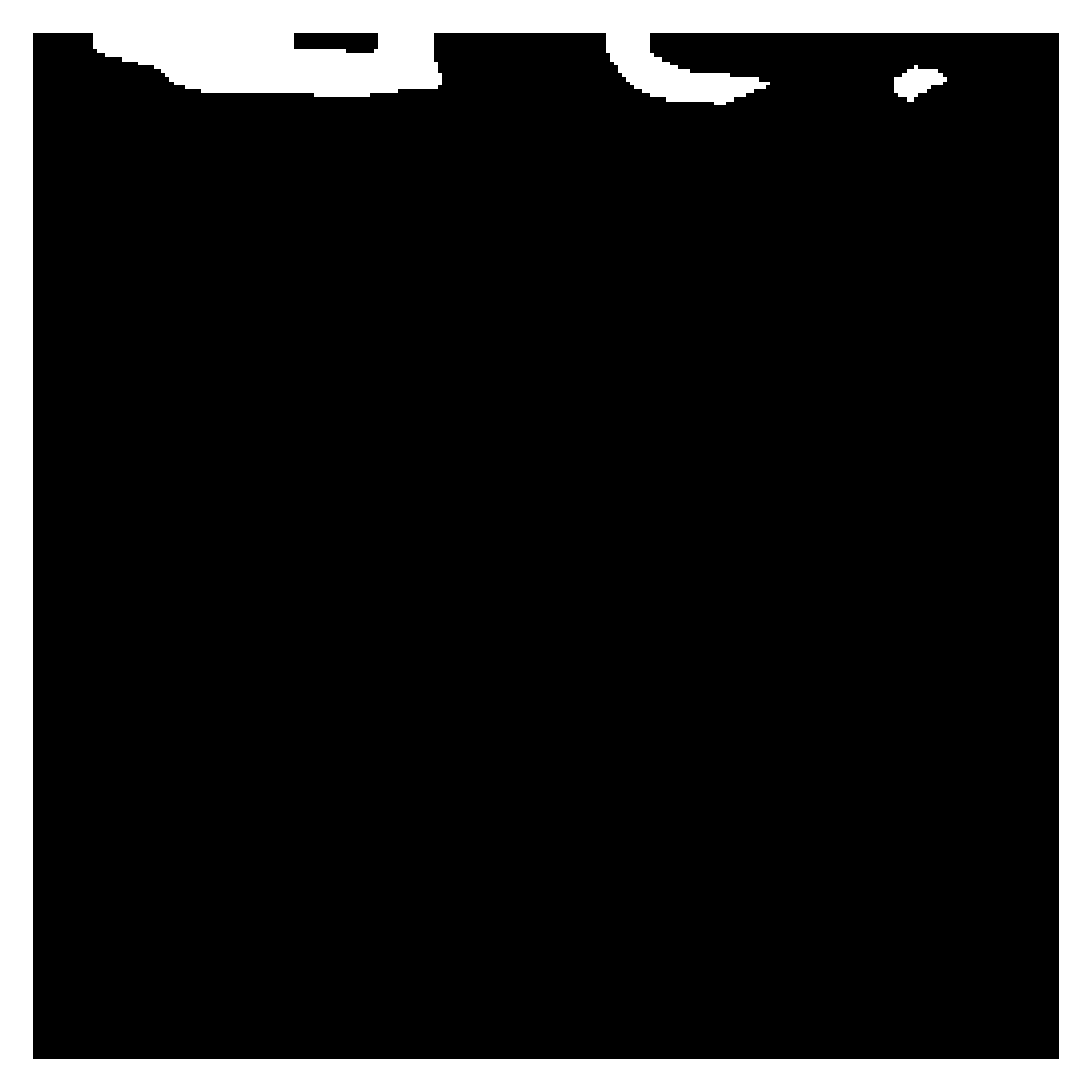}}
    \end{subfigure}

    \vspace{0.2cm} 

    \begin{subfigure}[b]{0.09\textwidth} 
        \subcaption*{CFD} 
    \end{subfigure}
    \begin{subfigure}[b]{0.1\textwidth}
        \adjustbox{trim=10 10 10 10,clip,width=1.6cm,height=1.6cm}{\includegraphics{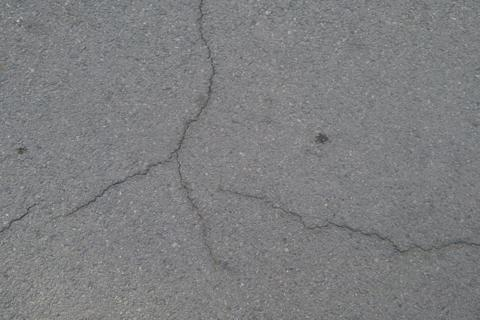}}
    \end{subfigure}
    \begin{subfigure}[b]{0.1\textwidth}
        \adjustbox{trim=10 10 10 10,clip,width=1.6cm,height=1.6cm}{\includegraphics{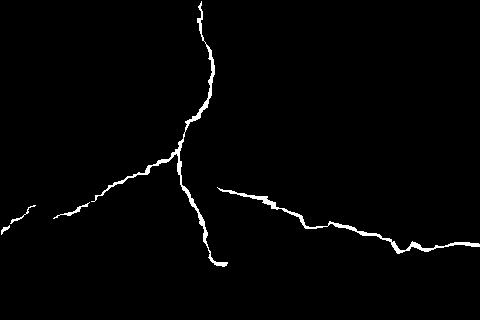}}
    \end{subfigure}
    \begin{subfigure}[b]{0.1\textwidth}
       \adjustbox{trim=10 10 10 10,clip,width=1.6cm,height=1.6cm}{\includegraphics{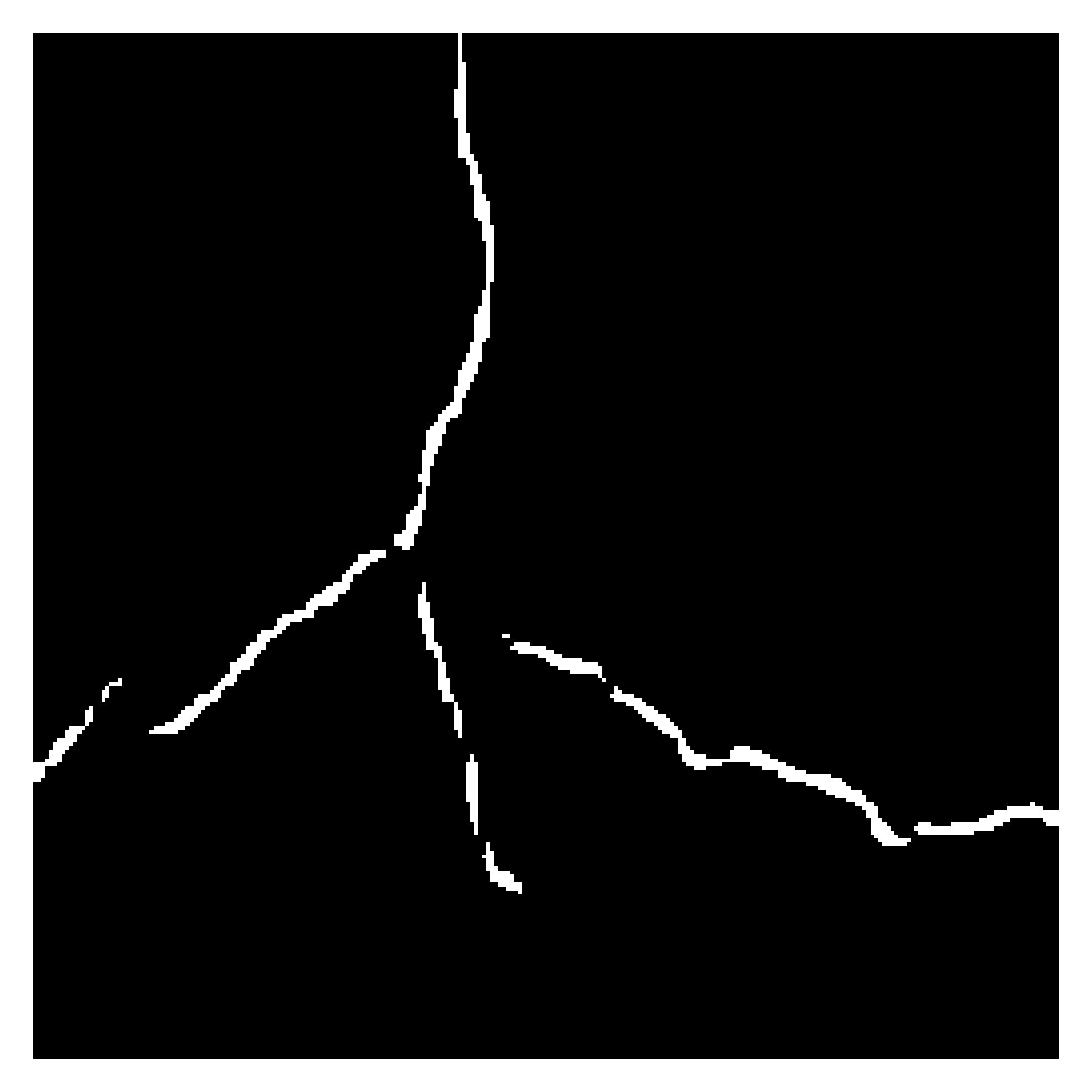}}
    \end{subfigure}
    \begin{subfigure}[b]{0.1\textwidth}
        \adjustbox{trim=10 10 10 10,clip,width=1.6cm,height=1.6cm}{\includegraphics{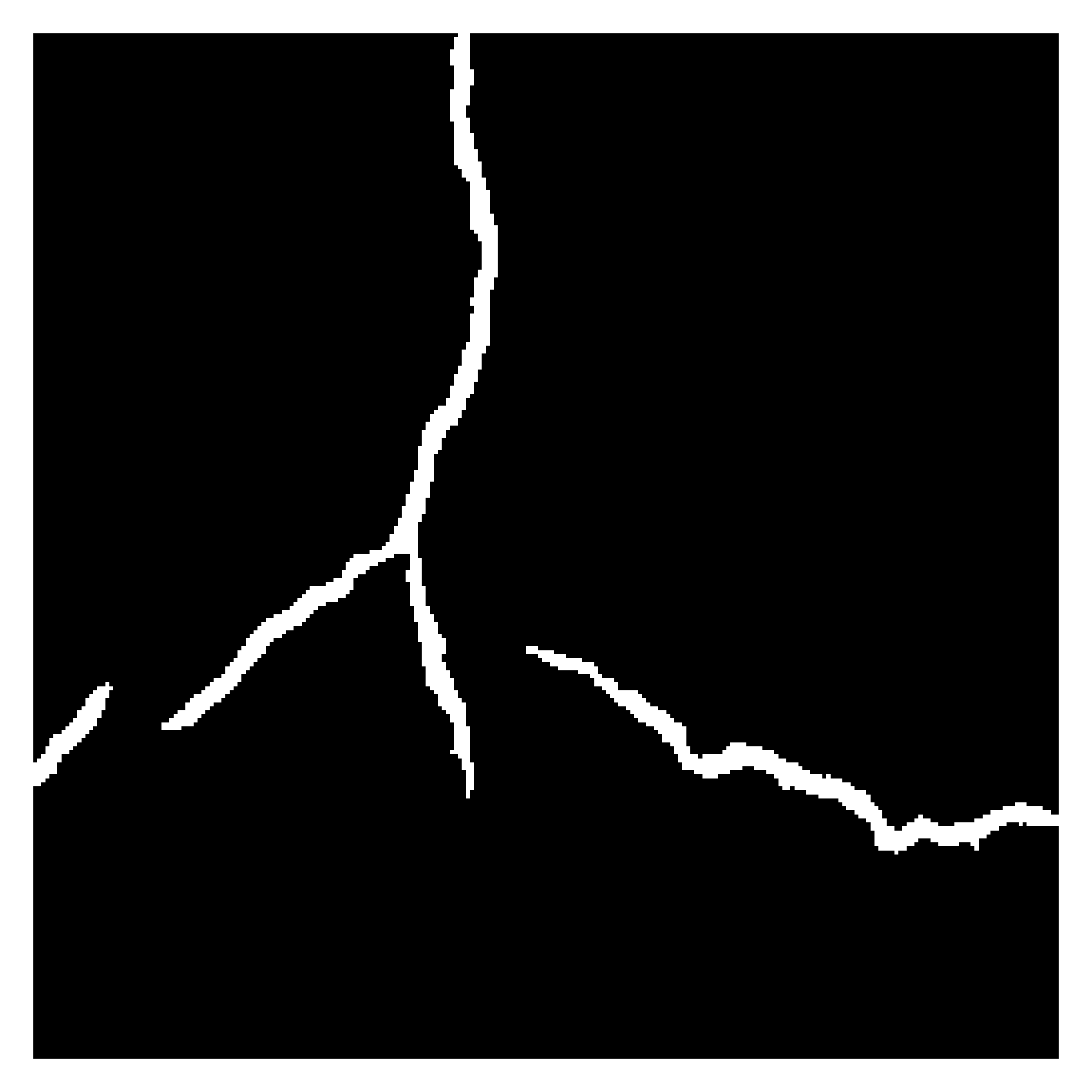}}
    \end{subfigure}
    \begin{subfigure}[b]{0.1\textwidth}
        \adjustbox{trim=10 10 10 10,clip,width=1.6cm,height=1.6cm}{\includegraphics{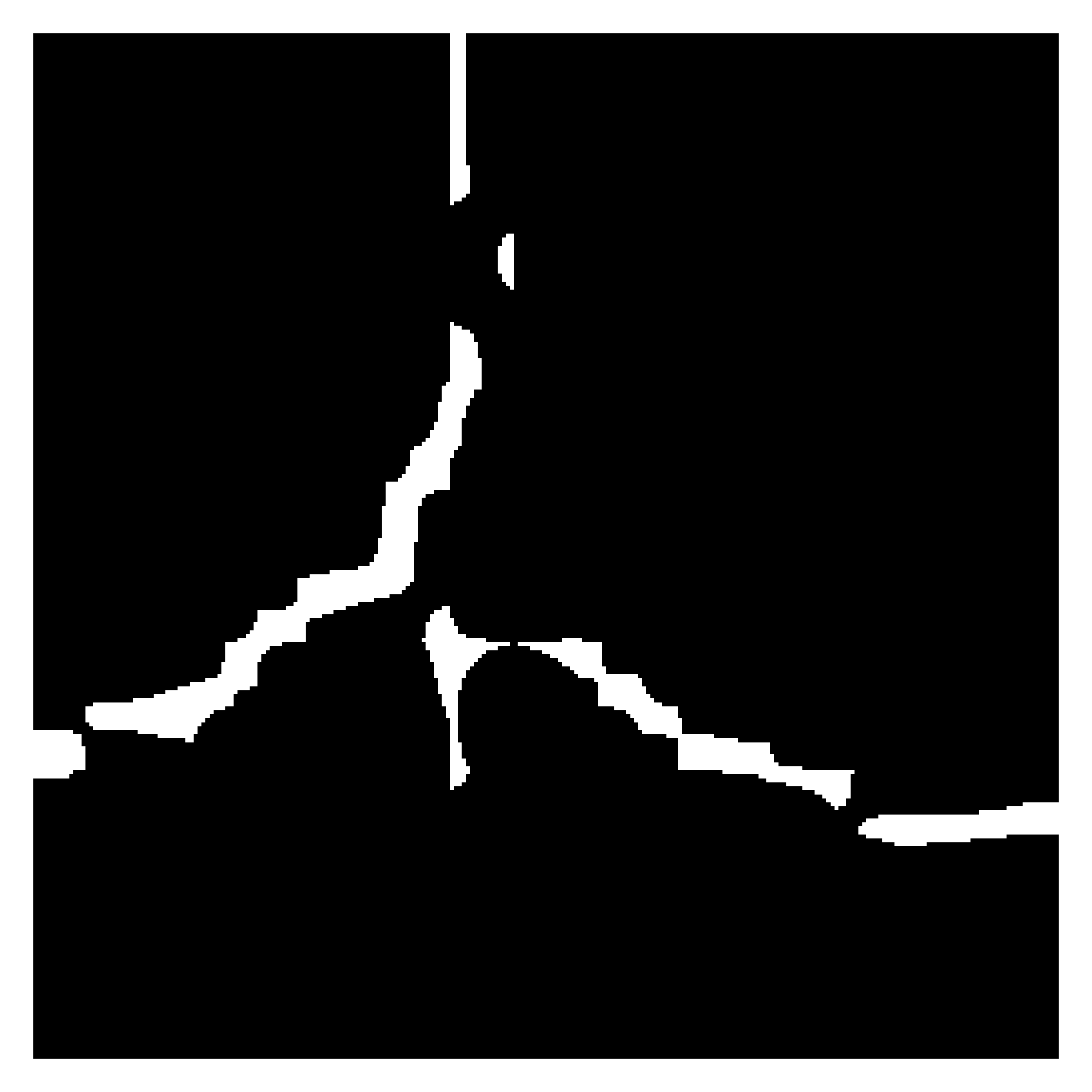}}
    \end{subfigure}
    \begin{subfigure}[b]{0.1\textwidth}
       \adjustbox{trim=10 10 10 10,clip,width=1.6cm,height=1.6cm}{\includegraphics{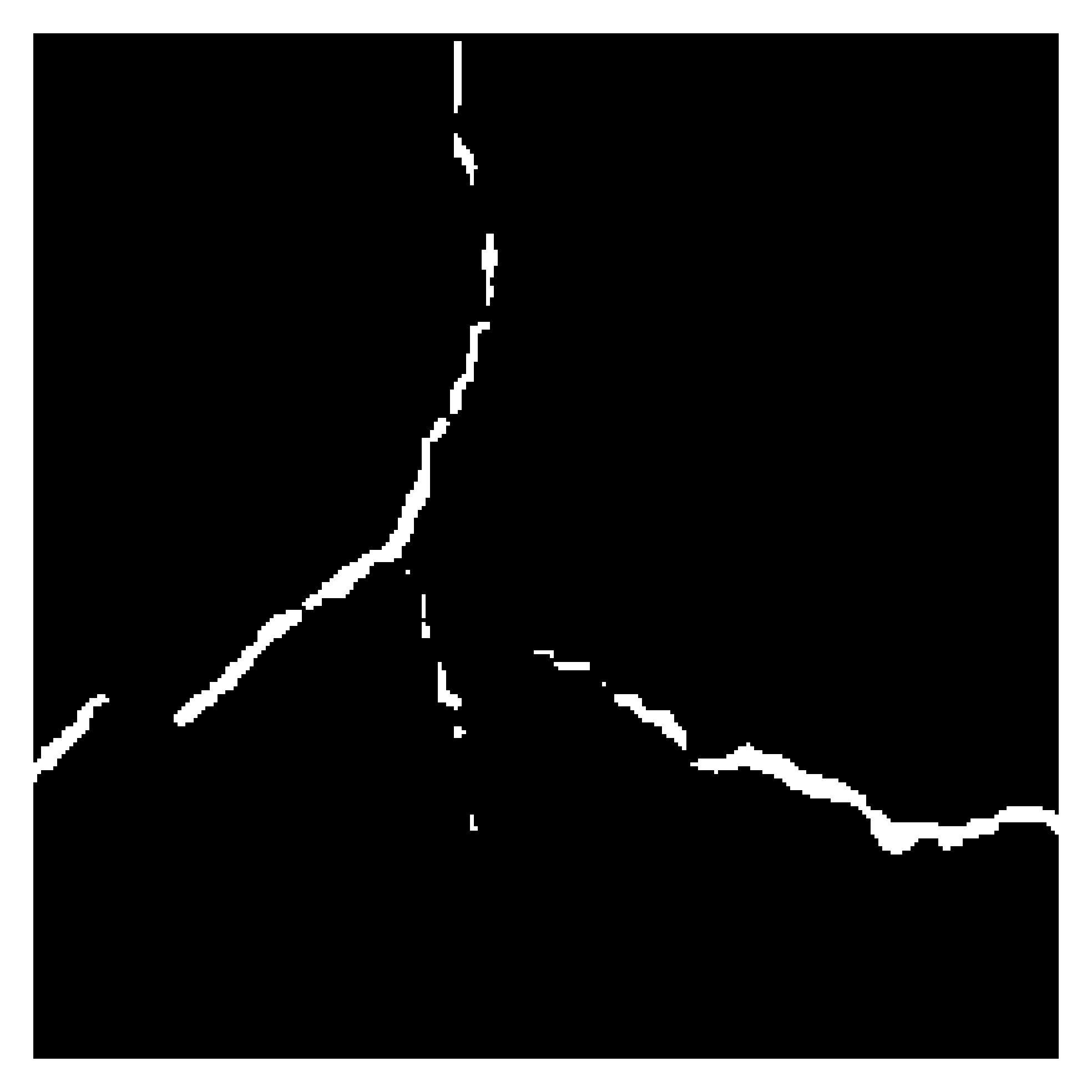}}
    \end{subfigure}
    \begin{subfigure}[b]{0.1\textwidth}
        \adjustbox{trim=10 10 10 10,clip,width=1.6cm,height=1.6cm}{\includegraphics{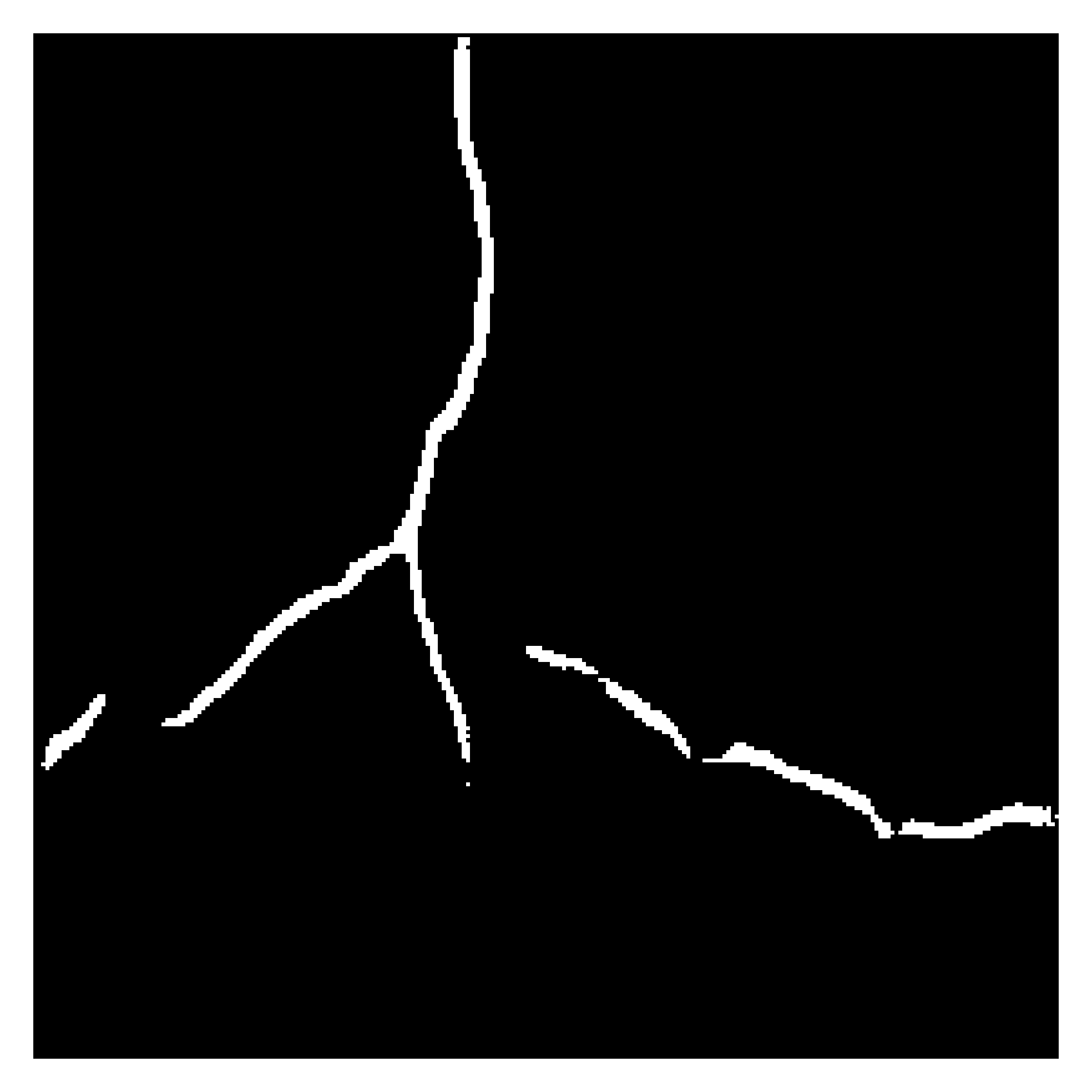}}
    \end{subfigure}
    \begin{subfigure}[b]{0.1\textwidth}
        \adjustbox{trim=10 10 10 10,clip,width=1.6cm,height=1.6cm}{\includegraphics{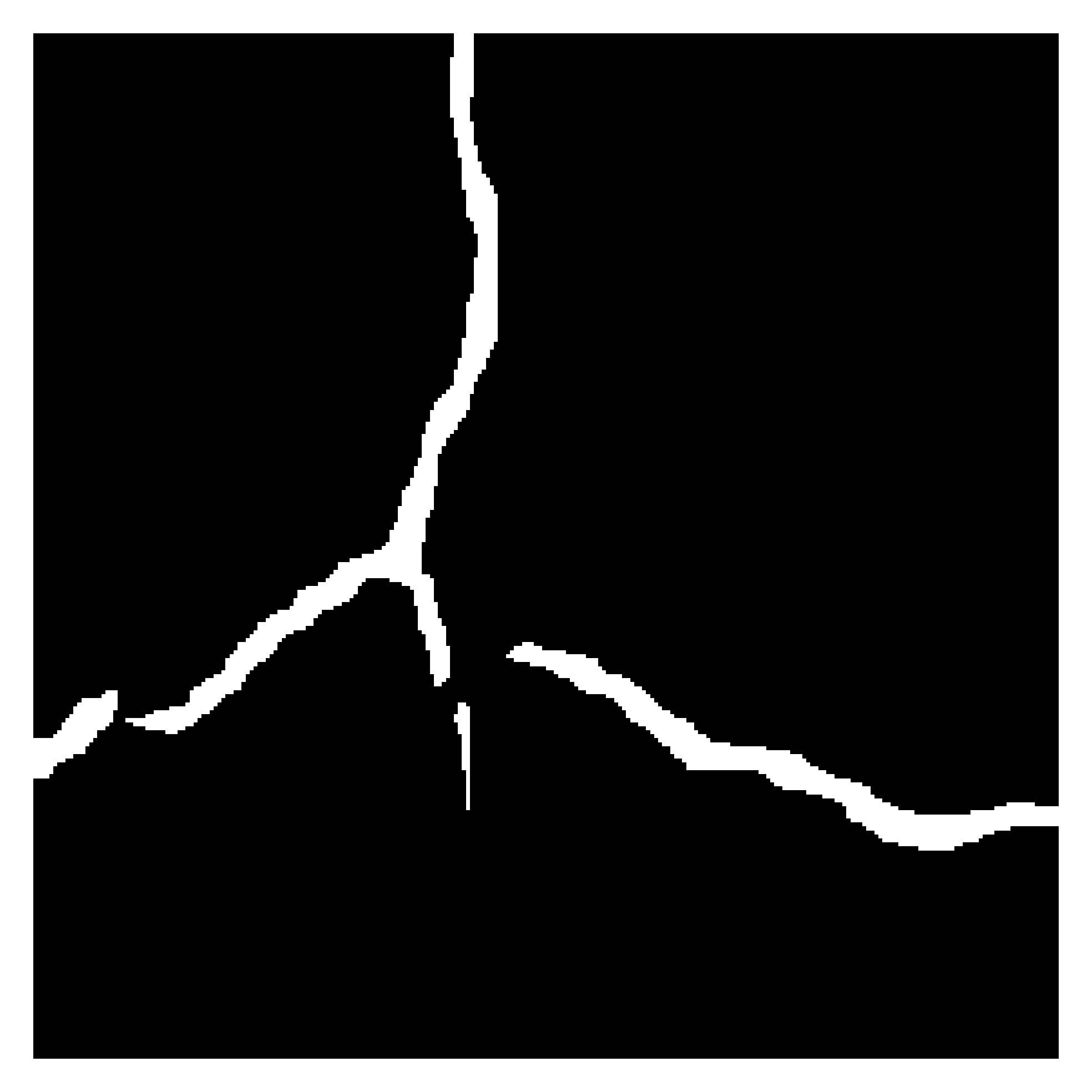}}
    \end{subfigure}

    \vspace{0.2cm}

    \begin{subfigure}[b]{0.09\textwidth} 
        \subcaption*{Crack500} 
    \end{subfigure}
    \begin{subfigure}[b]{0.1\textwidth}
        \adjustbox{trim=10 10 10 10,clip,width=1.6cm,height=1.6cm}{\includegraphics{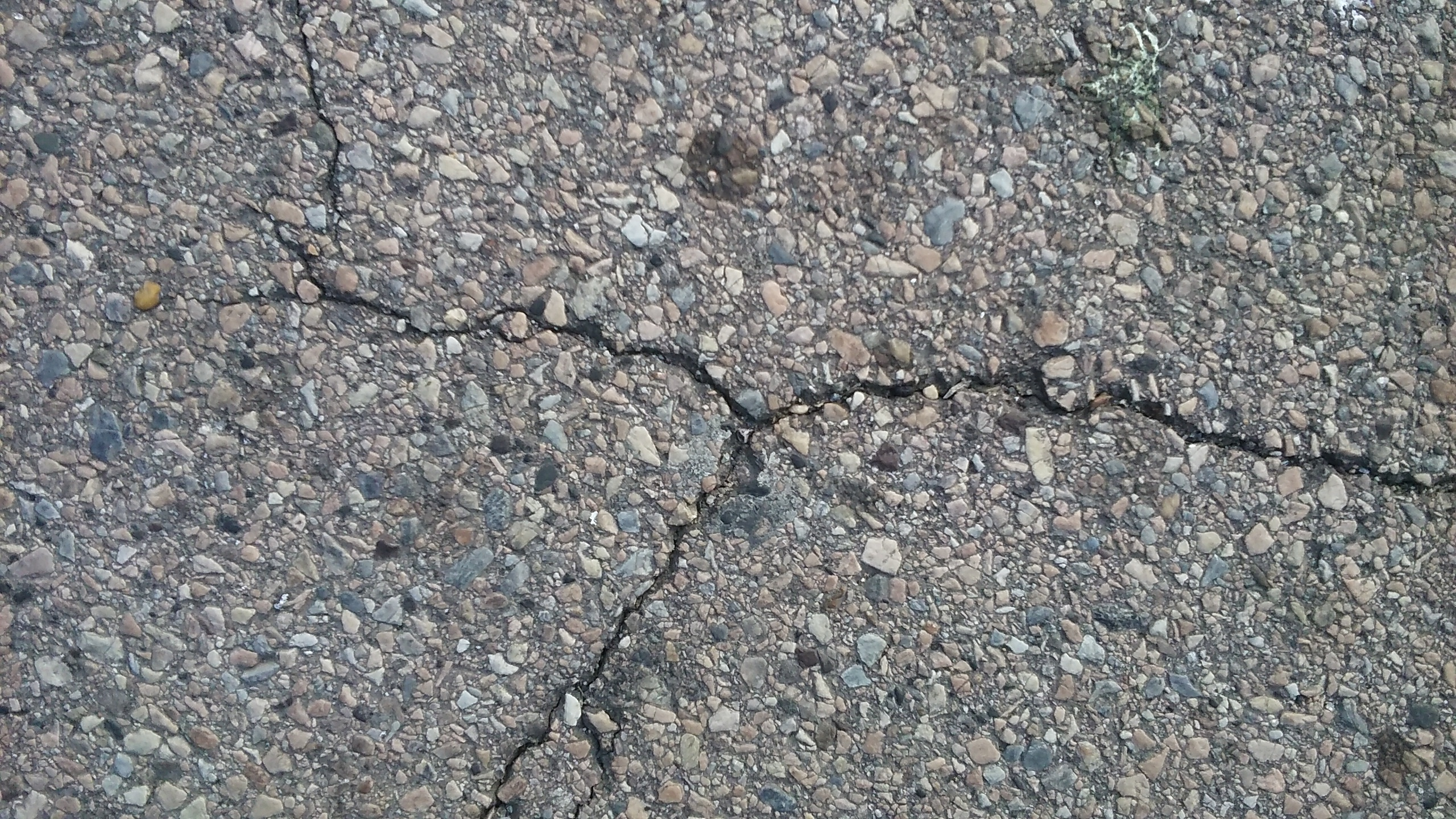}}
    \end{subfigure}
    \begin{subfigure}[b]{0.1\textwidth}
        \adjustbox{trim=10 10 10 10,clip,width=1.6cm,height=1.6cm}{\includegraphics{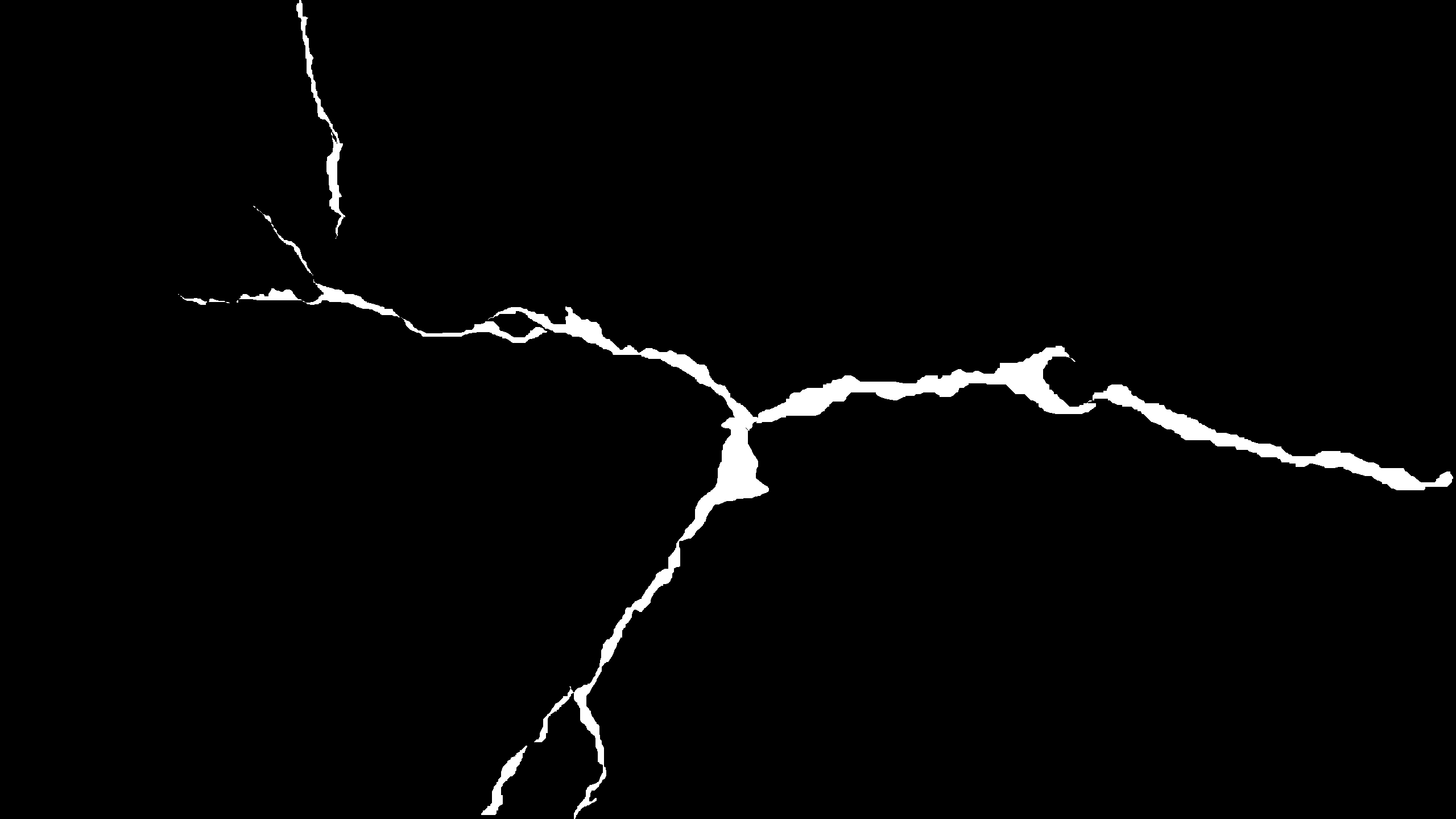}}
    \end{subfigure}
    \begin{subfigure}[b]{0.1\textwidth}
       \adjustbox{trim=10 10 10 10,clip,width=1.6cm,height=1.6cm}{\includegraphics{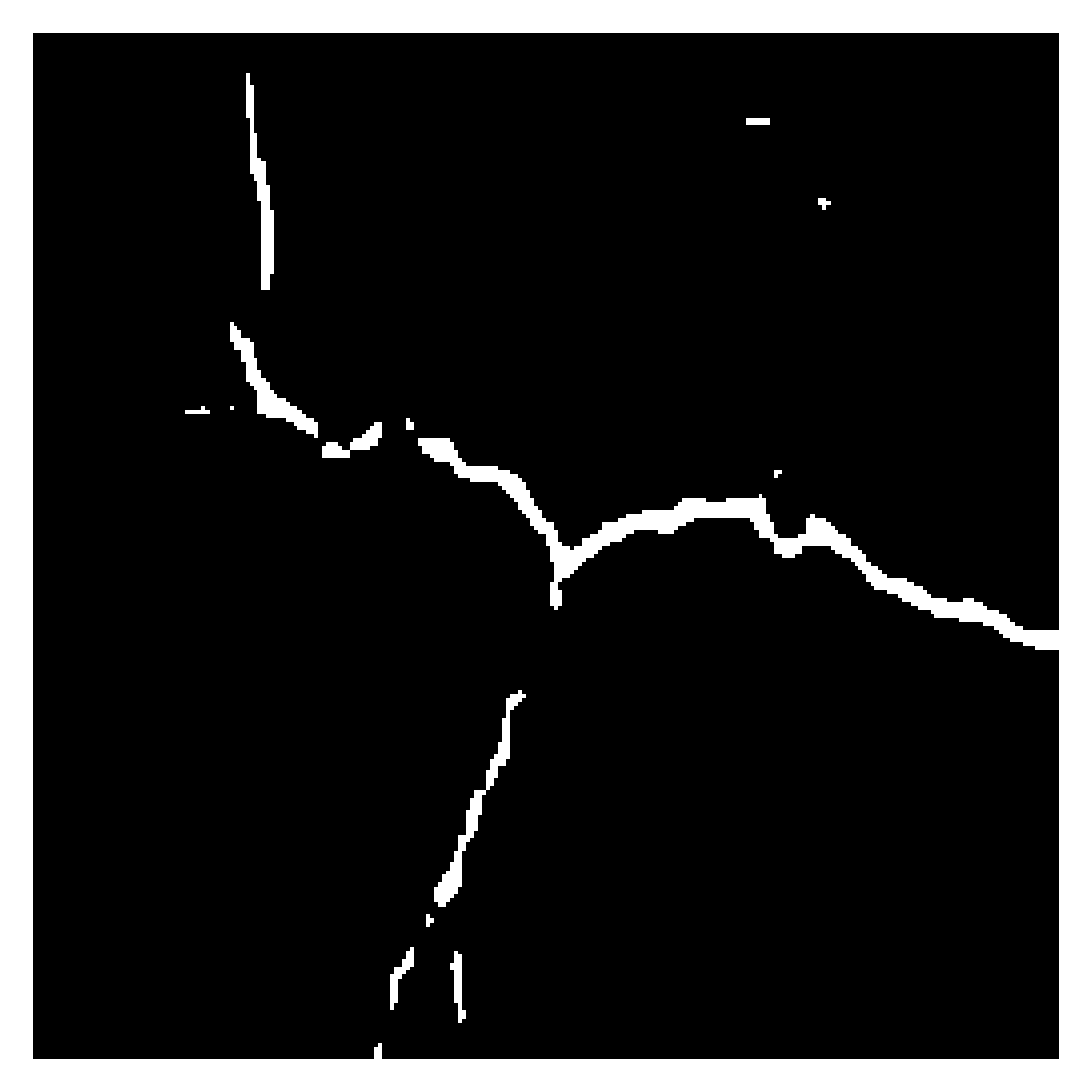}}
    \end{subfigure}
    \begin{subfigure}[b]{0.1\textwidth}
        \adjustbox{trim=10 10 10 10,clip,width=1.6cm,height=1.6cm}{\includegraphics{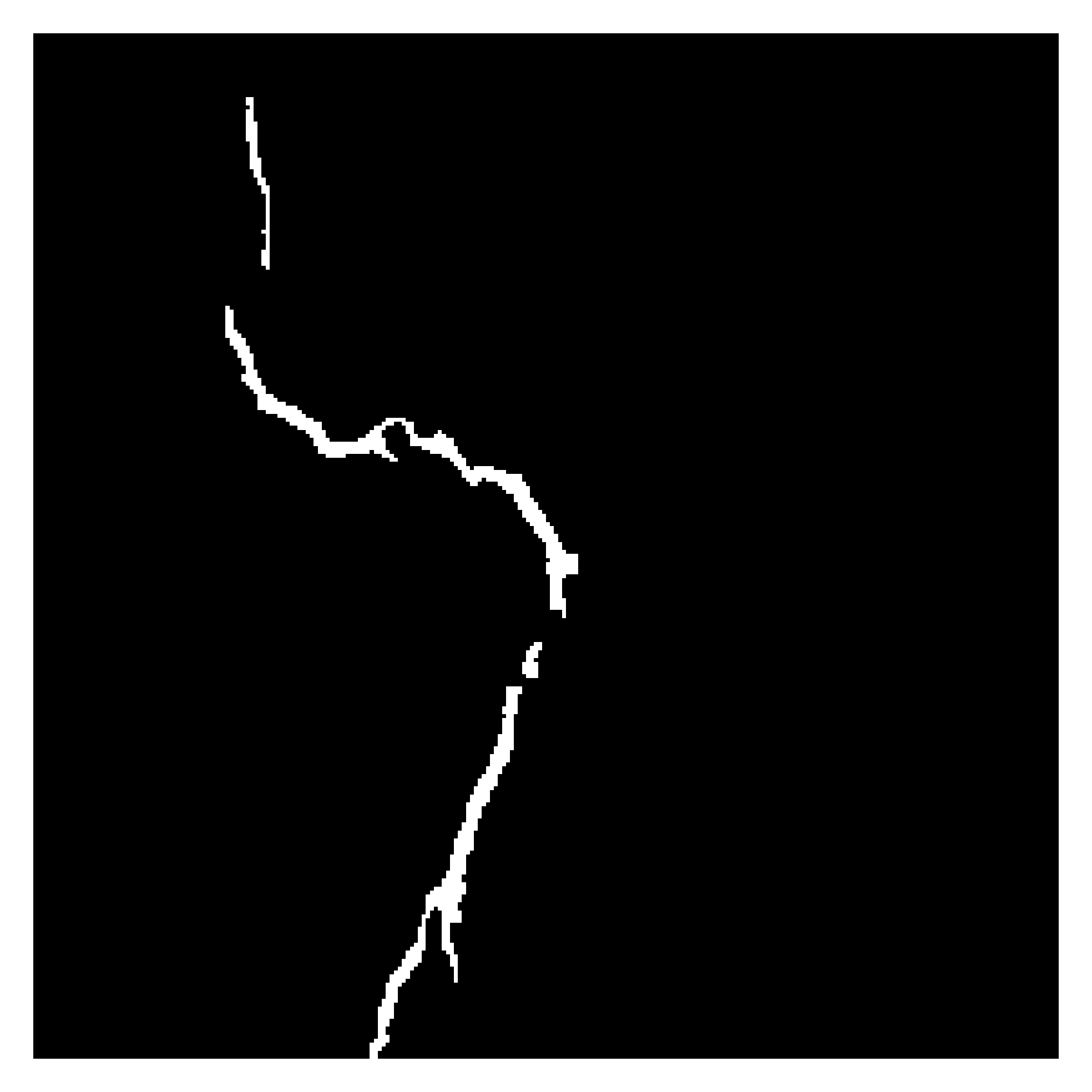}}
    \end{subfigure}
    \begin{subfigure}[b]{0.1\textwidth}
        \adjustbox{trim=10 10 10 10,clip,width=1.6cm,height=1.6cm}{\includegraphics{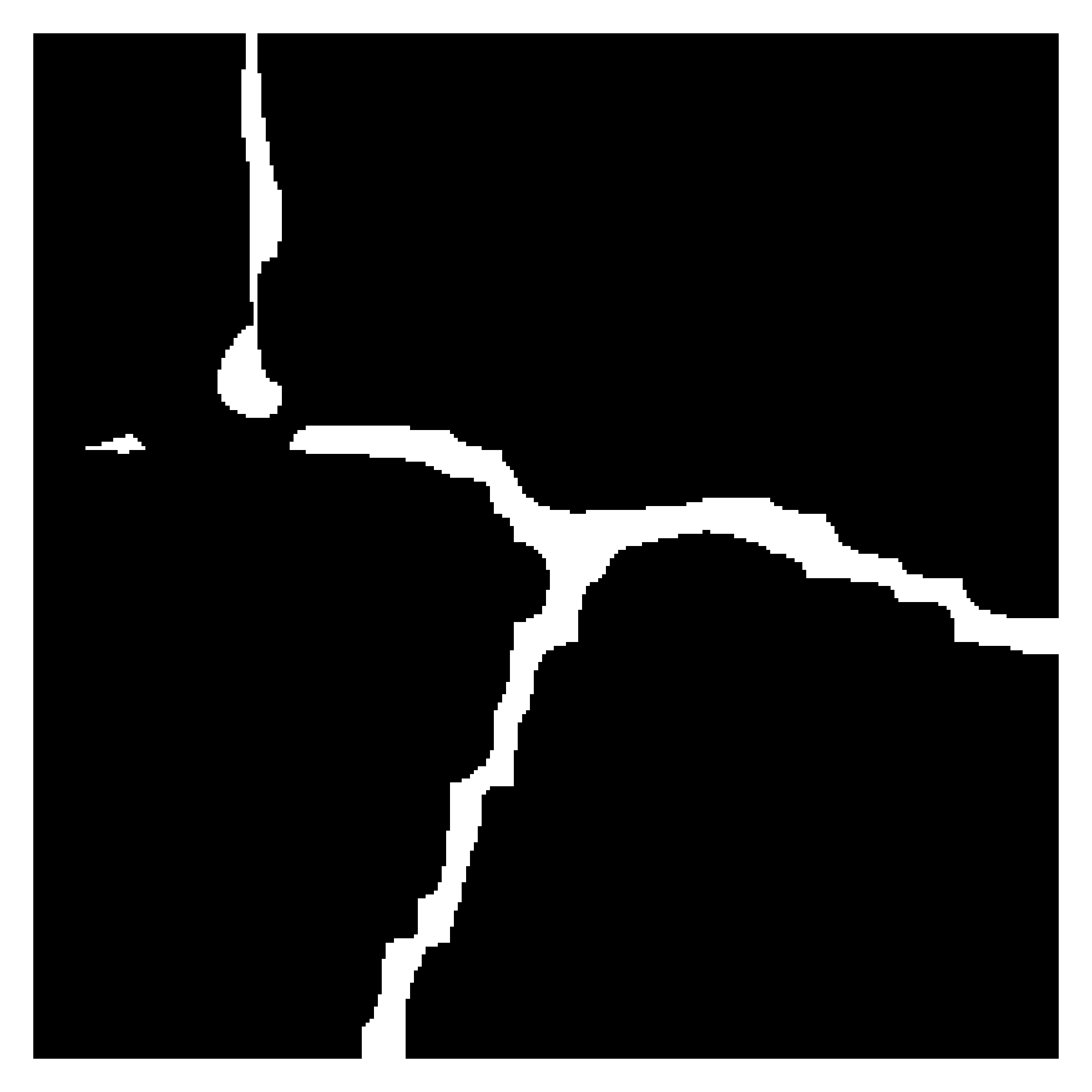}}
    \end{subfigure}
    \begin{subfigure}[b]{0.1\textwidth}
       \adjustbox{trim=10 10 10 10,clip,width=1.6cm,height=1.6cm}{\includegraphics{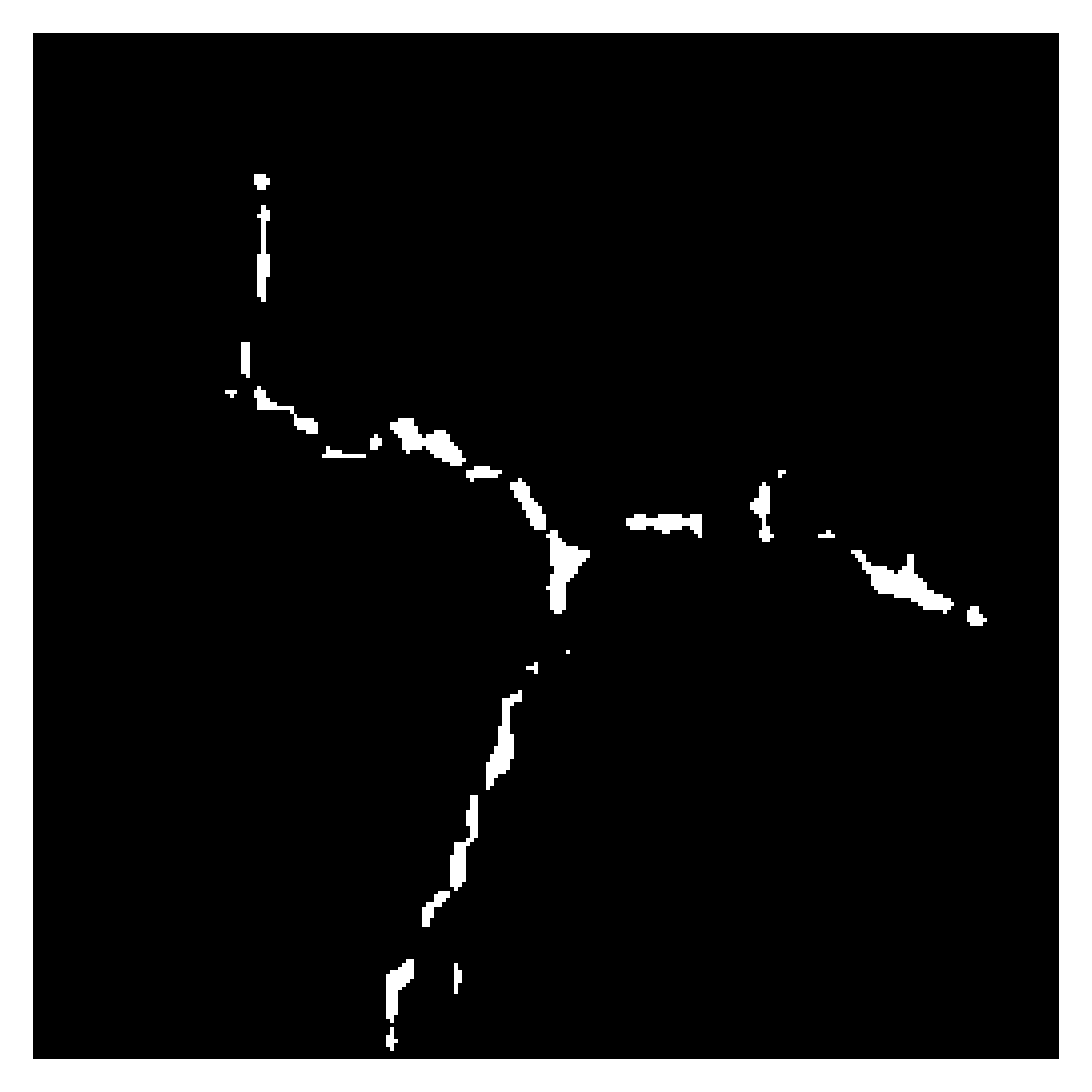}}
    \end{subfigure}
    \begin{subfigure}[b]{0.1\textwidth}
        \adjustbox{trim=10 10 10 10,clip,width=1.6cm,height=1.6cm}{\includegraphics{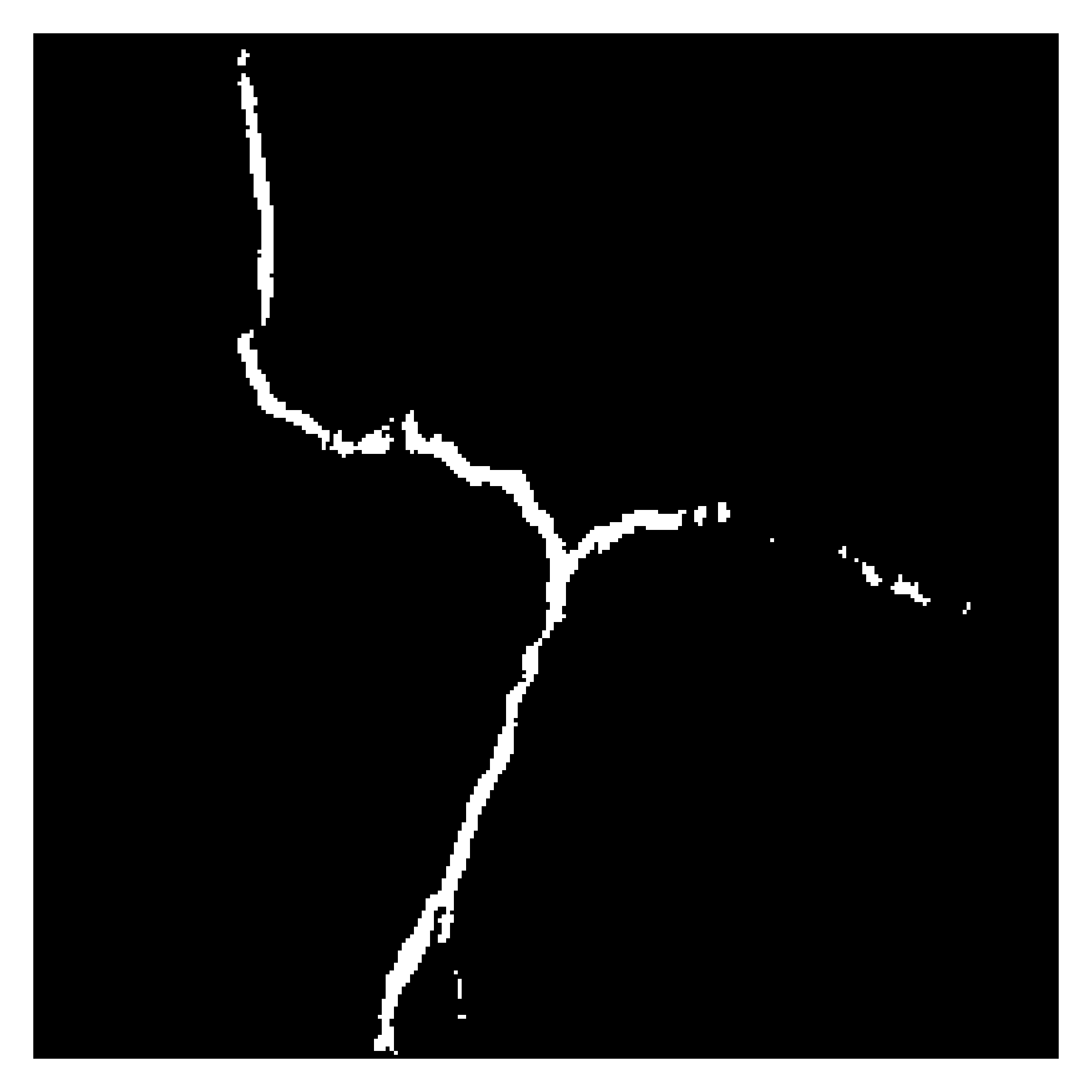}}
    \end{subfigure}
    \begin{subfigure}[b]{0.1\textwidth}
        \adjustbox{trim=10 10 10 10,clip,width=1.6cm,height=1.6cm}{\includegraphics{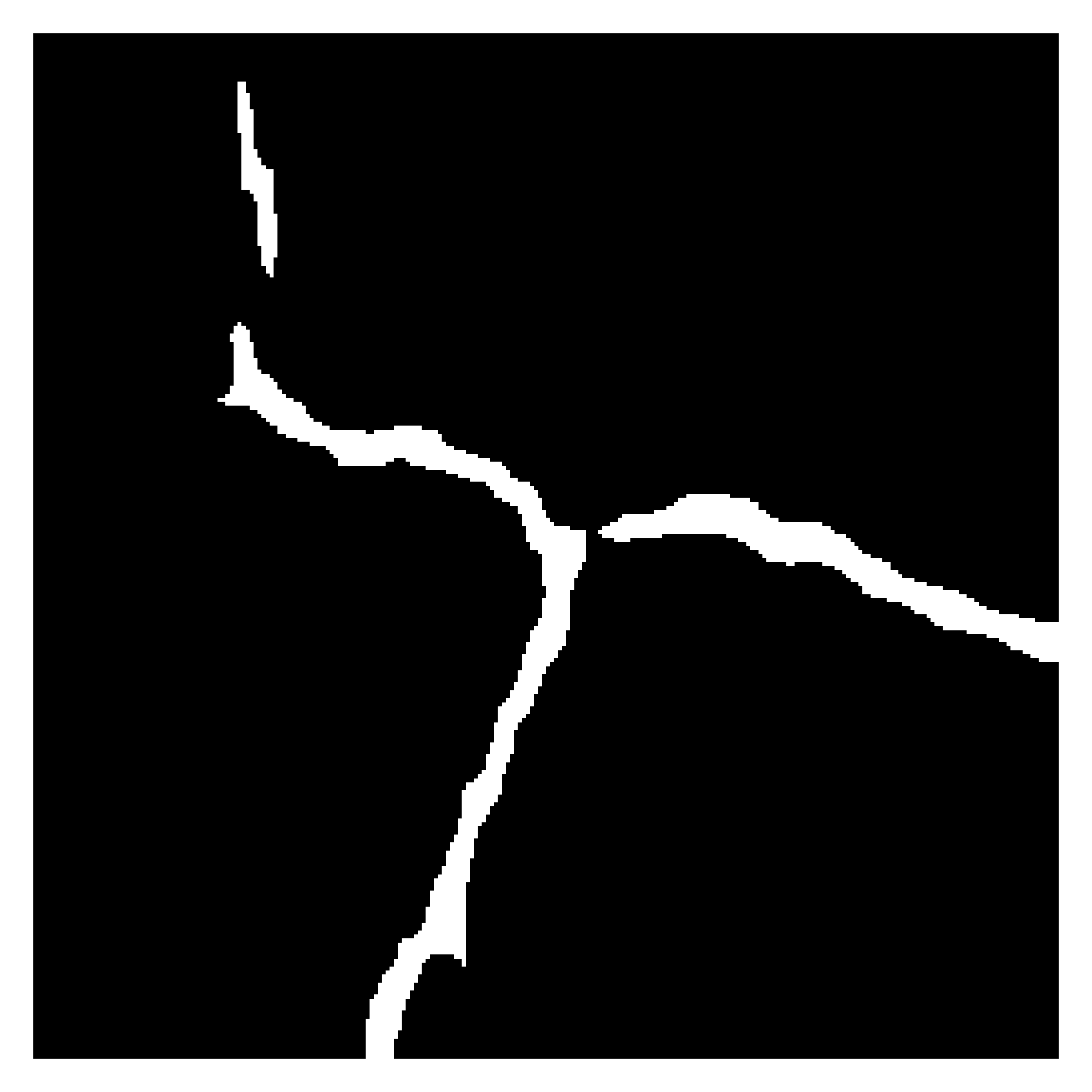}}
    \end{subfigure}

    \vspace{0.2cm}

    \begin{subfigure}[b]{0.09\textwidth} 
        \subcaption*{GAPS384} 
    \end{subfigure}
    \begin{subfigure}[b]{0.1\textwidth}
        \adjustbox{trim=10 10 10 10,clip,width=1.6cm,height=1.6cm}{\includegraphics{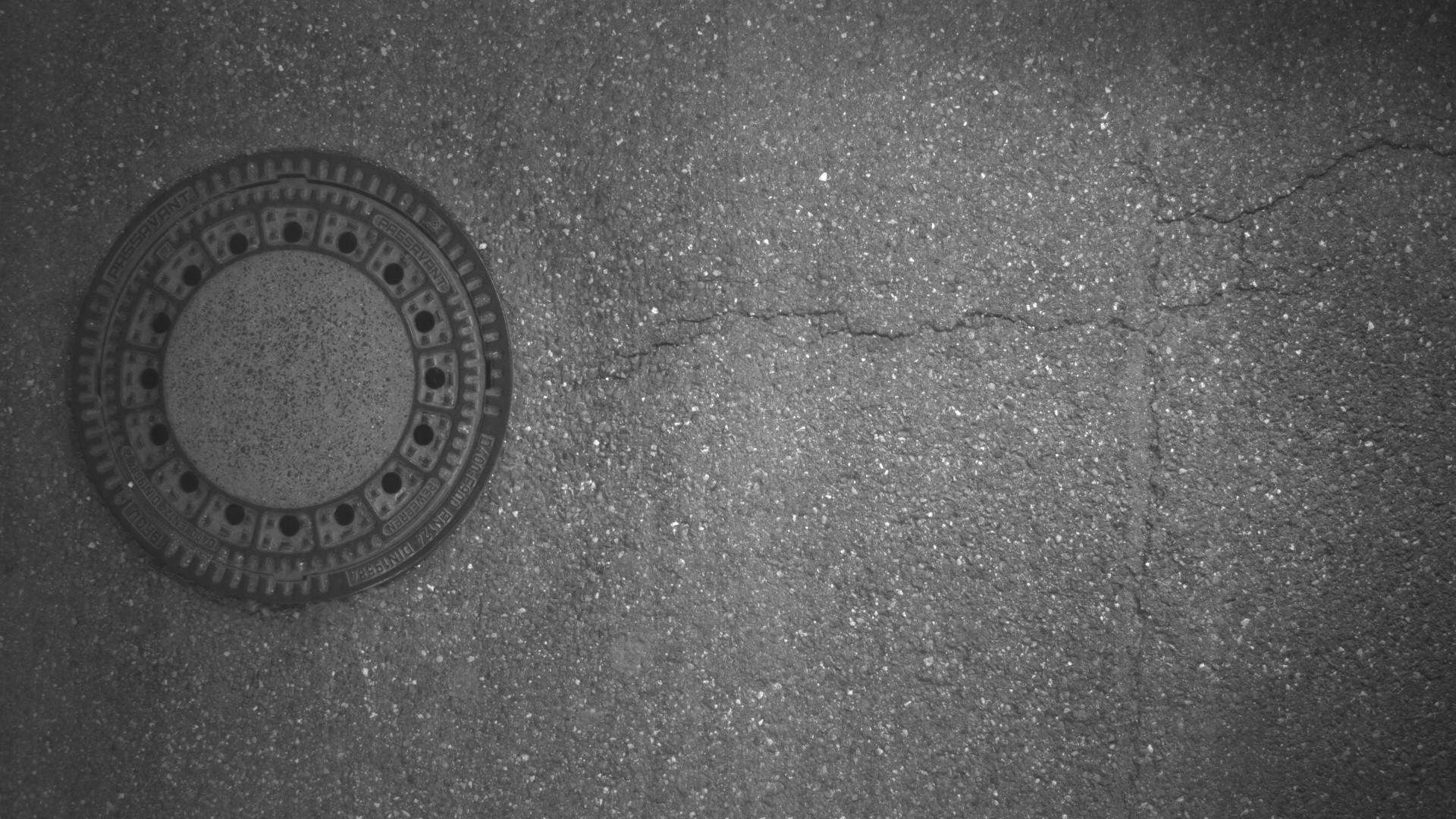}}
    \end{subfigure}
    \begin{subfigure}[b]{0.1\textwidth}
        \adjustbox{trim=10 10 10 10,clip,width=1.6cm,height=1.6cm}{\includegraphics{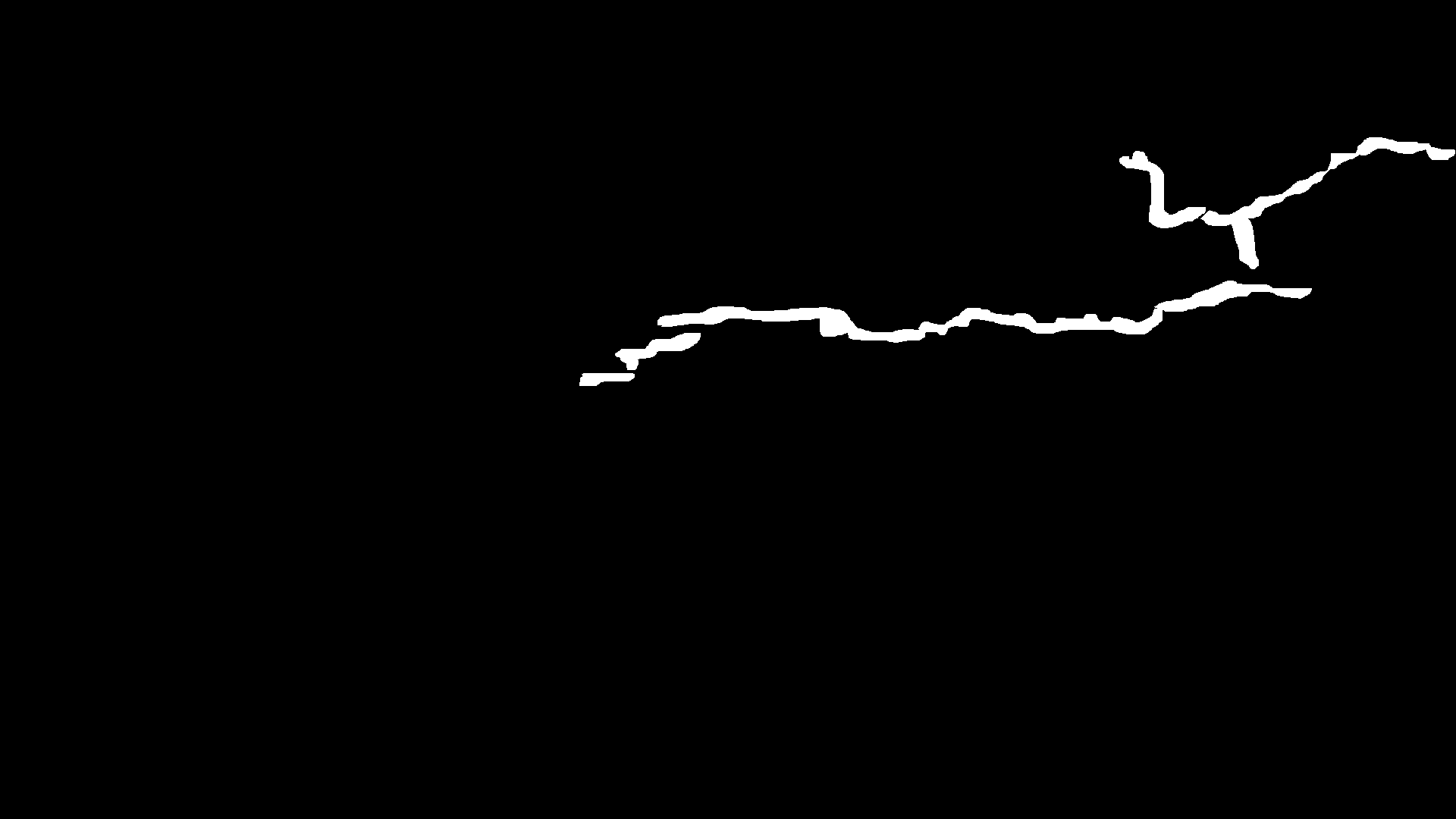}}
    \end{subfigure}
    \begin{subfigure}[b]{0.1\textwidth}
       \adjustbox{trim=10 10 10 10,clip,width=1.6cm,height=1.6cm}{\includegraphics{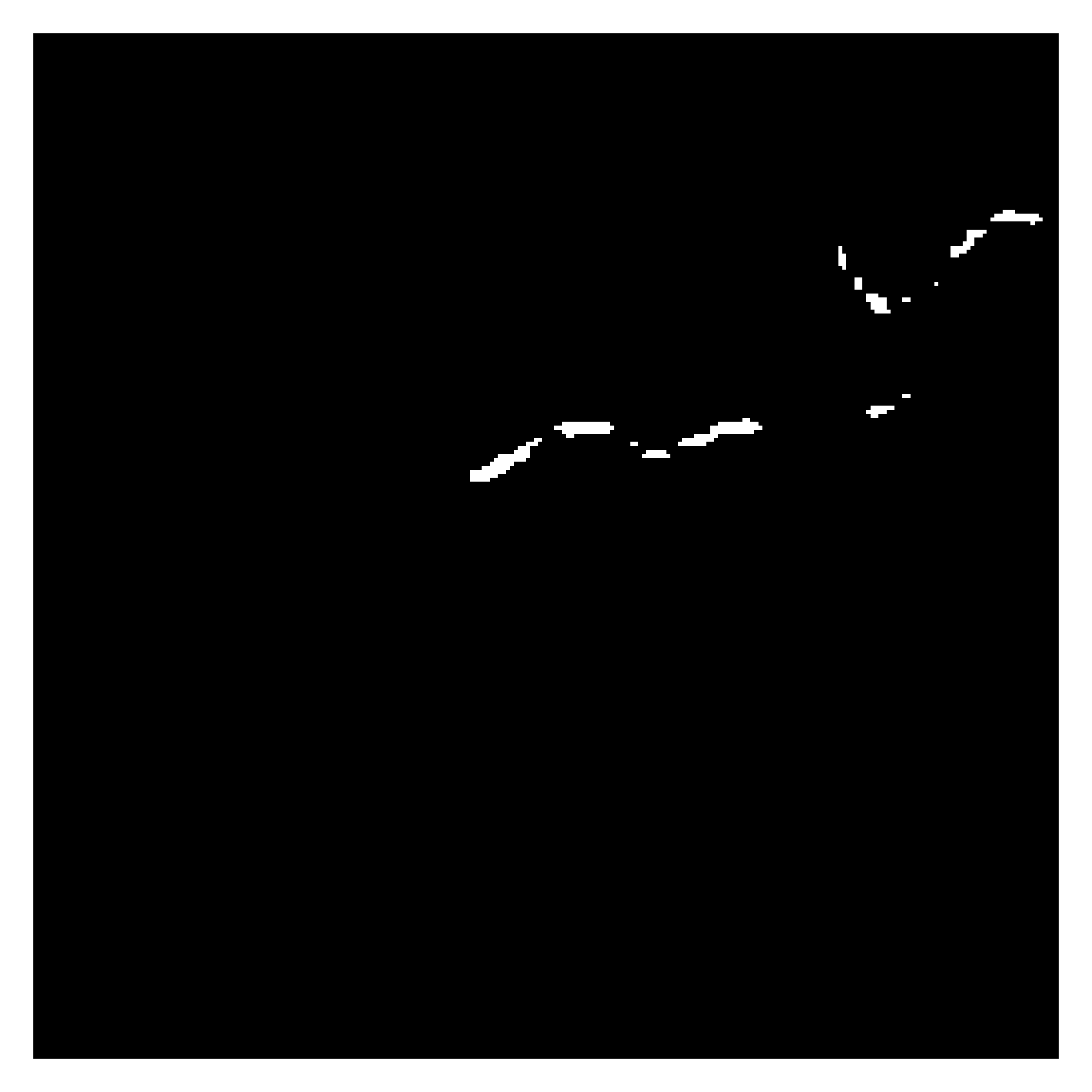}}
    \end{subfigure}
    \begin{subfigure}[b]{0.1\textwidth}
        \adjustbox{trim=10 10 10 10,clip,width=1.6cm,height=1.6cm}{\includegraphics{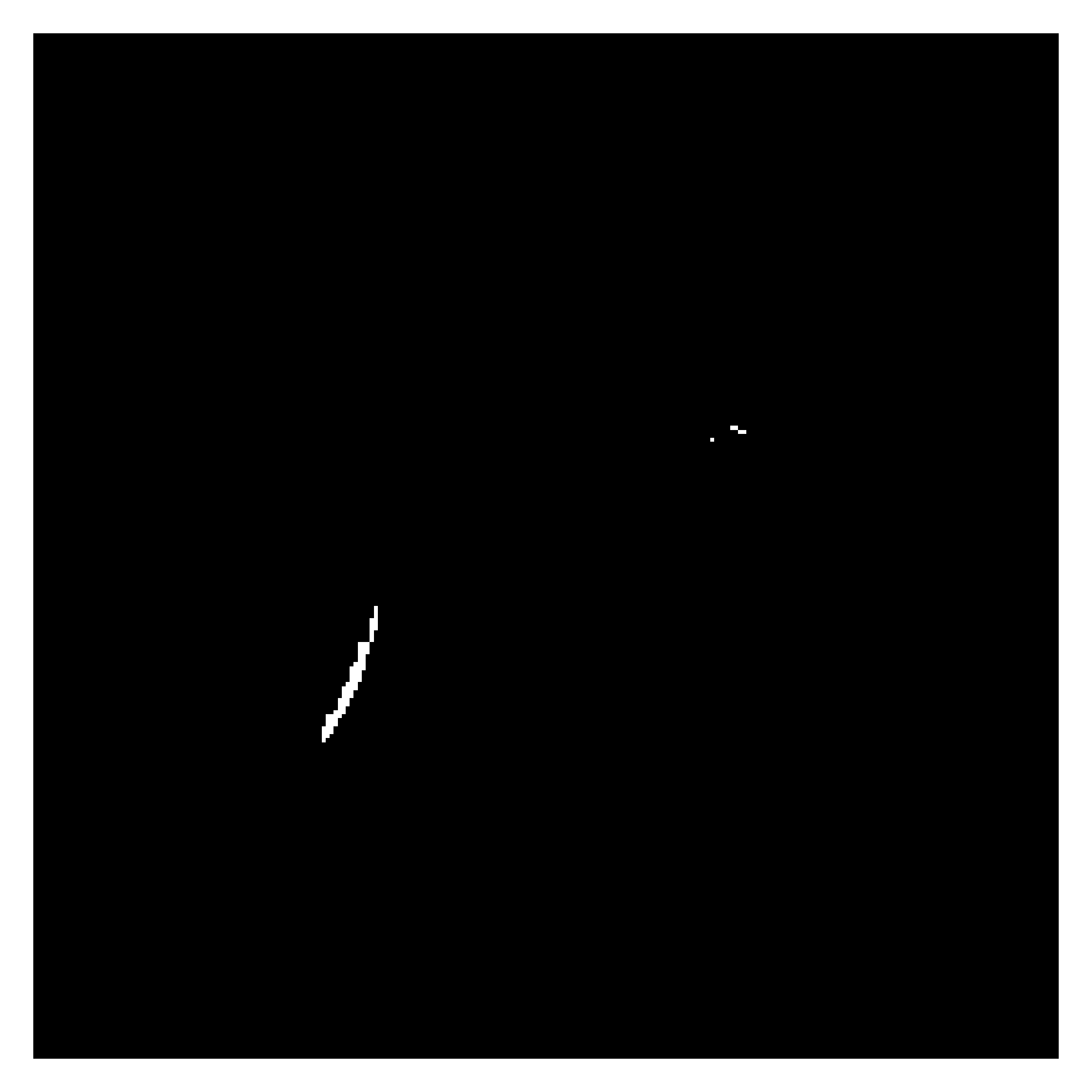}}
    \end{subfigure}
    \begin{subfigure}[b]{0.1\textwidth}
        \adjustbox{trim=10 10 10 10,clip,width=1.6cm,height=1.6cm}{\includegraphics{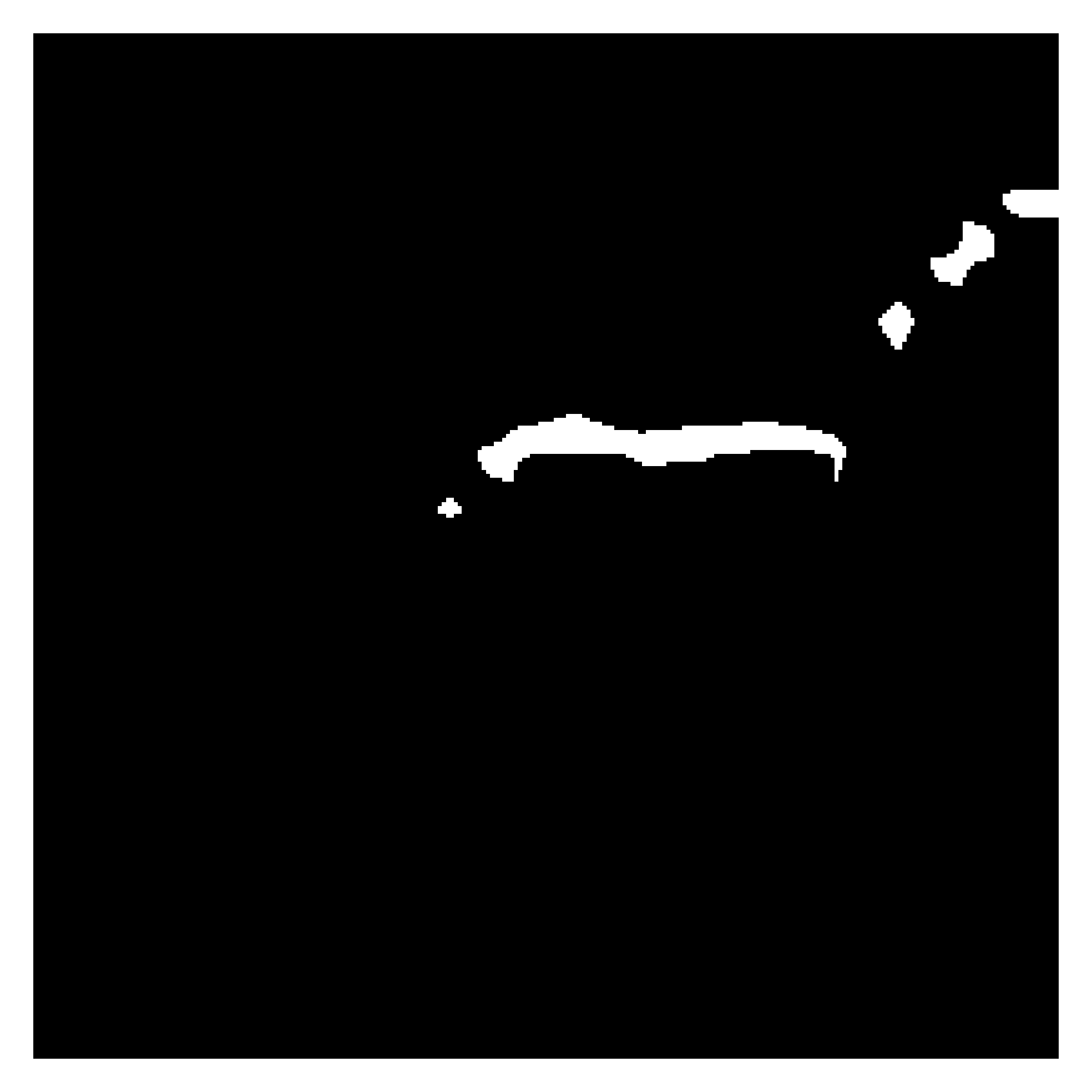}}
    \end{subfigure}
    \begin{subfigure}[b]{0.1\textwidth}
       \adjustbox{trim=10 10 10 10,clip,width=1.6cm,height=1.6cm}{\includegraphics{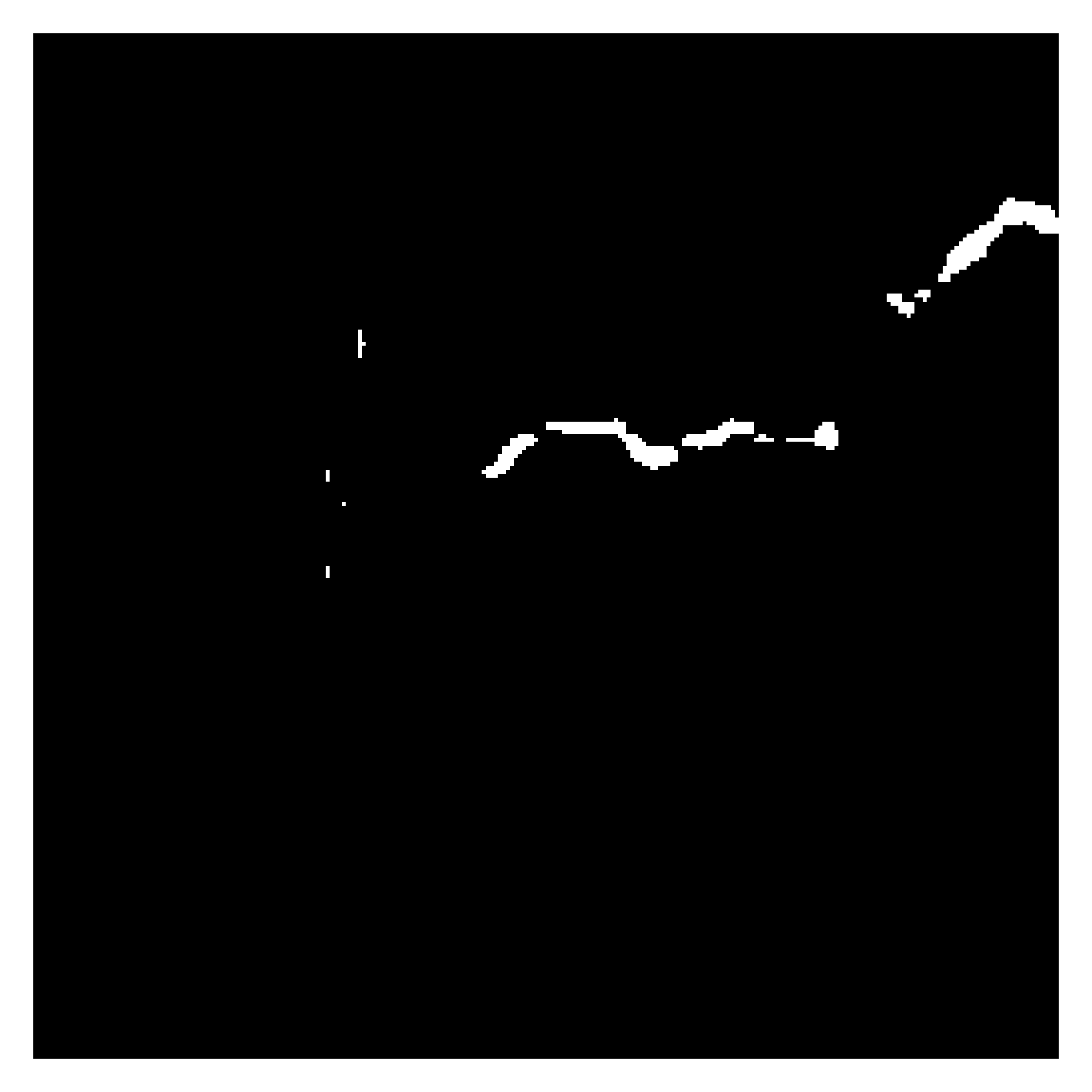}}
    \end{subfigure}
    \begin{subfigure}[b]{0.1\textwidth}
        \adjustbox{trim=10 10 10 10,clip,width=1.6cm,height=1.6cm}{\includegraphics{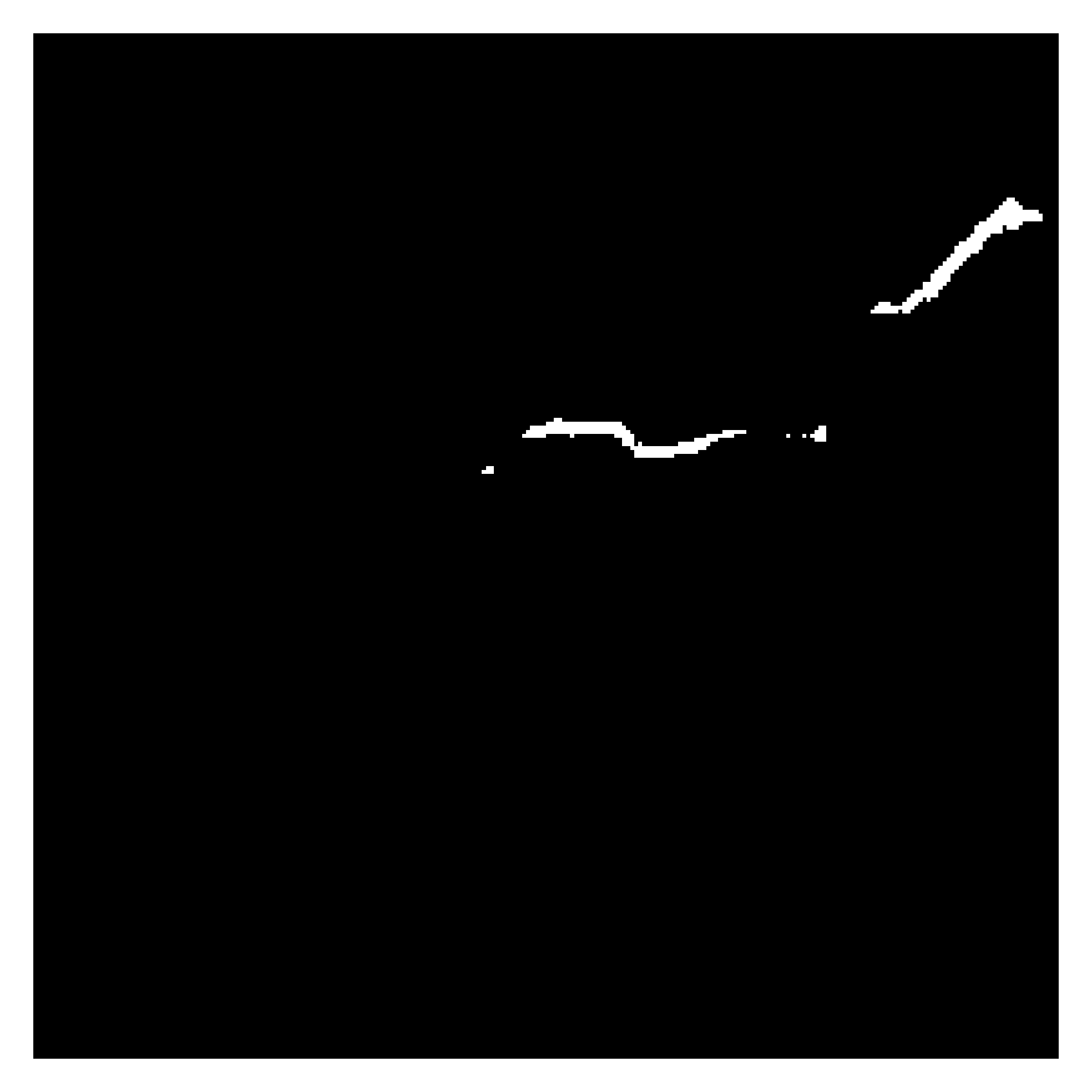}}
    \end{subfigure}
    \begin{subfigure}[b]{0.1\textwidth}
        \adjustbox{trim=10 10 10 10,clip,width=1.6cm,height=1.6cm}{\includegraphics{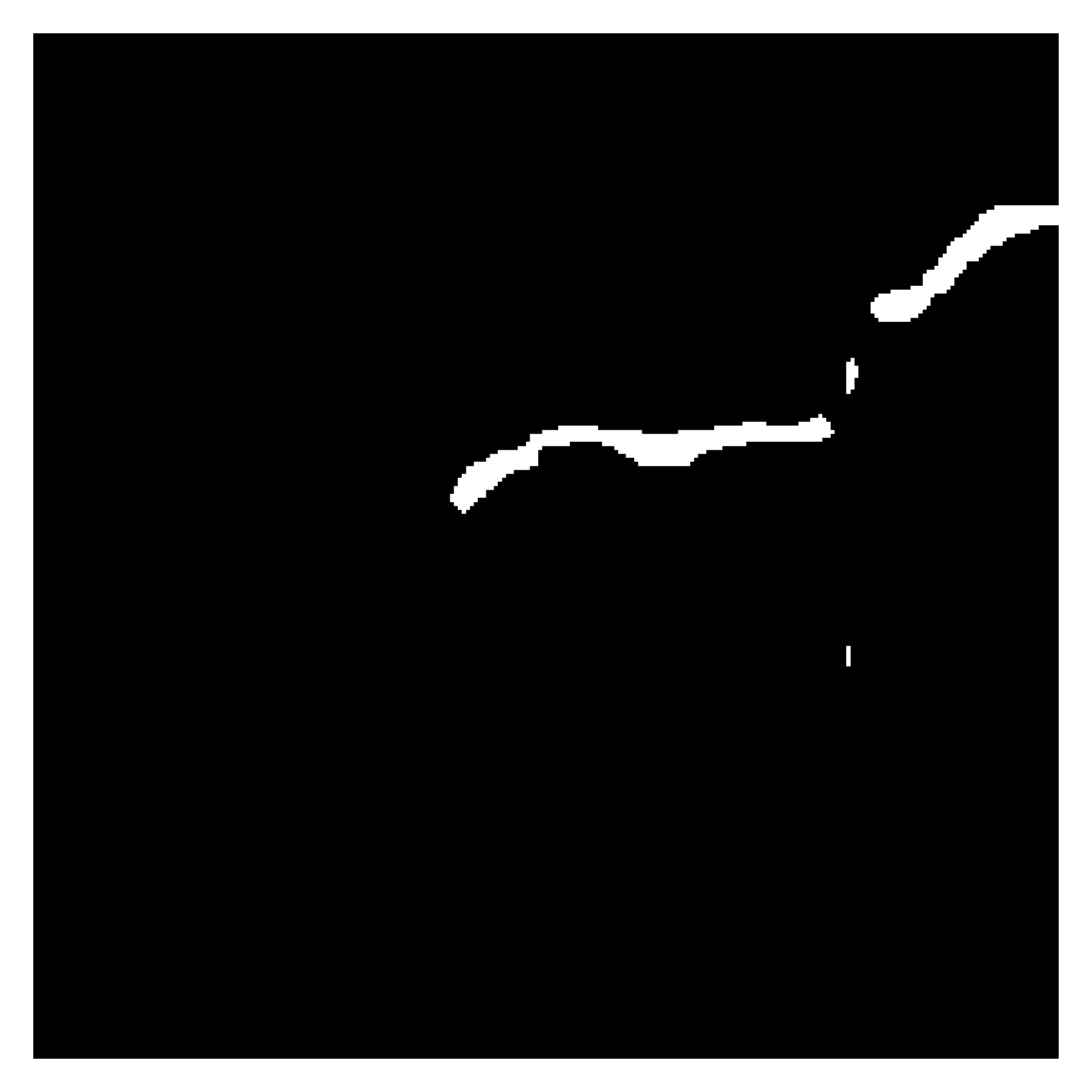}}
    \end{subfigure}
    
    \vspace{0.2cm}
    
    \begin{subfigure}[b]{0.09\textwidth} 
        \subcaption*{Khanh11k\_\\
        Rissbilder}
    \end{subfigure}
    \begin{subfigure}[b]{0.1\textwidth}
        \adjustbox{trim=10 10 10 10,clip,width=1.6cm,height=1.6cm}{\includegraphics{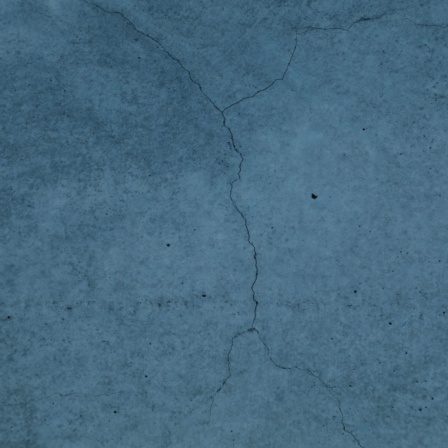}}
    \end{subfigure}
    \begin{subfigure}[b]{0.1\textwidth}
        \adjustbox{trim=10 10 10 10,clip,width=1.6cm,height=1.6cm}{\includegraphics{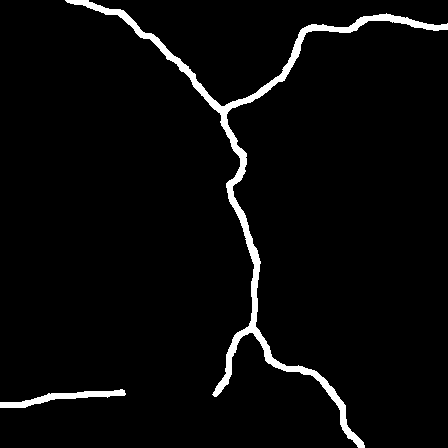}}
    \end{subfigure}
    \begin{subfigure}[b]{0.1\textwidth}
       \adjustbox{trim=10 10 10 10,clip,width=1.6cm,height=1.6cm}{\includegraphics{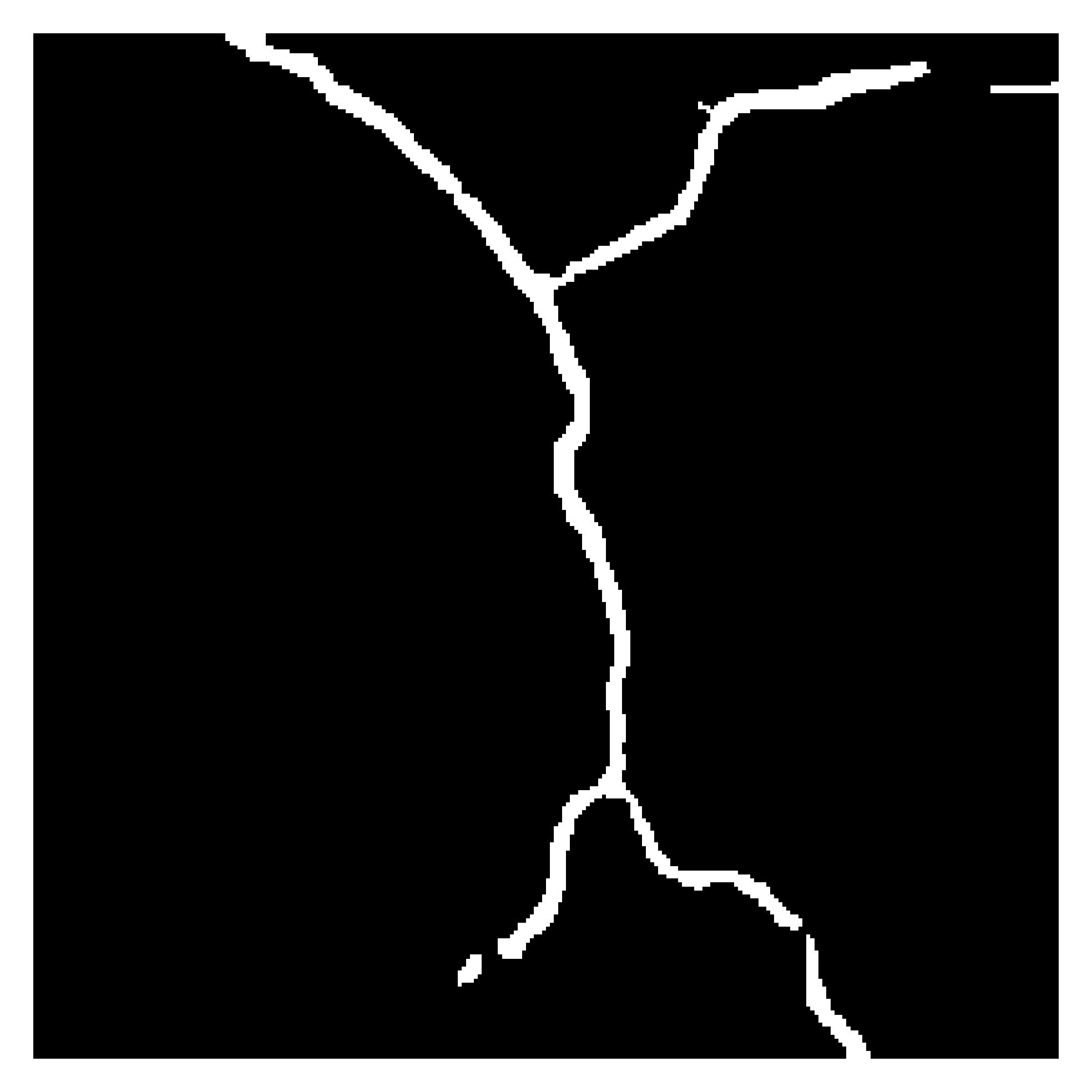}}
    \end{subfigure}
    \begin{subfigure}[b]{0.1\textwidth}
        \adjustbox{trim=10 10 10 10,clip,width=1.6cm,height=1.6cm}{\includegraphics{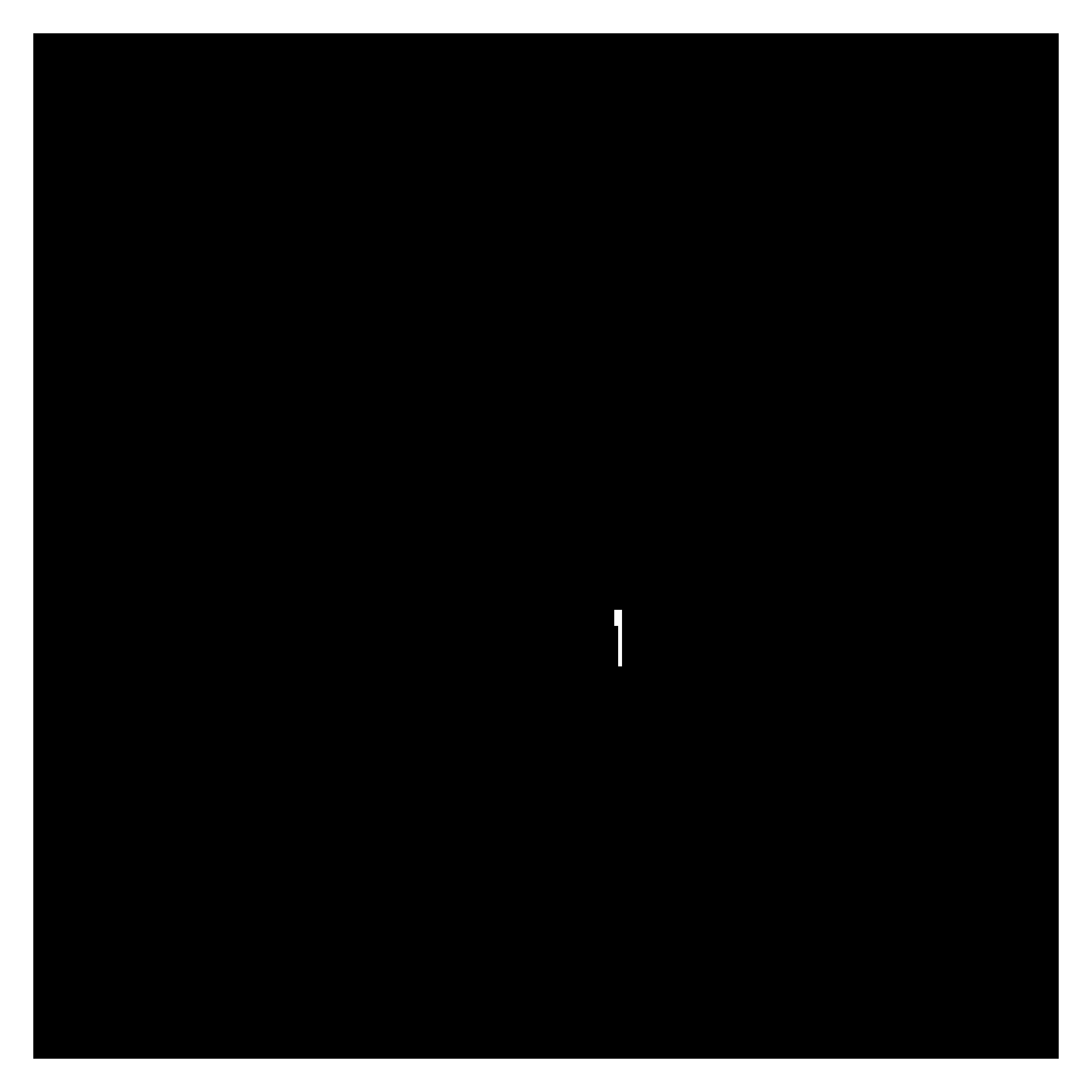}}
    \end{subfigure}
    \begin{subfigure}[b]{0.1\textwidth}
        \adjustbox{trim=10 10 10 10,clip,width=1.6cm,height=1.6cm}{\includegraphics{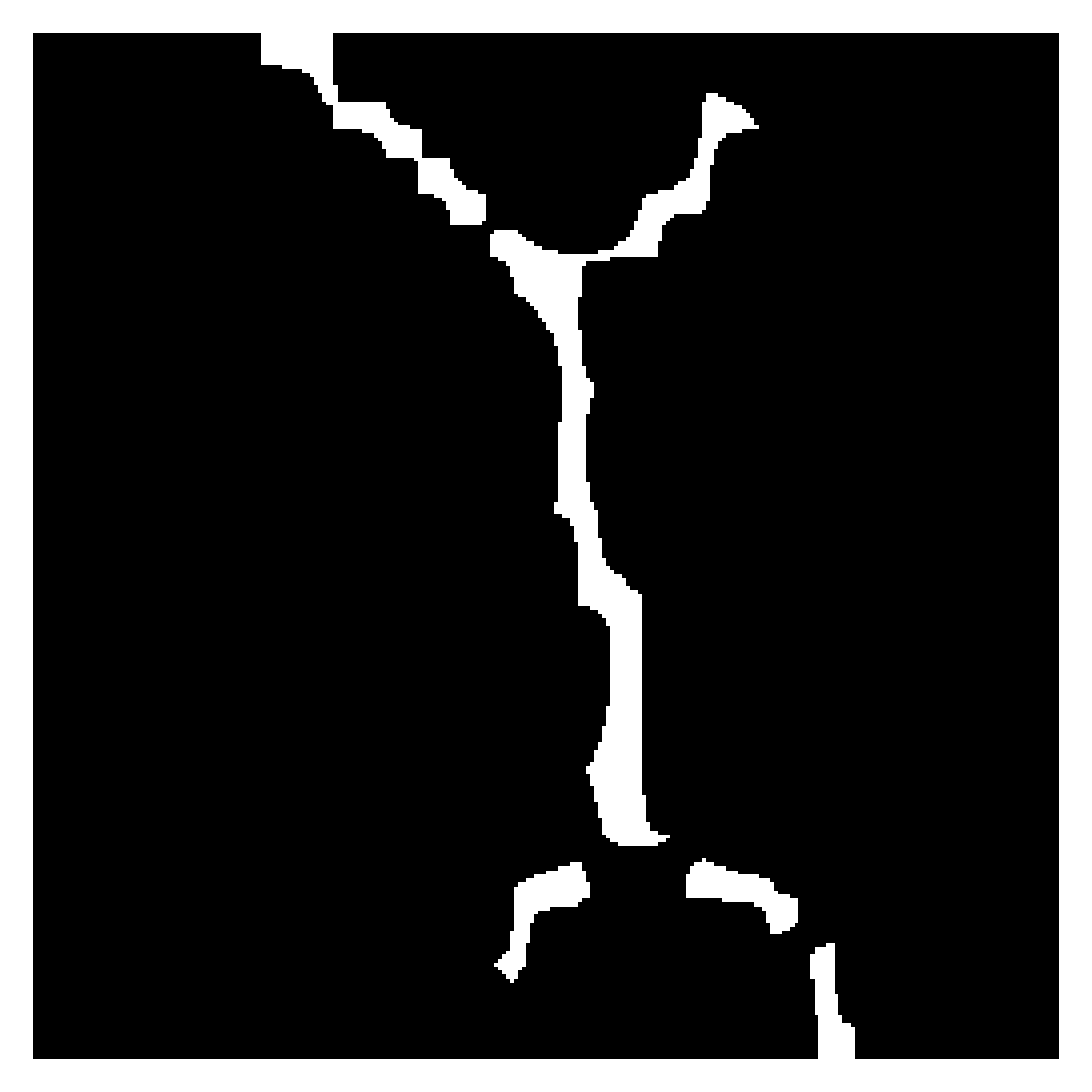}}
    \end{subfigure}
    \begin{subfigure}[b]{0.1\textwidth}
       \adjustbox{trim=10 10 10 10,clip,width=1.6cm,height=1.6cm}{\includegraphics{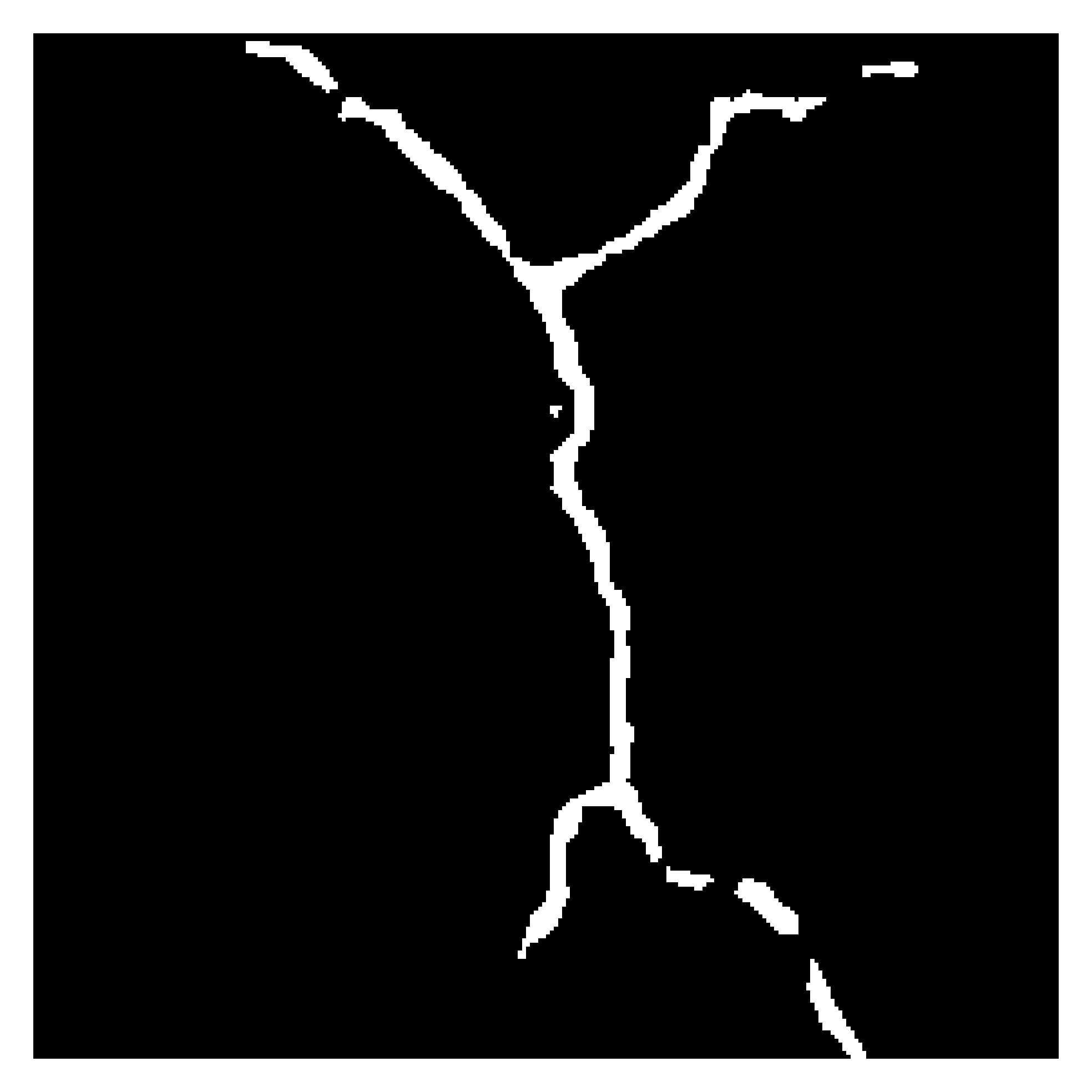}}
    \end{subfigure}
    \begin{subfigure}[b]{0.1\textwidth}
        \adjustbox{trim=10 10 10 10,clip,width=1.6cm,height=1.6cm}{\includegraphics{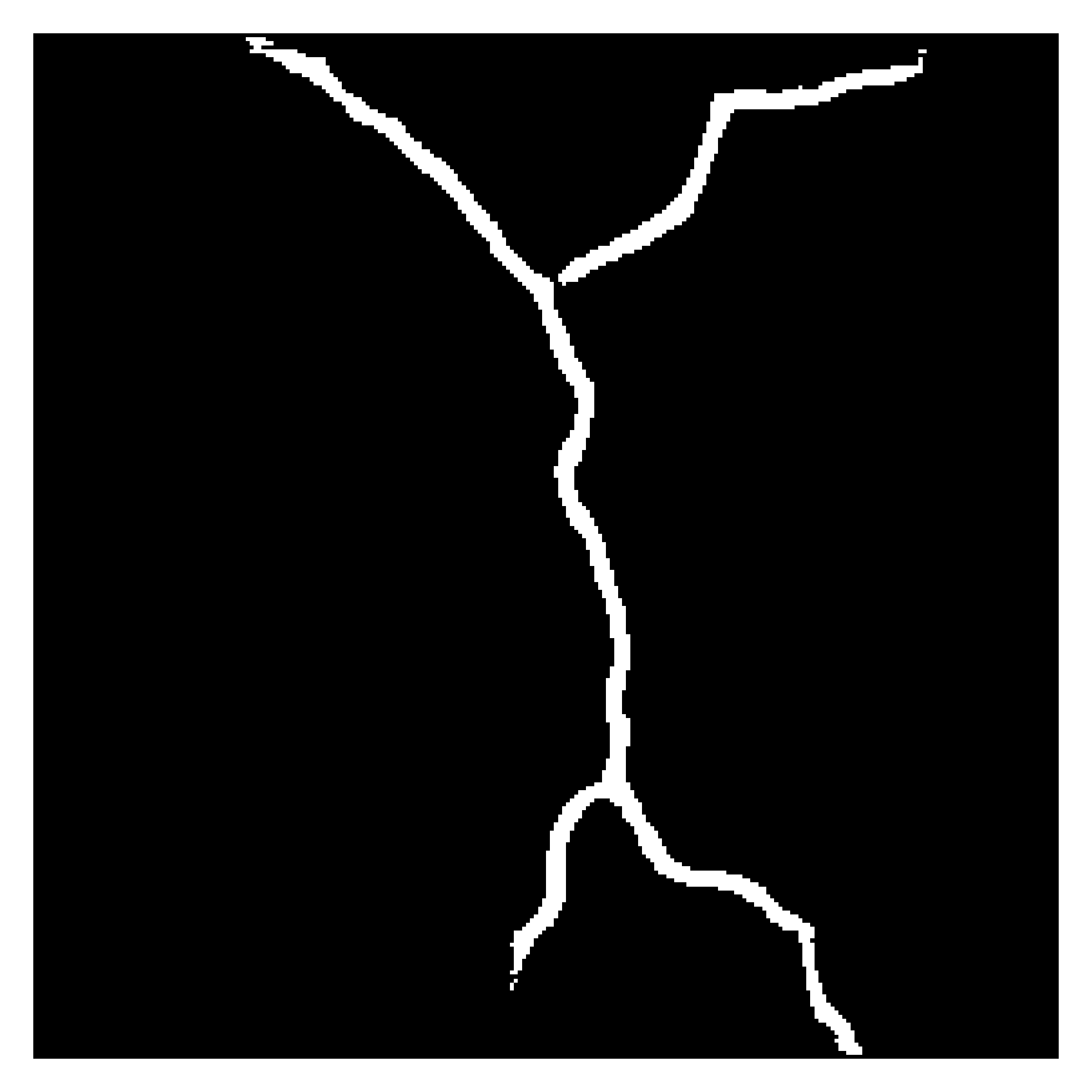}}
    \end{subfigure}
    \begin{subfigure}[b]{0.1\textwidth}
        \adjustbox{trim=10 10 10 10,clip,width=1.6cm,height=1.6cm}{\includegraphics{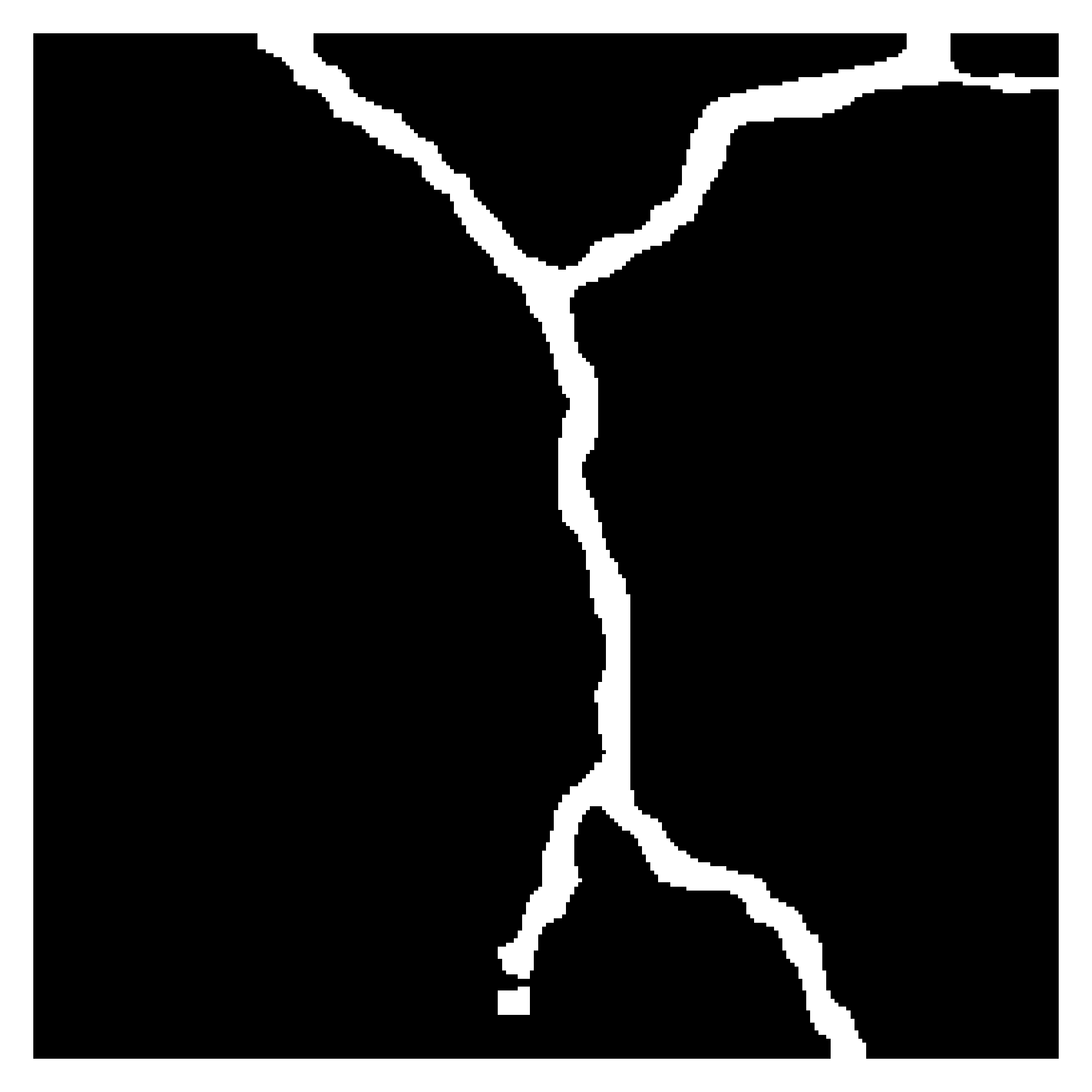}}
    \end{subfigure}
    
    \vspace{0.2cm}
    
    \begin{subfigure}[b]{0.09\textwidth} 
        \subcaption*{Khanh11k\_\\Volker} 
    \end{subfigure}
    \begin{subfigure}[b]{0.1\textwidth}
        \adjustbox{trim=10 10 10 10,clip,width=1.6cm,height=1.6cm}{\includegraphics{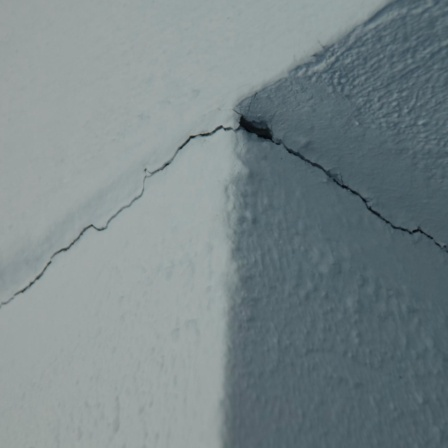}}
    \end{subfigure}
    \begin{subfigure}[b]{0.1\textwidth}
        \adjustbox{trim=10 10 10 10,clip,width=1.6cm,height=1.6cm}{\includegraphics{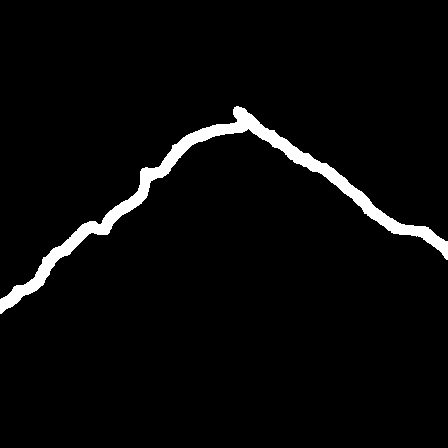}}
    \end{subfigure}
    \begin{subfigure}[b]{0.1\textwidth}
       \adjustbox{trim=10 10 10 10,clip,width=1.6cm,height=1.6cm}{\includegraphics{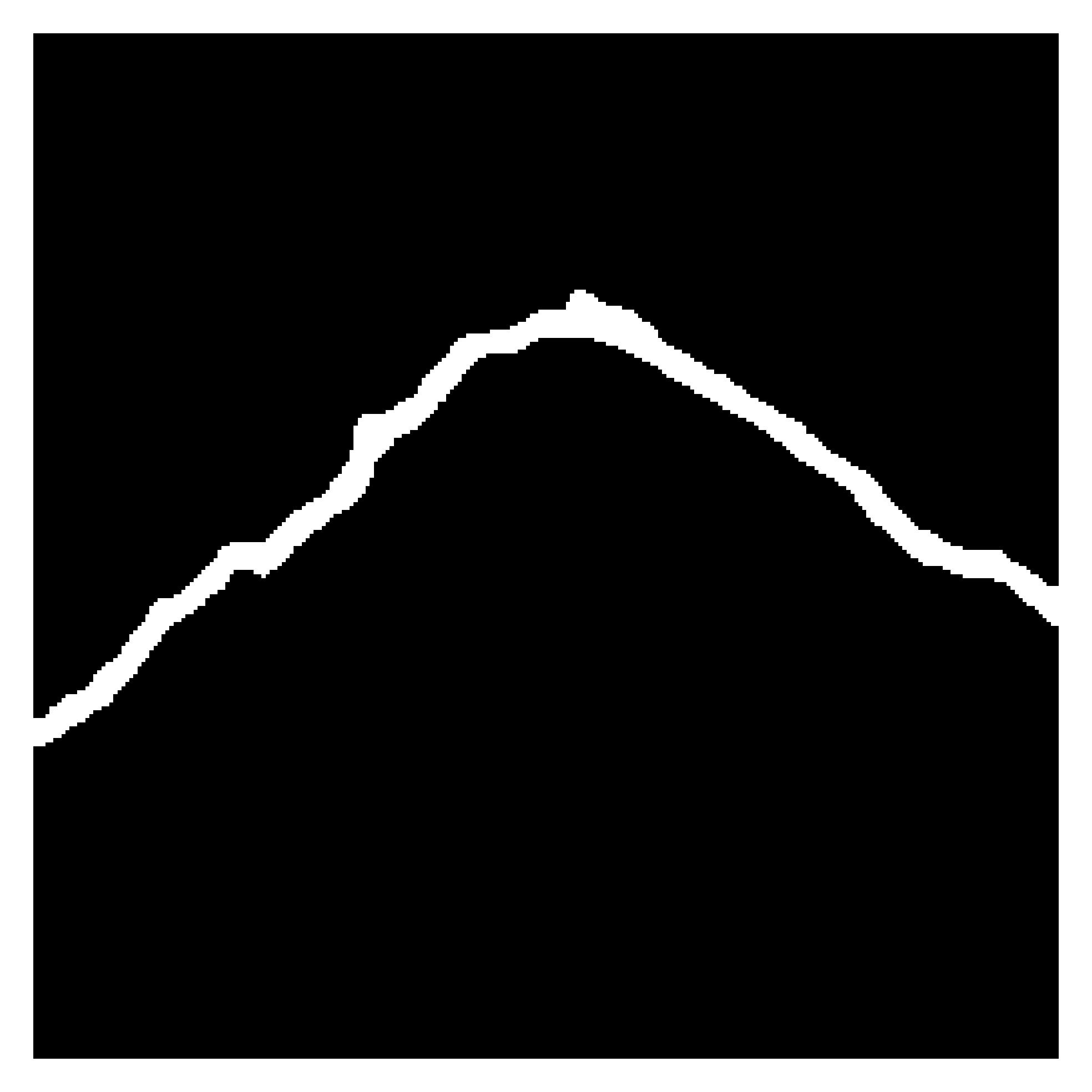}}
    \end{subfigure}
    \begin{subfigure}[b]{0.1\textwidth}
        \adjustbox{trim=10 10 10 10,clip,width=1.6cm,height=1.6cm}{\includegraphics{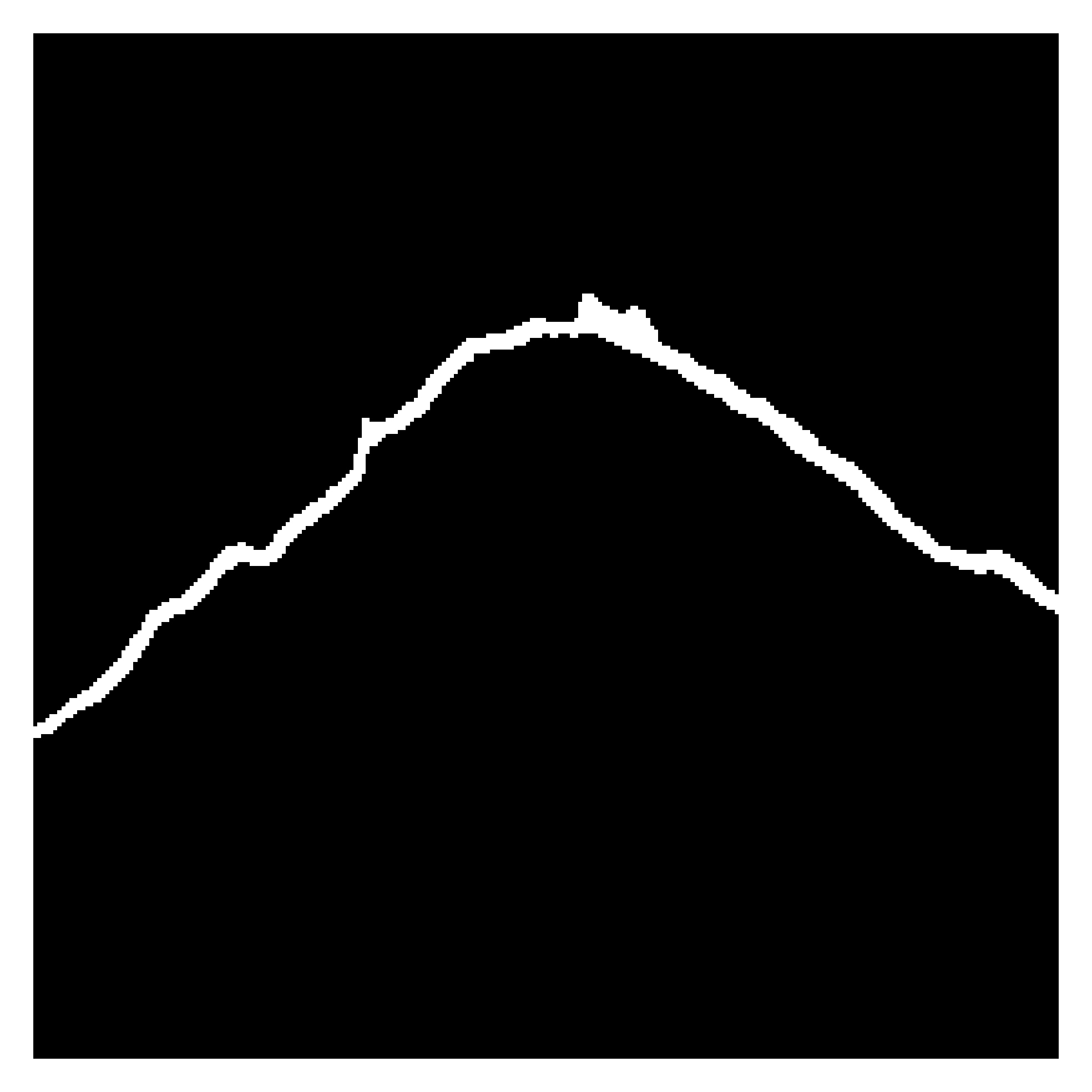}}
    \end{subfigure}
    \begin{subfigure}[b]{0.1\textwidth}
        \adjustbox{trim=10 10 10 10,clip,width=1.6cm,height=1.6cm}{\includegraphics{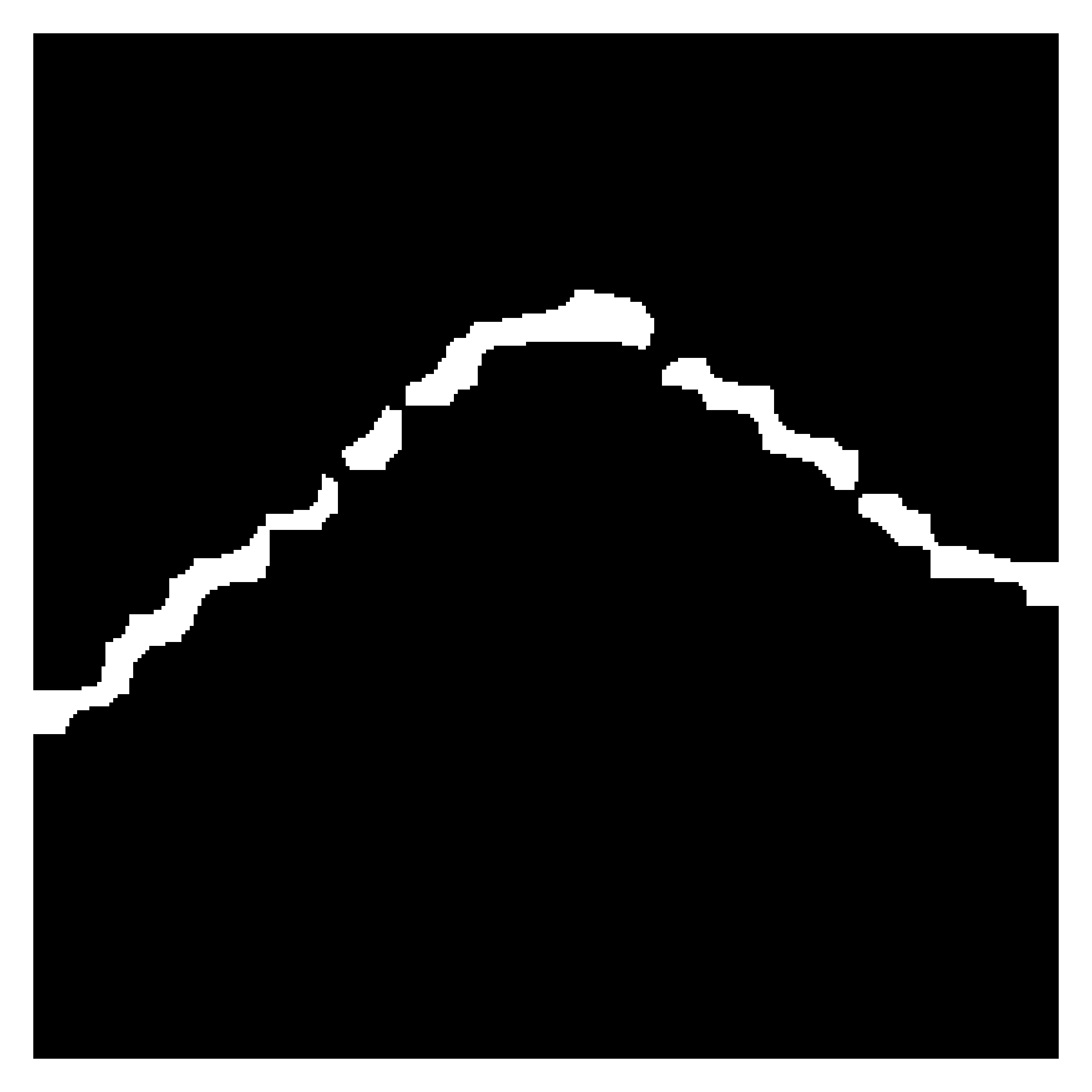}}
    \end{subfigure}
    \begin{subfigure}[b]{0.1\textwidth}
       \adjustbox{trim=10 10 10 10,clip,width=1.6cm,height=1.6cm}{\includegraphics{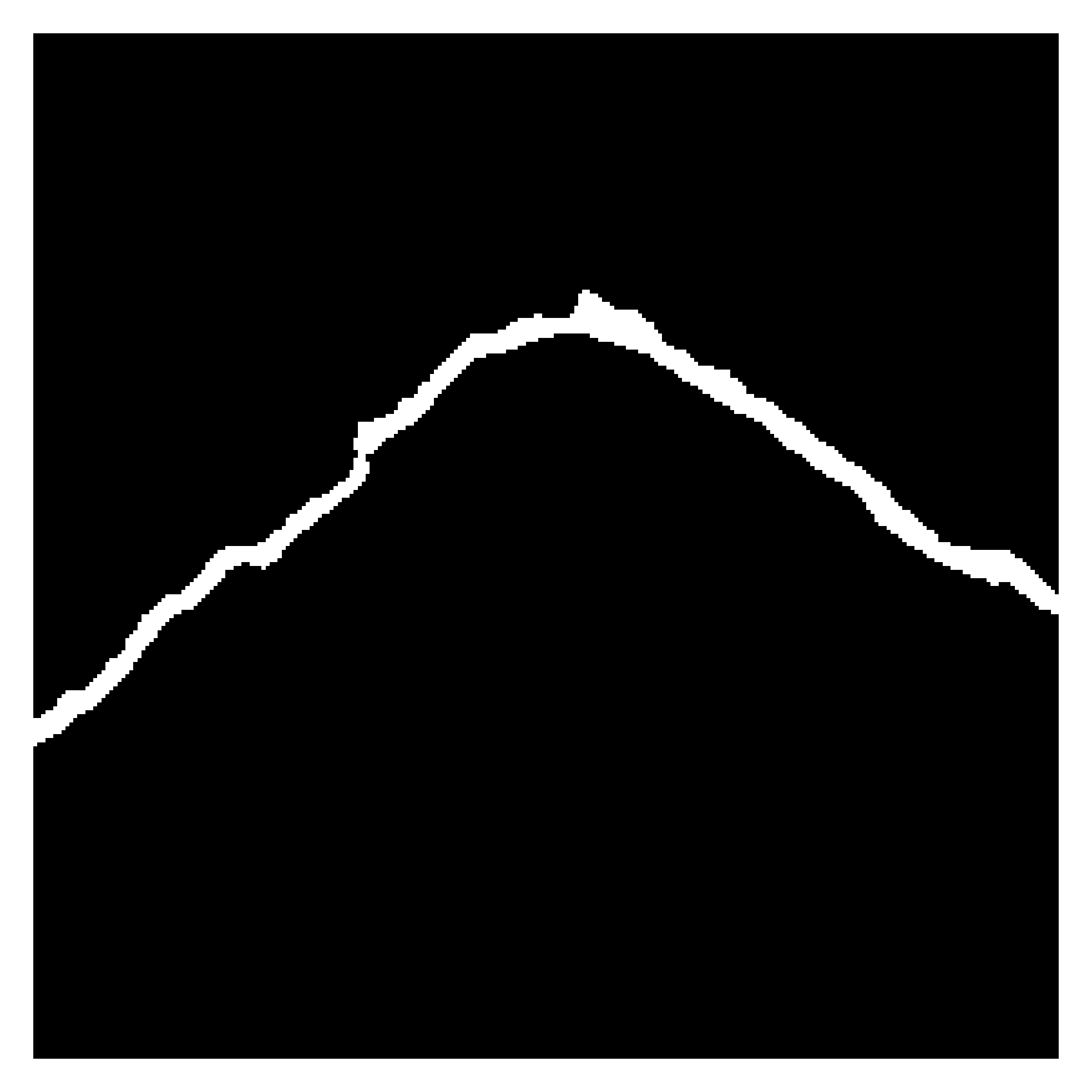}}
    \end{subfigure}
    \begin{subfigure}[b]{0.1\textwidth}
        \adjustbox{trim=10 10 10 10,clip,width=1.6cm,height=1.6cm}{\includegraphics{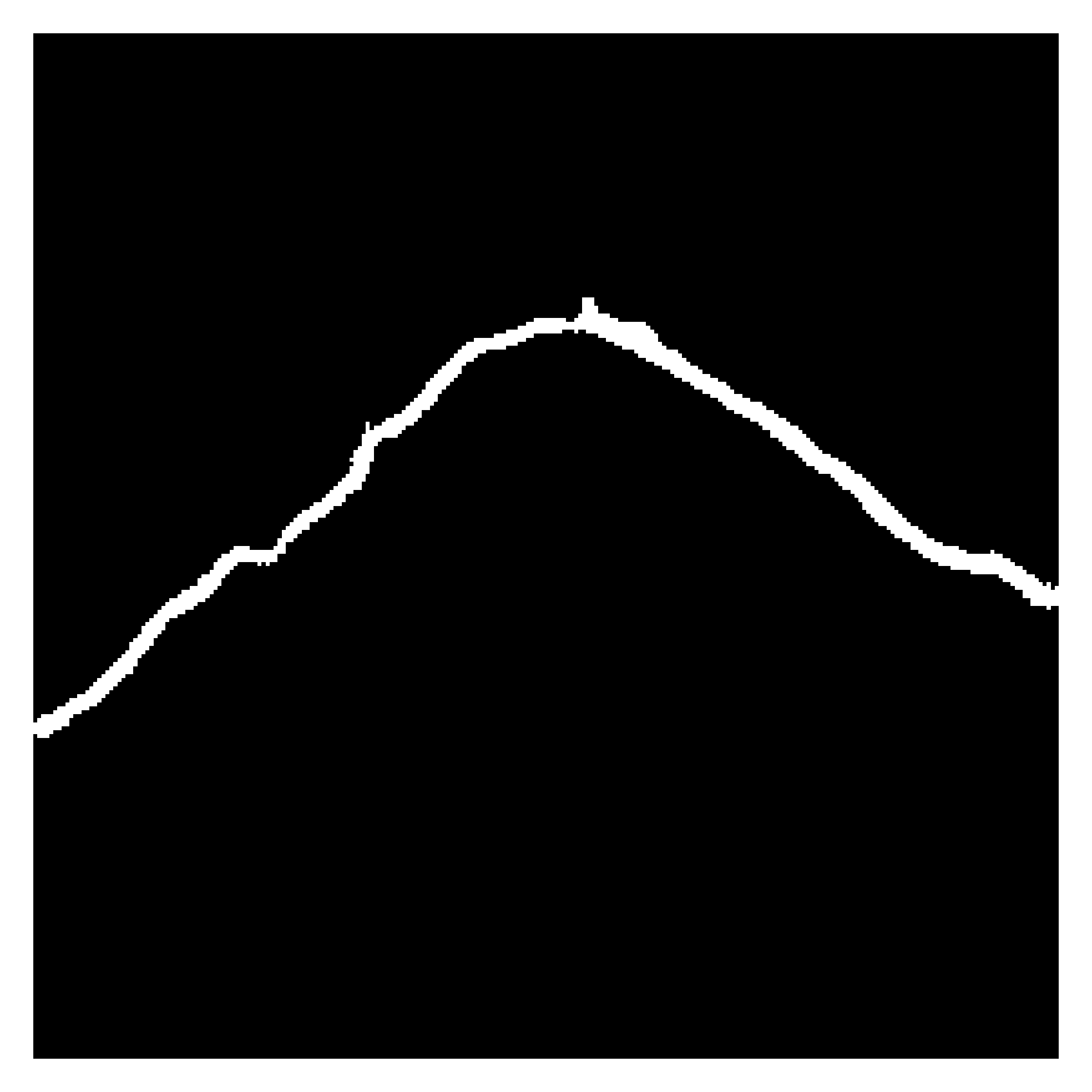}}
    \end{subfigure}
    \begin{subfigure}[b]{0.1\textwidth}
        \adjustbox{trim=10 10 10 10,clip,width=1.6cm,height=1.6cm}{\includegraphics{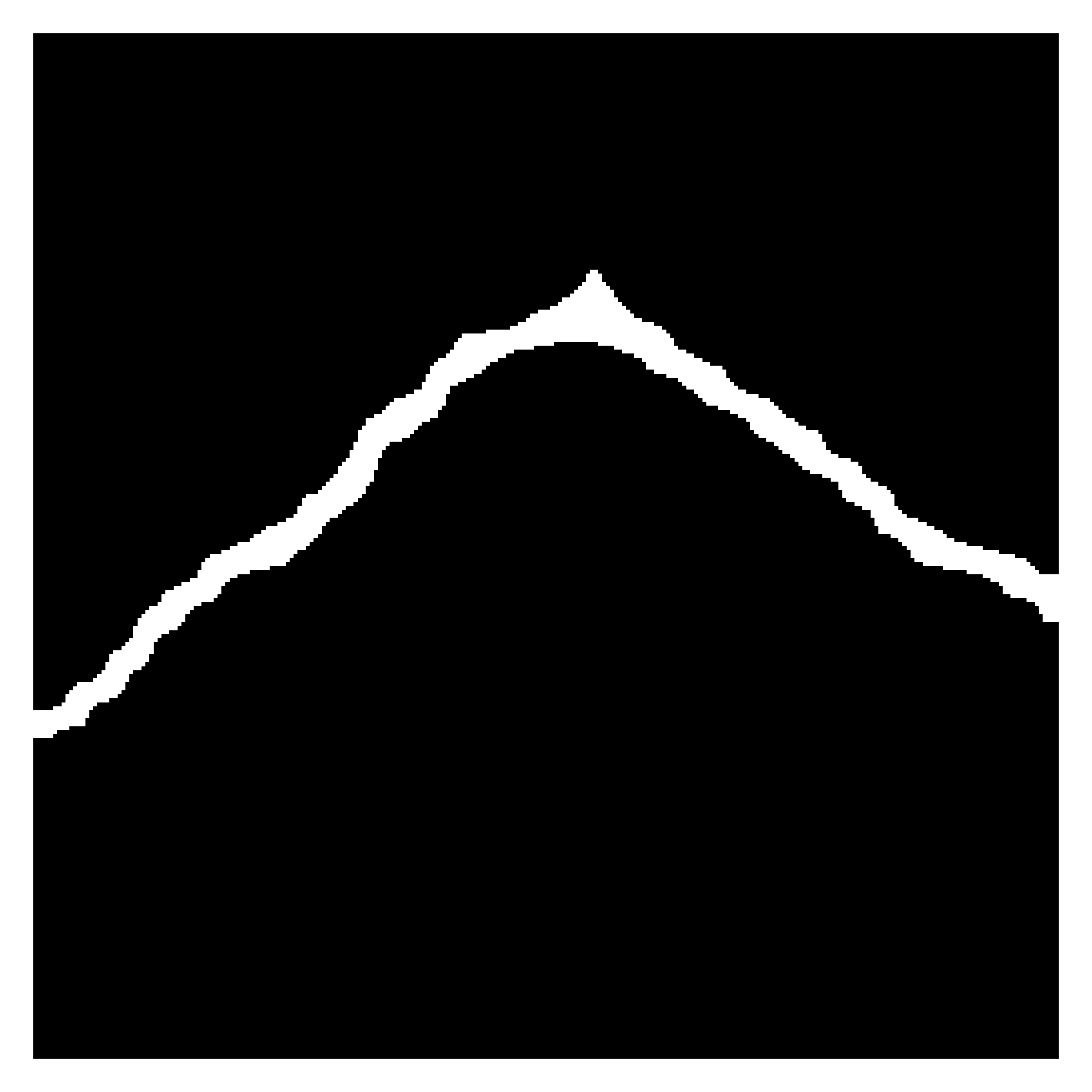}}
    \end{subfigure}

    \vspace{0.2cm}
    
    \begin{subfigure}[b]{0.09\textwidth} 
        \subcaption*{LCW} 
    \end{subfigure}
    \begin{subfigure}[b]{0.1\textwidth}
        \adjustbox{trim=10 10 10 10,clip,width=1.6cm,height=1.6cm}{\includegraphics{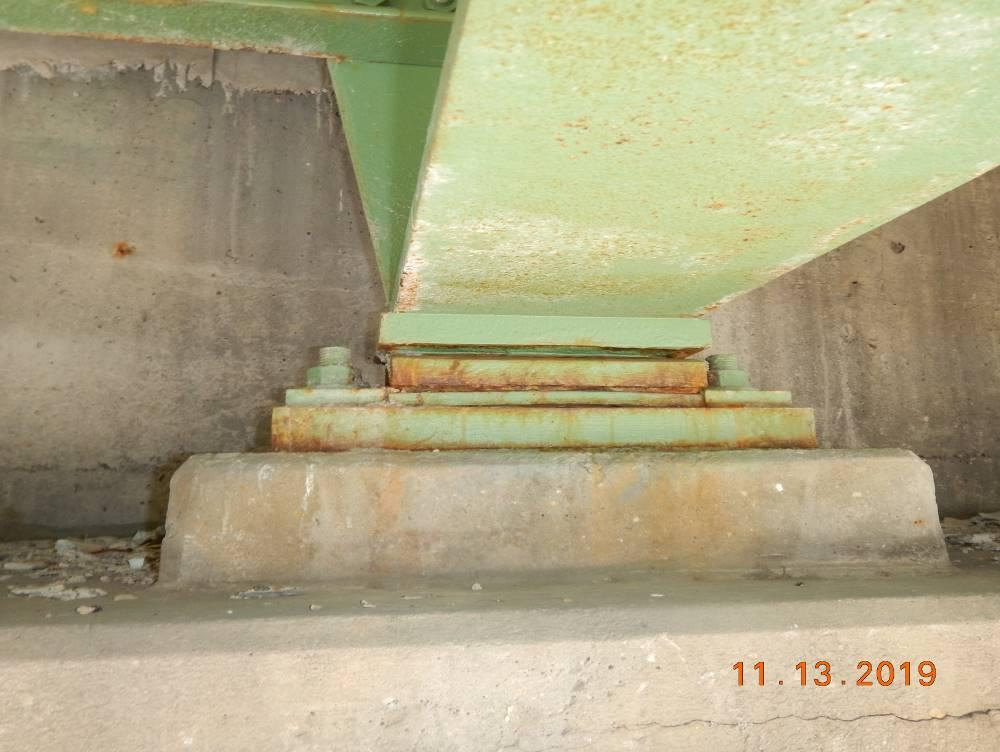}}
    \end{subfigure}
    \begin{subfigure}[b]{0.1\textwidth}
        \adjustbox{trim=10 10 10 10,clip,width=1.6cm,height=1.6cm}{\includegraphics{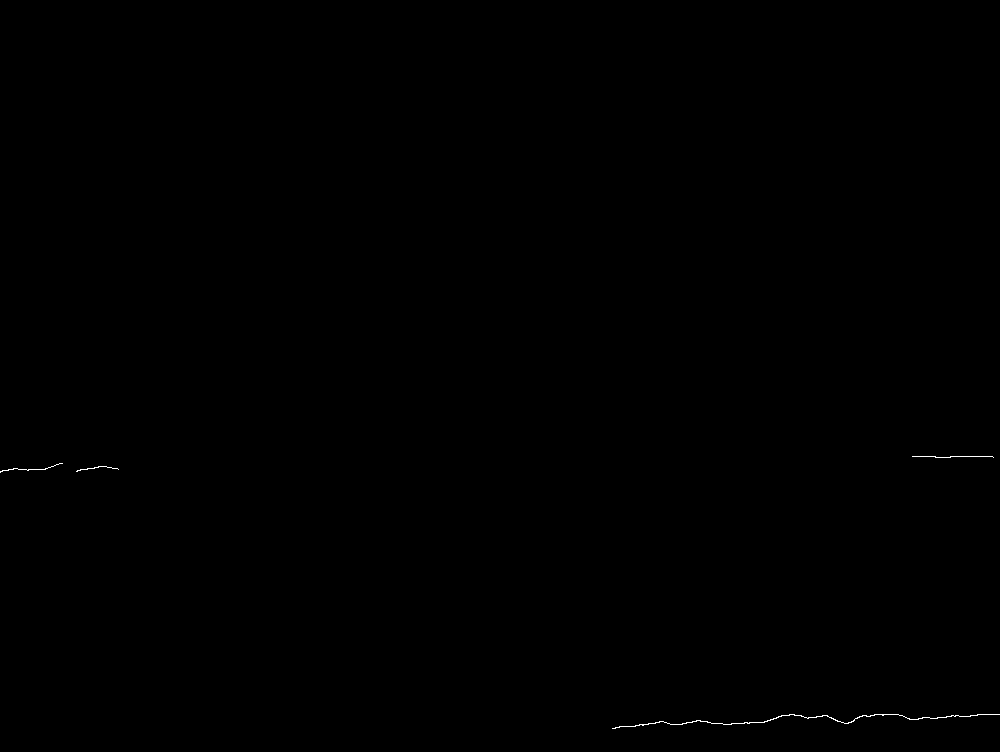}}
    \end{subfigure}
    \begin{subfigure}[b]{0.1\textwidth}
       \adjustbox{trim=10 10 10 10,clip,width=1.6cm,height=1.6cm}{\includegraphics{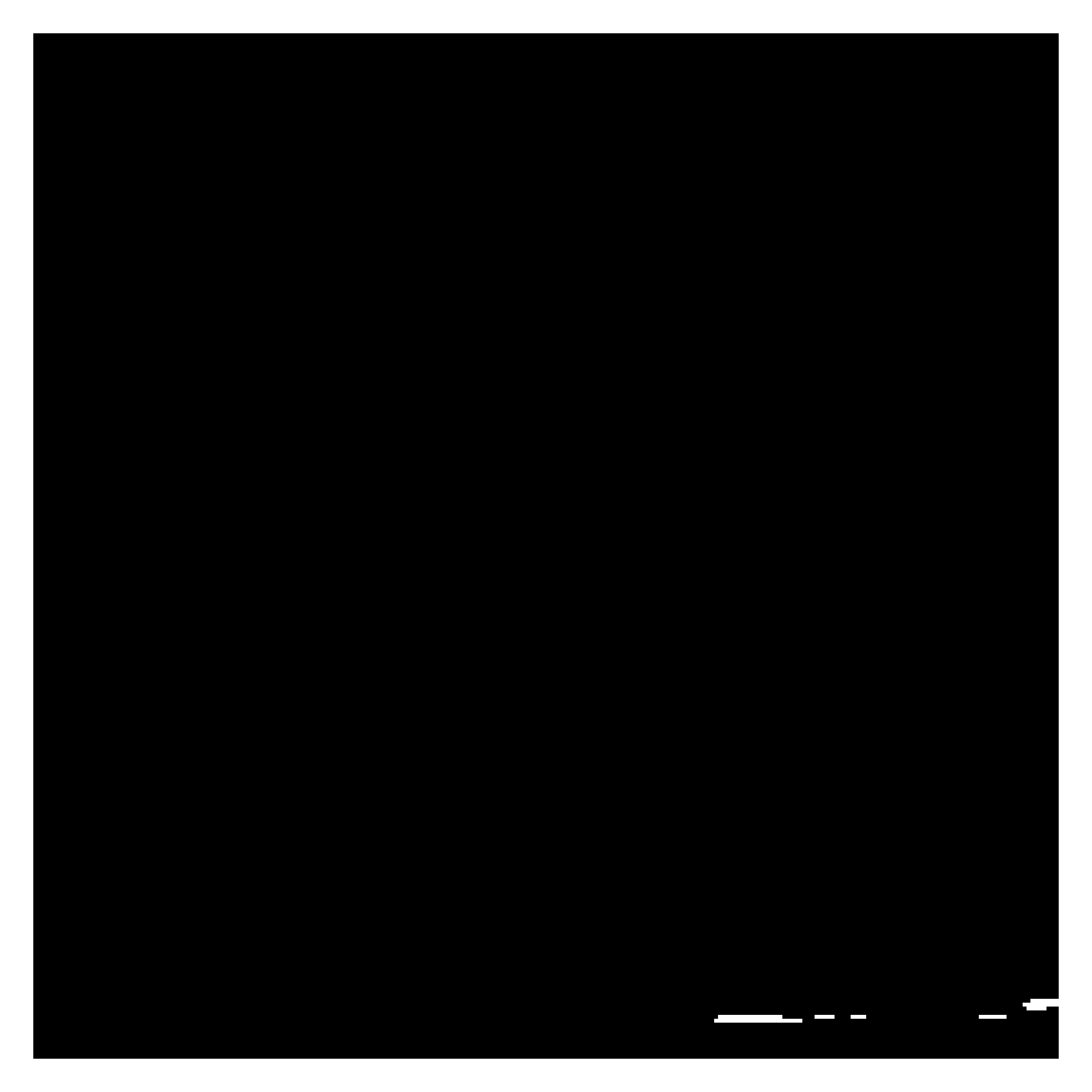}}
    \end{subfigure}
    \begin{subfigure}[b]{0.1\textwidth}
        \adjustbox{trim=10 10 10 10,clip,width=1.6cm,height=1.6cm}{\includegraphics{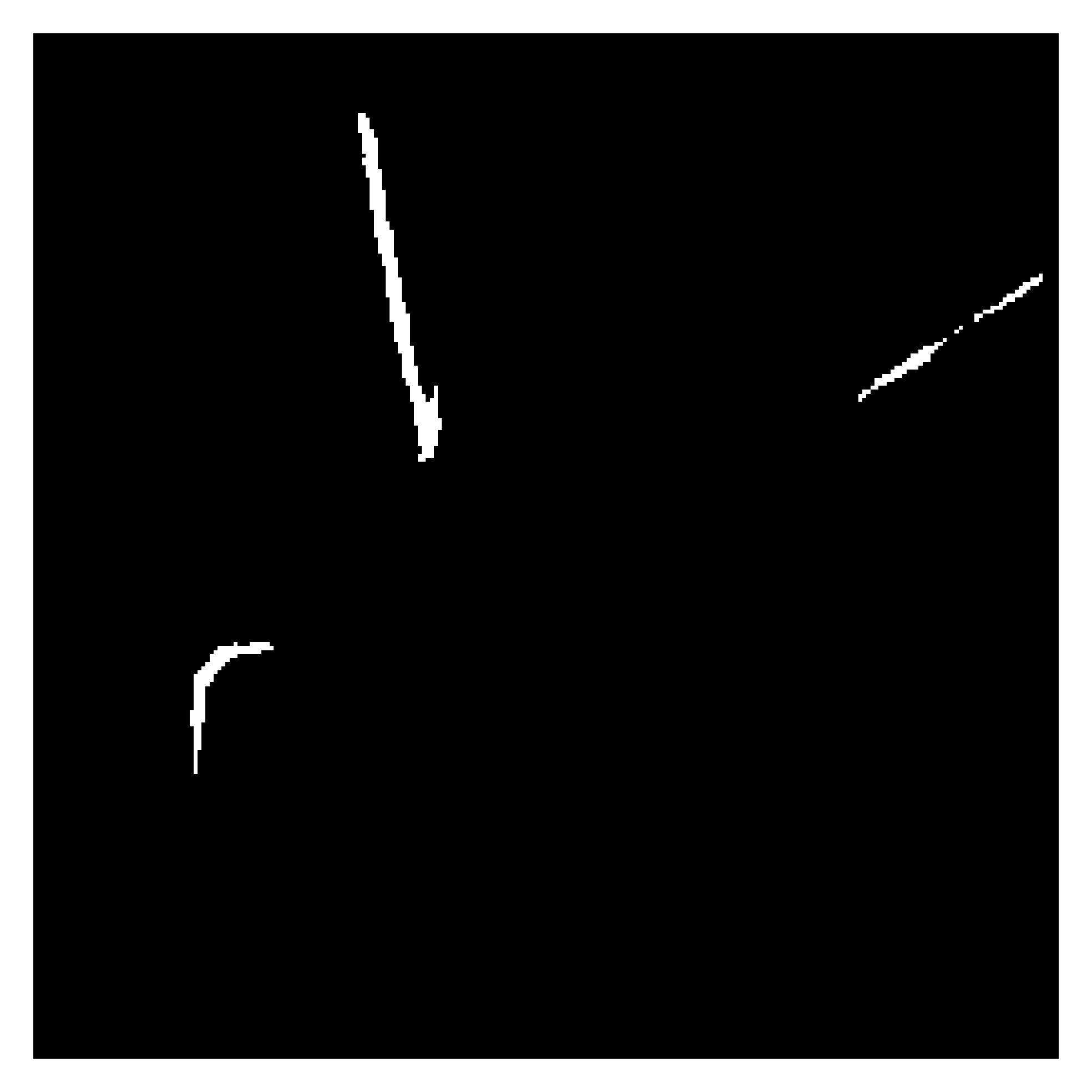}}
    \end{subfigure}
    \begin{subfigure}[b]{0.1\textwidth}
        \adjustbox{trim=10 10 10 10,clip,width=1.6cm,height=1.6cm}{\includegraphics{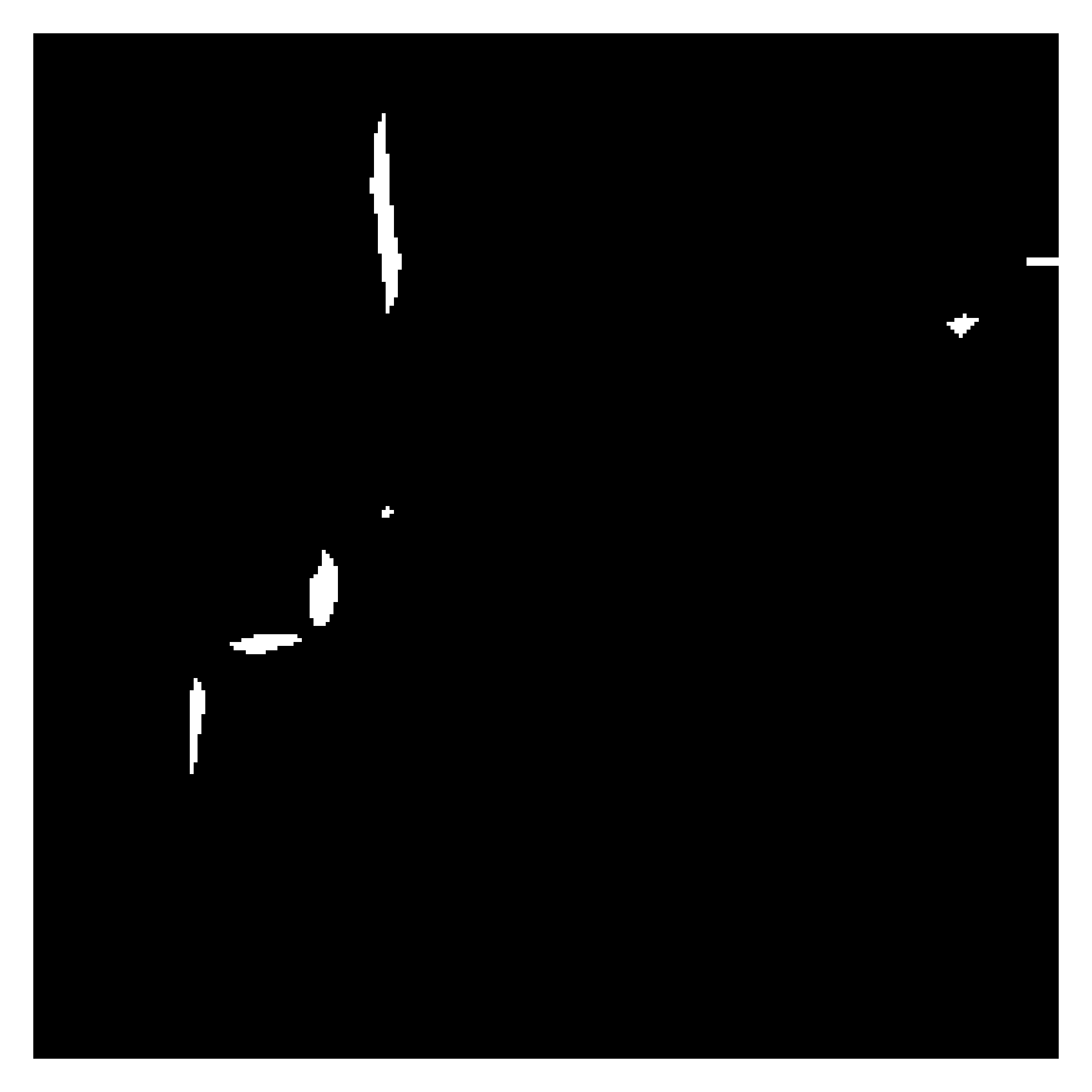}}
    \end{subfigure}
    \begin{subfigure}[b]{0.1\textwidth}
       \adjustbox{trim=10 10 10 10,clip,width=1.6cm,height=1.6cm}{\includegraphics{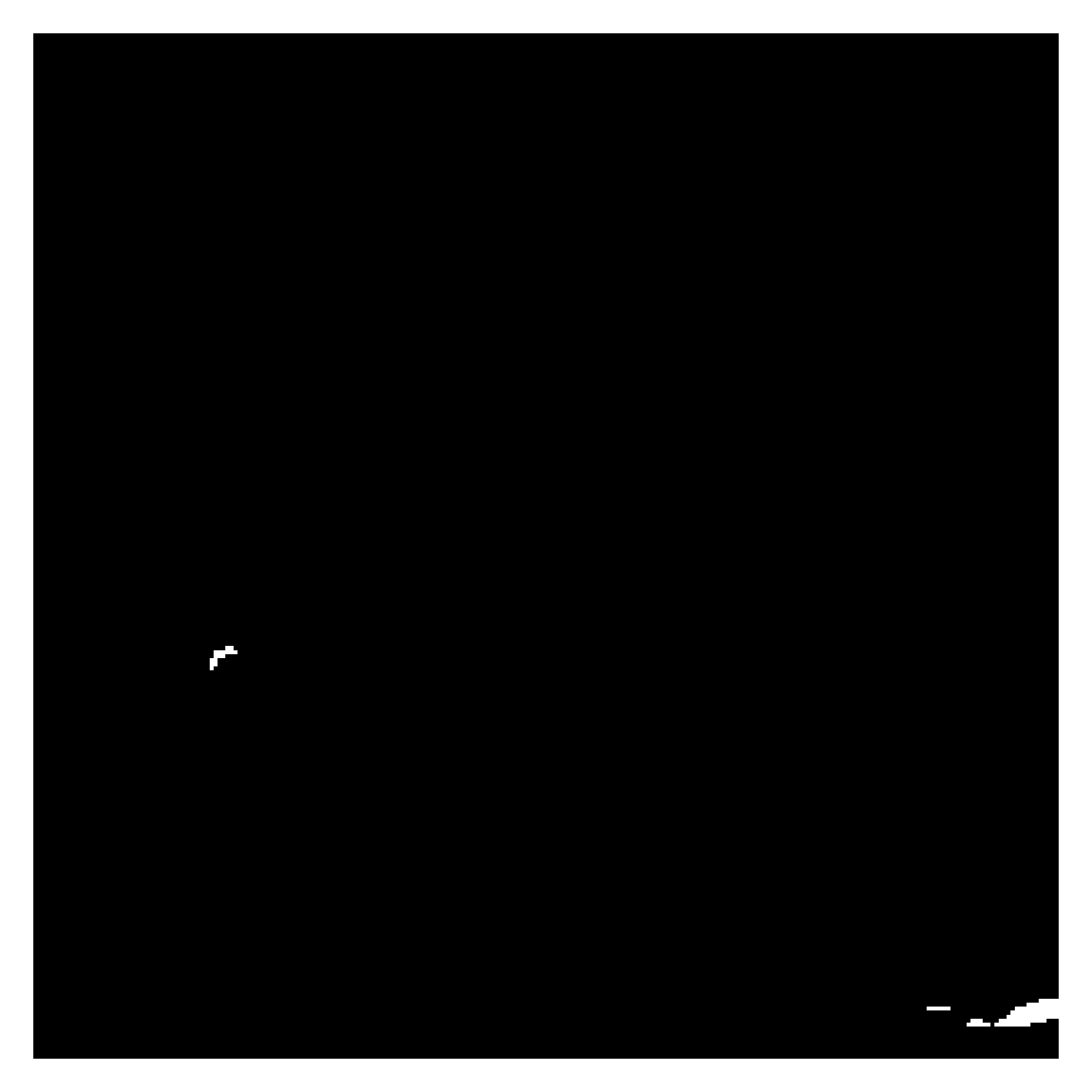}}
    \end{subfigure}
    \begin{subfigure}[b]{0.1\textwidth}
        \adjustbox{trim=10 10 10 10,clip,width=1.6cm,height=1.6cm}{\includegraphics{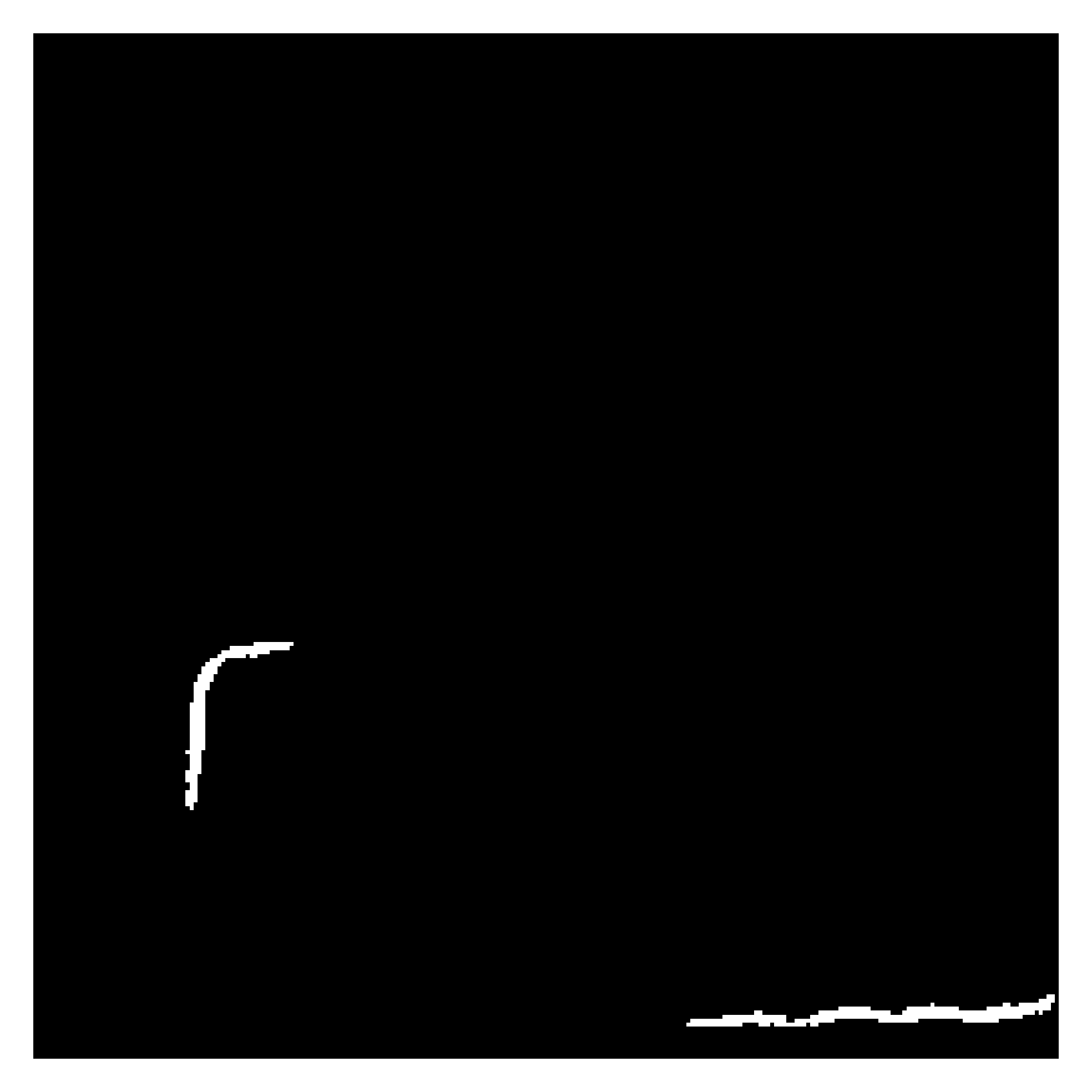}}
    \end{subfigure}
    \begin{subfigure}[b]{0.1\textwidth}
        \adjustbox{trim=10 10 10 10,clip,width=1.6cm,height=1.6cm}{\includegraphics{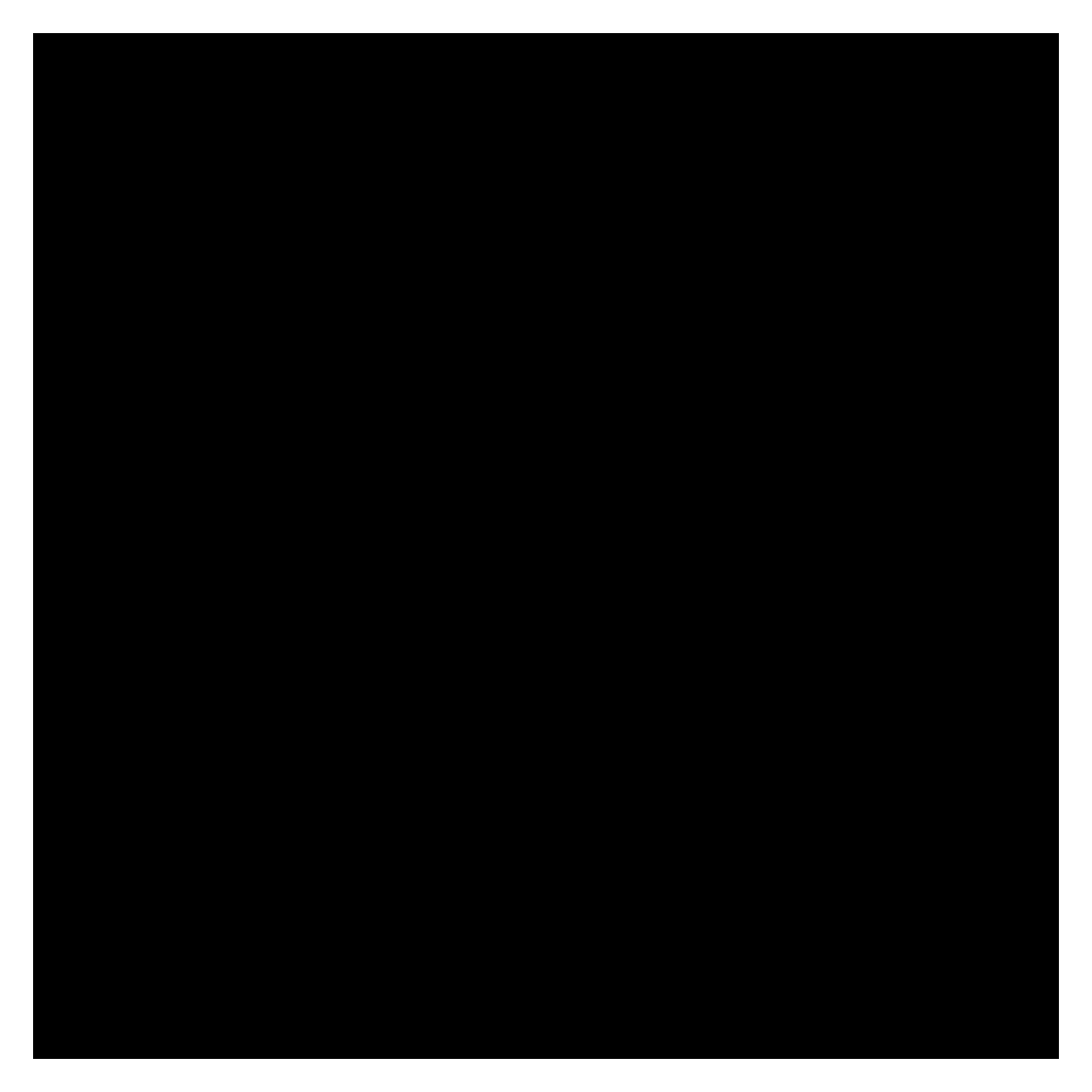}}
    \end{subfigure}
    
    \vspace{0.2cm}
    
    \begin{subfigure}[b]{0.09\textwidth} 
        \subcaption*{TopoDS} 
    \end{subfigure}
    \begin{subfigure}[b]{0.1\textwidth}
        \adjustbox{trim=10 10 10 10,clip,width=1.6cm,height=1.6cm}{\includegraphics{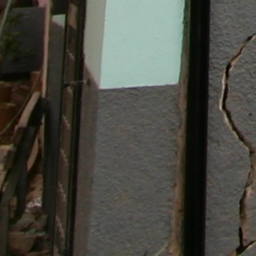}}
    \end{subfigure}
    \begin{subfigure}[b]{0.1\textwidth}
        \adjustbox{trim=10 10 10 10,clip,width=1.6cm,height=1.6cm}{\includegraphics{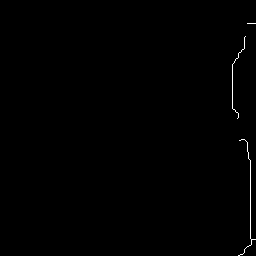}}
    \end{subfigure}
    \begin{subfigure}[b]{0.1\textwidth}
       \adjustbox{trim=10 10 10 10,clip,width=1.6cm,height=1.6cm}{\includegraphics{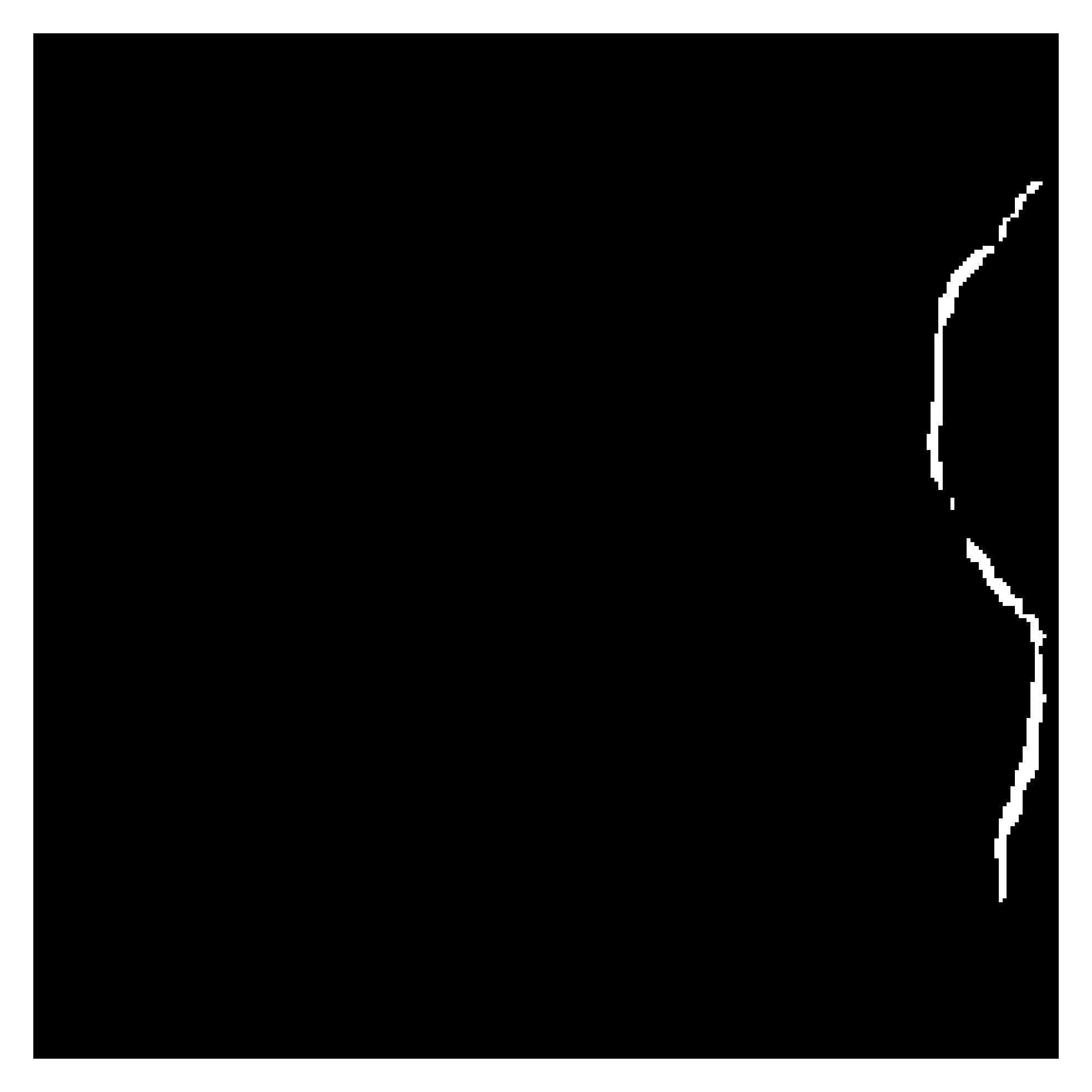}}
    \end{subfigure}
    \begin{subfigure}[b]{0.1\textwidth}
        \adjustbox{trim=10 10 10 10,clip,width=1.6cm,height=1.6cm}{\includegraphics{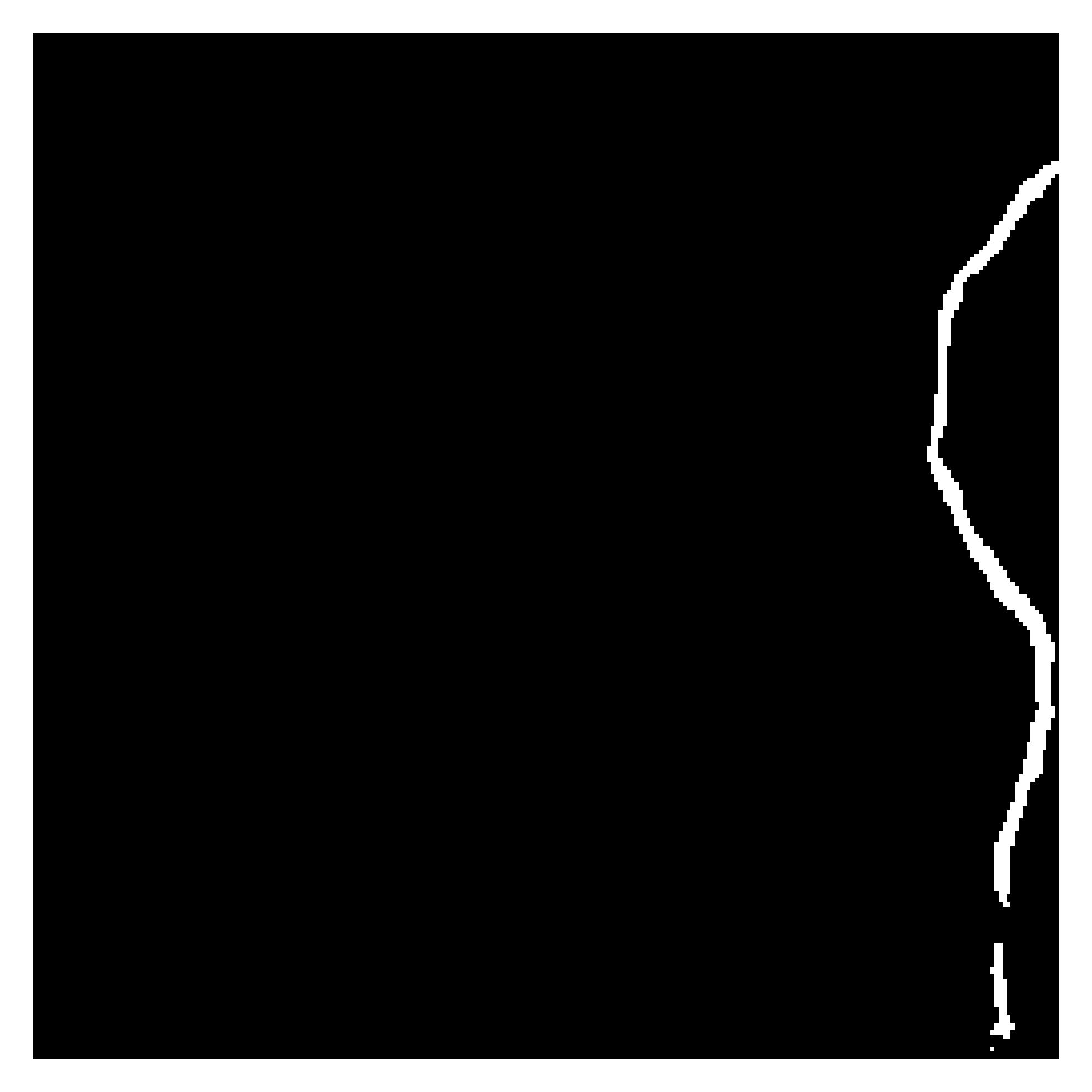}}
    \end{subfigure}
    \begin{subfigure}[b]{0.1\textwidth}
        \adjustbox{trim=10 10 10 10,clip,width=1.6cm,height=1.6cm}{\includegraphics{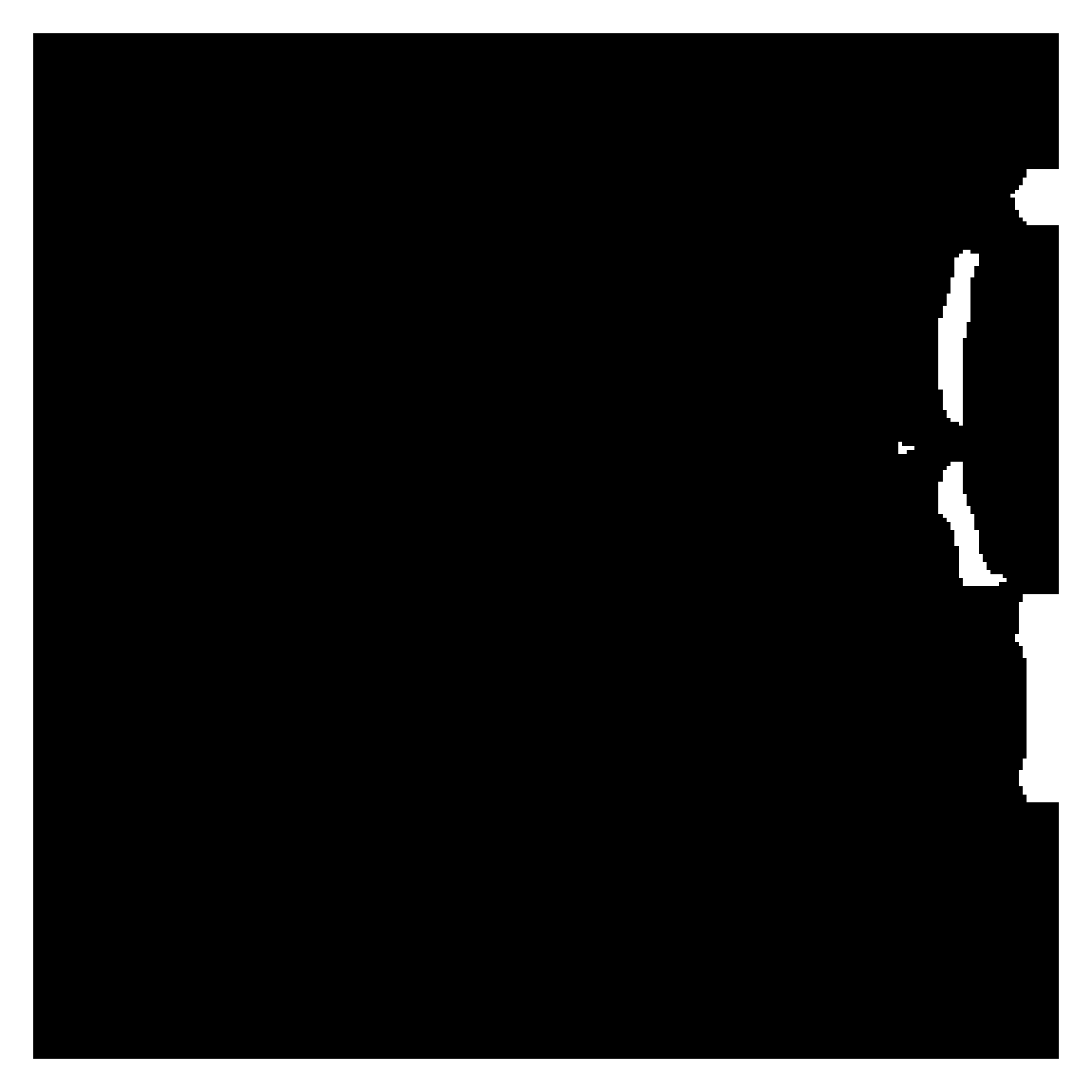}}
    \end{subfigure}
    \begin{subfigure}[b]{0.1\textwidth}
       \adjustbox{trim=10 10 10 10,clip,width=1.6cm,height=1.6cm}{\includegraphics{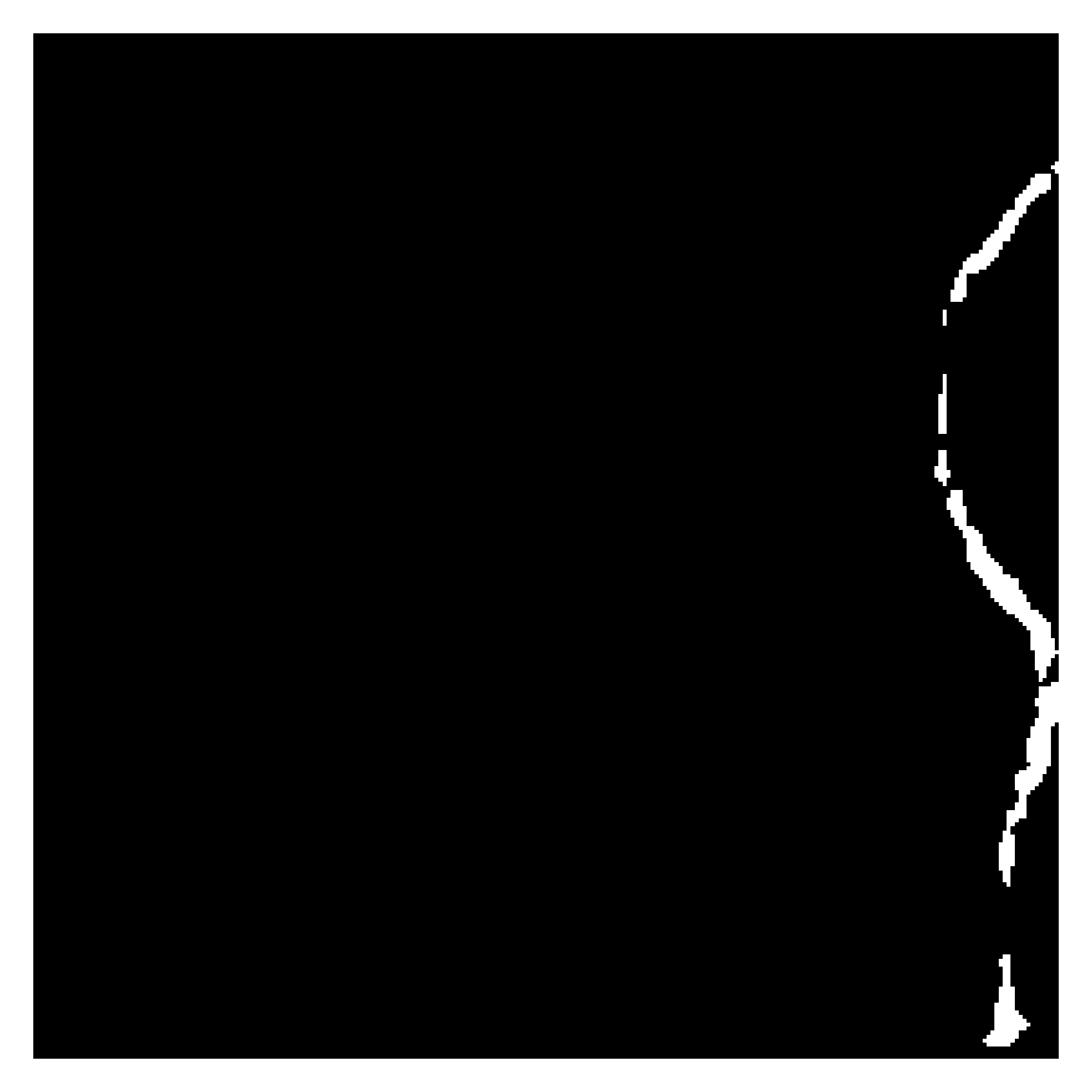}}
    \end{subfigure}
    \begin{subfigure}[b]{0.1\textwidth}
        \adjustbox{trim=10 10 10 10,clip,width=1.6cm,height=1.6cm}{\includegraphics{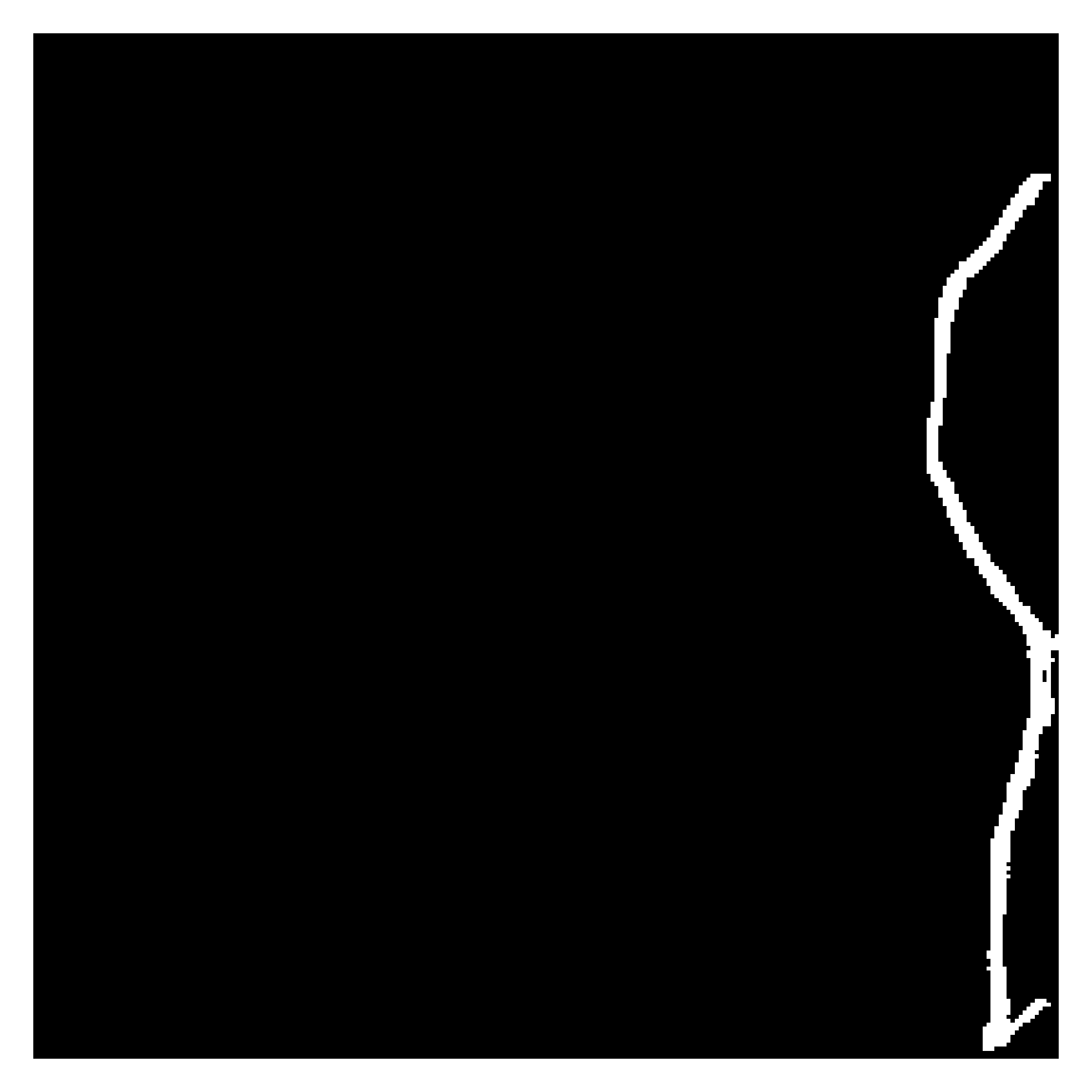}}
    \end{subfigure}
    \begin{subfigure}[b]{0.1\textwidth}
        \adjustbox{trim=10 10 10 10,clip,width=1.6cm,height=1.6cm}{\includegraphics{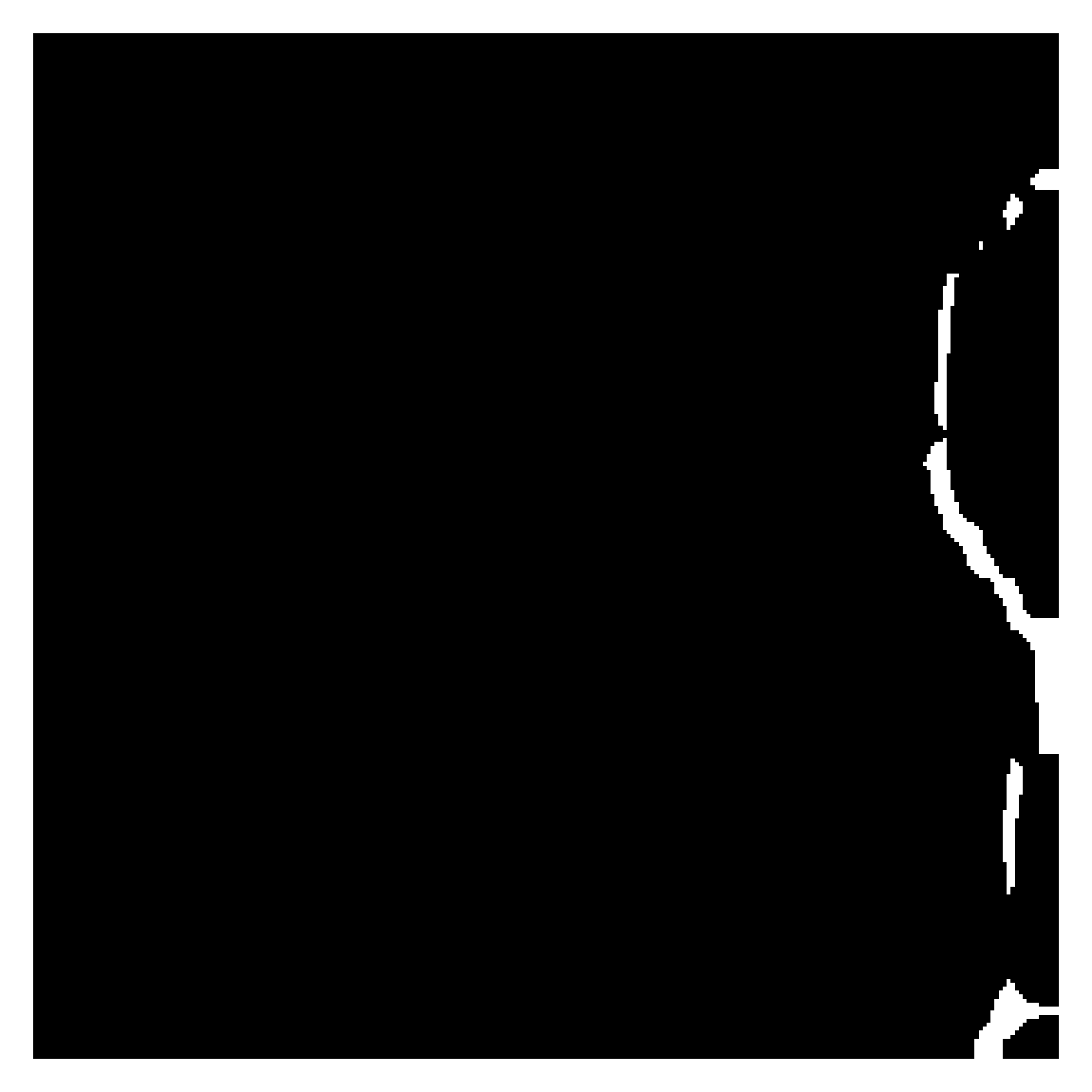}}
    \end{subfigure}
    
    \caption{Qualitative comparison of the binary mask predictions of the fine-tuned segmentation models.}
    \label{fig:qualitative_comparison}
\end{figure}

Qualitative assessments were also conducted. As illustrated in Figure \ref{fig:qualitative_comparison}, SAC predicts more accurate segmentation masks, showing fewer false positives and improved crack continuity, which are crucial for subsequent crack measurement and structural assessment.

\begin{table}[h]
\centering
\caption{Performance results of the full-tuned segmentation models on OmniCrack30k test set.}
\label{table:fulltuning results}
\small
\renewcommand{\arraystretch}{1.25}
\begin{tabular}{l l l l l l l}
    \hline\hline
    \multicolumn{1}{c}{Model} & 
    \multicolumn{1}{c}{\parbox[t]{2cm}{\# Tunable \\Parameters}} & 
    \multicolumn{1}{c}{\parbox[t]{2cm}{Training Time\\(min/it)}} & 
    \multicolumn{1}{c}{\parbox[t]{1.5cm}{Precision \\(\%)}} &
    \multicolumn{1}{c}{\parbox[t]{1.5cm}{Recall (\%)}} & 
    \multicolumn{1}{c}{\parbox[t]{1.5cm}{F1-Score (\%)}} & 
    \multicolumn{1}{c}{\parbox[t]{1.5cm}{IoU \\(\%)}} \\
    \hline\hline
    DeepCrackZ & 31M & 82.8 & 61.95 & 54.91 & 58.19 & 41.05 \\
    DeepCrackL & 15M & 7.1 & 67.54 & 44.87 & 53.9 & 36.9  \\
    CrackFormer & 5M & 35.1 & 48.63 & 48.30 & 48.45 & 31.98 \\
    \hline
    \textbf{SegFormer} & \textbf{3.7M} & \textbf{5.1} & 58.81 & \textbf{61.26} & \textbf{59.98} & \textbf{42.85} \\
    U-Net & 32.5M & 6.8 & \textbf{71} & 43.99 & 54.28 & 37.27 \\
    DeepLabv3+-ResNet50 & 42M & 12.9 & 57.93 & 52.91 & 55.27 & 38.21 \\
    DeepLav3+-ResNet101 & 61M & 18.0 & 60.03 & 53.44 & 56.52 & 39.41\\
    \hline
\end{tabular}
\normalsize
\end{table}

\begin{table}[h]
\centering
\caption{Performance results of the fine-tuned segmentation models on OmniCrack30k test set. DeepCrackZ does not include any normalization layers.}
\label{table:fine-tuning results}
\small
\renewcommand{\arraystretch}{1.25}
\begin{tabular}{p{2.5cm} l l l l l l l}
    \hline\hline
    \multicolumn{1}{c}{Model} &  
    \multicolumn{1}{c}{\parbox[t]{1.5cm}{Tuned \\ Params}} &
    \multicolumn{1}{c}{\parbox[t]{1.5cm}{\# Tuned \\Params}} &
    \multicolumn{1}{c}{\parbox[t]{2cm}{Training Time\\(min/it)}} & 
    \multicolumn{1}{c}{\parbox[t]{1.5cm}{Precision \\(\%)}} &
    \multicolumn{1}{c}{\parbox[t]{1.5cm}{Recall (\%)}} & 
    \multicolumn{1}{c}{\parbox[t]{1.5cm}{F1-Score (\%)}} & 
    \multicolumn{1}{c}{\parbox[t]{1cm}{IoU \\(\%)}} \\
    \hline
    DeepCrackZ & - $^{\ast}$ & - & - & - & - & - & - \\
    DeepCrackL & BN & 8K  & 4.7 & 55.07 & 38.28 & 45.16 & 29.17\\
    CrackFormer & GN & 40K & 30.8 & 38.84 & 50.5 & 43.9 & 28.13 \\
    \hline
    SegFormer & BN \& LN & 7.6K & 4.3 & 55.18 & 50.69 & 52.82 & 35.91 \\
    U-Net & BN & 55K & 4.6 & 61.97 & 49.17 & 54.82 & 37.77 \\
    DeepLabv3+-ResNet50 & BN & 57K & 7.5 & 52.62 & 53.28 & 52.93 & 36.01 \\
    DeepLav3+-ResNet101 & BN & 110K & 12.0 & 50.59 & 58.14 & 54.09 & 37.09\\
    \textbf{SAC} & \textbf{LN} & \textbf{41K} & \textbf{12.3} & \textbf{64.16} & \textbf{58.6} & \textbf{61.22} & \textbf{44.13} \\
    \hline
    \multicolumn{8}{l}{\footnotesize BN: Batch Normalization, GN: Group Normalization, LN: Layer Normalization}\\
    \hline\hline
\end{tabular}
\normalsize
\end{table}

\subsection{Zero-Shot Evaluations}
\begin{figure}[h]
    \centering
    \begin{subfigure}{0.49\textwidth}
        \centering
        \includegraphics[width=\linewidth]{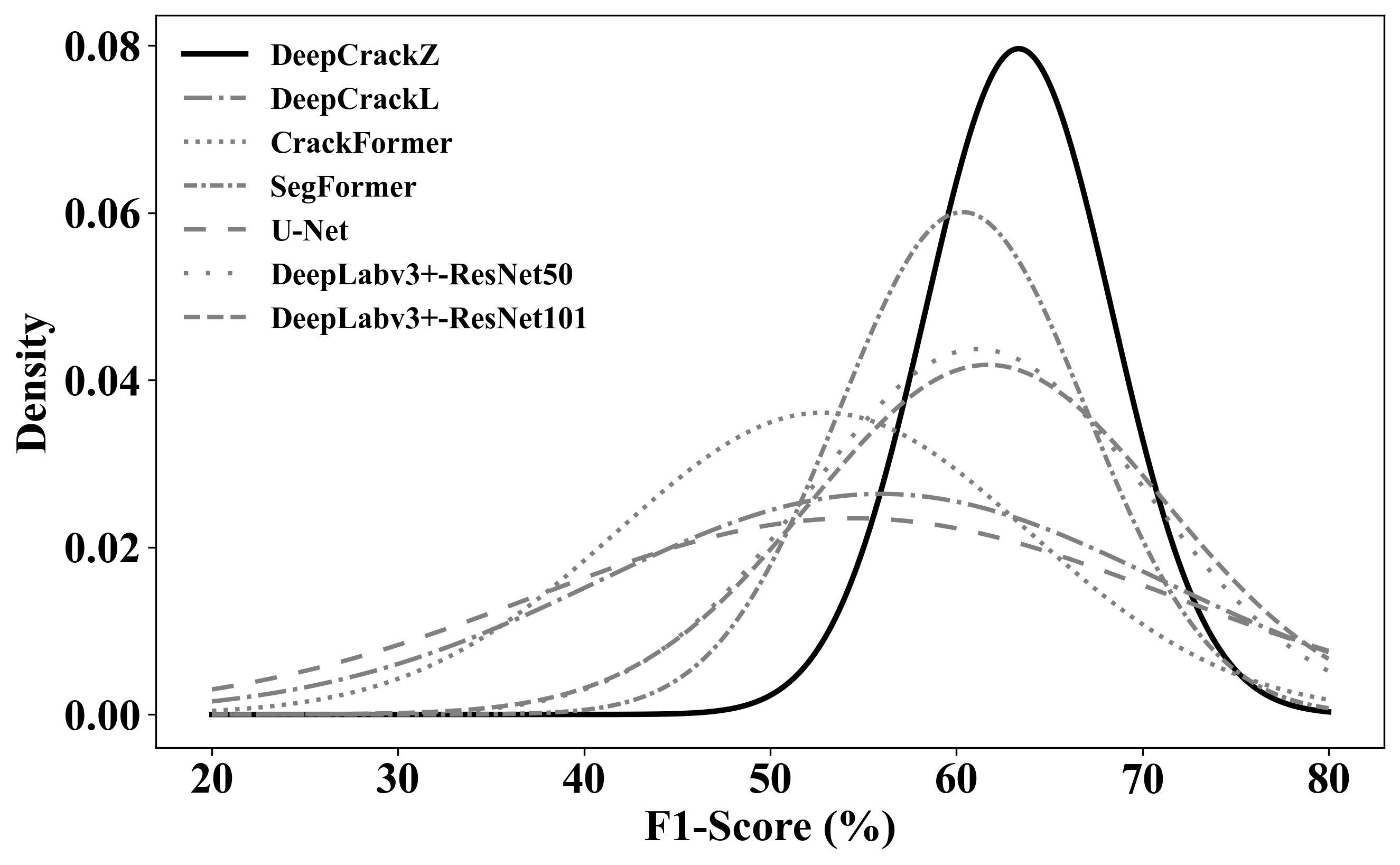}
        \caption*{(a)}
    \end{subfigure}
    \begin{subfigure}{0.49\textwidth}
        \centering
        \includegraphics[width=\linewidth]{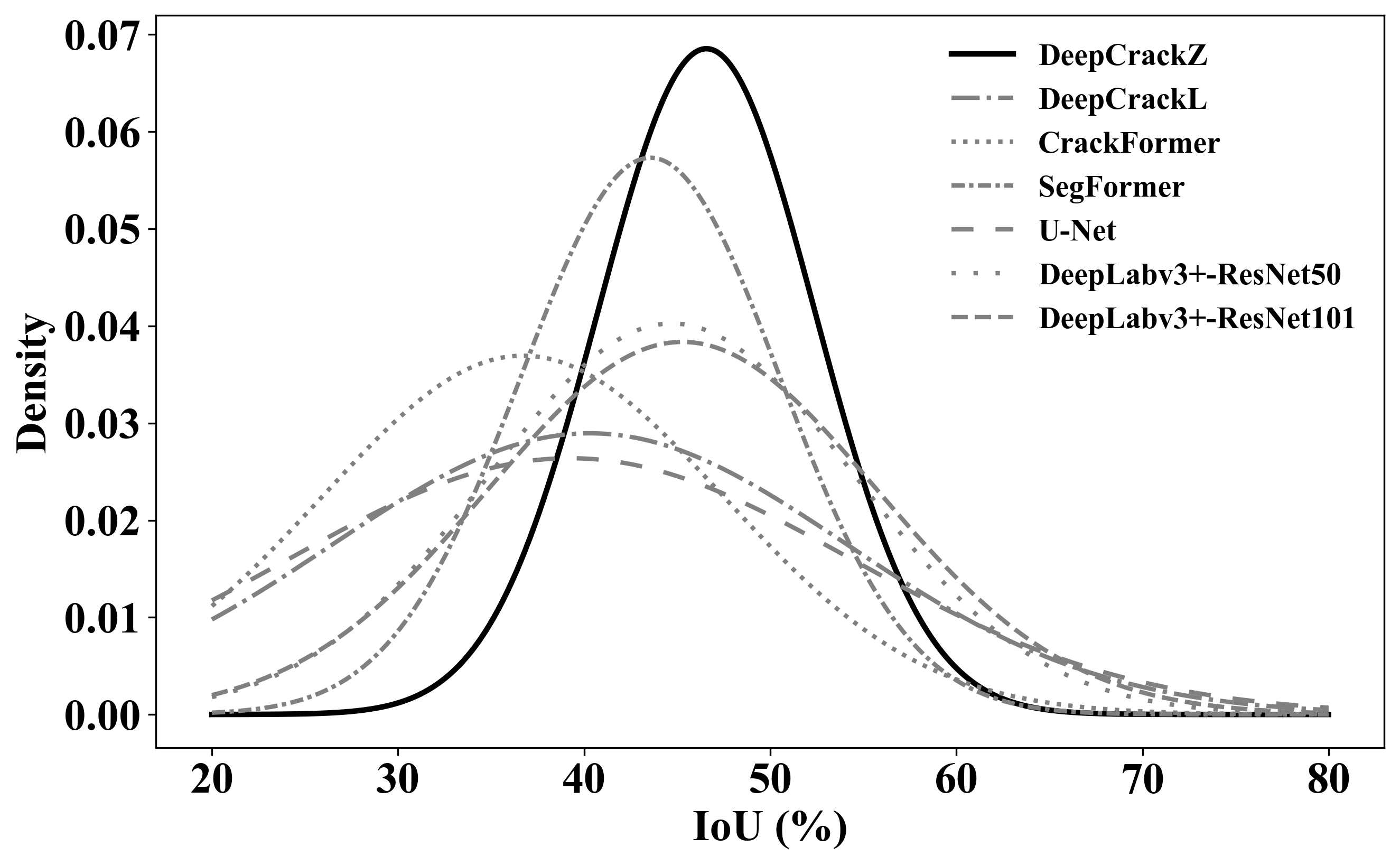}
        \caption*{(b)}
    \end{subfigure}

    \vspace{0.05cm}

    \begin{subfigure}{0.49\textwidth}
        \centering
        \includegraphics[width=\linewidth]{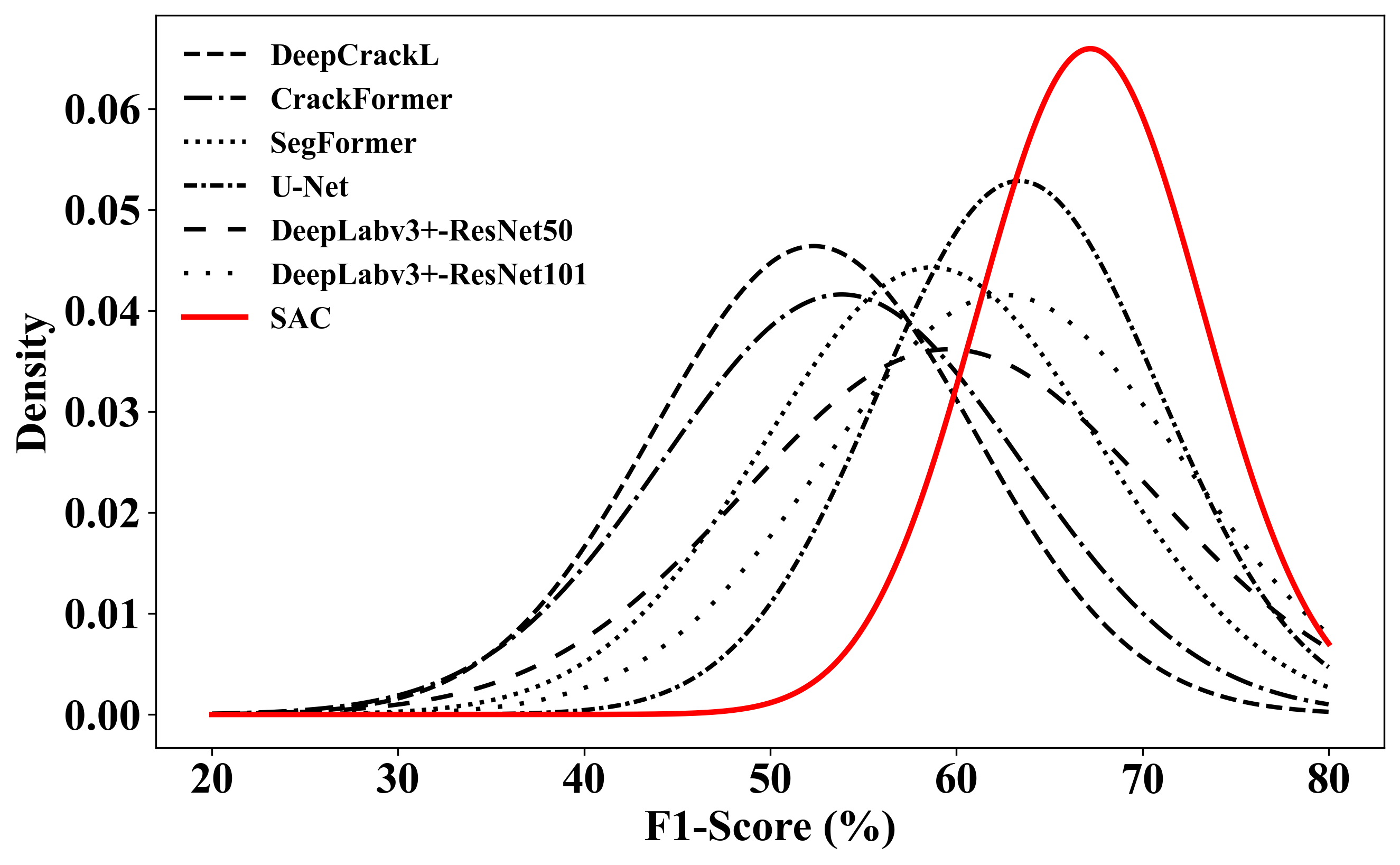}
        \caption*{(c)}
    \end{subfigure}
    \begin{subfigure}{0.49\textwidth}
        \centering
        \includegraphics[width=\linewidth]{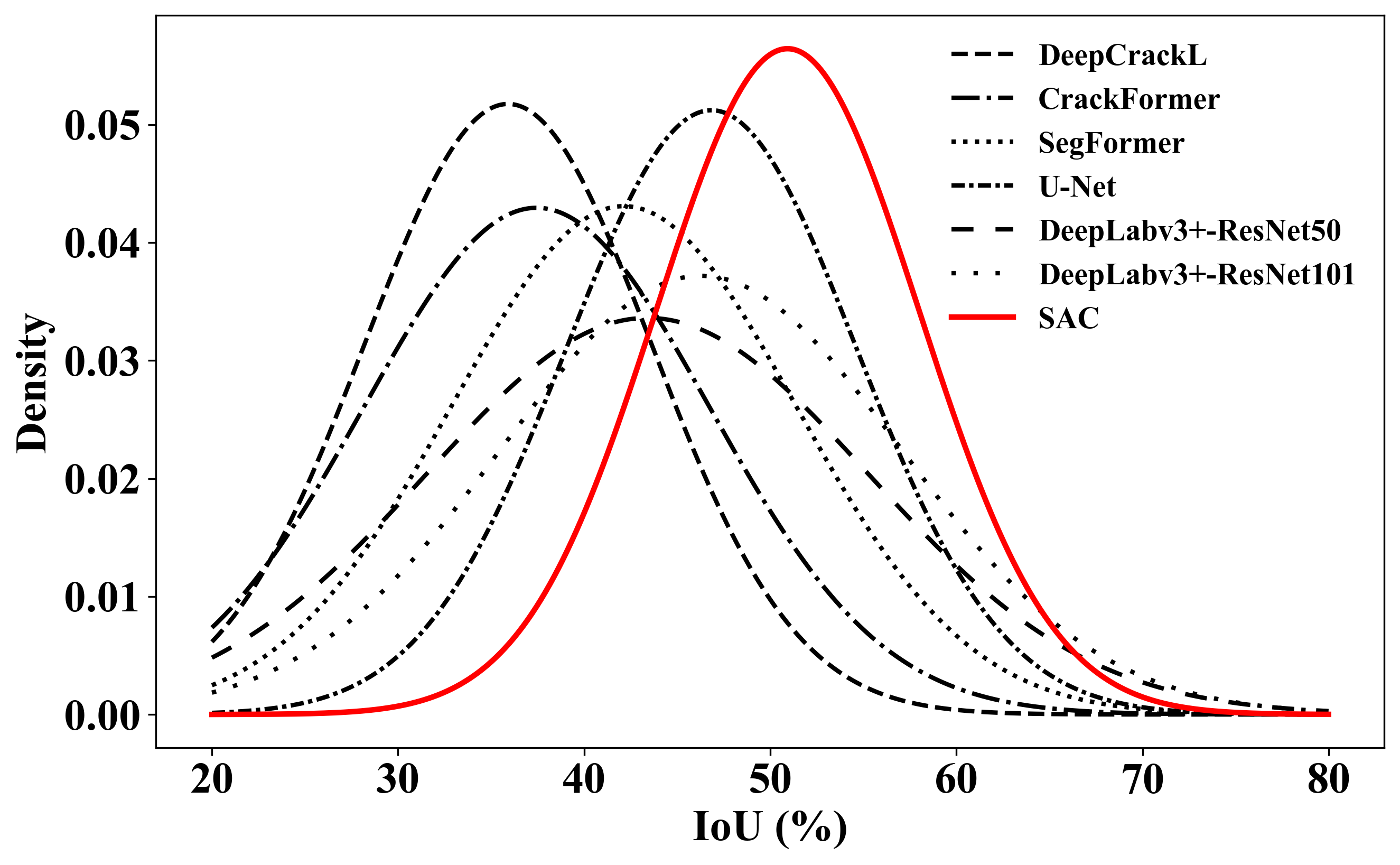}
        \caption*{(d)}
    \end{subfigure}

    \caption{Normal Distributions of the zero-shot evaluation of the segmentation models across three datasets. The x-axis represents the average performance in terms of IoU or F1-score. (a) and (b) correspond to the full-tuned models and (c) and (d) depict the performance of the fine-tuned models. The figures indicate that the proposed fine-tuning method enhances the zero-shot performance of the models.}
    \label{fig:zeroshot}
\end{figure}

The results of the zero-shot evaluations on the three datasets, Road420, Facade390, and Concrete3k, using full and fine-tuned segmentation methods are presented in Tables \ref{table:Fulltuning ZeroShot Results} and \ref{table:Finetuning ZeroShot Results}, respectively. The SAC model showed superior performance in comparison to the other models, achieving the highest F1-score (67.20\%) and IoU (50.93\%), along with the lowest standard deviation in both metrics (6.05 and 7.07, respectively). This highlights its strong generalization capability and robustness across various scenes and environments. Furthermore, as illustrated in Figure \ref{fig:zeroshot}, the performance of the methods improves with the fine-tuning method, showing that optimizing the normalization parameters not only achieves comparable performances but also enhances generalization.
Moreover, it can be seen that while SegFormer achieved the highest performance in full tuning setting, DeepLabv3+-ResNet101 achieved the highest zero shot performance across the three datasets with an average F1-score of 61.70\% (standard deviation: 9.54) and an average IoU of 45.28\% (standard deviation: 10.40). This underscores the importance of generalization capability in segmentation models, specially for real-world applications where models must adapt to unseen conditions with minimal fine-tuning.

\begin{figure}[h]
    \captionsetup[subfigure]{labelformat=empty, font=footnotesize, justification=centering, position=top} 
    \centering
    \begin{subfigure}[b]{0.1\textwidth}
        \subcaption*{Input Image} 
        \adjustbox{trim=10 10 10 10,clip,width=1.6cm,height=1.6cm}{\includegraphics{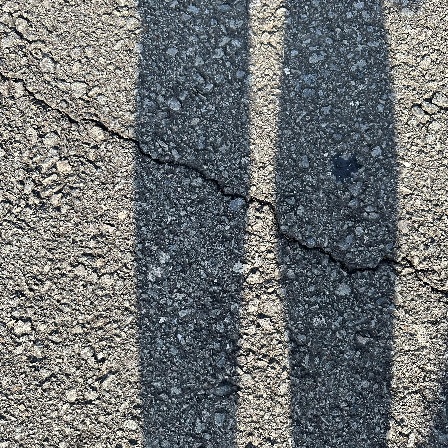}}
    \end{subfigure}
    \begin{subfigure}[b]{0.1\textwidth}
        \subcaption*{GroundTruth} 
        \adjustbox{trim=10 10 10 10,clip,width=1.6cm,height=1.6cm}{\includegraphics{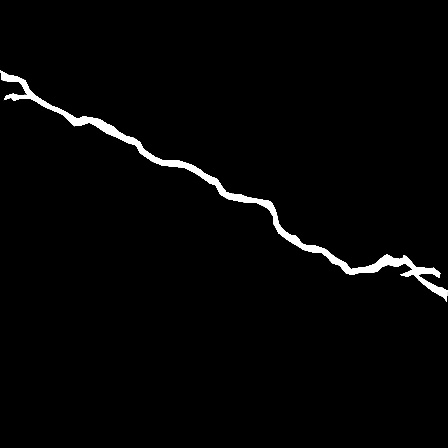}}
    \end{subfigure}  
    \begin{subfigure}[b]{0.1\textwidth}
        \subcaption*{SAC} 
        \adjustbox{trim=10 10 10 10,clip,width=1.6cm,height=1.6cm}{\includegraphics{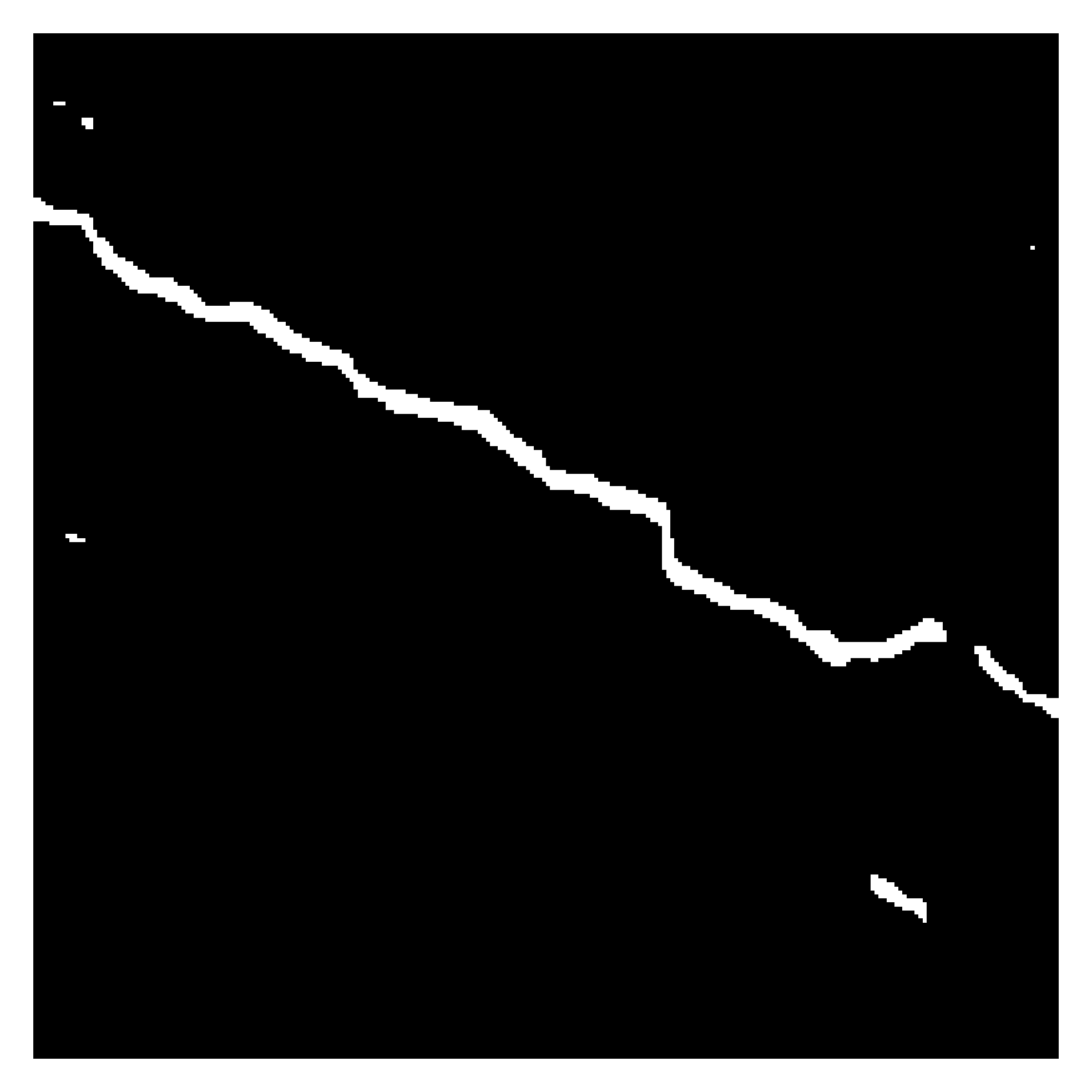}}
    \end{subfigure}
    \begin{subfigure}[b]{0.1\textwidth}
        \subcaption*{DeepCrackL} 
        \adjustbox{trim=10 10 10 10,clip,width=1.6cm,height=1.6cm}{\includegraphics{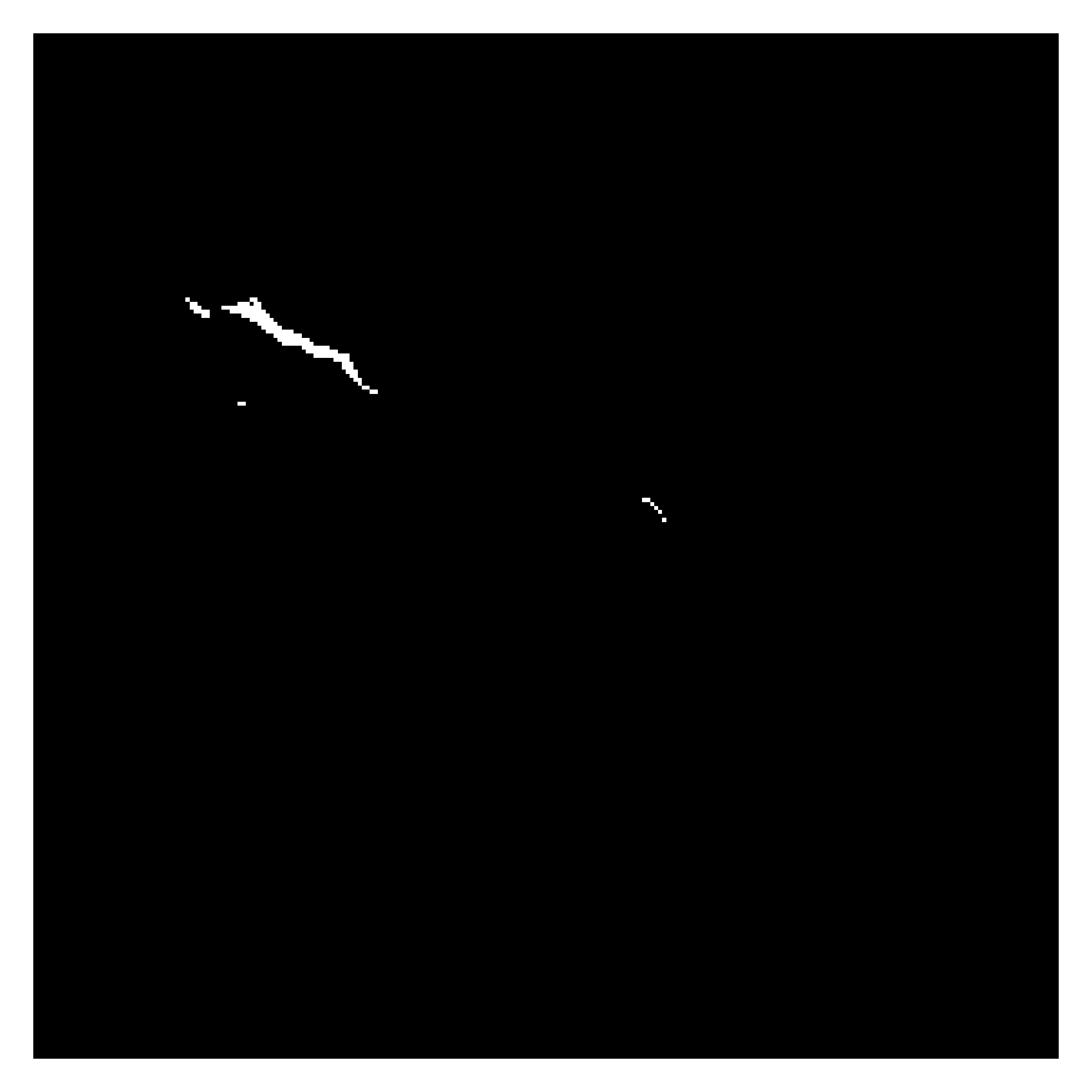}}
    \end{subfigure}
    \begin{subfigure}[b]{0.1\textwidth}
        \subcaption*{CrackFormer} 
        \adjustbox{trim=10 10 10 10,clip,width=1.6cm,height=1.6cm}{\includegraphics{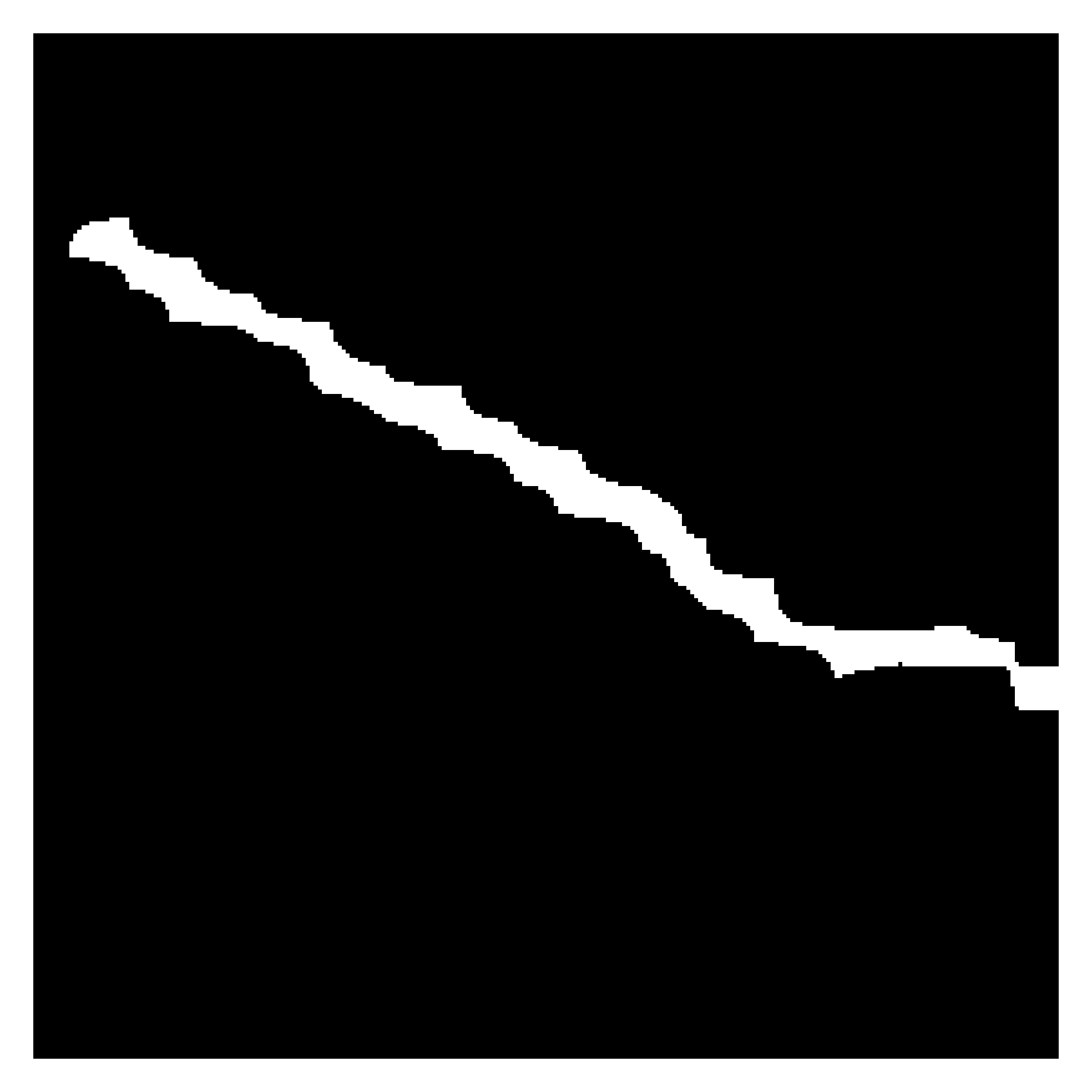}}
    \end{subfigure}
    \begin{subfigure}[b]{0.1\textwidth}
        \subcaption*{SegFormer} 
        \adjustbox{trim=10 10 10 10,clip,width=1.6cm,height=1.6cm}{\includegraphics{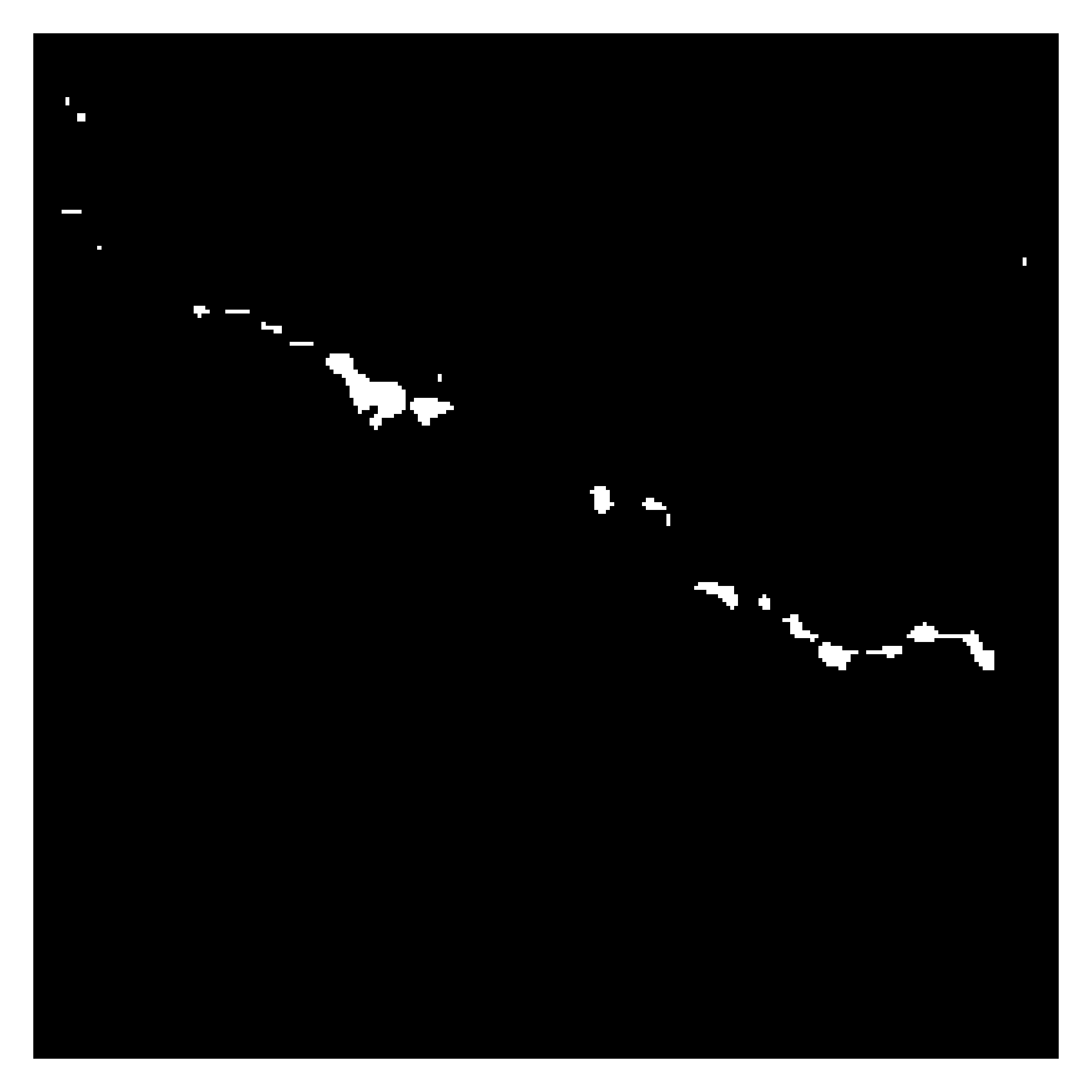}}
    \end{subfigure}
    \begin{subfigure}[b]{0.1\textwidth}
        \subcaption*{U-Net} 
        \adjustbox{trim=10 10 10 10,clip,width=1.6cm,height=1.6cm}{\includegraphics{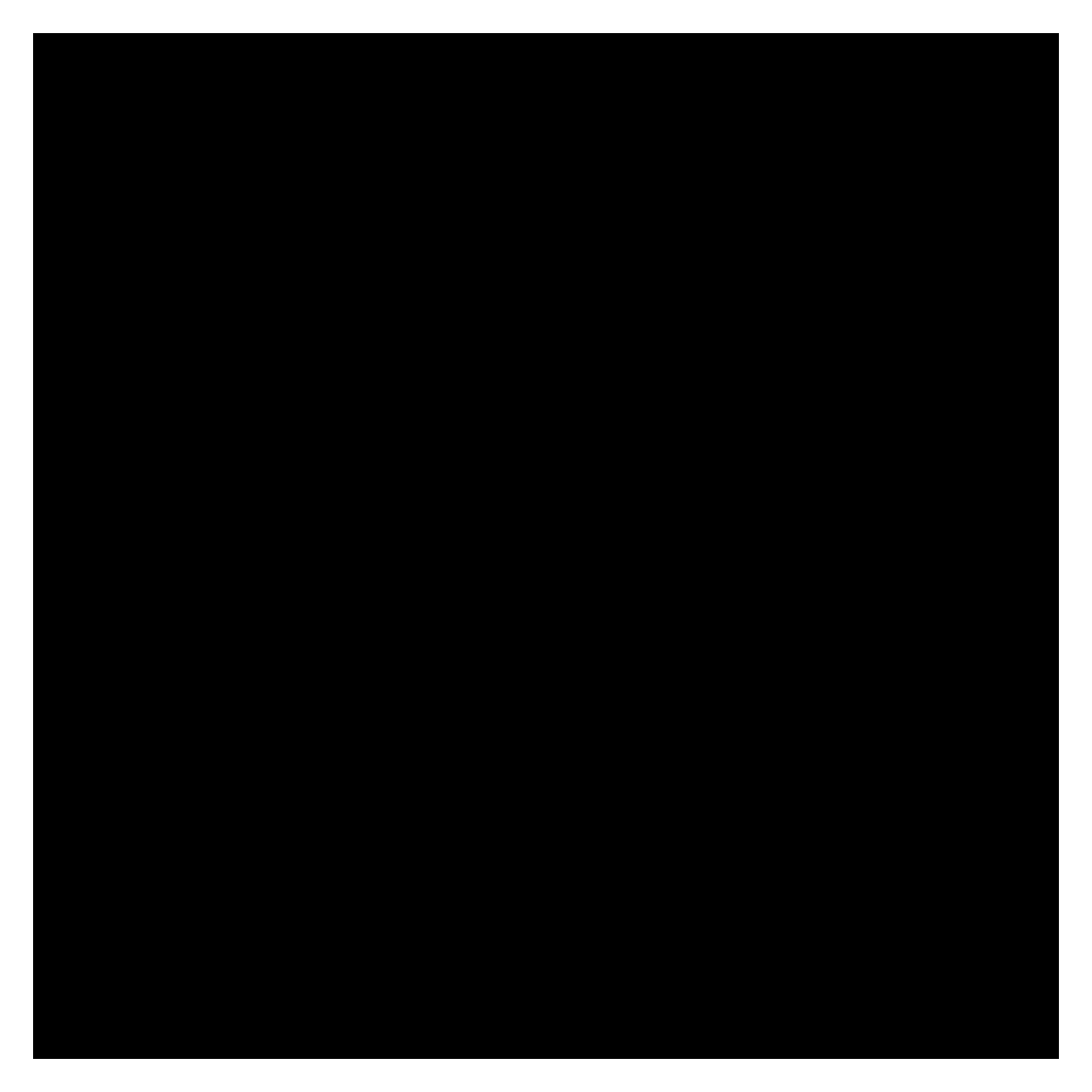}}
    \end{subfigure}
    \begin{subfigure}[b]{0.1\textwidth}
        \subcaption*{DeepLabv3+} 
        \adjustbox{trim=10 10 10 10,clip,width=1.6cm,height=1.6cm}{\includegraphics{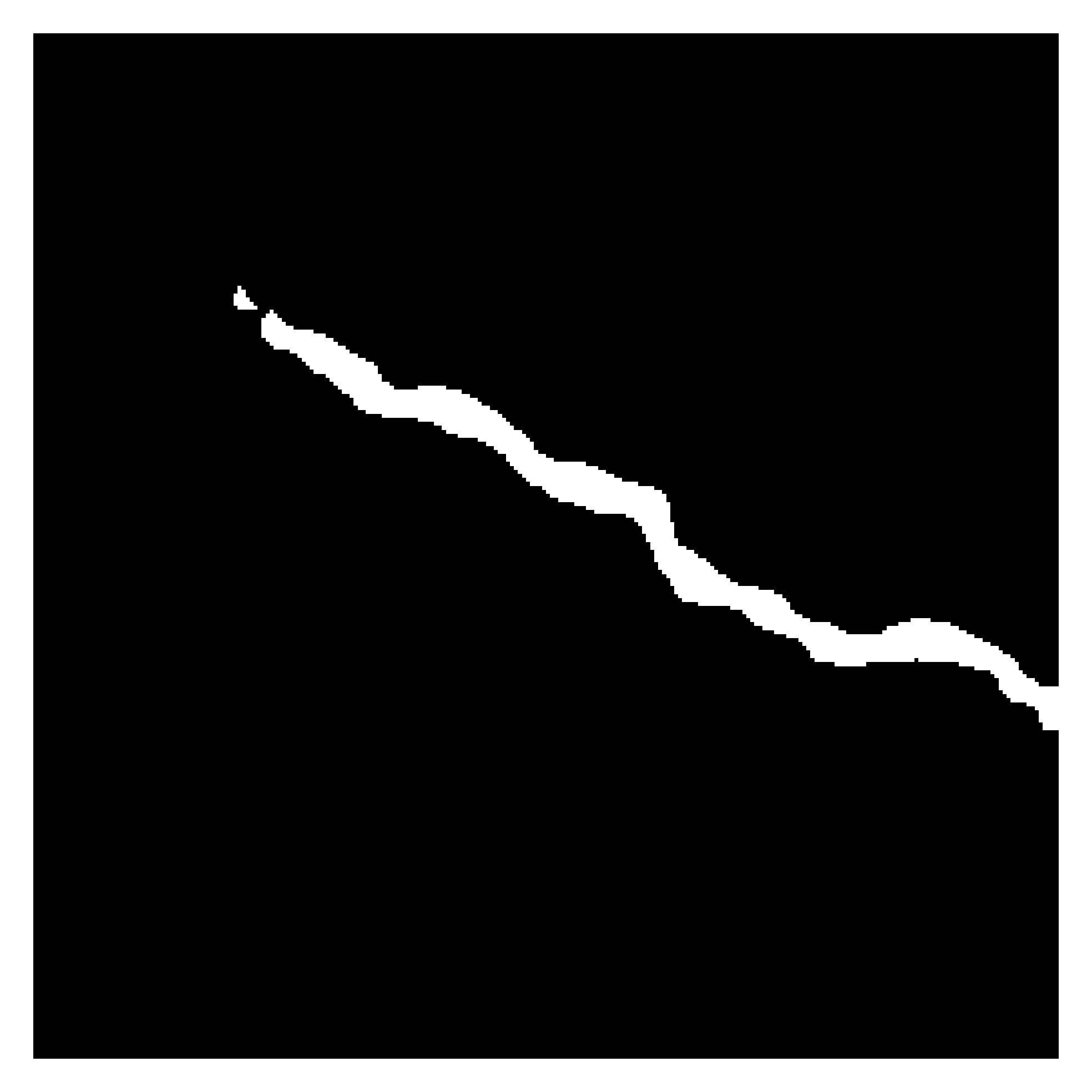}}
    \end{subfigure}

    \vspace{0.2cm}
    
    \begin{subfigure}[b]{0.1\textwidth}
        \adjustbox{trim=10 10 10 10,clip,width=1.6cm,height=1.6cm}{\includegraphics{figures/road420_zero_shot/1/2023_10_30_15_54_IMG_5916.jpg}}
    \end{subfigure}
    \begin{subfigure}[b]{0.1\textwidth}
        \adjustbox{trim=10 10 10 10,clip,width=1.6cm,height=1.6cm}{\includegraphics{figures/road420_zero_shot/1/2023_10_30_15_54_IMG_5916_mask.jpg}}
    \end{subfigure}  
    \begin{subfigure}[b]{0.1\textwidth}
        \adjustbox{trim=10 10 10 10,clip,width=1.6cm,height=1.6cm}{\includegraphics{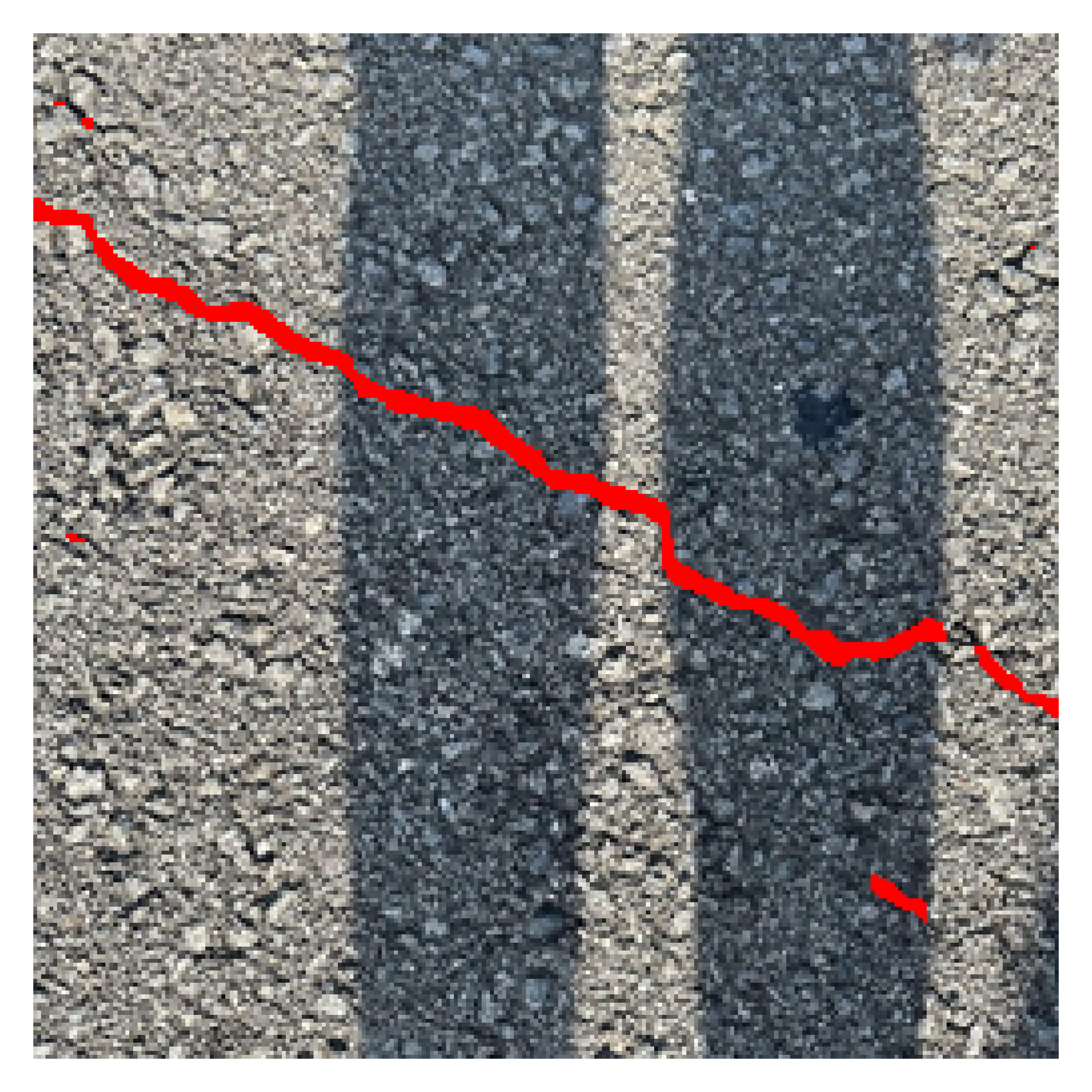}}
    \end{subfigure}
    \begin{subfigure}[b]{0.1\textwidth}
        \adjustbox{trim=10 10 10 10,clip,width=1.6cm,height=1.6cm}{\includegraphics{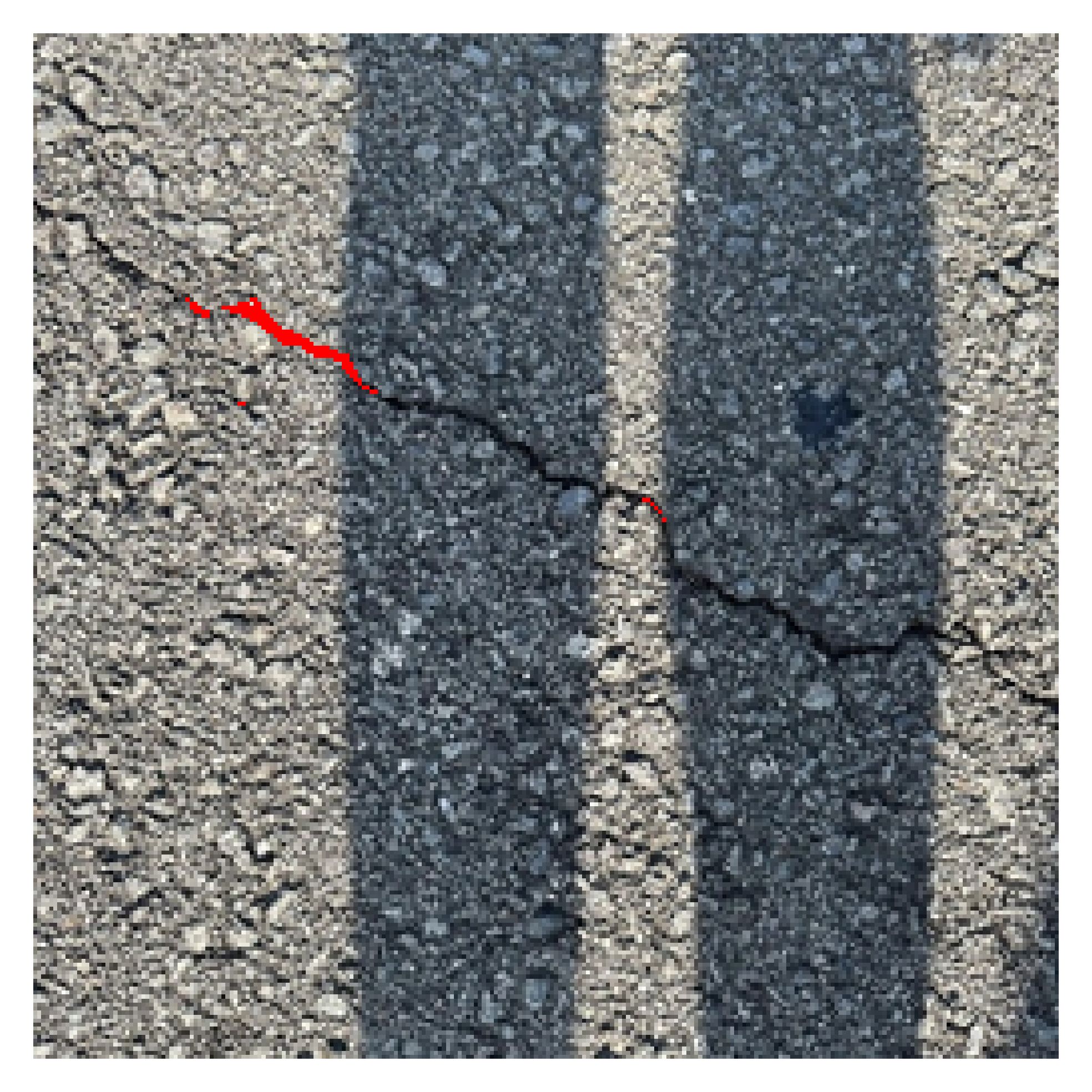}}
    \end{subfigure}
    \begin{subfigure}[b]{0.1\textwidth}
        \adjustbox{trim=10 10 10 10,clip,width=1.6cm,height=1.6cm}{\includegraphics{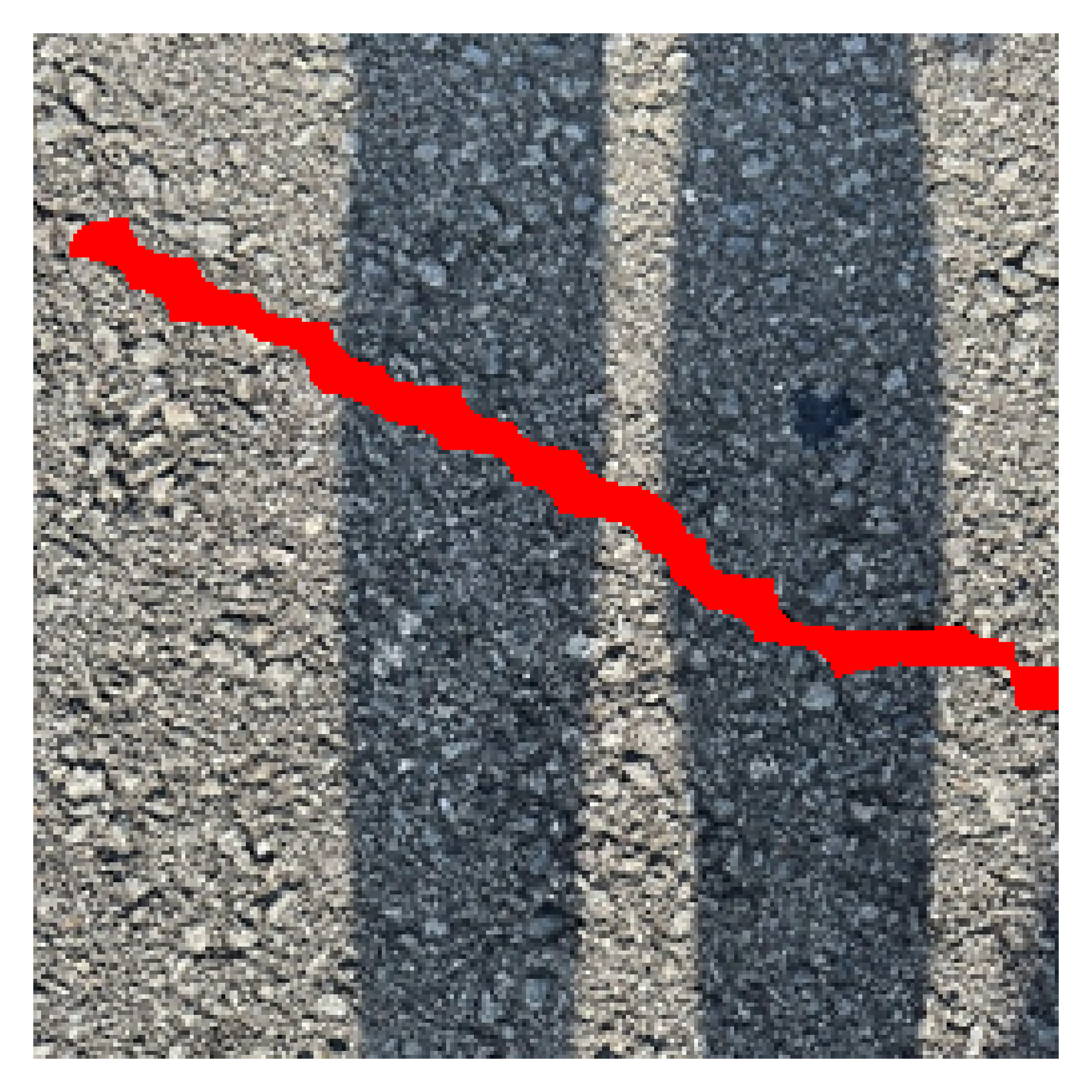}}
    \end{subfigure}
    \begin{subfigure}[b]{0.1\textwidth}
        \adjustbox{trim=10 10 10 10,clip,width=1.6cm,height=1.6cm}{\includegraphics{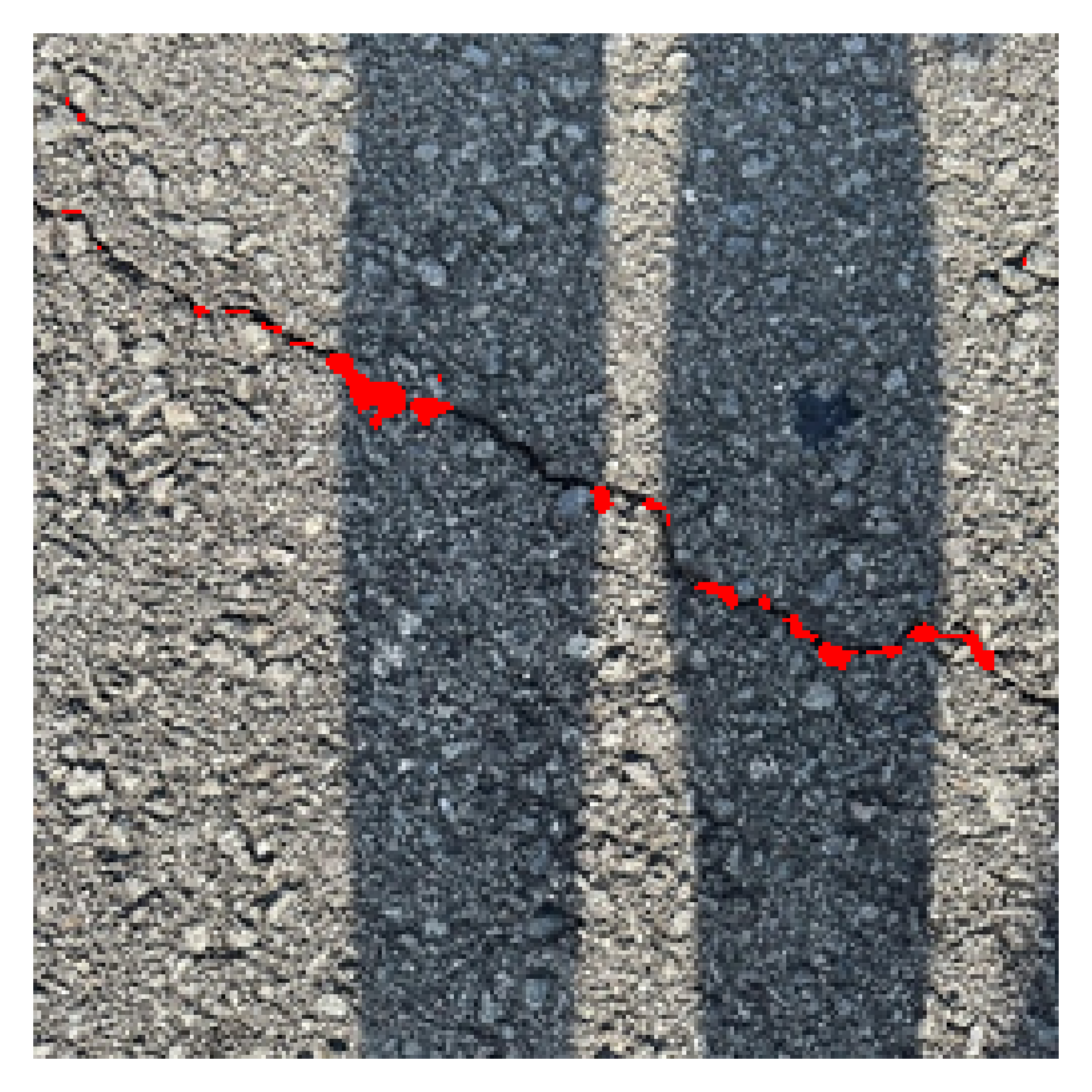}}
    \end{subfigure}
    \begin{subfigure}[b]{0.1\textwidth}
        \adjustbox{trim=10 10 10 10,clip,width=1.6cm,height=1.6cm}{\includegraphics{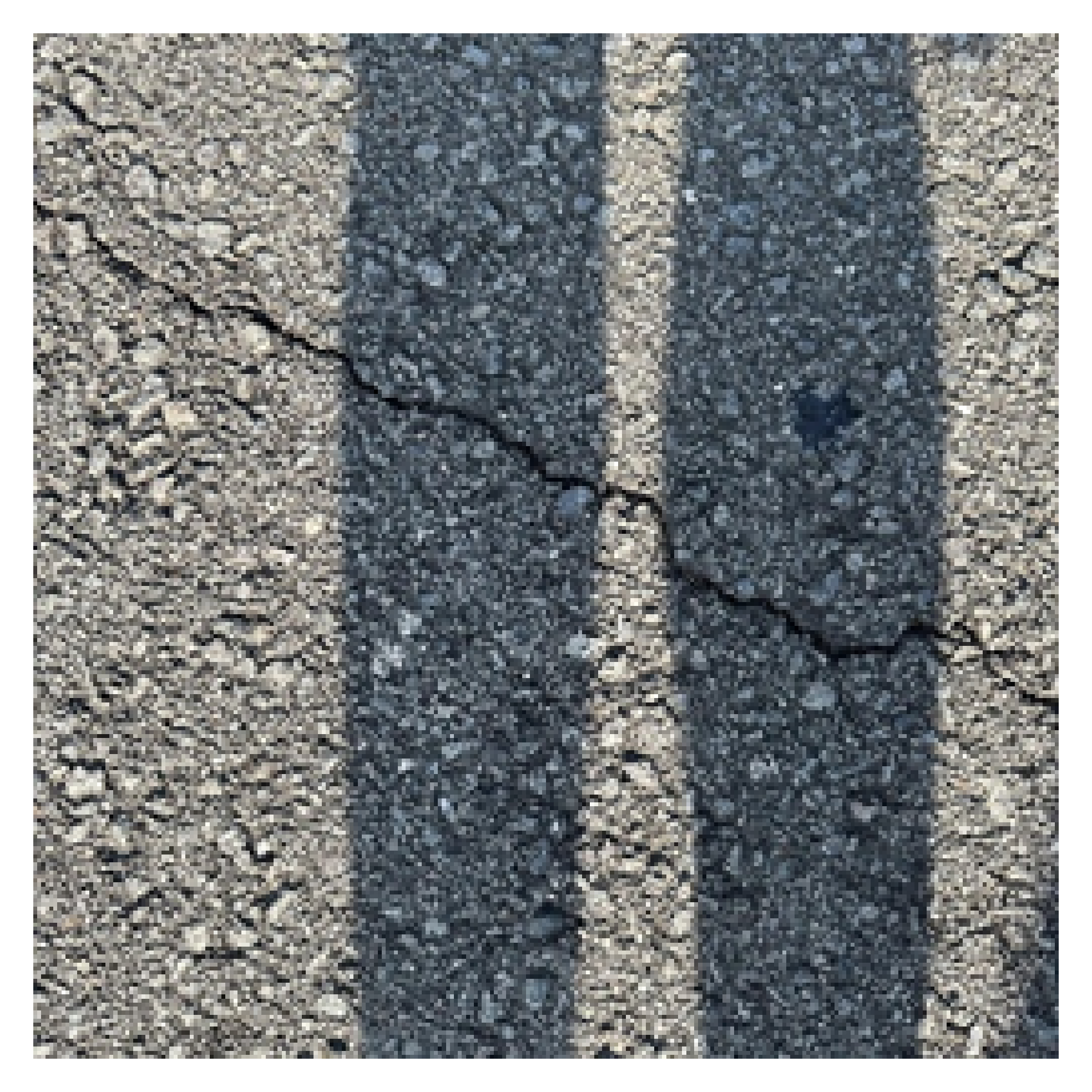}}
    \end{subfigure}
    \begin{subfigure}[b]{0.1\textwidth}
        \adjustbox{trim=10 10 10 10,clip,width=1.6cm,height=1.6cm}{\includegraphics{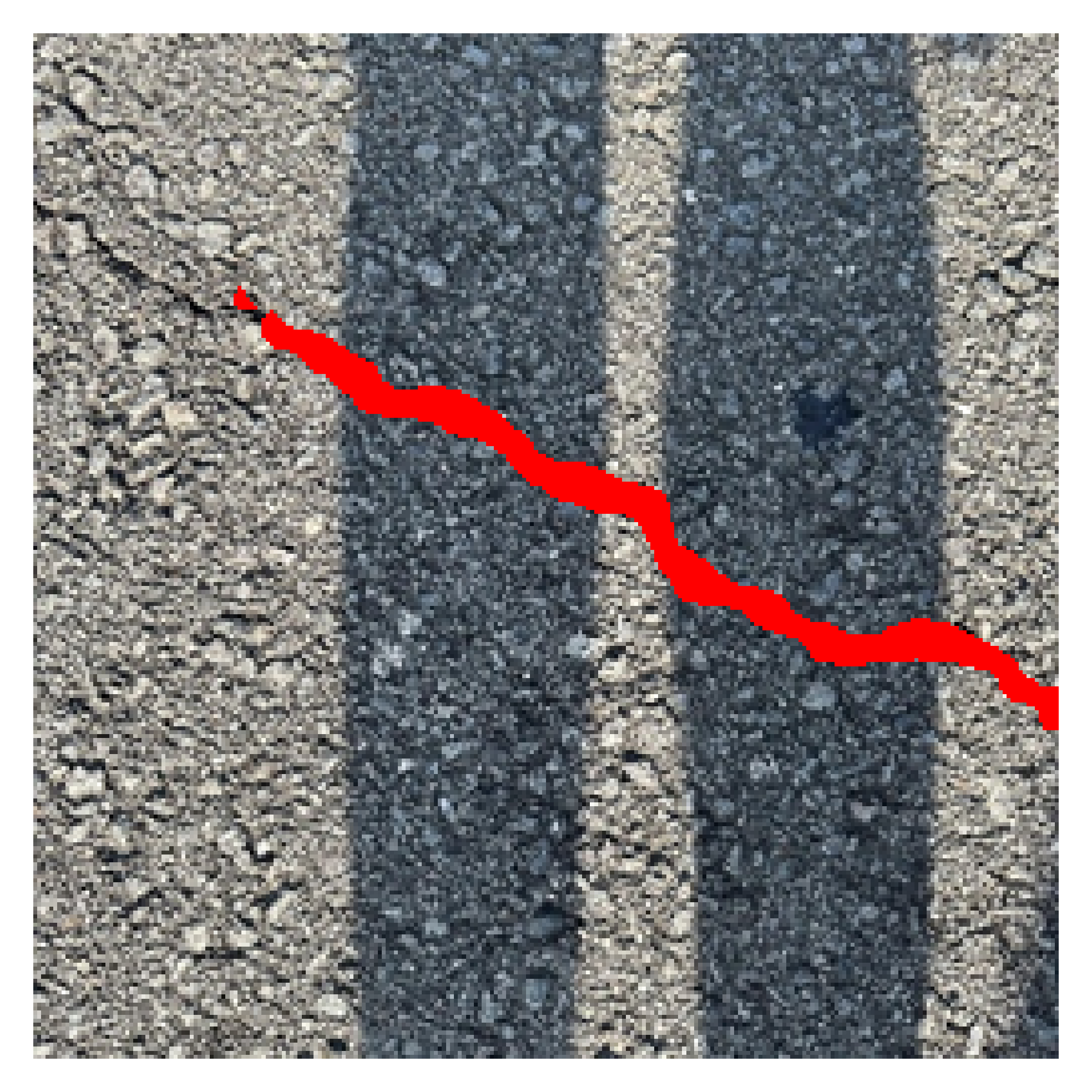}}
    \end{subfigure}
    
    \vspace{0.5cm}

    \begin{subfigure}[b]{0.1\textwidth}
        \adjustbox{trim=10 10 10 10,clip,width=1.6cm,height=1.6cm}{\includegraphics{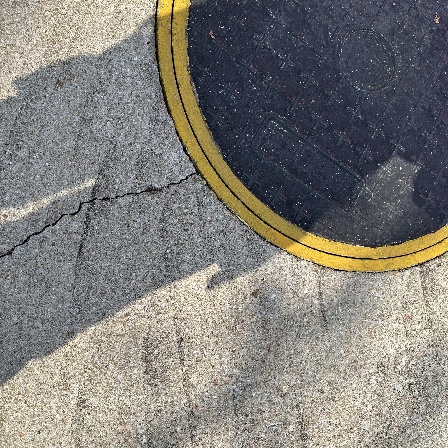}}
    \end{subfigure}
    \begin{subfigure}[b]{0.1\textwidth}
        \adjustbox{trim=10 10 10 10,clip,width=1.6cm,height=1.6cm}{\includegraphics{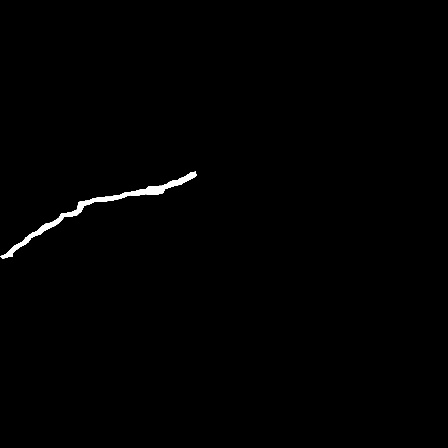}}
    \end{subfigure}  
    \begin{subfigure}[b]{0.1\textwidth}
        \adjustbox{trim=10 10 10 10,clip,width=1.6cm,height=1.6cm}{\includegraphics{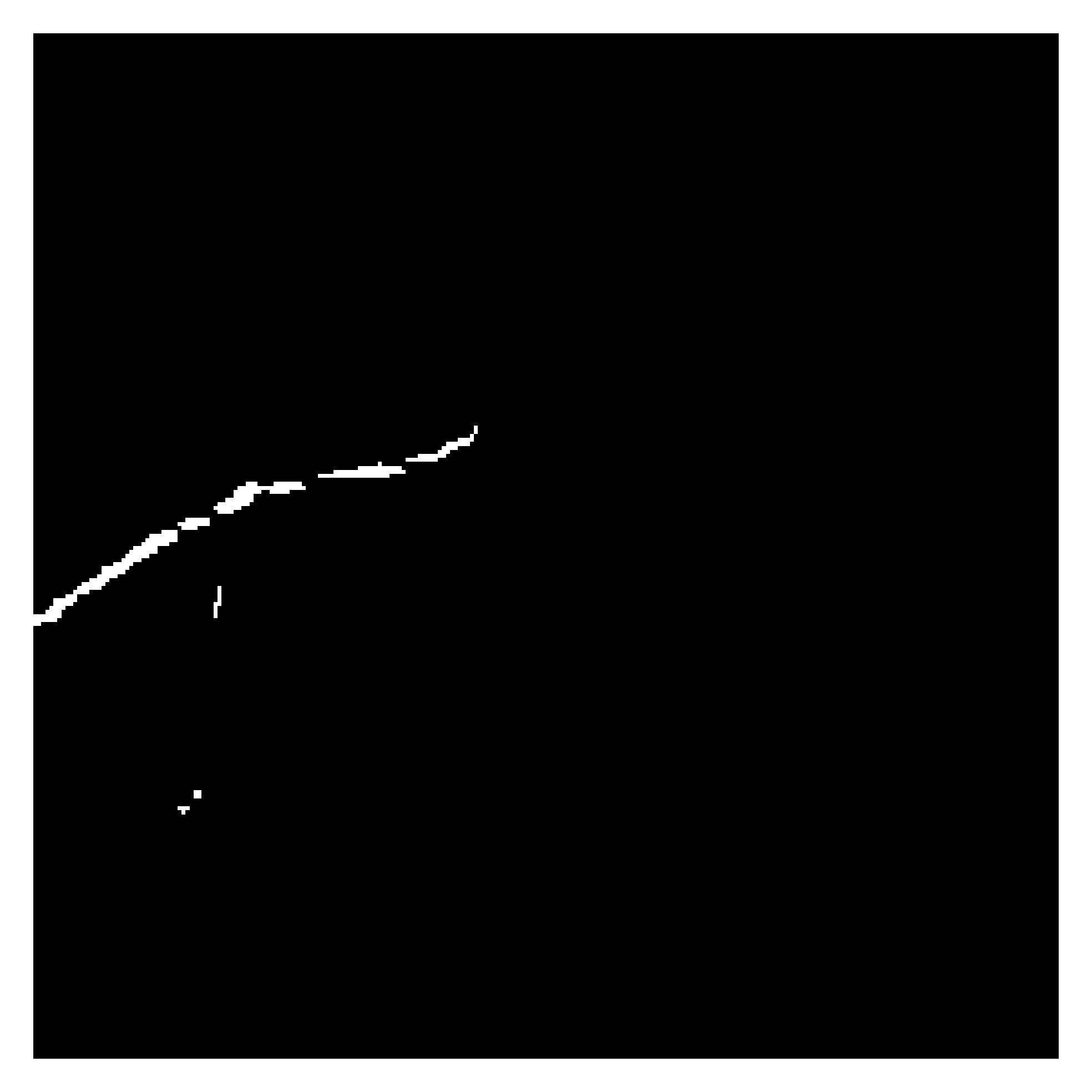}}
    \end{subfigure}
    \begin{subfigure}[b]{0.1\textwidth}
        \adjustbox{trim=10 10 10 10,clip,width=1.6cm,height=1.6cm}{\includegraphics{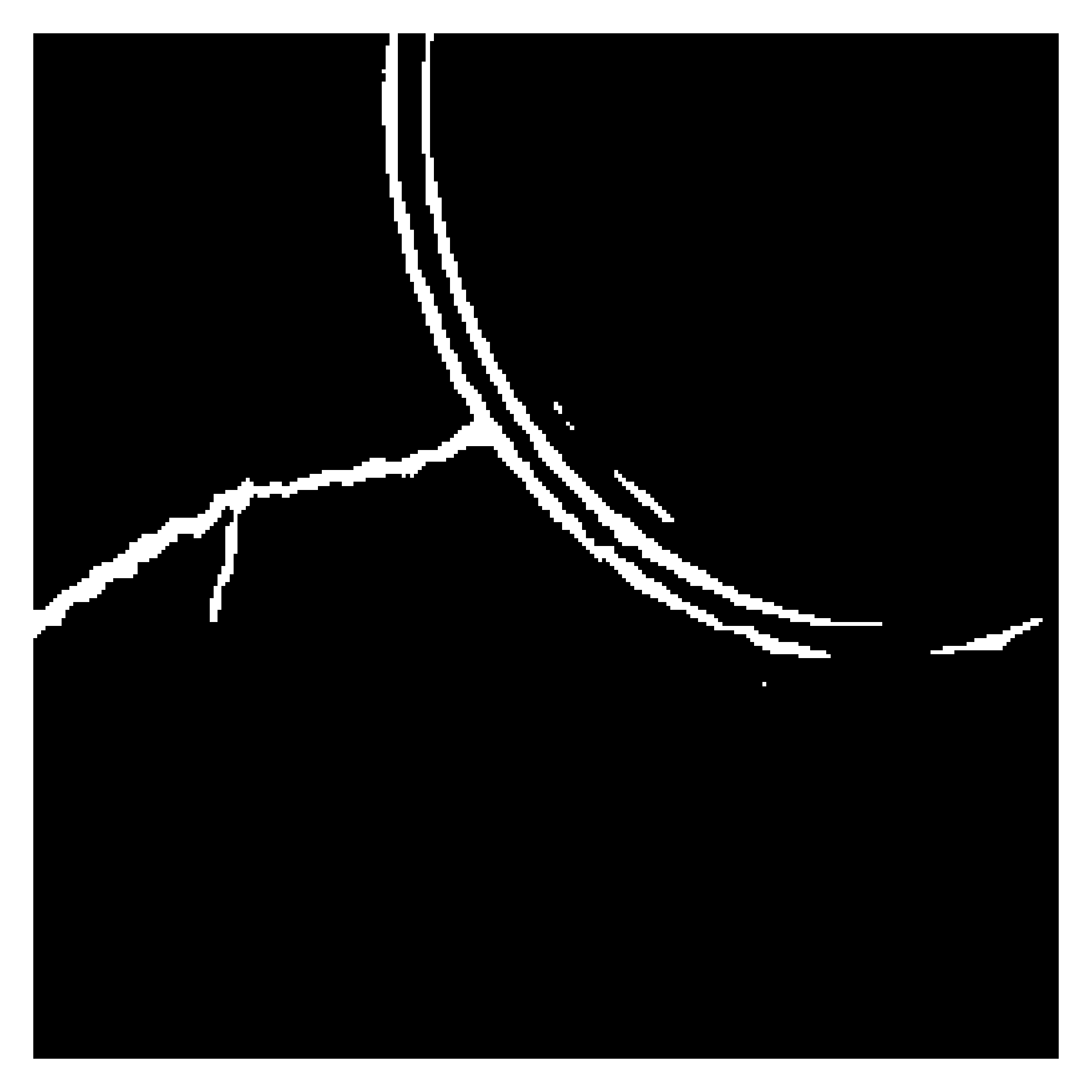}}
    \end{subfigure}
    \begin{subfigure}[b]{0.1\textwidth}
        \adjustbox{trim=10 10 10 10,clip,width=1.6cm,height=1.6cm}{\includegraphics{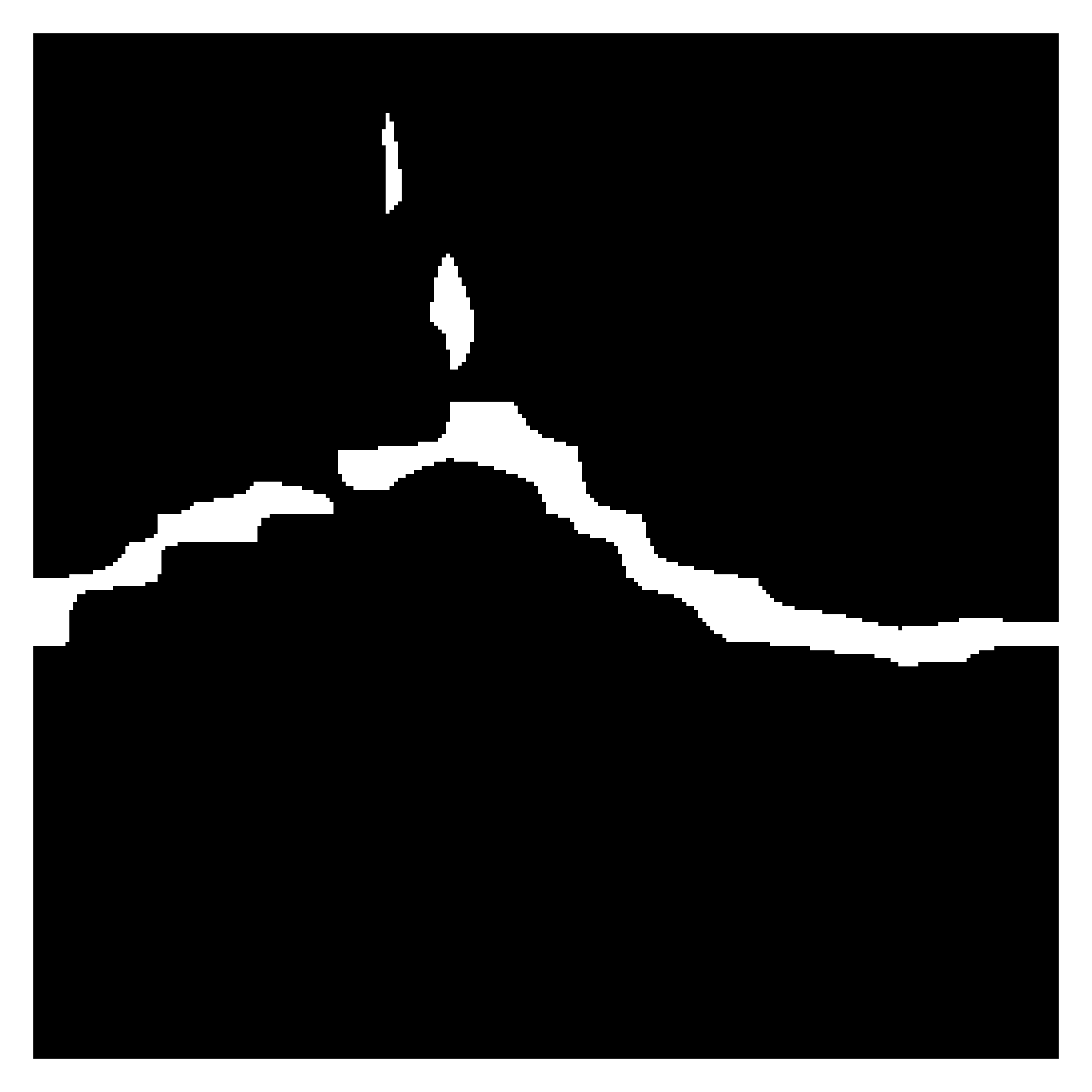}}
    \end{subfigure}
    \begin{subfigure}[b]{0.1\textwidth}
        \adjustbox{trim=10 10 10 10,clip,width=1.6cm,height=1.6cm}{\includegraphics{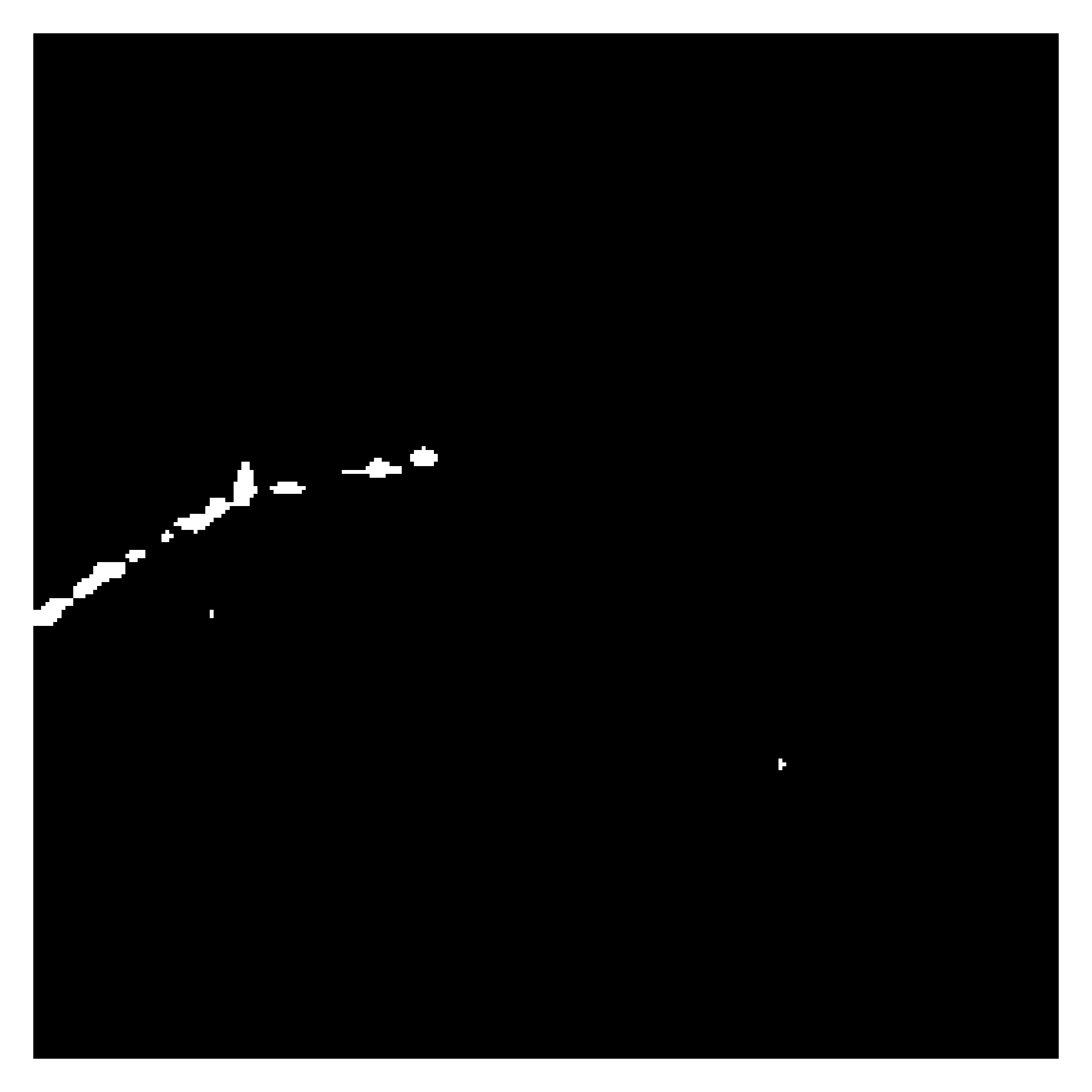}}
    \end{subfigure}
    \begin{subfigure}[b]{0.1\textwidth}
        \adjustbox{trim=10 10 10 10,clip,width=1.6cm,height=1.6cm}{\includegraphics{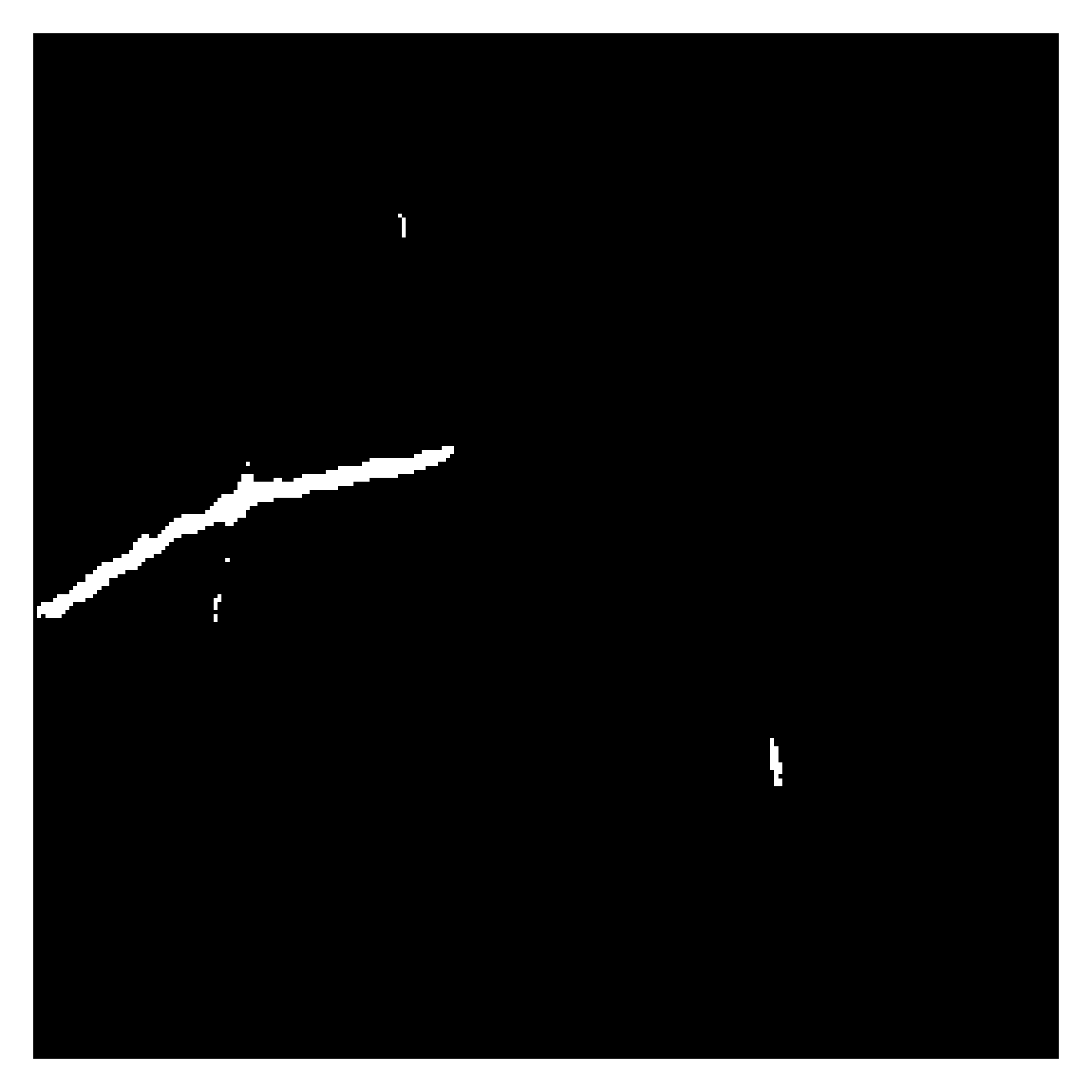}}
    \end{subfigure}
    \begin{subfigure}[b]{0.1\textwidth}
        \adjustbox{trim=10 10 10 10,clip,width=1.6cm,height=1.6cm}{\includegraphics{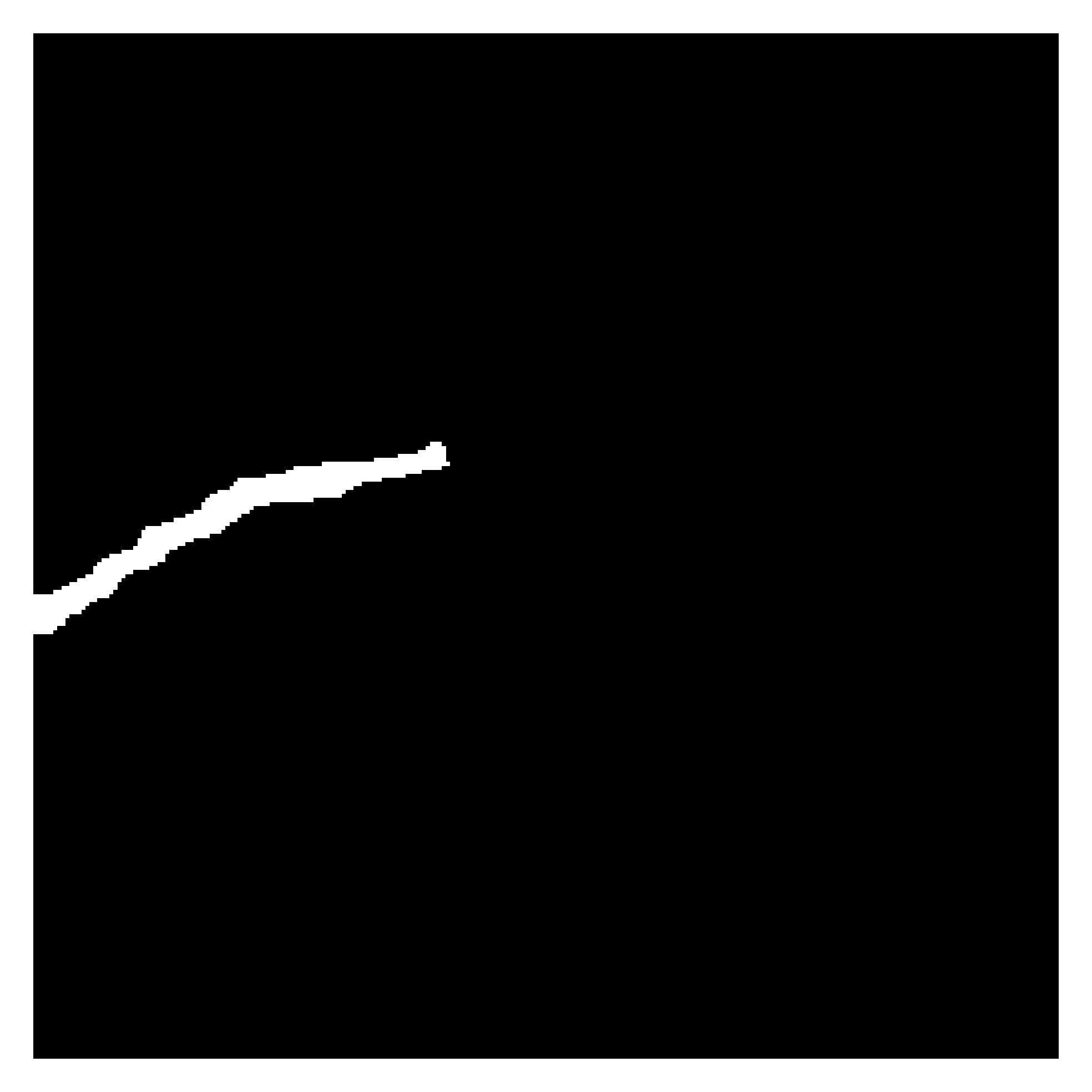}}
    \end{subfigure}

    \vspace{0.2cm}
    
    \begin{subfigure}[b]{0.1\textwidth}
        \adjustbox{trim=10 10 10 10,clip,width=1.6cm,height=1.6cm}{\includegraphics{figures/road420_zero_shot/3/2023_10_31_13_43_IMG_6217.jpg}}
    \end{subfigure}
    \begin{subfigure}[b]{0.1\textwidth}
        \adjustbox{trim=10 10 10 10,clip,width=1.6cm,height=1.6cm}{\includegraphics{figures/road420_zero_shot/3/2023_10_31_13_43_IMG_6217_mask.jpg}}
    \end{subfigure}  
    \begin{subfigure}[b]{0.1\textwidth}
        \adjustbox{trim=10 10 10 10,clip,width=1.6cm,height=1.6cm}{\includegraphics{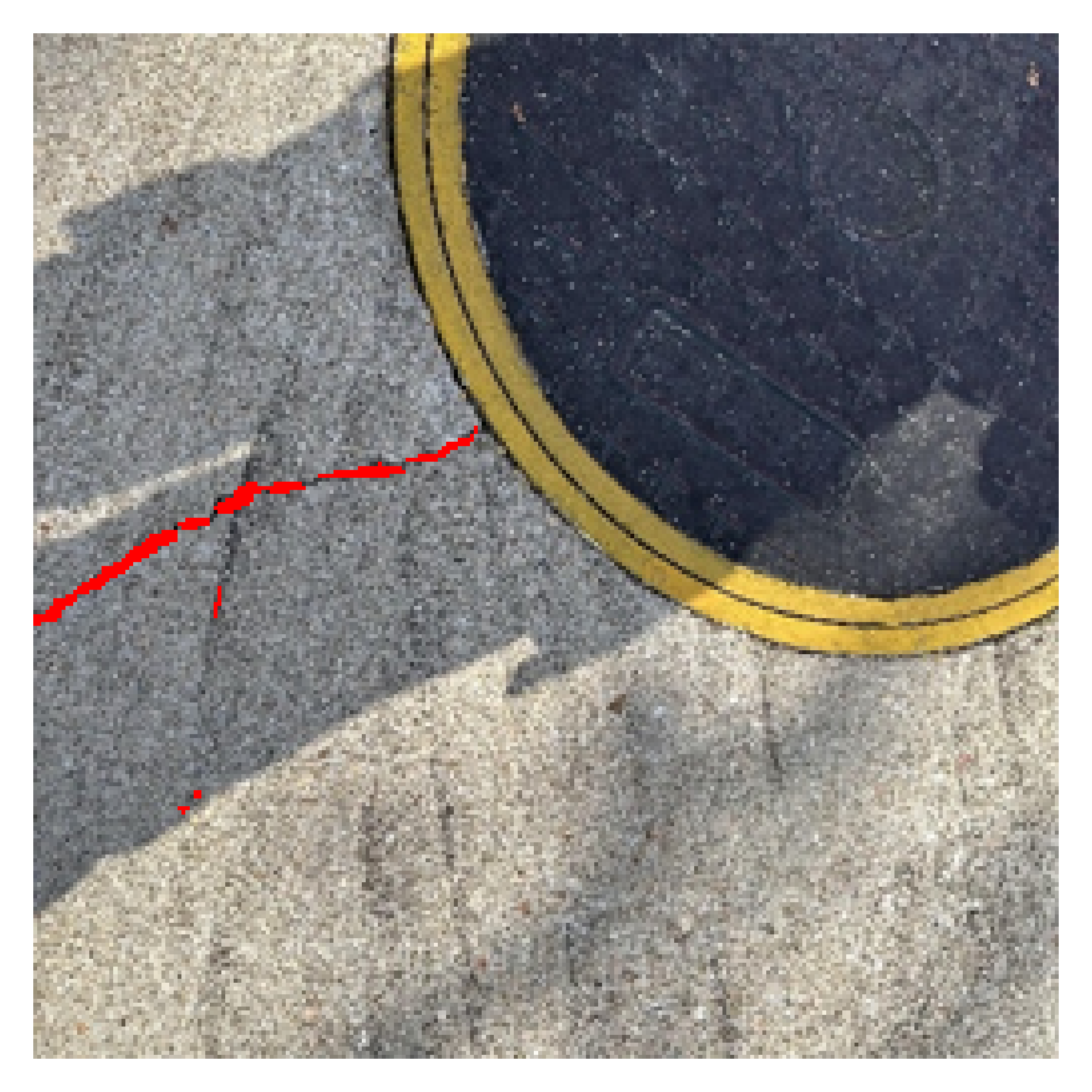}}
    \end{subfigure}
    \begin{subfigure}[b]{0.1\textwidth}
        \adjustbox{trim=10 10 10 10,clip,width=1.6cm,height=1.6cm}{\includegraphics{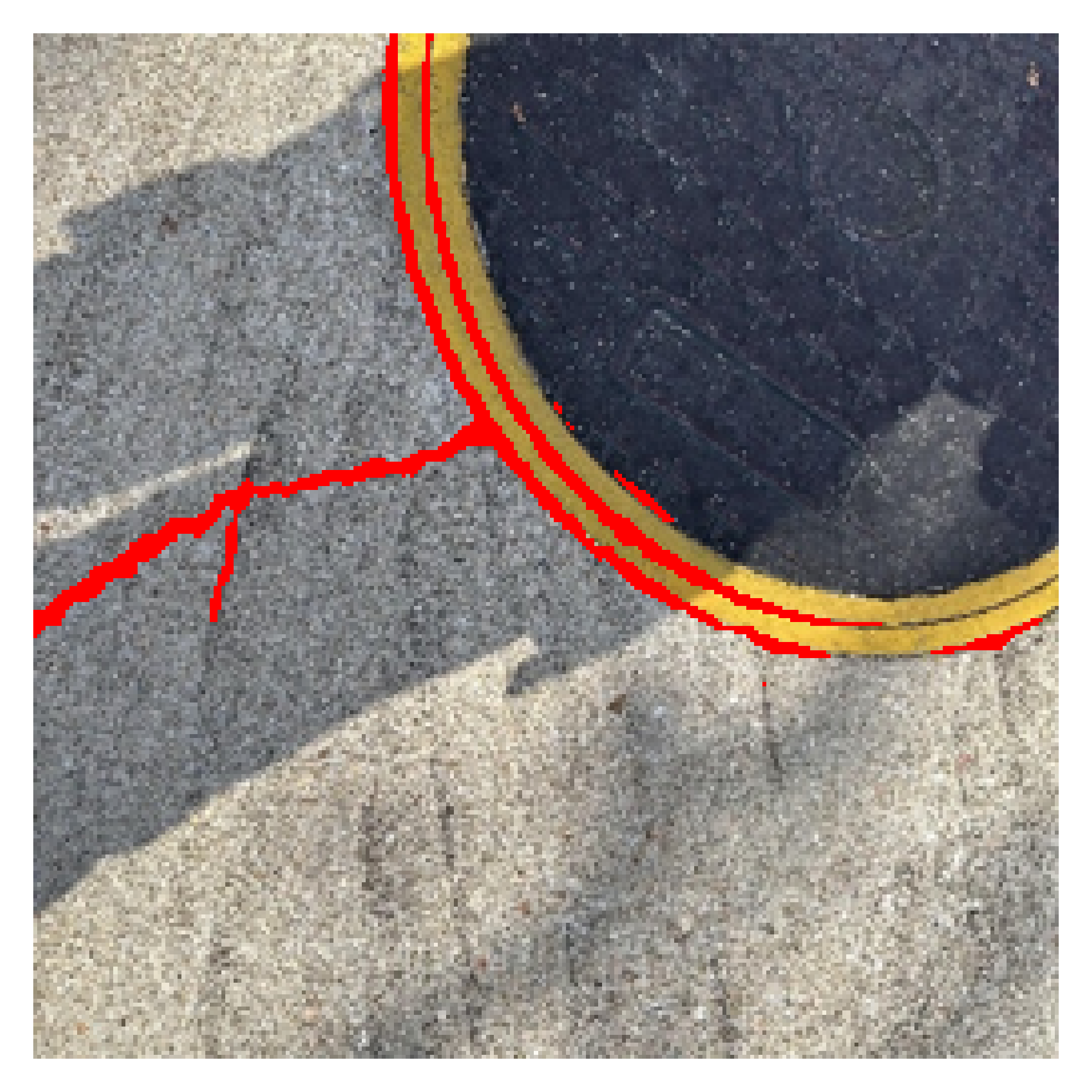}}
    \end{subfigure}
    \begin{subfigure}[b]{0.1\textwidth}
        \adjustbox{trim=10 10 10 10,clip,width=1.6cm,height=1.6cm}{\includegraphics{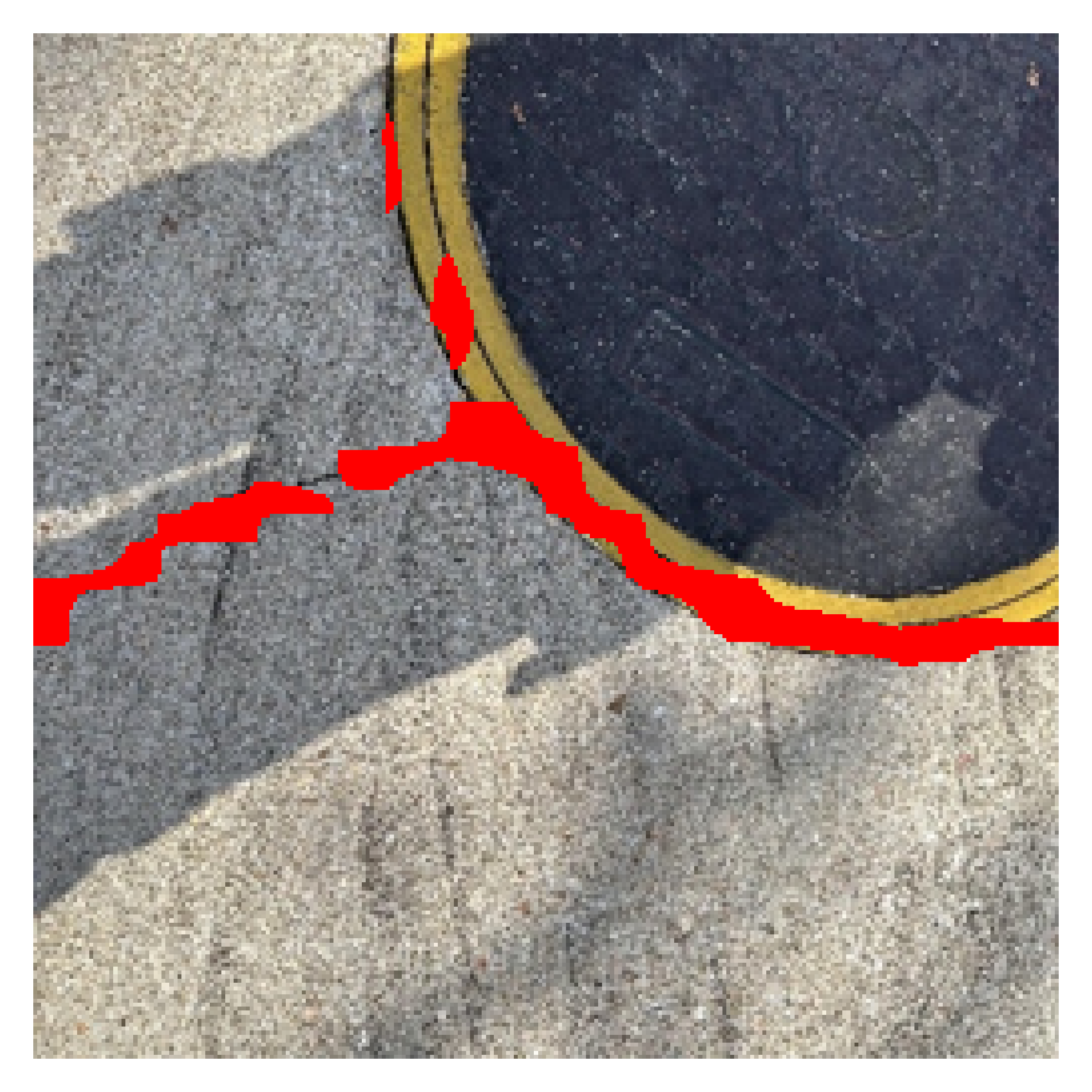}}
    \end{subfigure}
    \begin{subfigure}[b]{0.1\textwidth}
        \adjustbox{trim=10 10 10 10,clip,width=1.6cm,height=1.6cm}{\includegraphics{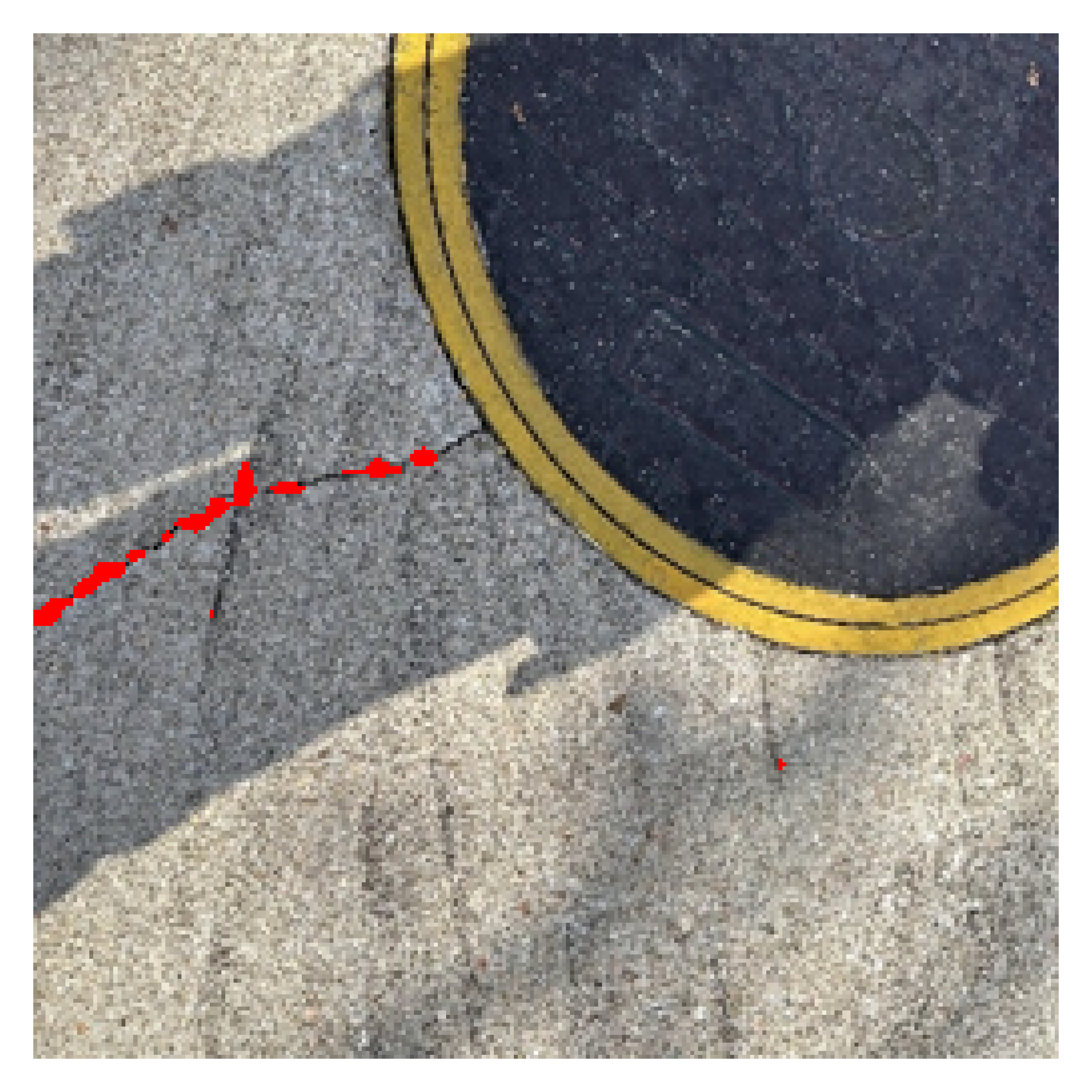}}
    \end{subfigure}
    \begin{subfigure}[b]{0.1\textwidth}
        \adjustbox{trim=10 10 10 10,clip,width=1.6cm,height=1.6cm}{\includegraphics{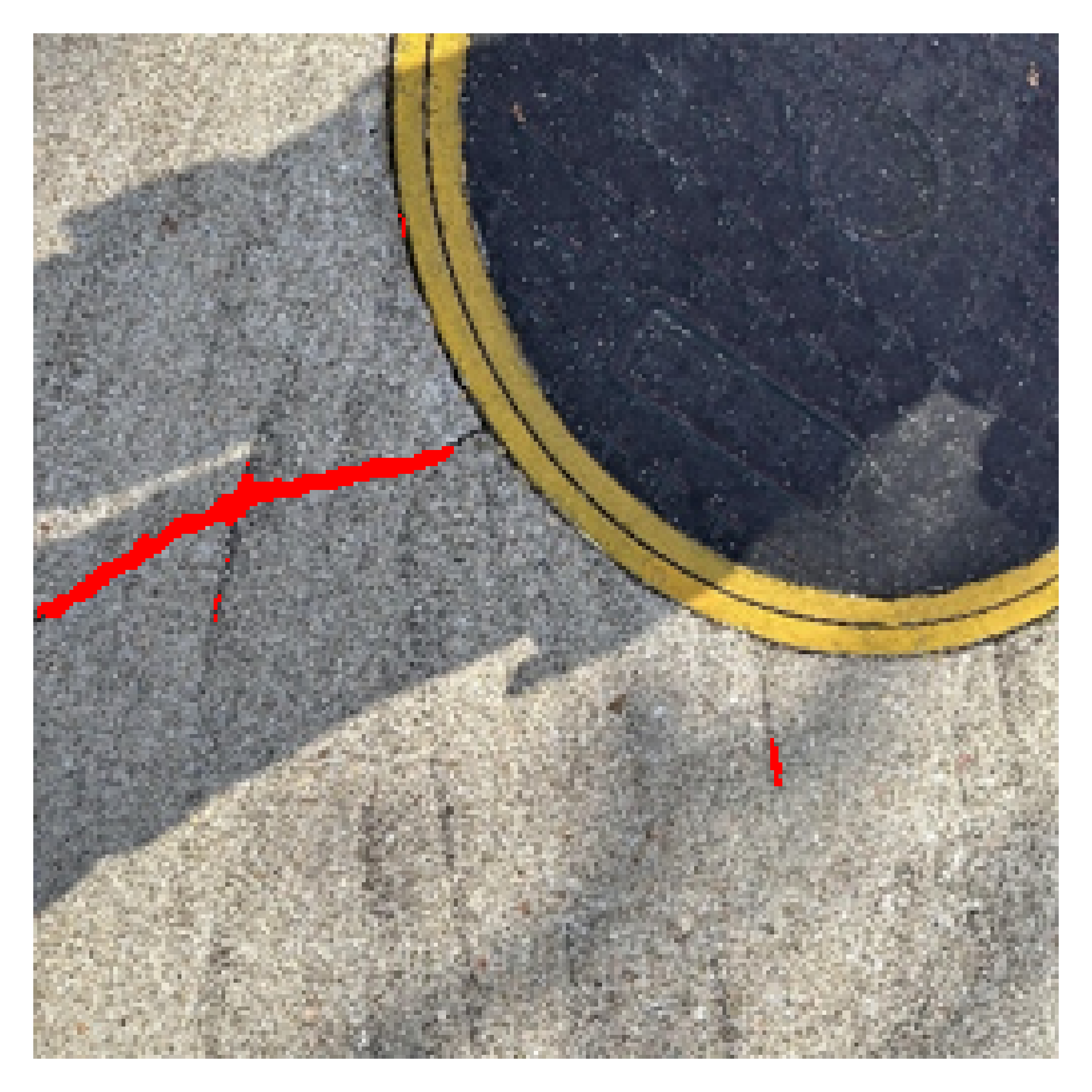}}
    \end{subfigure}
    \begin{subfigure}[b]{0.1\textwidth}
        \adjustbox{trim=10 10 10 10,clip,width=1.6cm,height=1.6cm}{\includegraphics{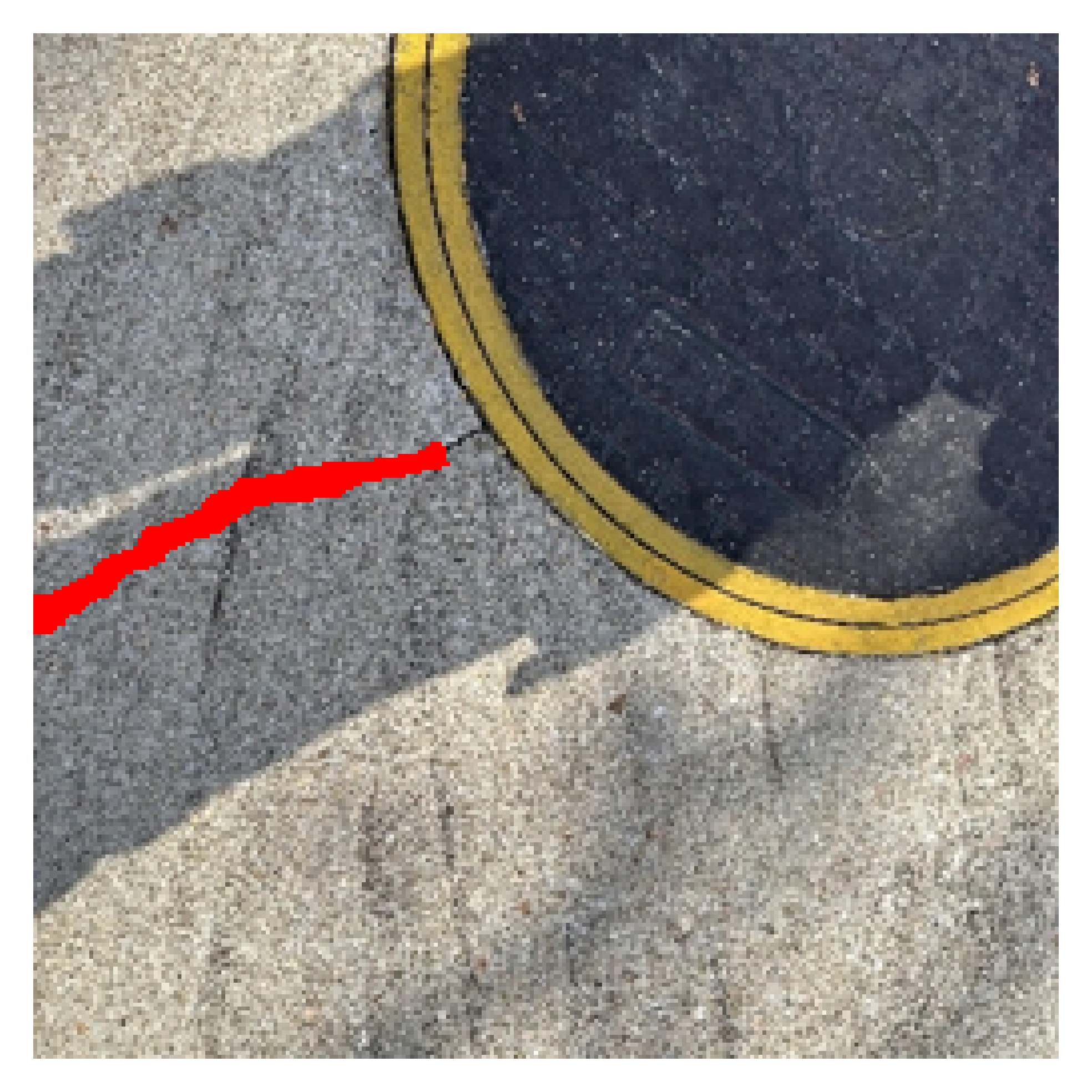}}
    \end{subfigure}
    
    \vspace{0.5cm}

    \begin{subfigure}[b]{0.1\textwidth}
        \adjustbox{trim=10 10 10 10,clip,width=1.6cm,height=1.6cm}{\includegraphics{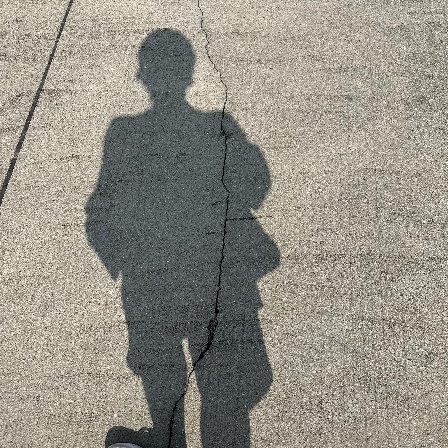}}
    \end{subfigure}
    \begin{subfigure}[b]{0.1\textwidth}
        \adjustbox{trim=10 10 10 10,clip,width=1.6cm,height=1.6cm}{\includegraphics{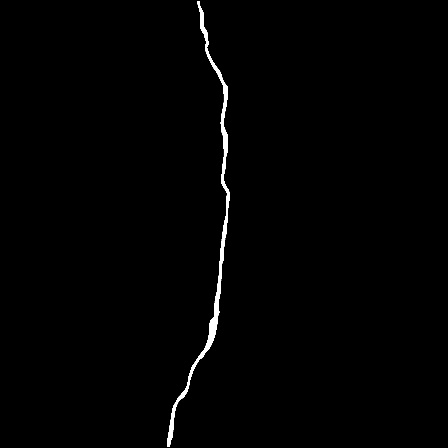}}
    \end{subfigure}  
    \begin{subfigure}[b]{0.1\textwidth}
        \adjustbox{trim=10 10 10 10,clip,width=1.6cm,height=1.6cm}{\includegraphics{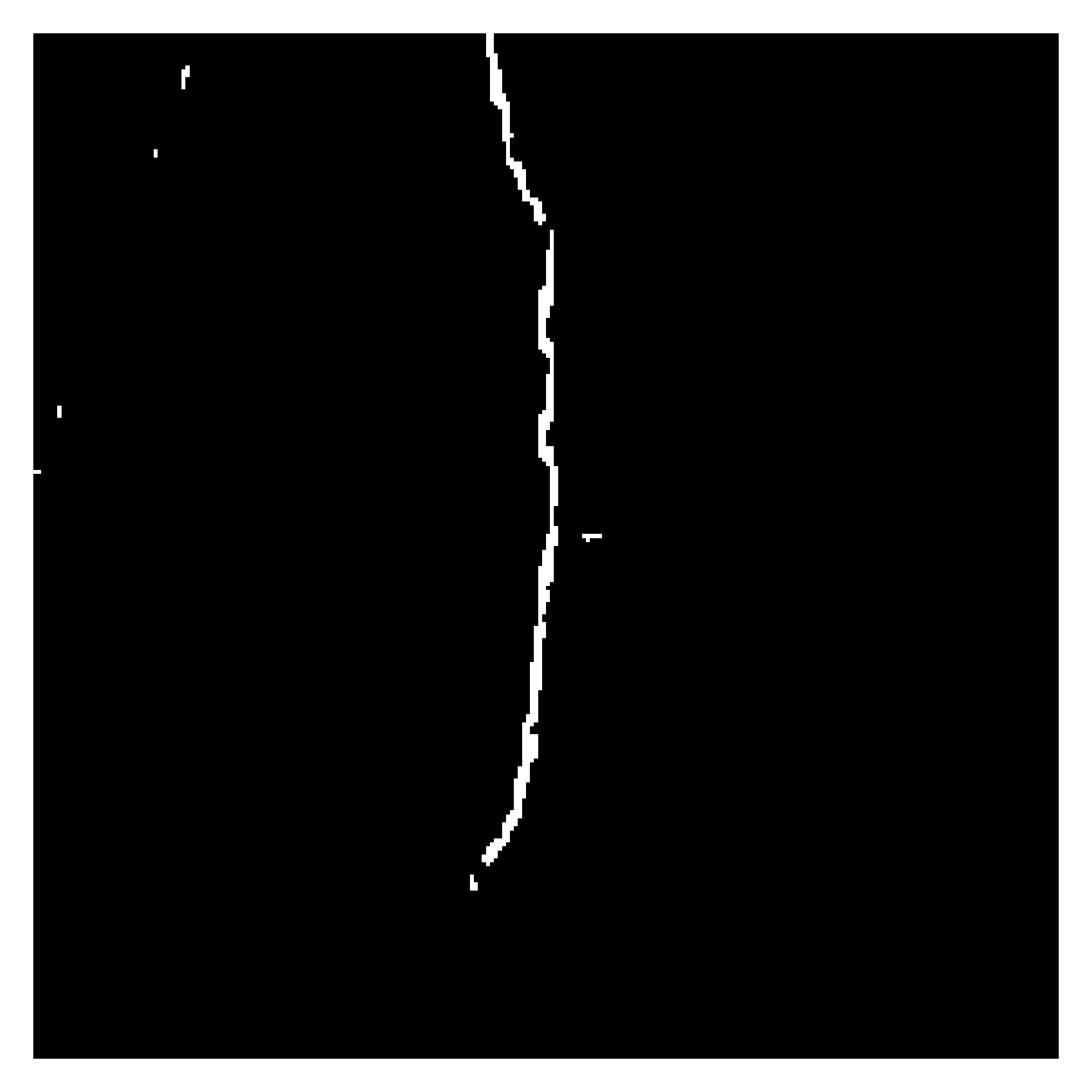}}
    \end{subfigure}
    \begin{subfigure}[b]{0.1\textwidth}
        \adjustbox{trim=10 10 10 10,clip,width=1.6cm,height=1.6cm}{\includegraphics{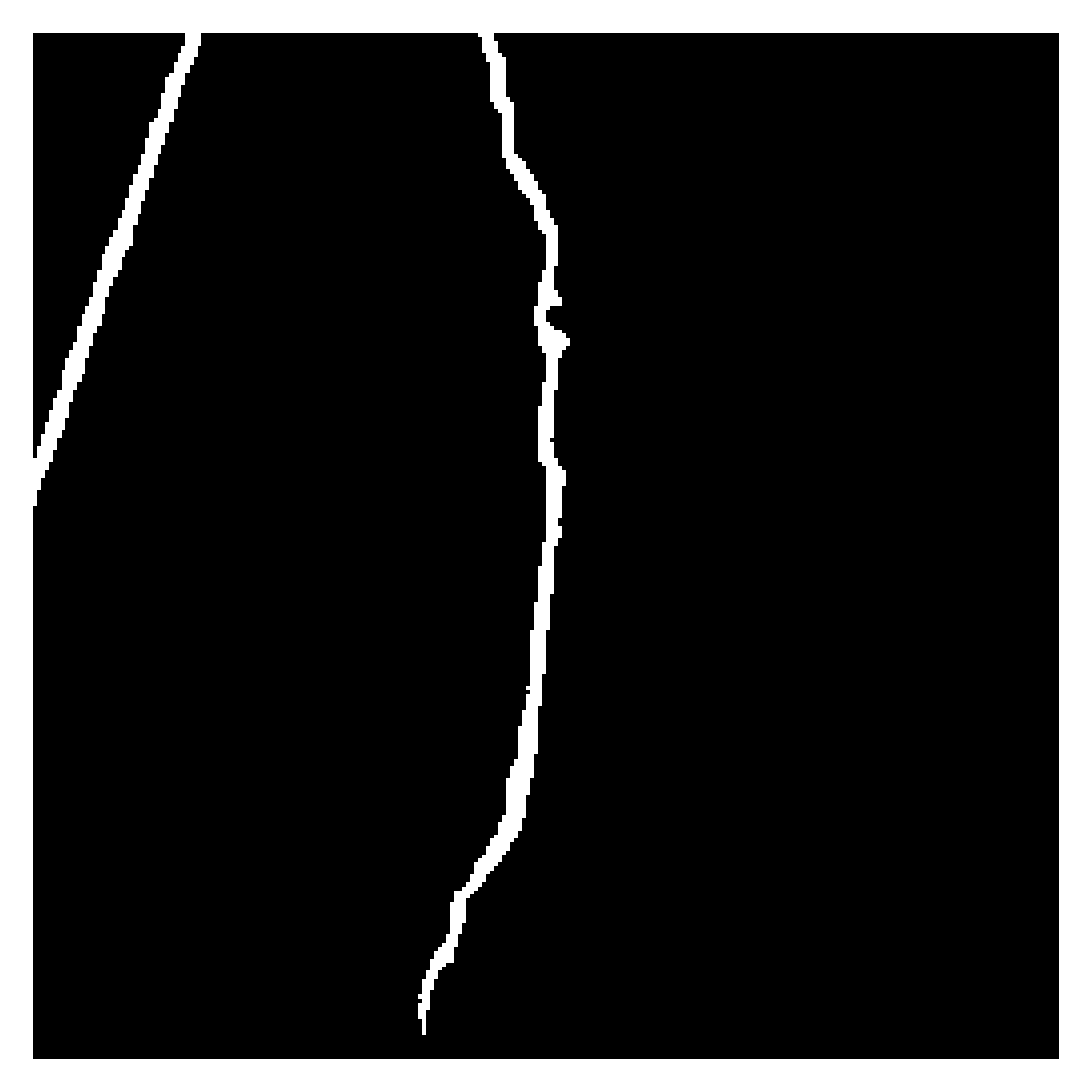}}
    \end{subfigure}
    \begin{subfigure}[b]{0.1\textwidth}
        \adjustbox{trim=10 10 10 10,clip,width=1.6cm,height=1.6cm}{\includegraphics{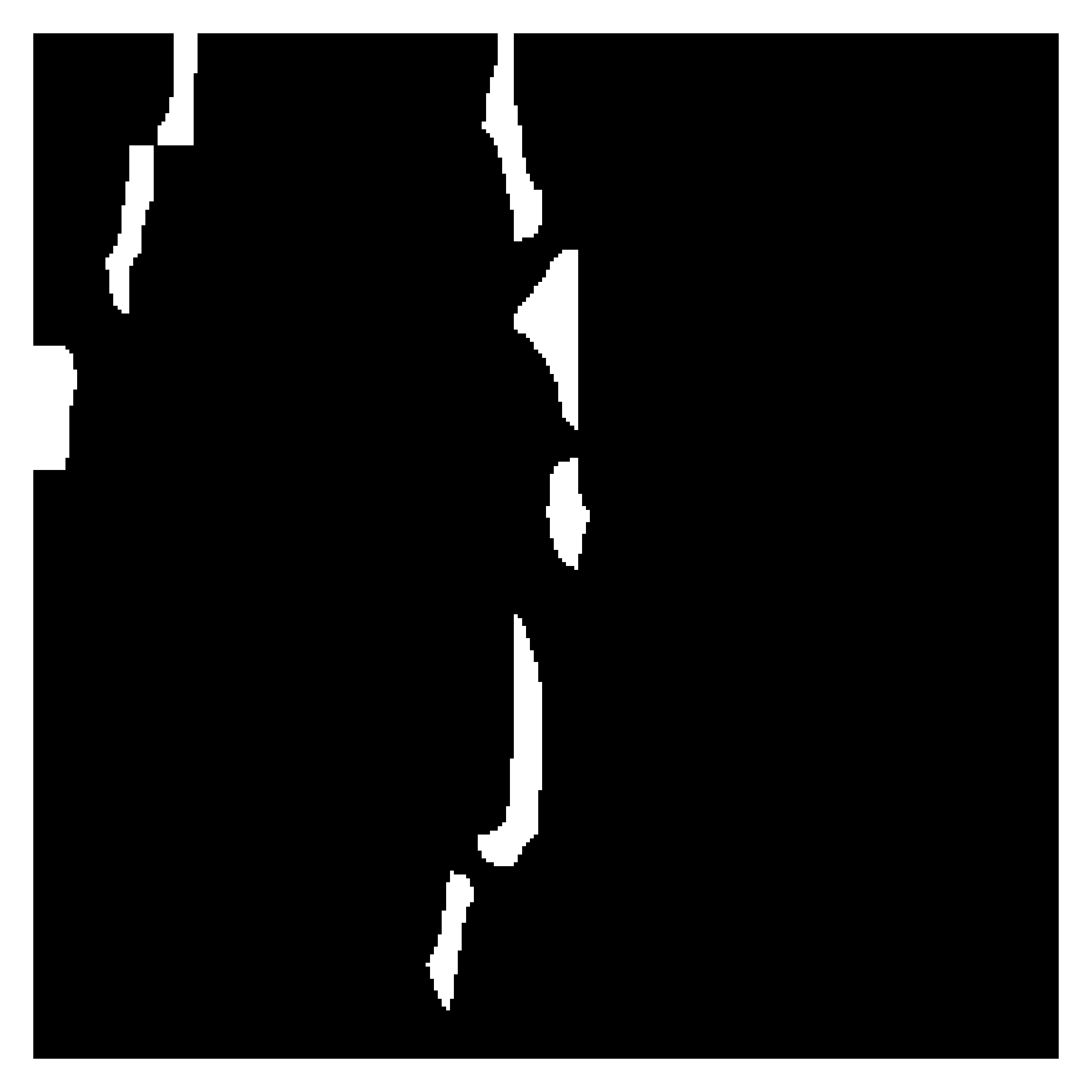}}
    \end{subfigure}
    \begin{subfigure}[b]{0.1\textwidth}
        \adjustbox{trim=10 10 10 10,clip,width=1.6cm,height=1.6cm}{\includegraphics{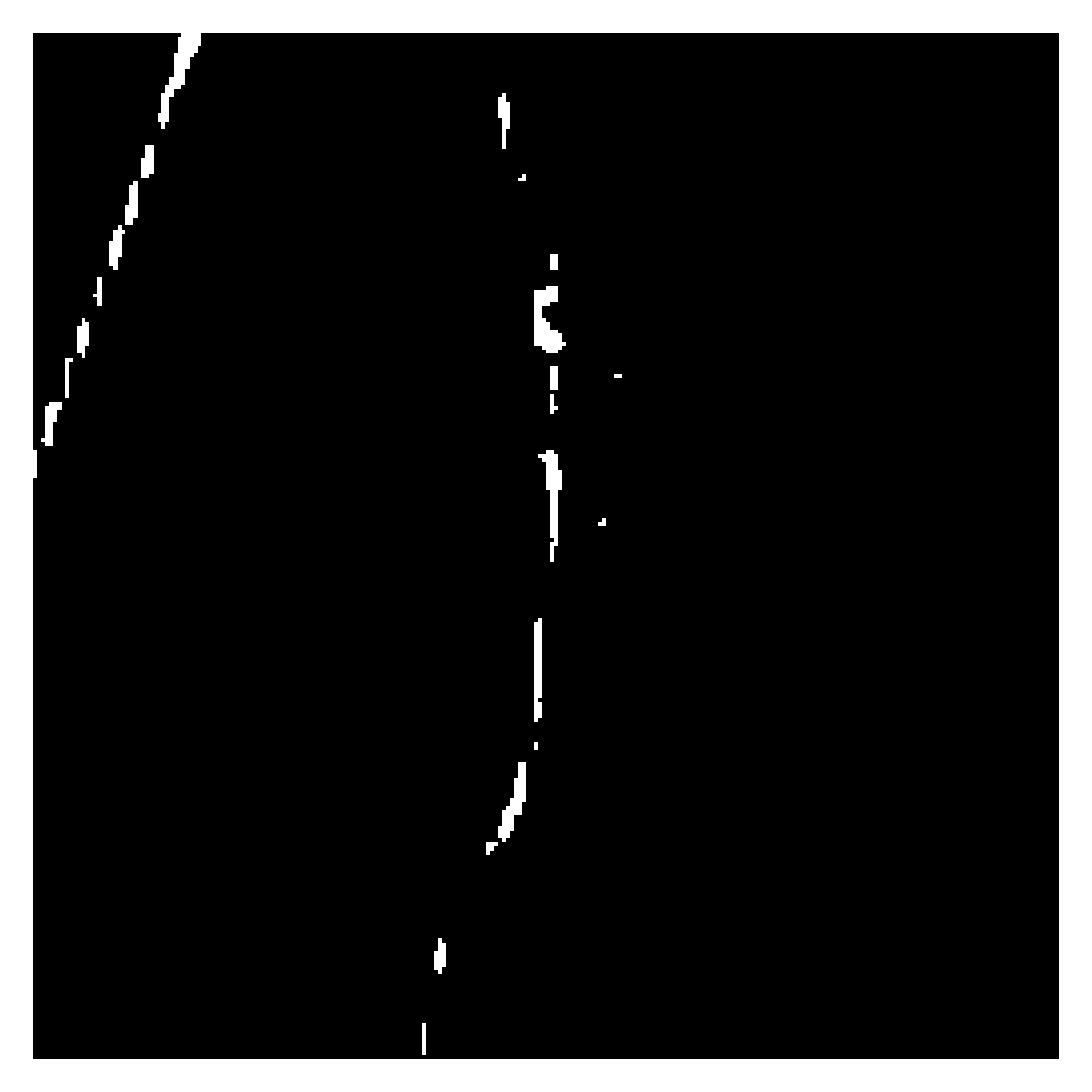}}
    \end{subfigure}
    \begin{subfigure}[b]{0.1\textwidth}
        \adjustbox{trim=10 10 10 10,clip,width=1.6cm,height=1.6cm}{\includegraphics{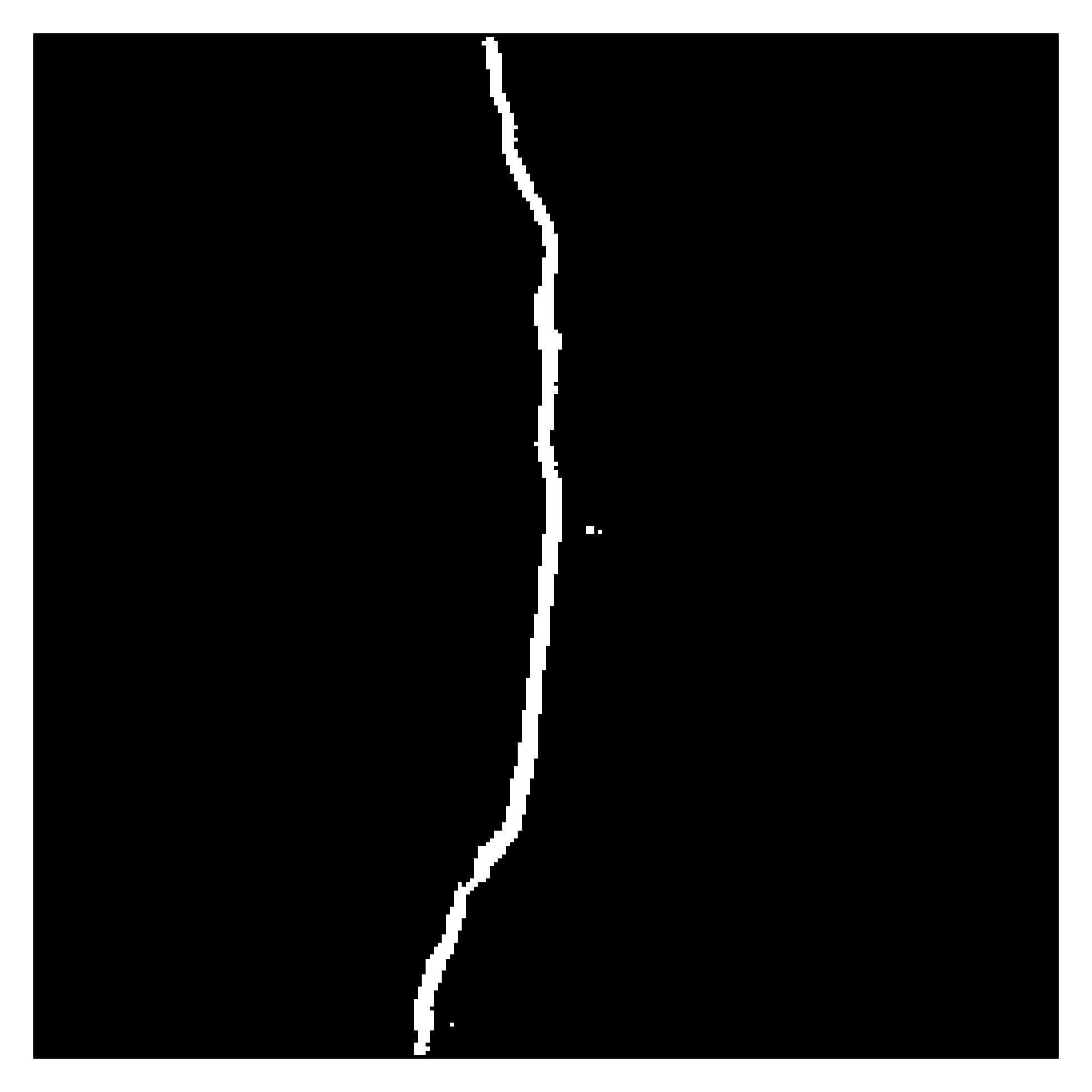}}
    \end{subfigure}
    \begin{subfigure}[b]{0.1\textwidth}
        \adjustbox{trim=10 10 10 10,clip,width=1.6cm,height=1.6cm}{\includegraphics{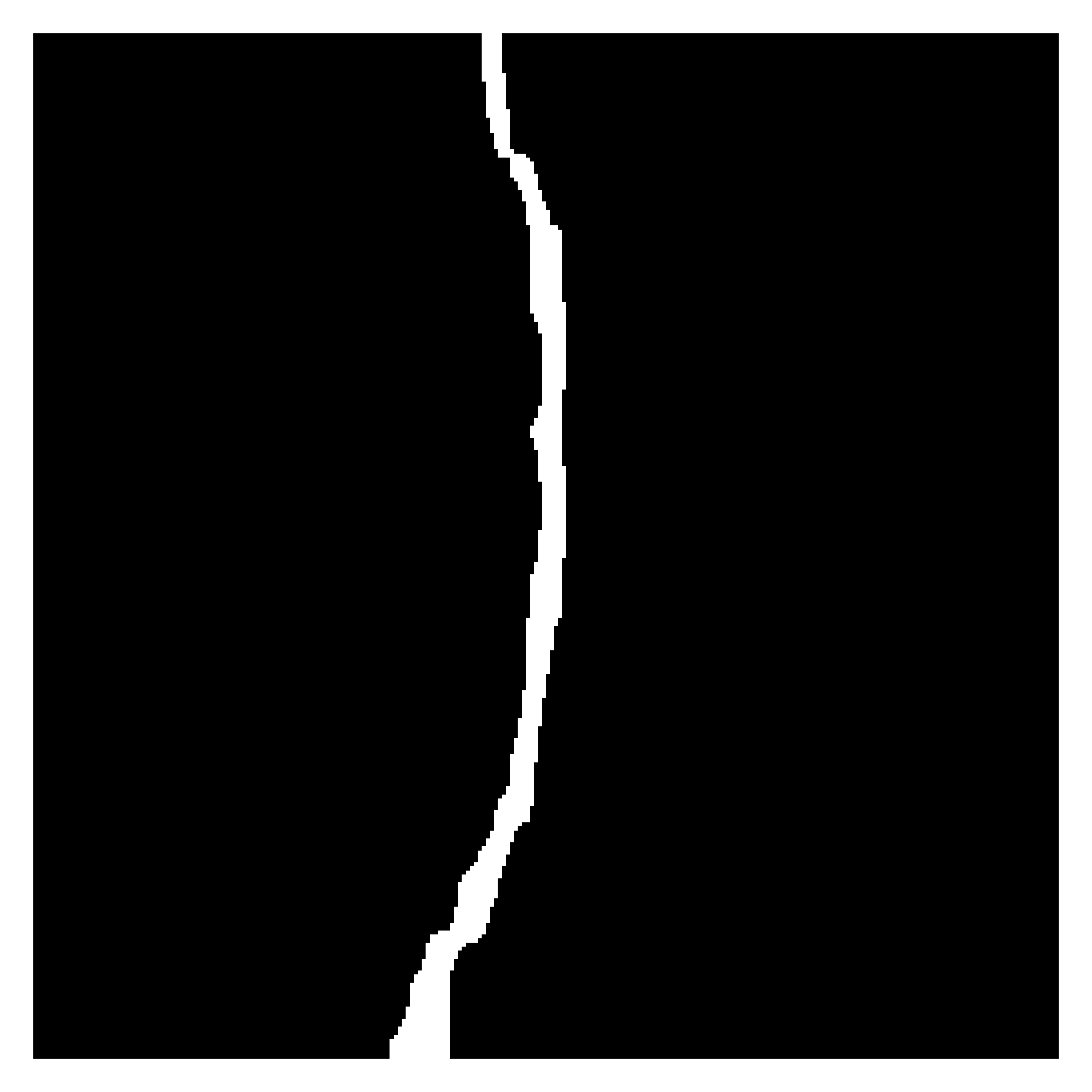}}
    \end{subfigure}

    \vspace{0.2cm}
    
    \begin{subfigure}[b]{0.1\textwidth}
        \adjustbox{trim=10 10 10 10,clip,width=1.6cm,height=1.6cm}{\includegraphics{figures/road420_zero_shot/2/2023_10_31_13_39_IMG_6214.jpg}}
    \end{subfigure}
    \begin{subfigure}[b]{0.1\textwidth}
        \adjustbox{trim=10 10 10 10,clip,width=1.6cm,height=1.6cm}{\includegraphics{figures/road420_zero_shot/2/2023_10_31_13_39_IMG_6214_mask.jpg}}
    \end{subfigure}  
    \begin{subfigure}[b]{0.1\textwidth}
        \adjustbox{trim=10 10 10 10,clip,width=1.6cm,height=1.6cm}{\includegraphics{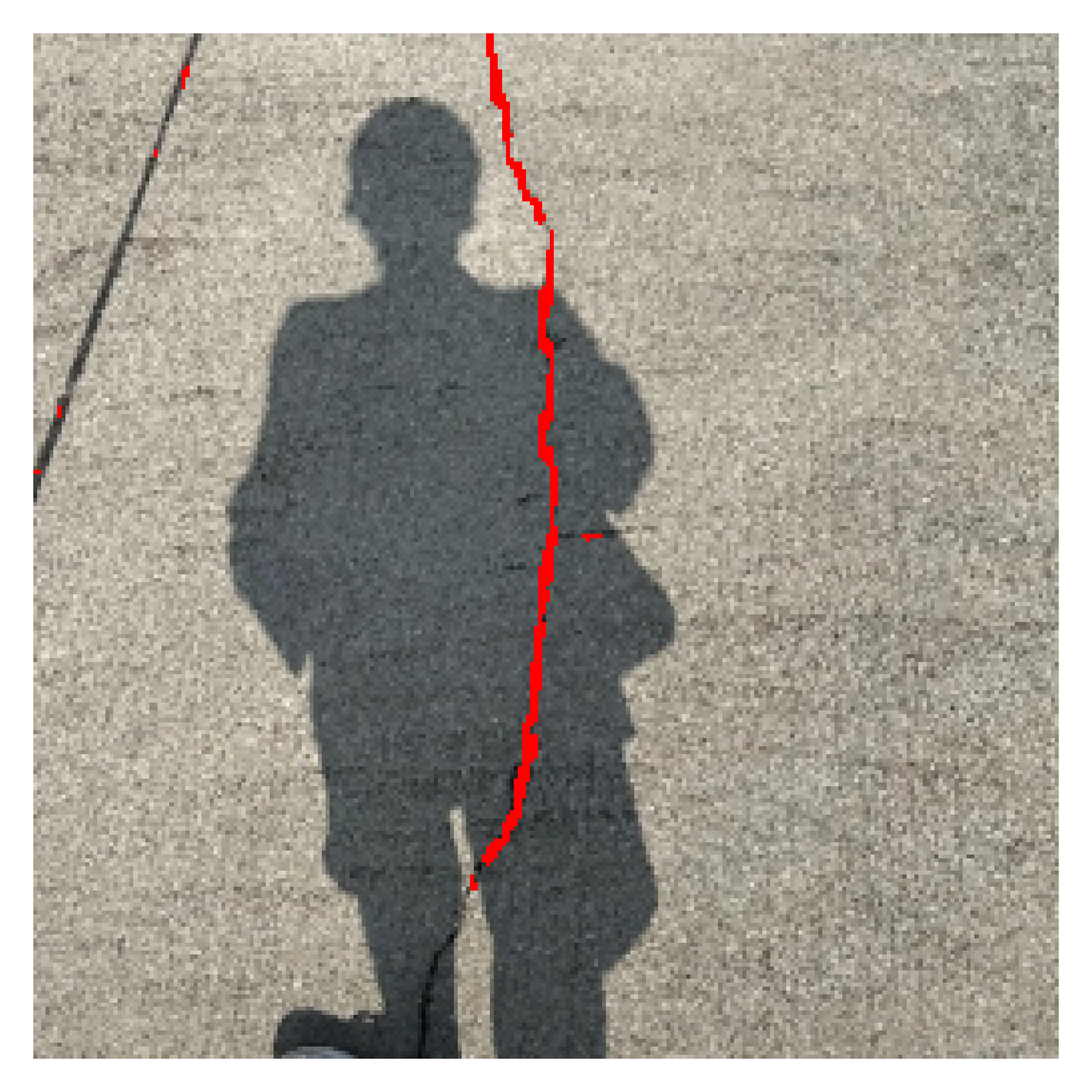}}
    \end{subfigure}
    \begin{subfigure}[b]{0.1\textwidth}
        \adjustbox{trim=10 10 10 10,clip,width=1.6cm,height=1.6cm}{\includegraphics{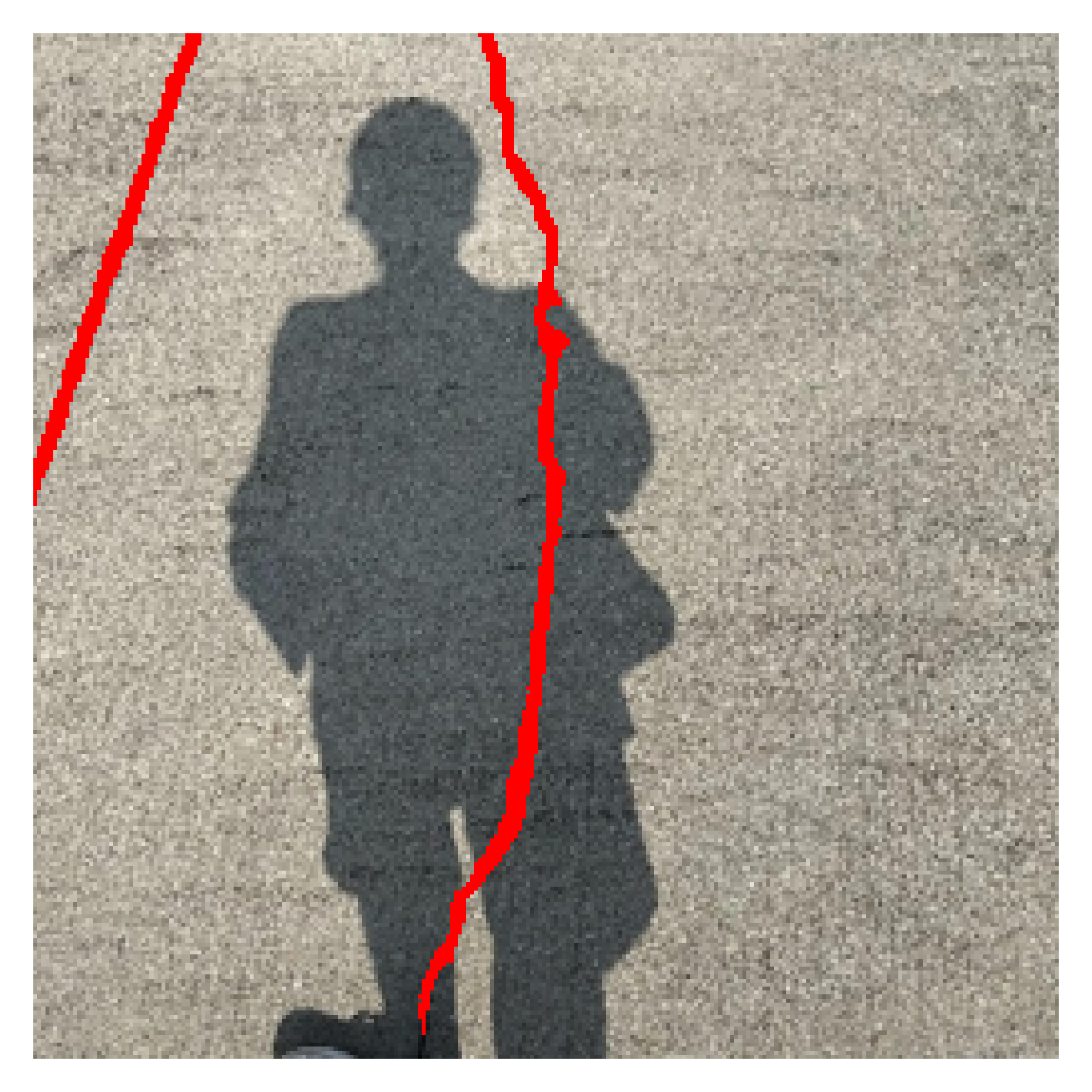}}
    \end{subfigure}
    \begin{subfigure}[b]{0.1\textwidth}
        \adjustbox{trim=10 10 10 10,clip,width=1.6cm,height=1.6cm}{\includegraphics{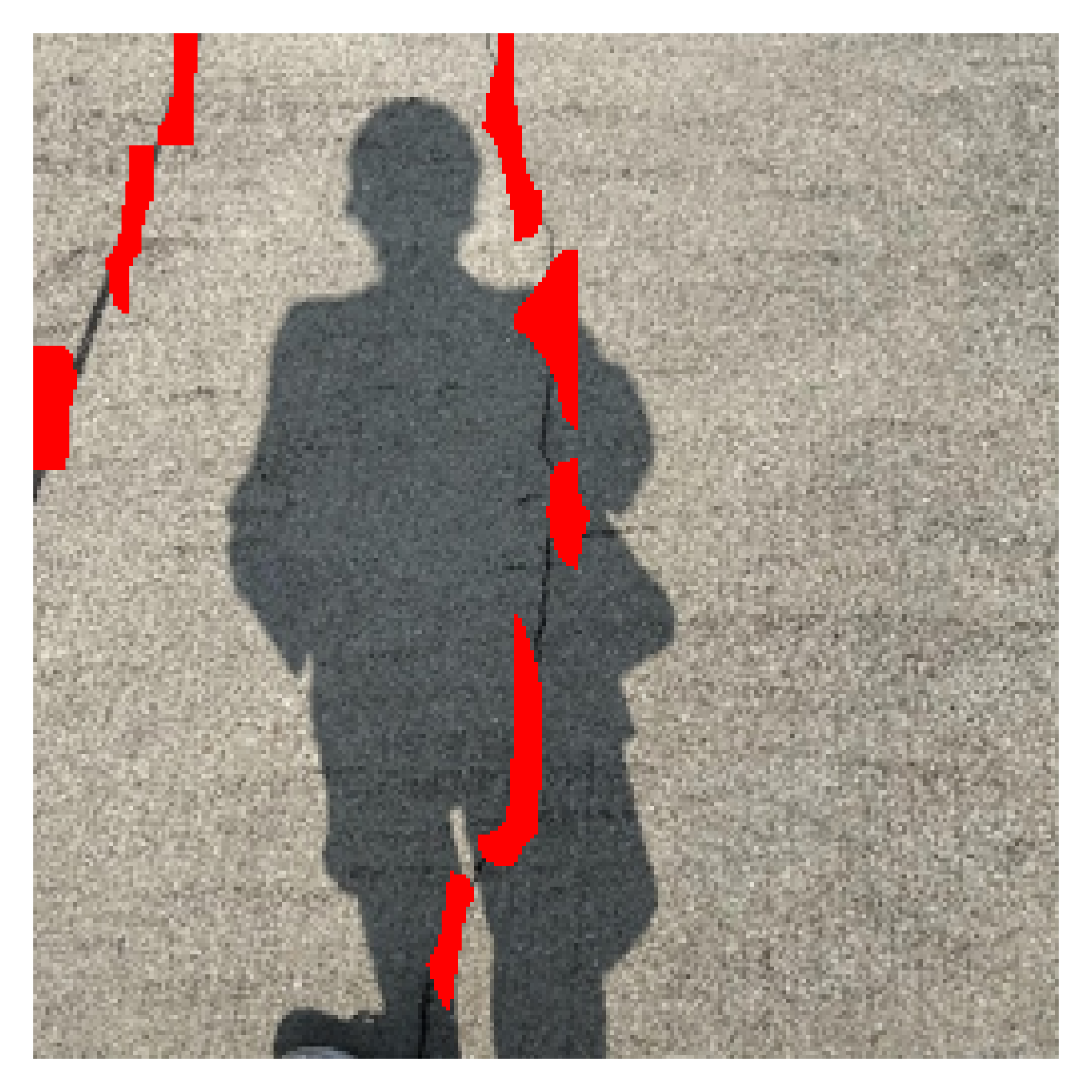}}
    \end{subfigure}
    \begin{subfigure}[b]{0.1\textwidth}
        \adjustbox{trim=10 10 10 10,clip,width=1.6cm,height=1.6cm}{\includegraphics{figures/road420_zero_shot/2/deepcrackl_finetuned_zero_shot_overlaymask_2023_10_31_13_39_IMG_6214.png}}
    \end{subfigure}
    \begin{subfigure}[b]{0.1\textwidth}
        \adjustbox{trim=10 10 10 10,clip,width=1.6cm,height=1.6cm}{\includegraphics{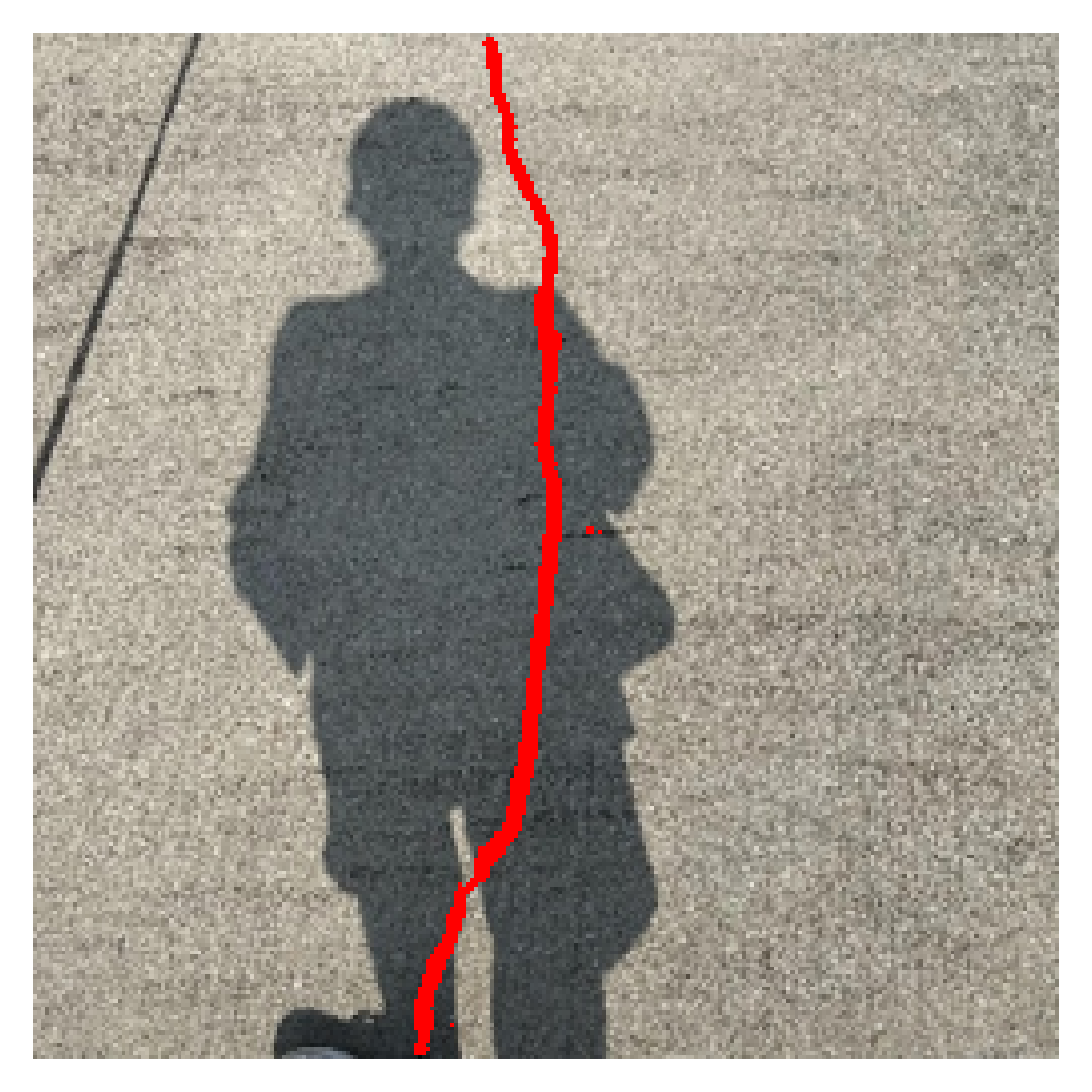}}
    \end{subfigure}
    \begin{subfigure}[b]{0.1\textwidth}
        \adjustbox{trim=10 10 10 10,clip,width=1.6cm,height=1.6cm}{\includegraphics{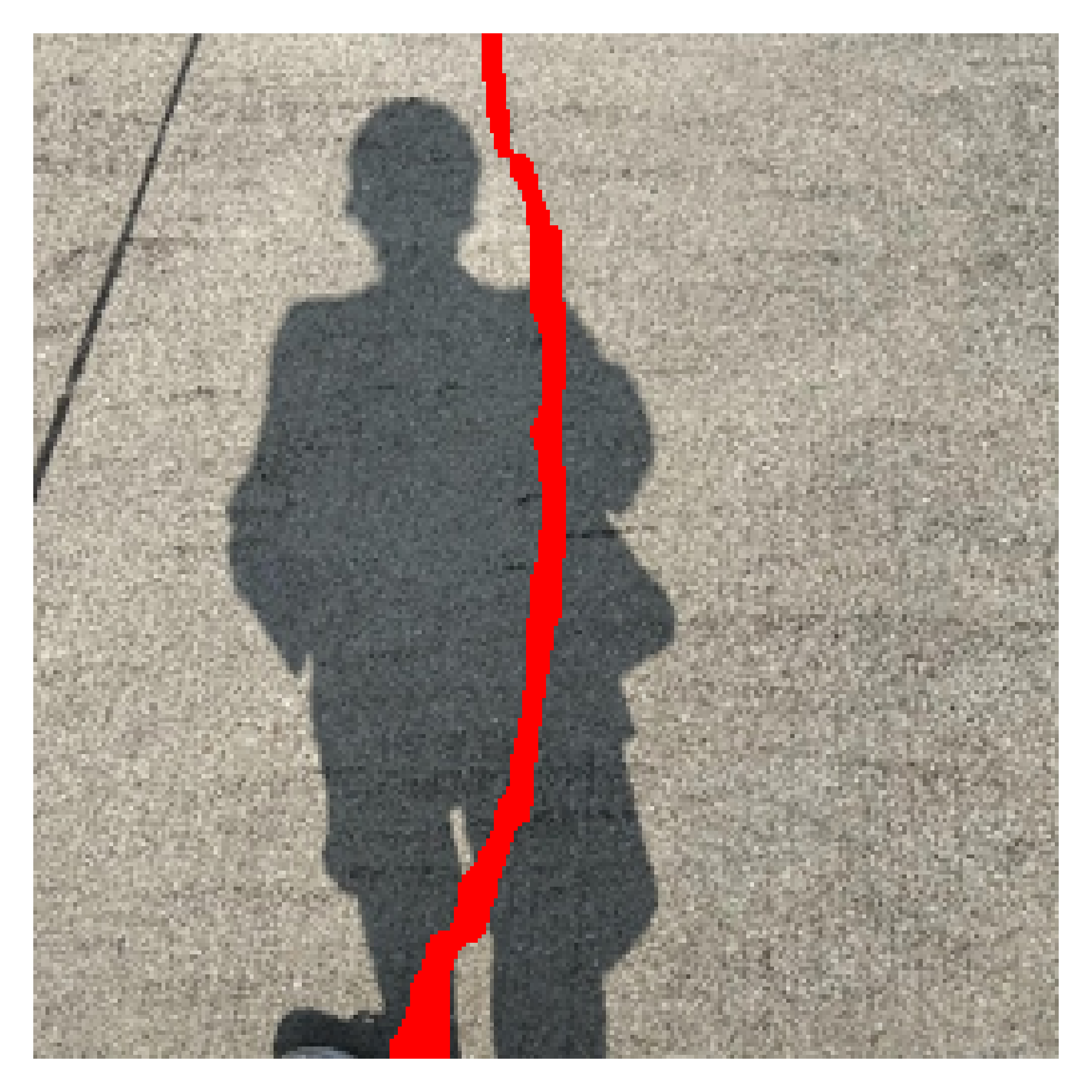}}
    \end{subfigure}
    \vspace{0.05cm}
    \vspace{0.2cm}
    \caption{Zero-shot performance of the fine-tuned models on Road420 Dataset}
    \label{fig:zero_shot_road420_qualitative}
\end{figure}

\begin{figure}[h]
    \captionsetup[subfigure]{labelformat=empty, font=footnotesize, justification=centering, position=top} 
    \centering
    \begin{subfigure}[b]{0.1\textwidth}
        \subcaption*{Input Image} 
        \adjustbox{trim=10 10 10 10,clip,width=1.6cm,height=1.6cm}{\includegraphics{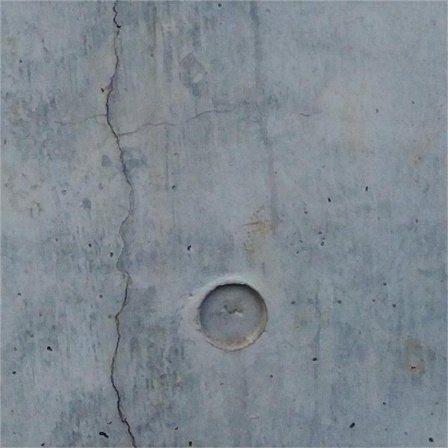}}
    \end{subfigure}
    \begin{subfigure}[b]{0.1\textwidth}
        \subcaption*{GroundTruth} 
        \adjustbox{trim=10 10 10 10,clip,width=1.6cm,height=1.6cm}{\includegraphics{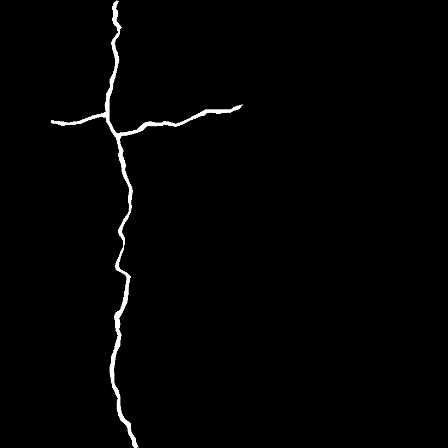}}
    \end{subfigure}
    \begin{subfigure}[b]{0.1\textwidth}
        \subcaption*{SAC} 
        \adjustbox{trim=10 10 10 10,clip,width=1.6cm,height=1.6cm}{\includegraphics{figures/facade390_zero_shot/1/DJ_Wall_2_mask.jpg}}
    \end{subfigure}
    \begin{subfigure}[b]{0.1\textwidth}
        \subcaption*{DeepCrackL} 
        \adjustbox{trim=10 10 10 10,clip,width=1.6cm,height=1.6cm}{\includegraphics{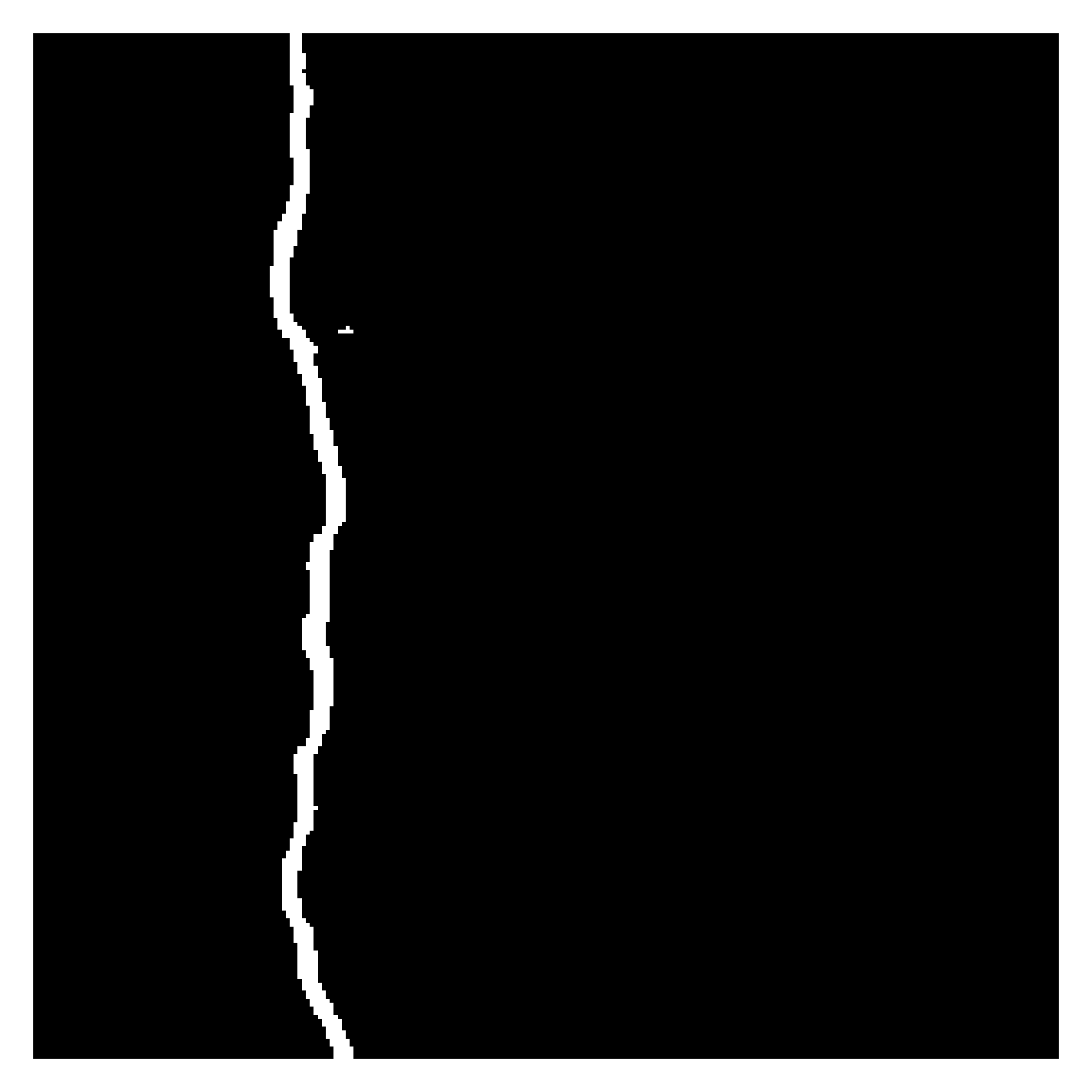}}
    \end{subfigure}
    \begin{subfigure}[b]{0.1\textwidth}
        \subcaption*{CrackFormer} 
        \adjustbox{trim=10 10 10 10,clip,width=1.6cm,height=1.6cm}{\includegraphics{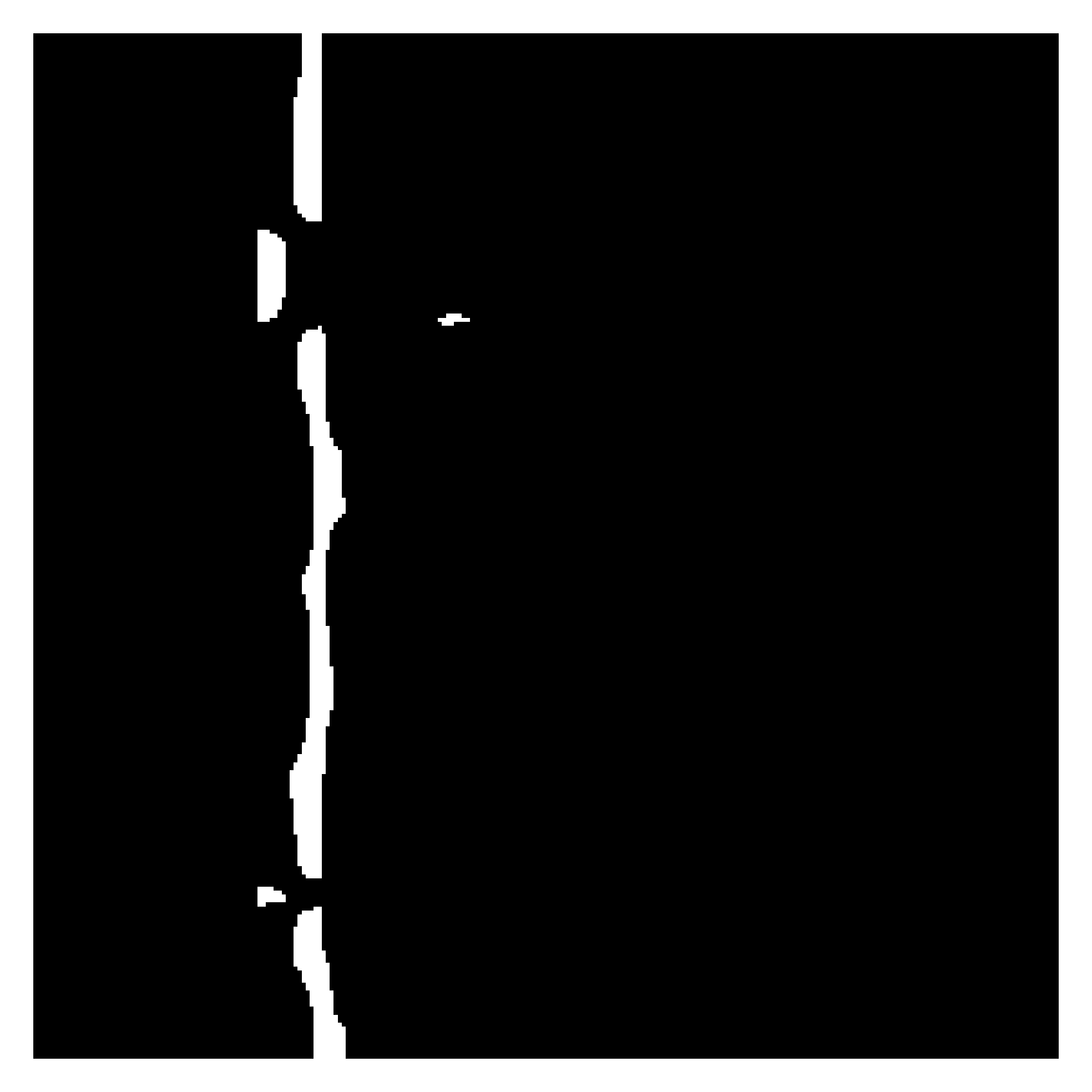}}
    \end{subfigure}
    \begin{subfigure}[b]{0.1\textwidth}
        \subcaption*{SegFormer} 
        \adjustbox{trim=10 10 10 10,clip,width=1.6cm,height=1.6cm}{\includegraphics{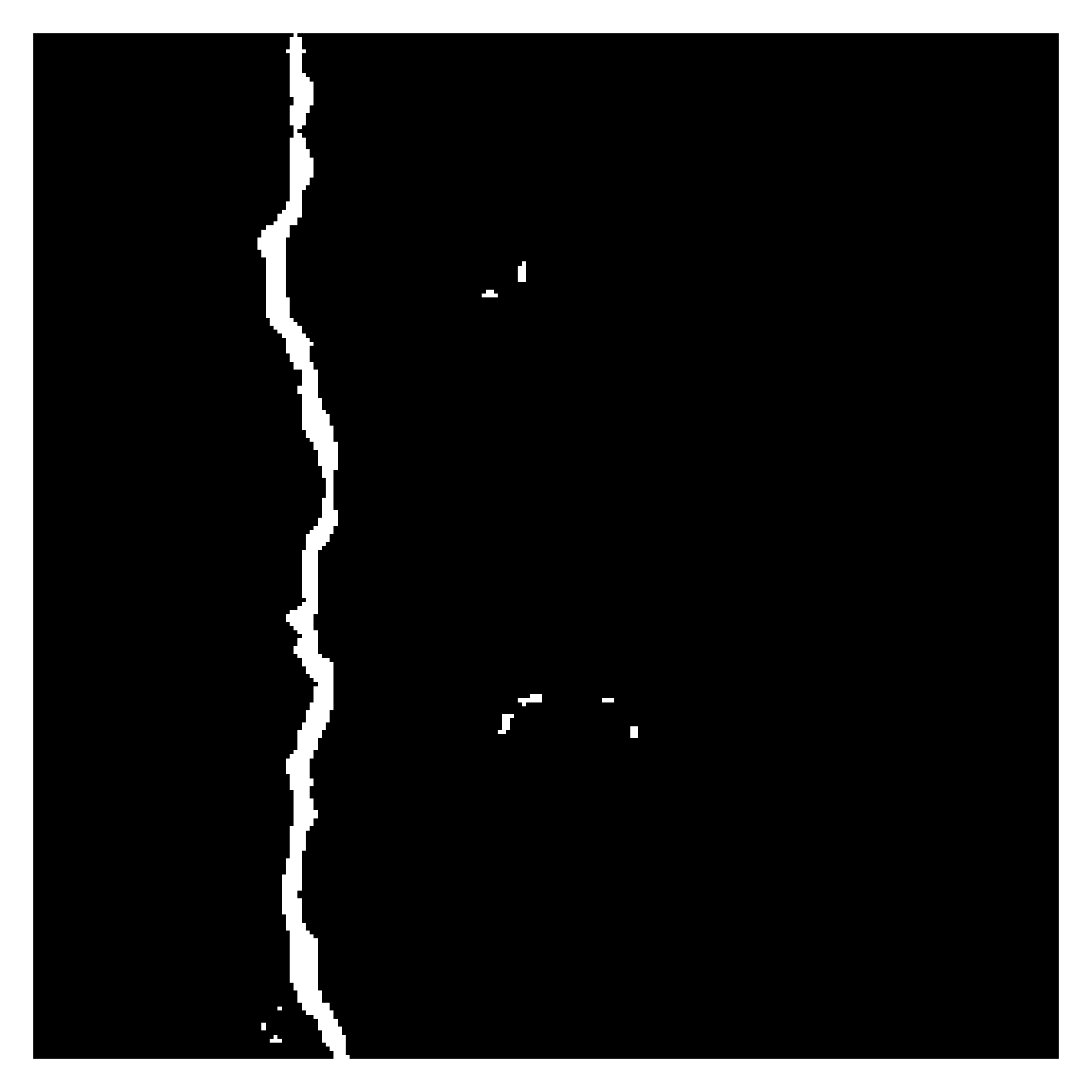}}
    \end{subfigure}
    \begin{subfigure}[b]{0.1\textwidth}
        \subcaption*{U-Net} 
        \adjustbox{trim=10 10 10 10,clip,width=1.6cm,height=1.6cm}{\includegraphics{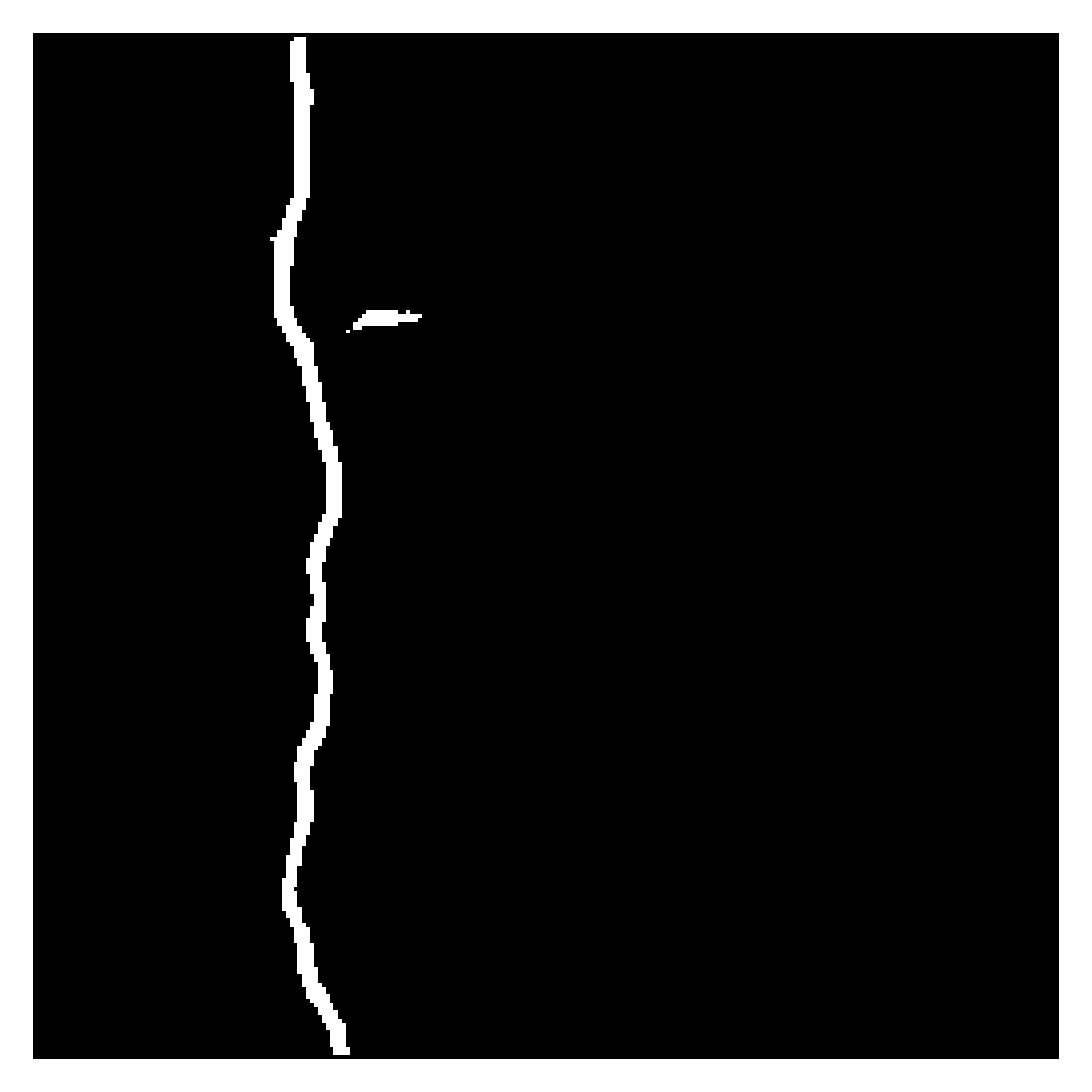}}
    \end{subfigure}
    \begin{subfigure}[b]{0.1\textwidth}
        \subcaption*{DeepLabv3+} 
        \adjustbox{trim=10 10 10 10,clip,width=1.6cm,height=1.6cm}{\includegraphics{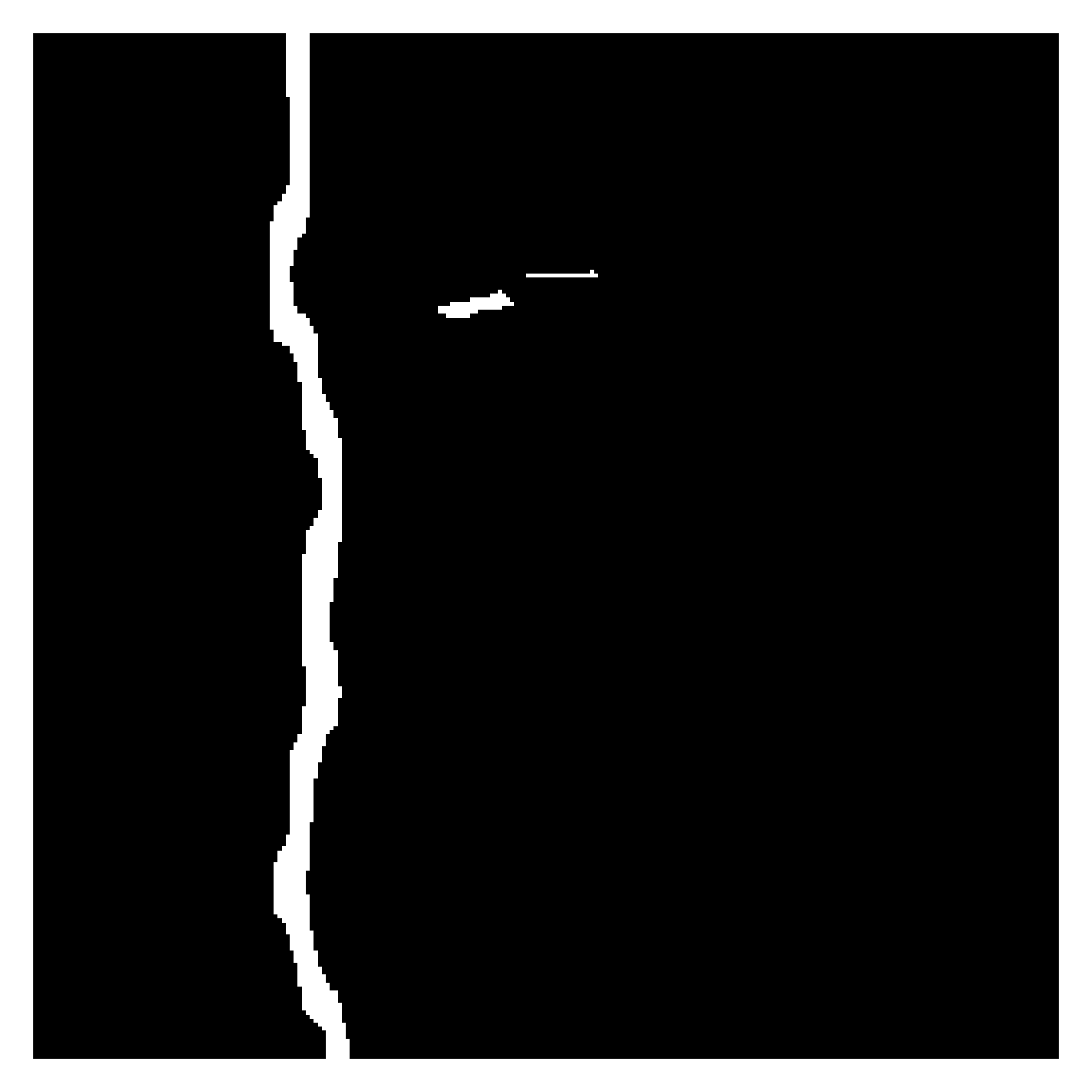}}
    \end{subfigure}

    \vspace{0.2cm}
    
    \begin{subfigure}[b]{0.1\textwidth}
        \adjustbox{trim=10 10 10 10,clip,width=1.6cm,height=1.6cm}{\includegraphics{figures/facade390_zero_shot/1/DJ_Wall_2.jpg}}
    \end{subfigure}
    \begin{subfigure}[b]{0.1\textwidth}
        \adjustbox{trim=10 10 10 10,clip,width=1.6cm,height=1.6cm}{\includegraphics{figures/facade390_zero_shot/1/DJ_Wall_2_mask.jpg}}
    \end{subfigure}  
    \begin{subfigure}[b]{0.1\textwidth}
        \adjustbox{trim=10 10 10 10,clip,width=1.6cm,height=1.6cm}{\includegraphics{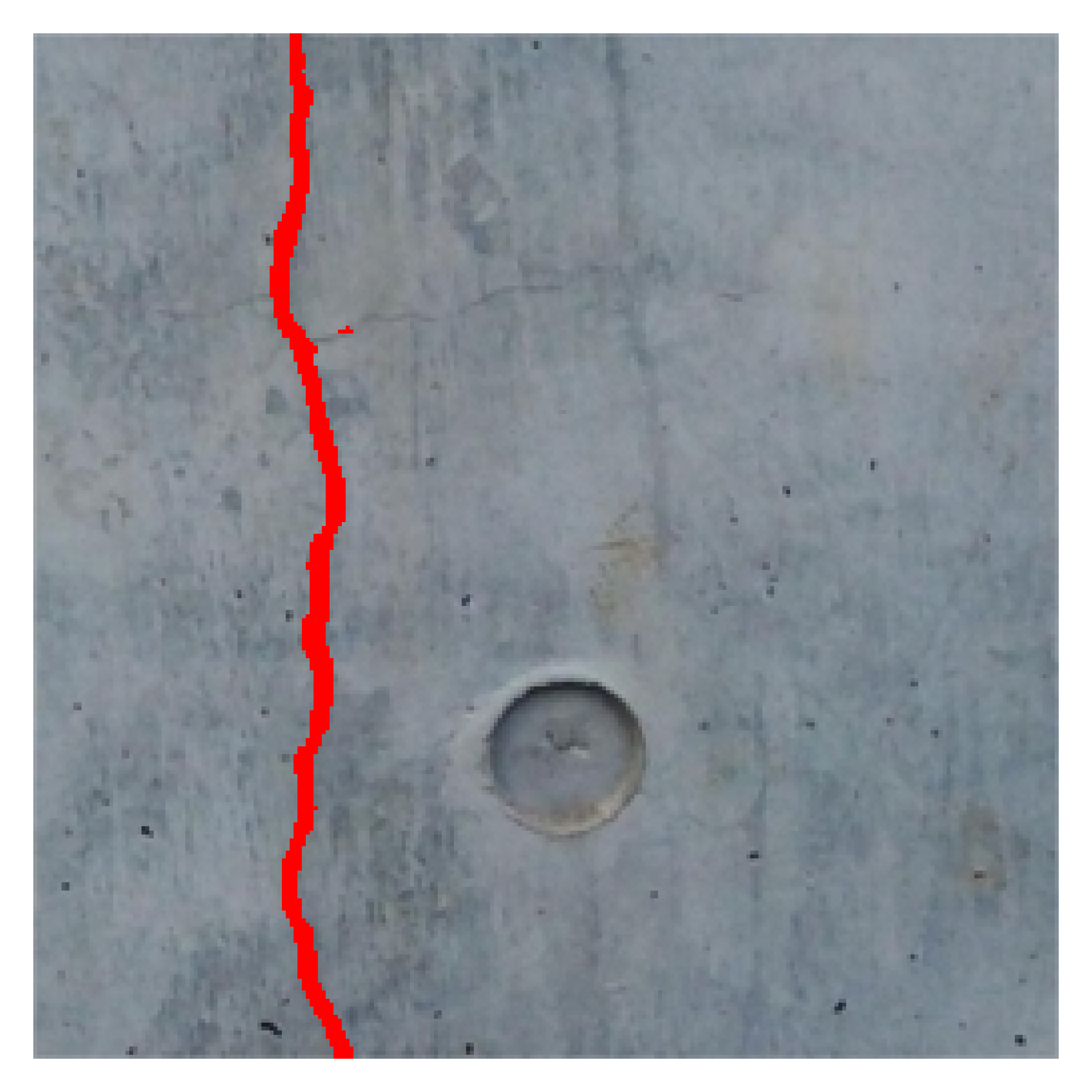}}
    \end{subfigure}
    \begin{subfigure}[b]{0.1\textwidth}
        \adjustbox{trim=10 10 10 10,clip,width=1.6cm,height=1.6cm}{\includegraphics{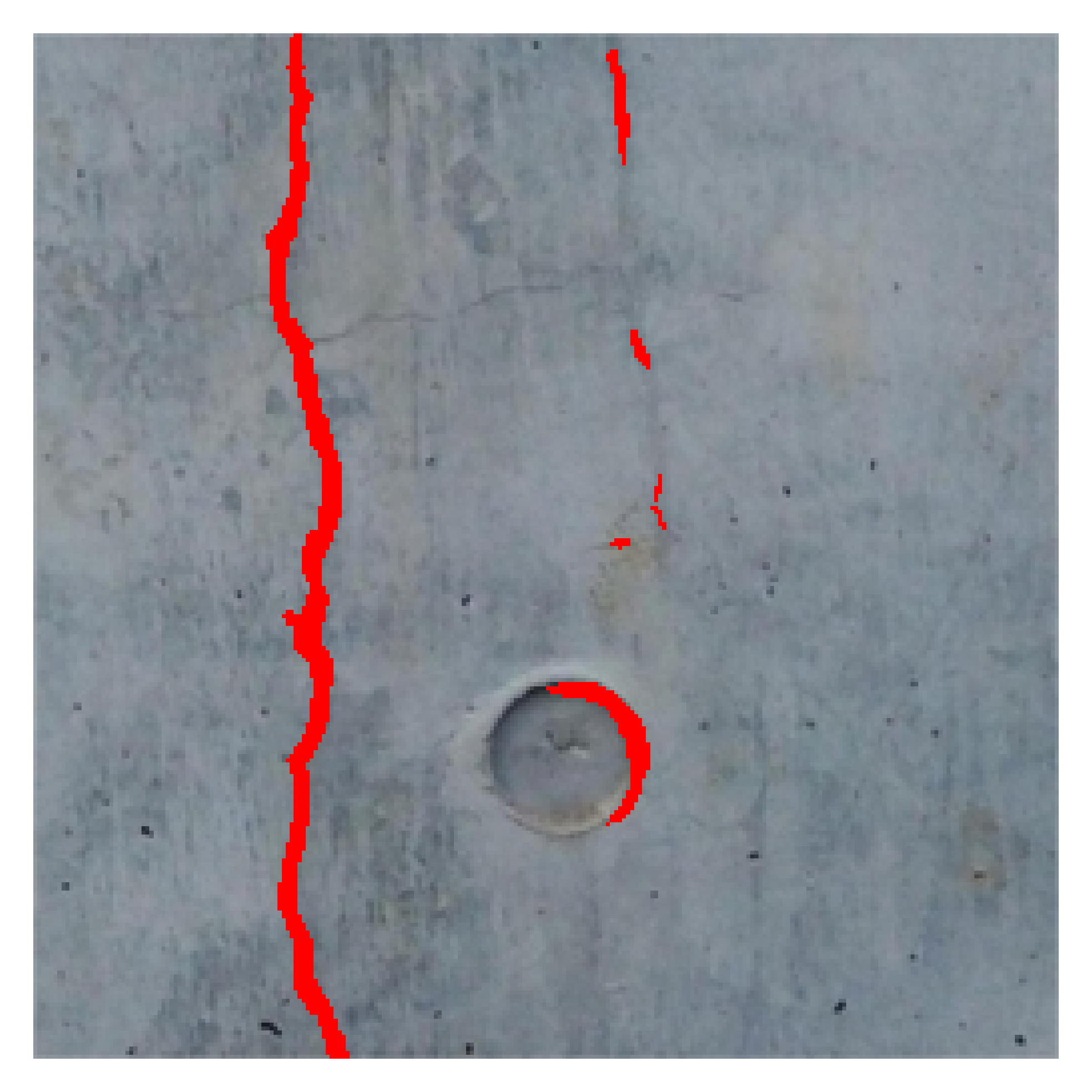}}
    \end{subfigure}
    \begin{subfigure}[b]{0.1\textwidth}
        \adjustbox{trim=10 10 10 10,clip,width=1.6cm,height=1.6cm}{\includegraphics{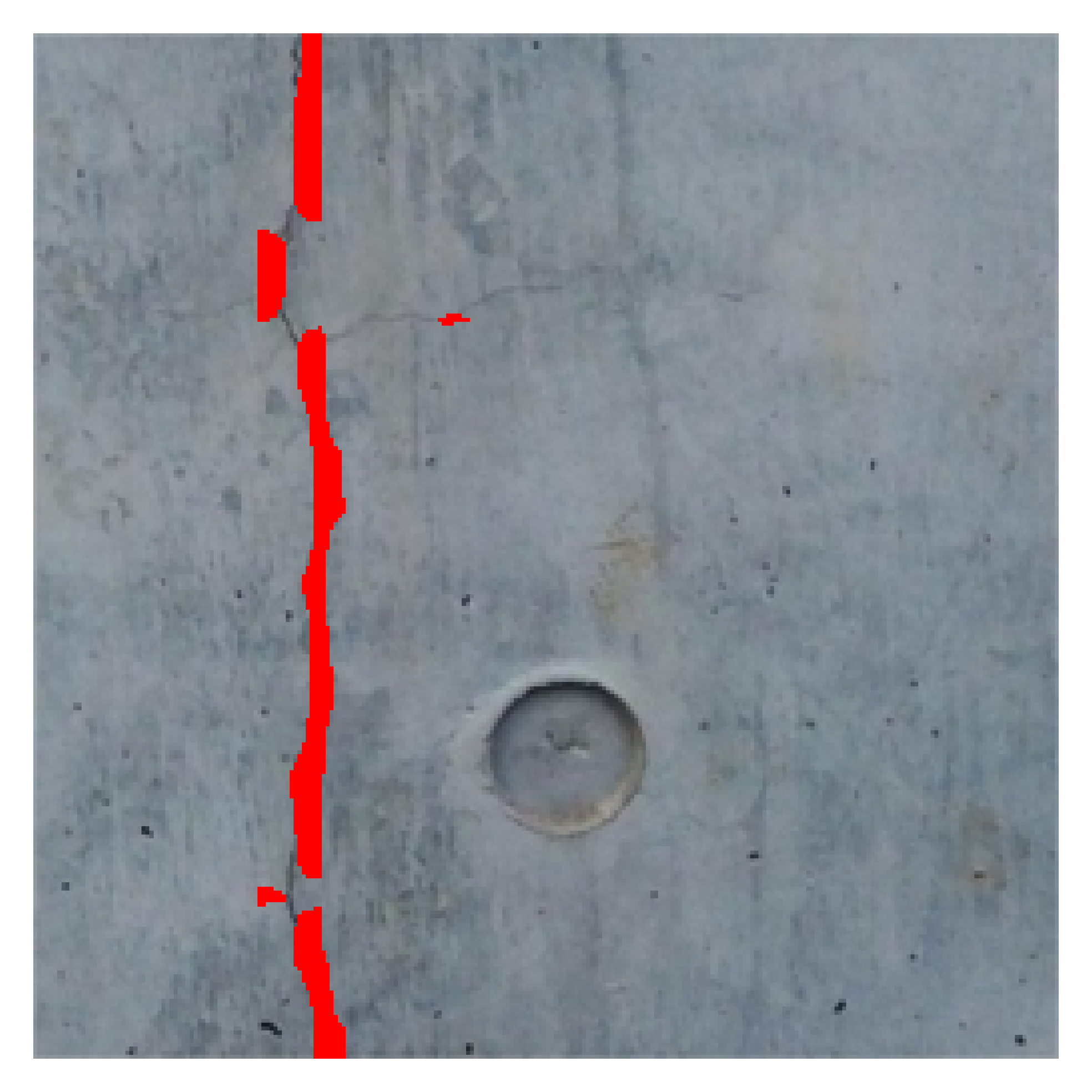}}
    \end{subfigure}
    \begin{subfigure}[b]{0.1\textwidth}
        \adjustbox{trim=10 10 10 10,clip,width=1.6cm,height=1.6cm}{\includegraphics{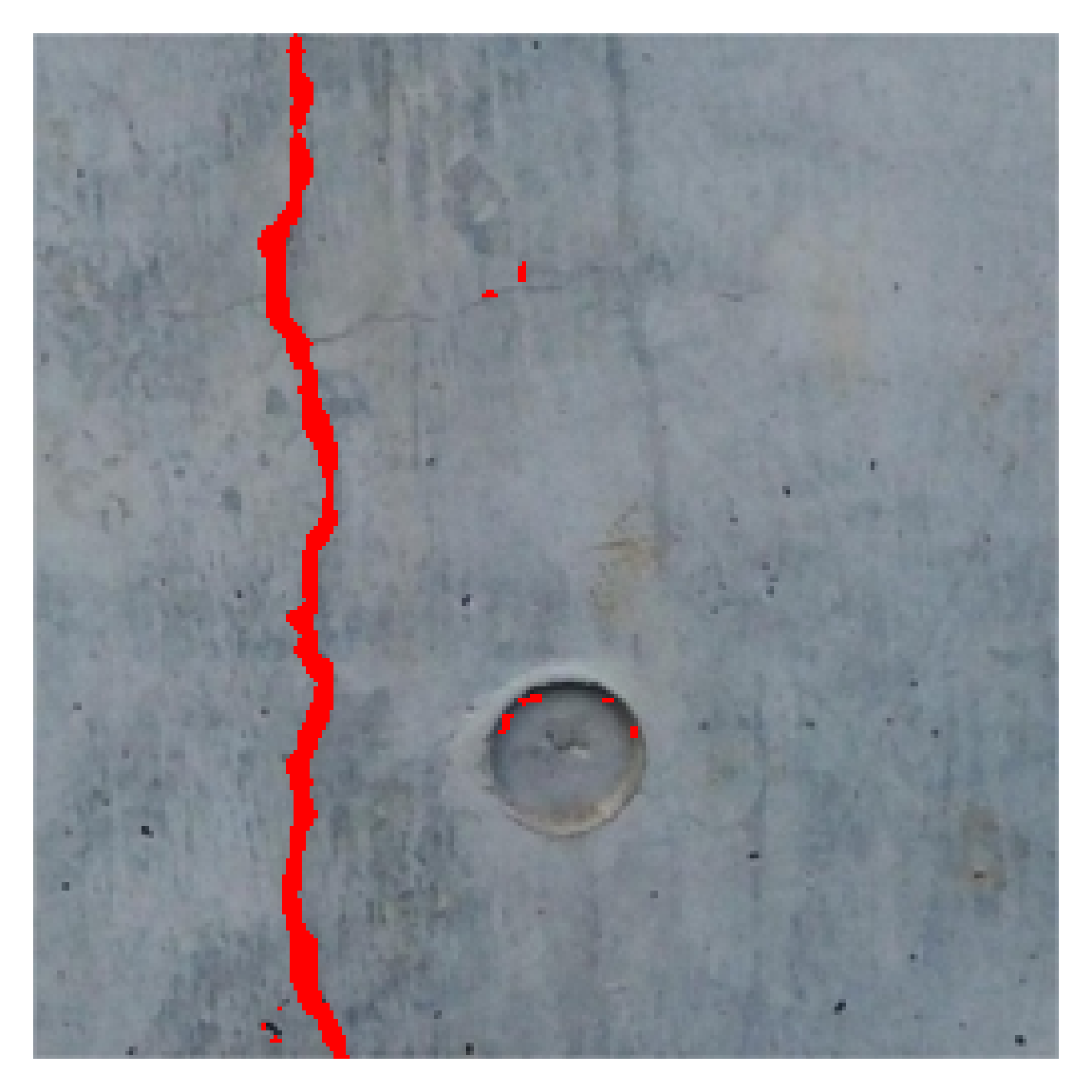}}
    \end{subfigure}
    \begin{subfigure}[b]{0.1\textwidth}
        \adjustbox{trim=10 10 10 10,clip,width=1.6cm,height=1.6cm}{\includegraphics{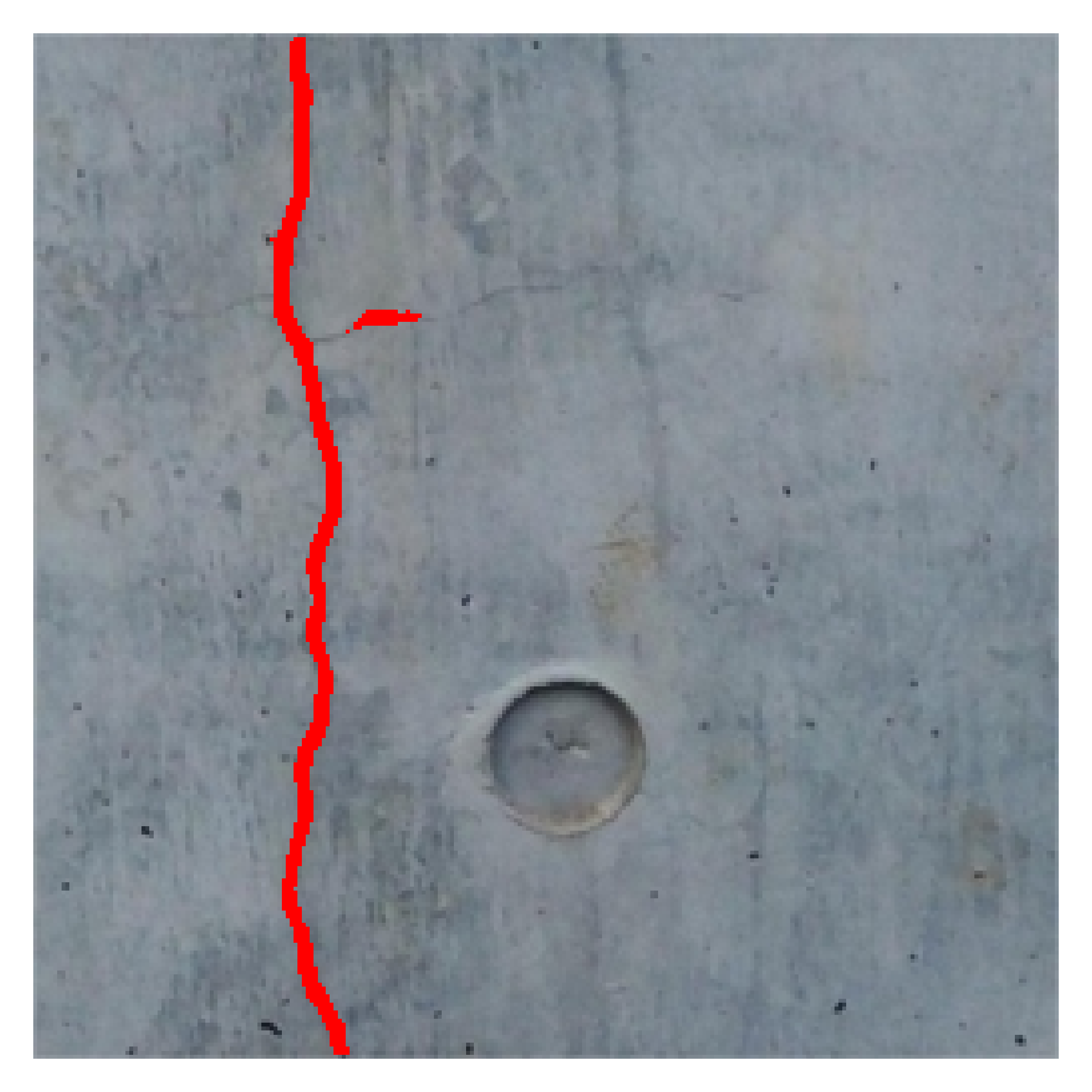}}
    \end{subfigure}
    \begin{subfigure}[b]{0.1\textwidth}
        \adjustbox{trim=10 10 10 10,clip,width=1.6cm,height=1.6cm}{\includegraphics{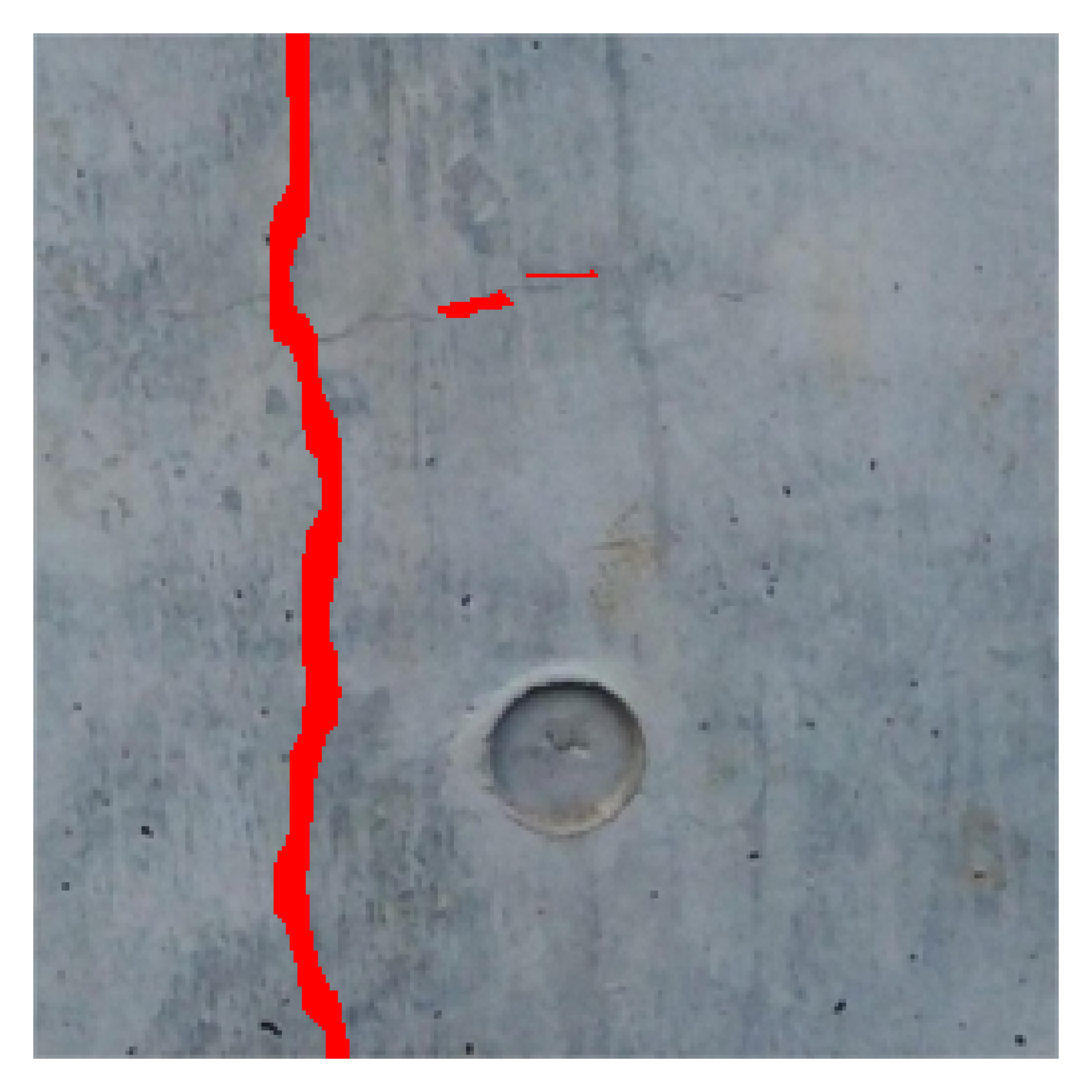}}
    \end{subfigure}

    \vspace{0.5cm}

    \begin{subfigure}[b]{0.1\textwidth}
        \adjustbox{trim=10 10 10 10,clip,width=1.6cm,height=1.6cm}{\includegraphics{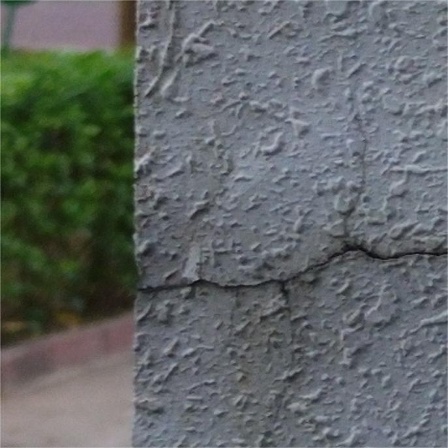}}
    \end{subfigure}
    \begin{subfigure}[b]{0.1\textwidth}
        \adjustbox{trim=10 10 10 10,clip,width=1.6cm,height=1.6cm}{\includegraphics{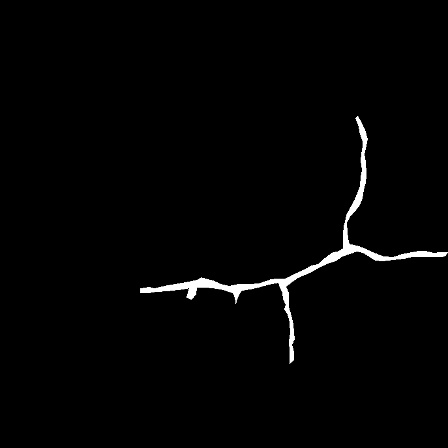}}
    \end{subfigure}  
    \begin{subfigure}[b]{0.1\textwidth}
        \adjustbox{trim=10 10 10 10,clip,width=1.6cm,height=1.6cm}{\includegraphics{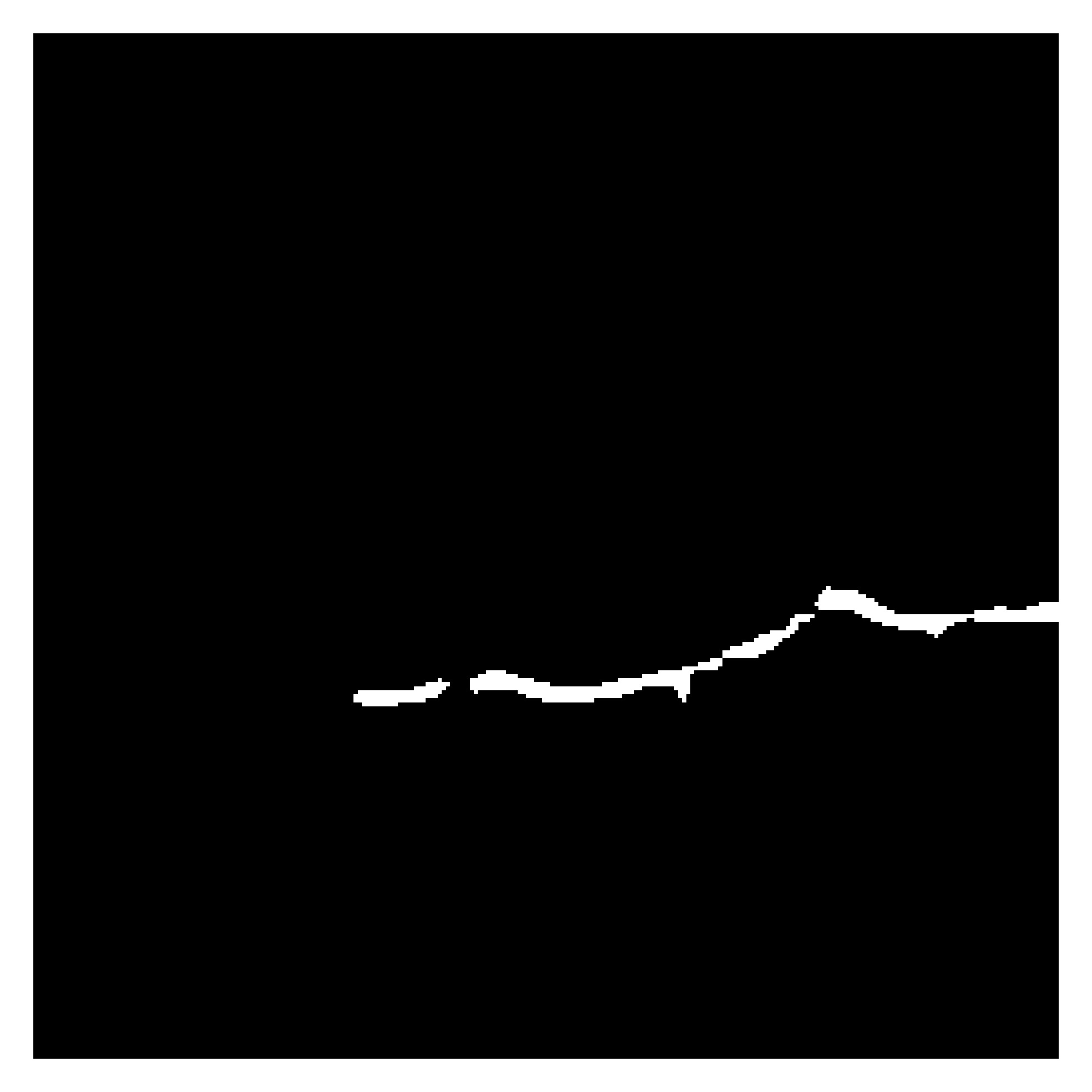}}
    \end{subfigure}
    \begin{subfigure}[b]{0.1\textwidth}
        \adjustbox{trim=10 10 10 10,clip,width=1.6cm,height=1.6cm}{\includegraphics{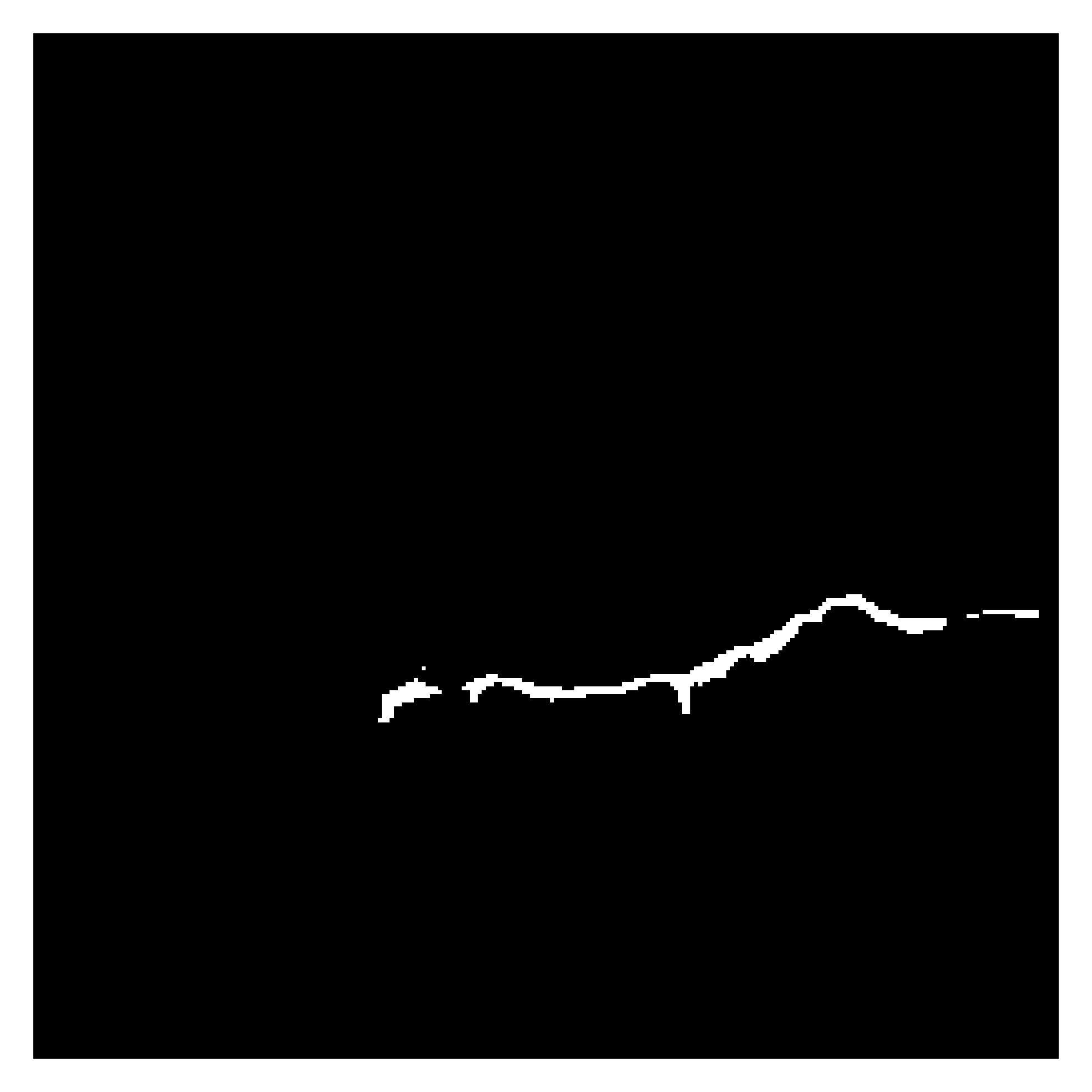}}
    \end{subfigure}
    \begin{subfigure}[b]{0.1\textwidth}
        \adjustbox{trim=10 10 10 10,clip,width=1.6cm,height=1.6cm}{\includegraphics{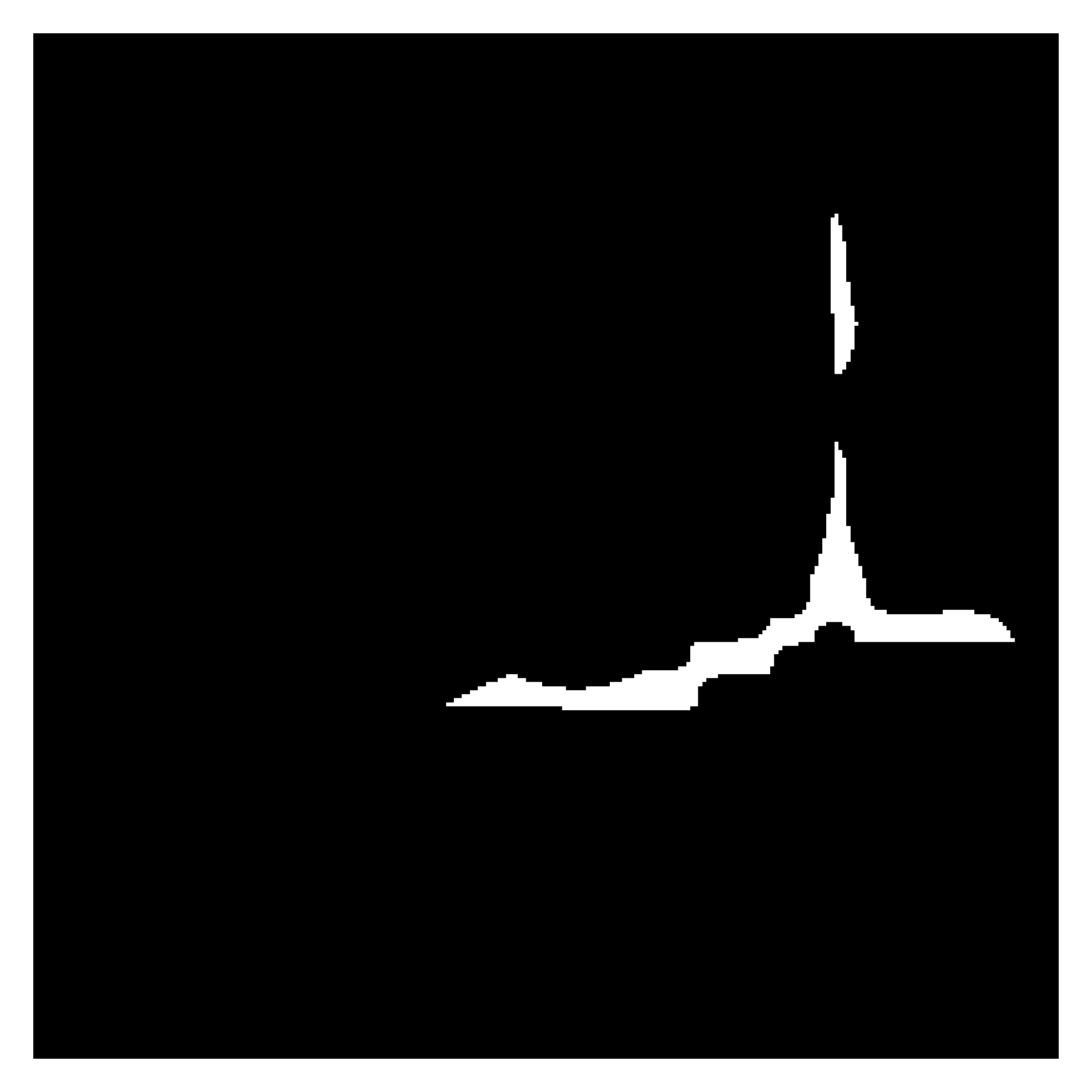}}
    \end{subfigure}
    \begin{subfigure}[b]{0.1\textwidth}
        \adjustbox{trim=10 10 10 10,clip,width=1.6cm,height=1.6cm}{\includegraphics{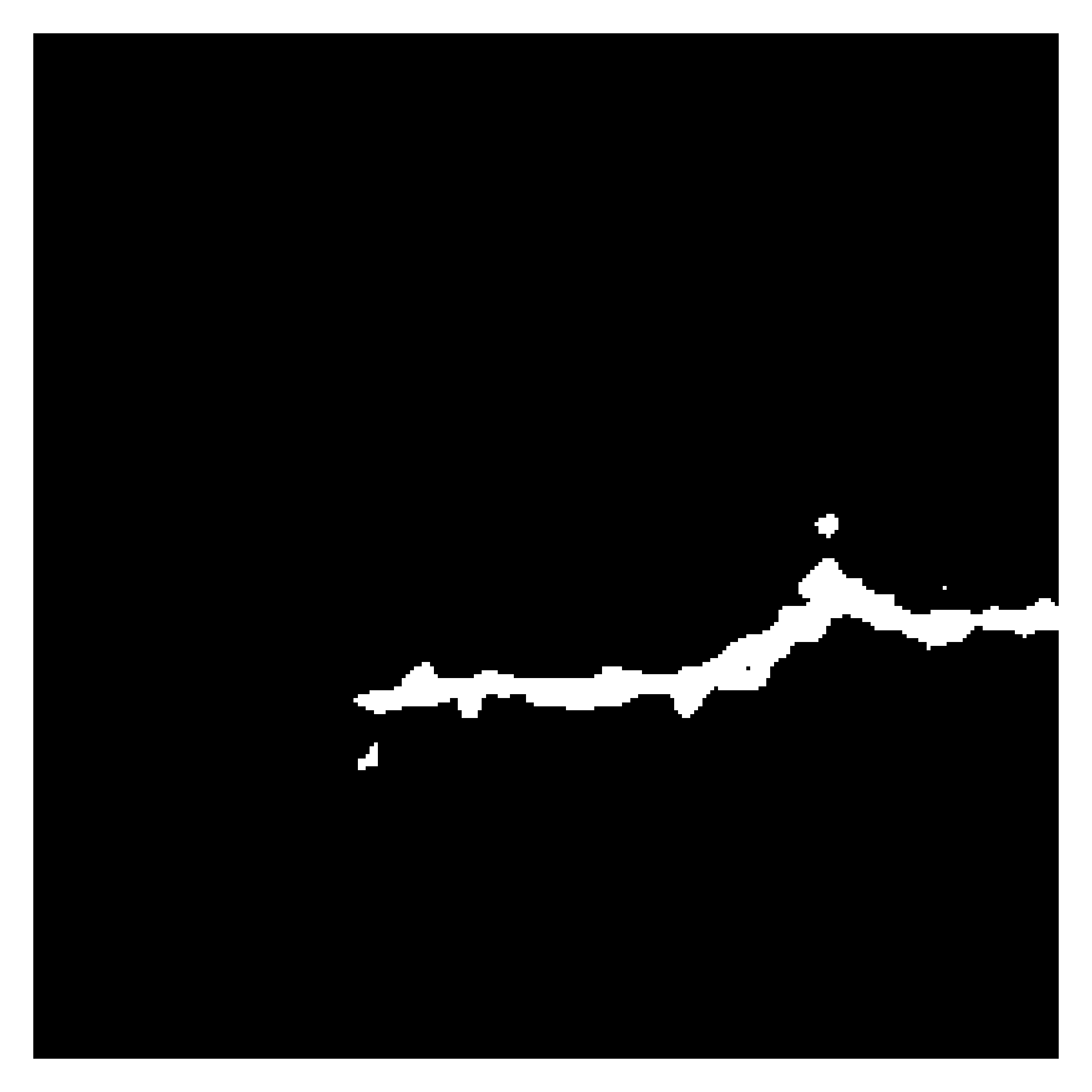}}
    \end{subfigure}
    \begin{subfigure}[b]{0.1\textwidth}
        \adjustbox{trim=10 10 10 10,clip,width=1.6cm,height=1.6cm}{\includegraphics{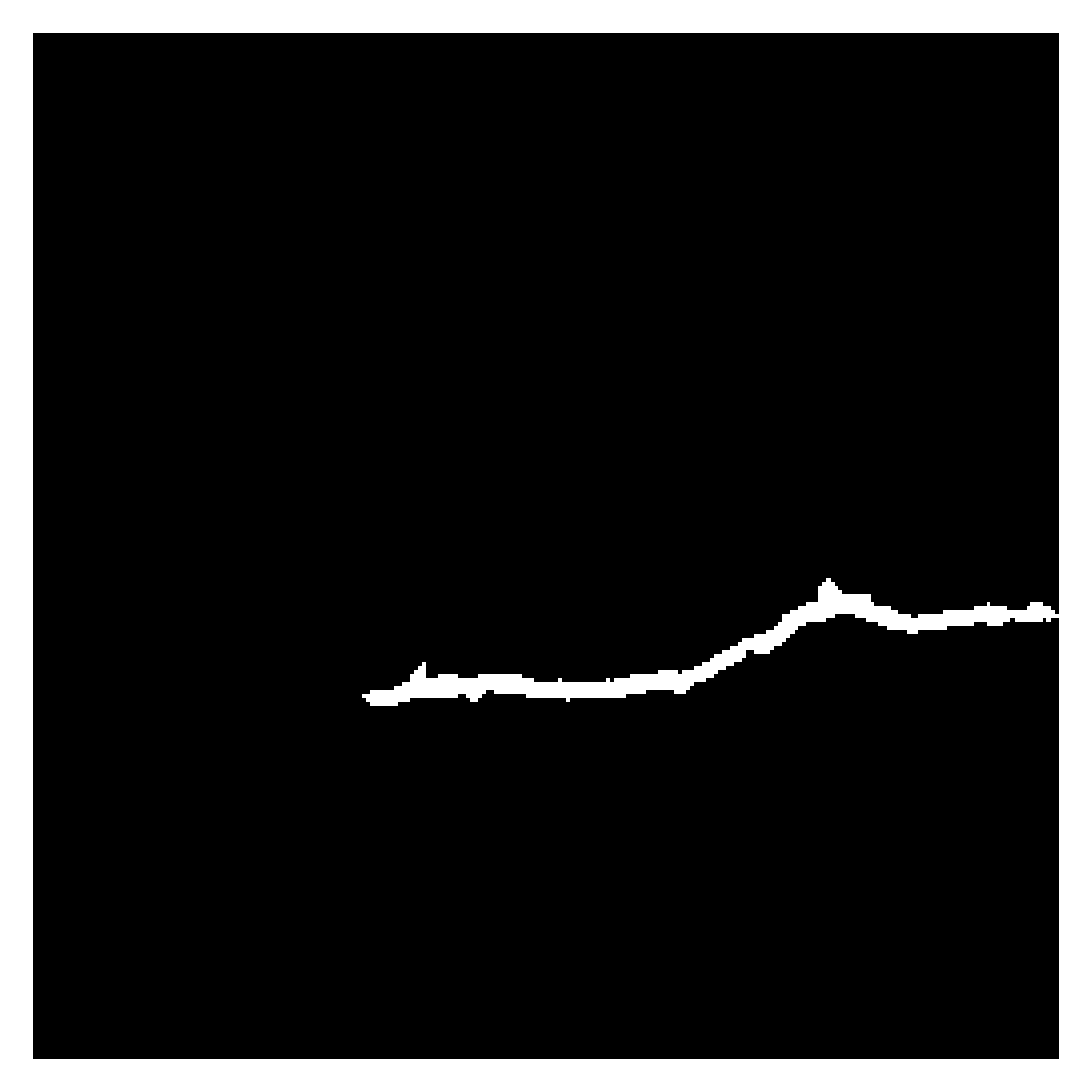}}
    \end{subfigure}
    \begin{subfigure}[b]{0.1\textwidth}
        \adjustbox{trim=10 10 10 10,clip,width=1.6cm,height=1.6cm}{\includegraphics{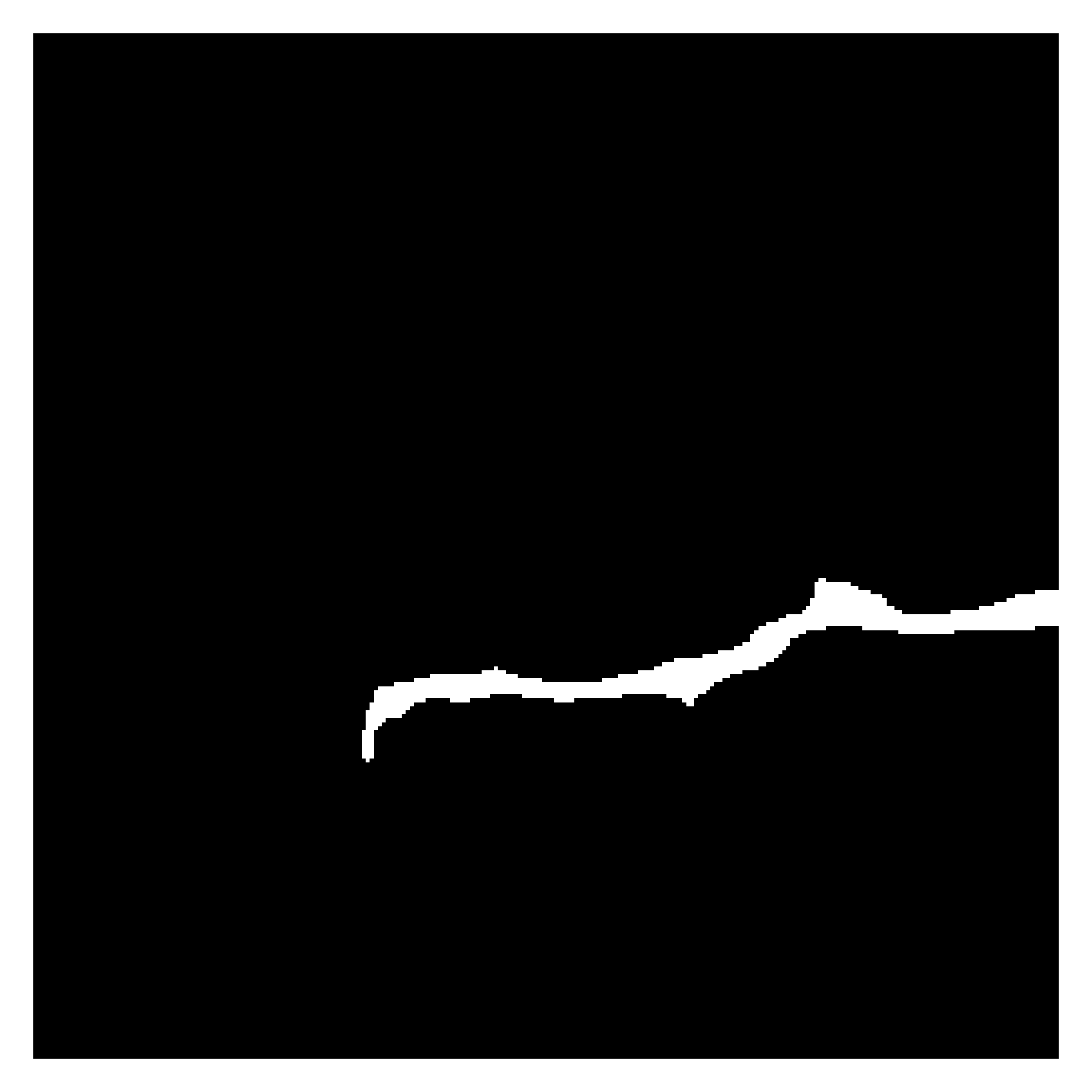}}
    \end{subfigure}

    \vspace{0.2cm}
    
    \begin{subfigure}[b]{0.1\textwidth}
        \adjustbox{trim=10 10 10 10,clip,width=1.6cm,height=1.6cm}{\includegraphics{figures/facade390_zero_shot/2/DJ_Wall_334.jpg}}
    \end{subfigure}
    \begin{subfigure}[b]{0.1\textwidth}
        \adjustbox{trim=10 10 10 10,clip,width=1.6cm,height=1.6cm}{\includegraphics{figures/facade390_zero_shot/2/DJ_Wall_334_mask.jpg}}
    \end{subfigure}  
    \begin{subfigure}[b]{0.1\textwidth}
        \adjustbox{trim=10 10 10 10,clip,width=1.6cm,height=1.6cm}{\includegraphics{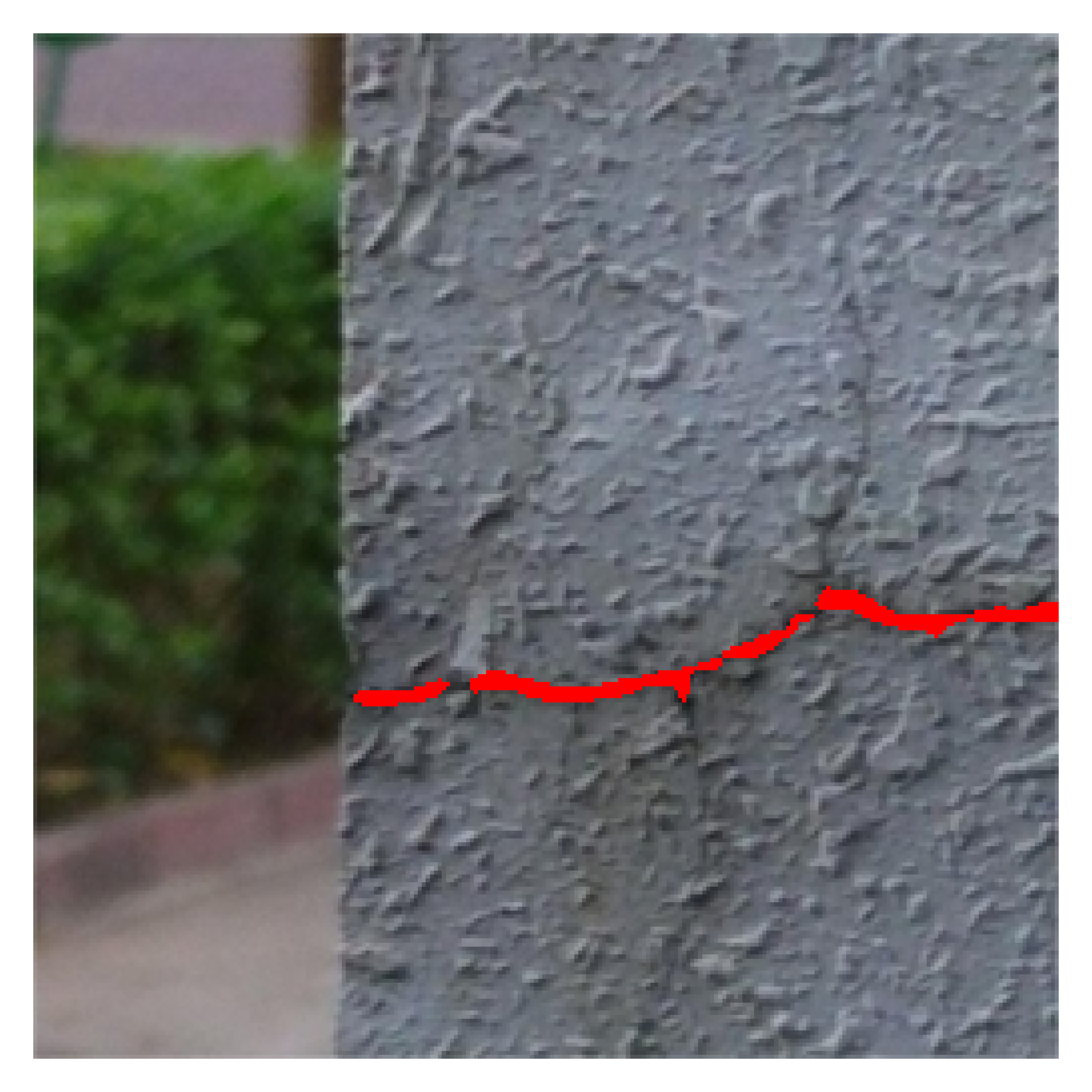}}
    \end{subfigure}
    \begin{subfigure}[b]{0.1\textwidth}
        \adjustbox{trim=10 10 10 10,clip,width=1.6cm,height=1.6cm}{\includegraphics{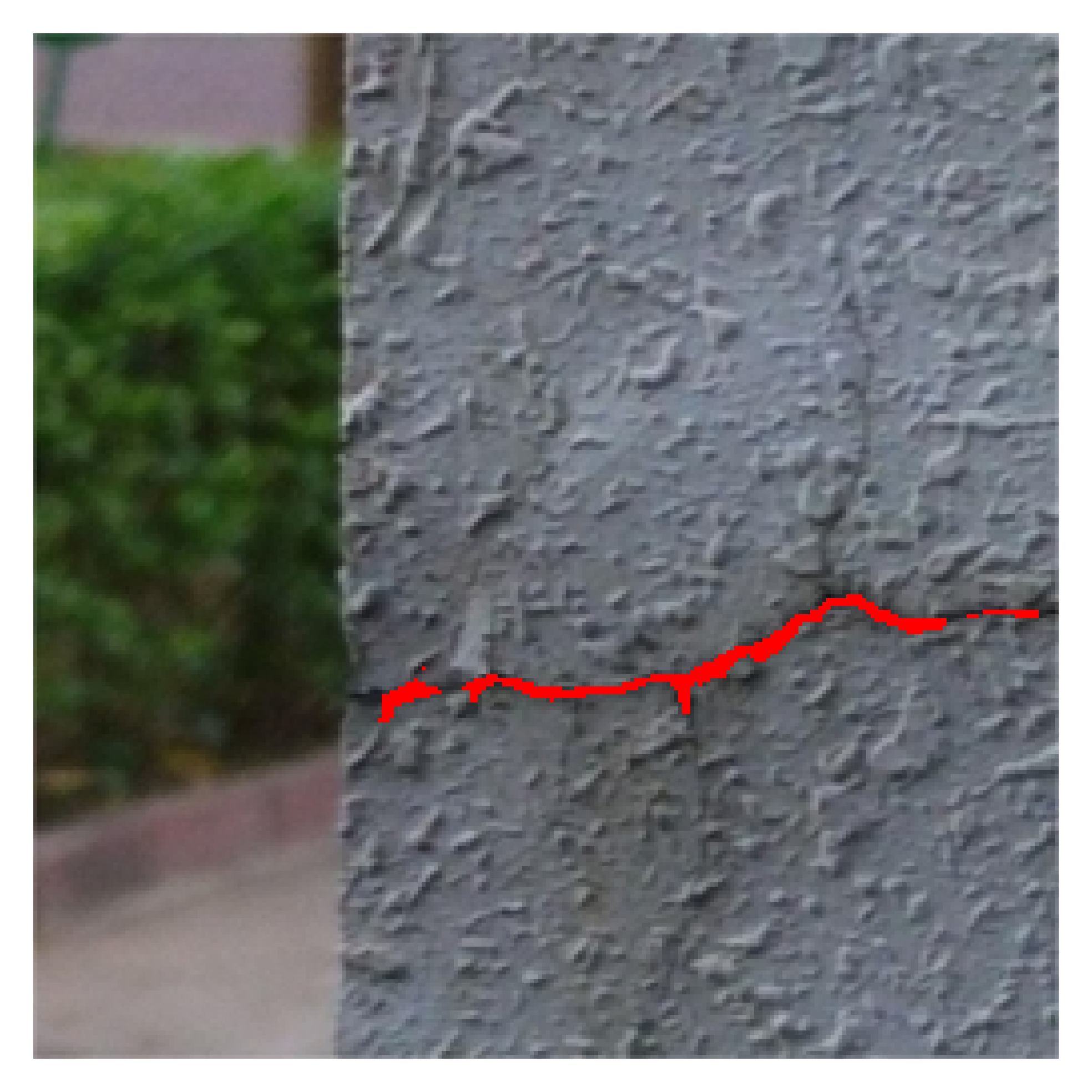}}
    \end{subfigure}
    \begin{subfigure}[b]{0.1\textwidth}
        \adjustbox{trim=10 10 10 10,clip,width=1.6cm,height=1.6cm}{\includegraphics{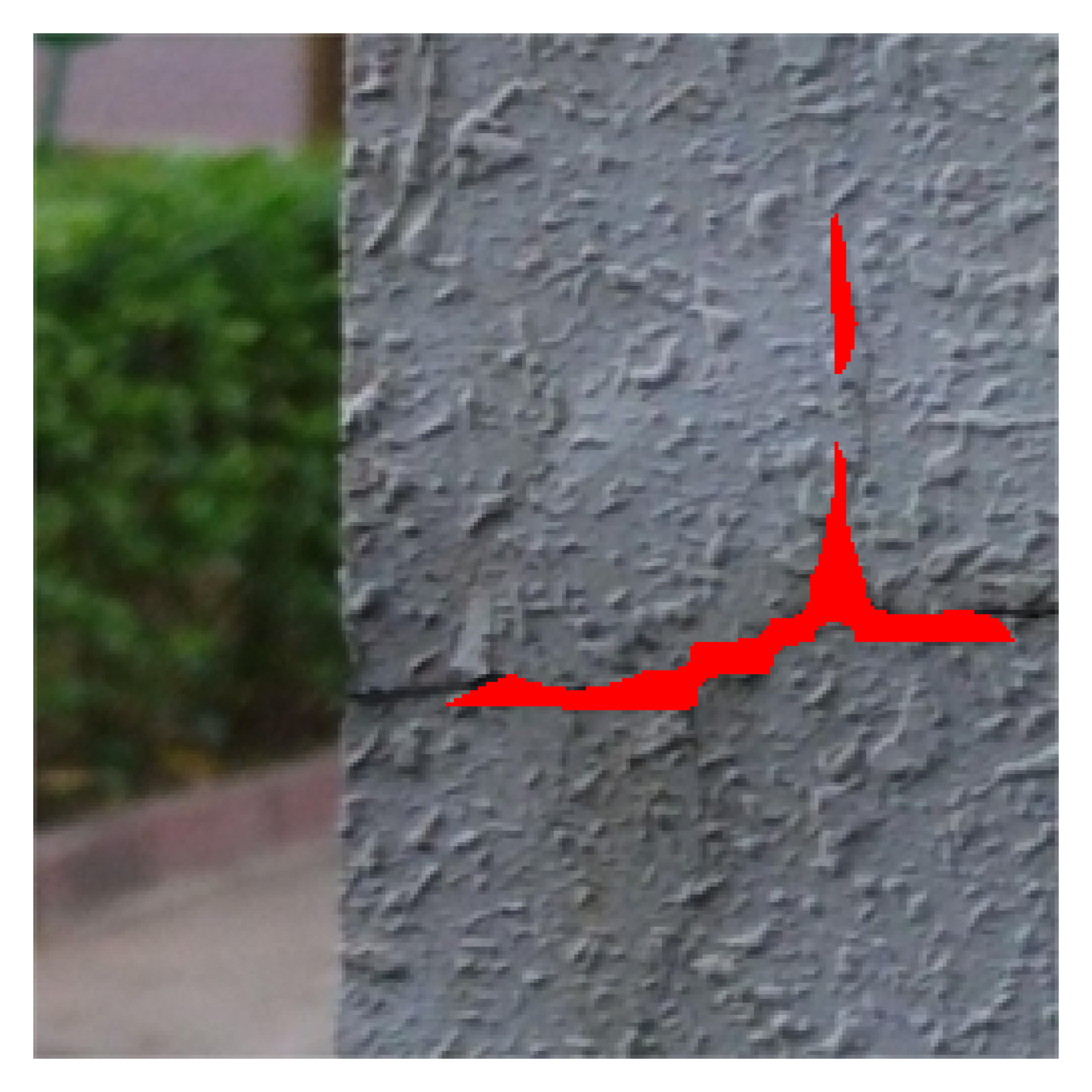}}
    \end{subfigure}
    \begin{subfigure}[b]{0.1\textwidth}
        \adjustbox{trim=10 10 10 10,clip,width=1.6cm,height=1.6cm}{\includegraphics{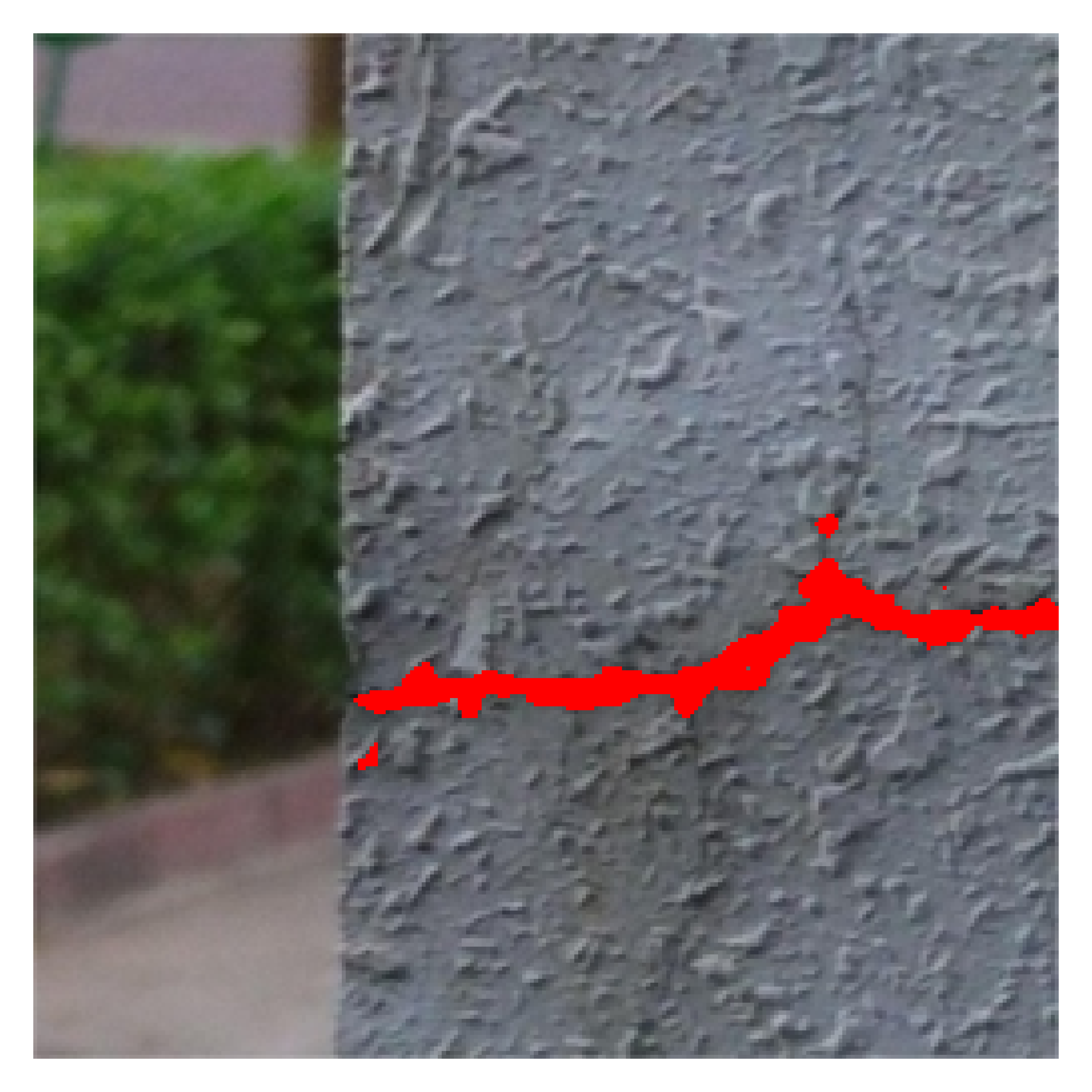}}
    \end{subfigure}
    \begin{subfigure}[b]{0.1\textwidth}
        \adjustbox{trim=10 10 10 10,clip,width=1.6cm,height=1.6cm}{\includegraphics{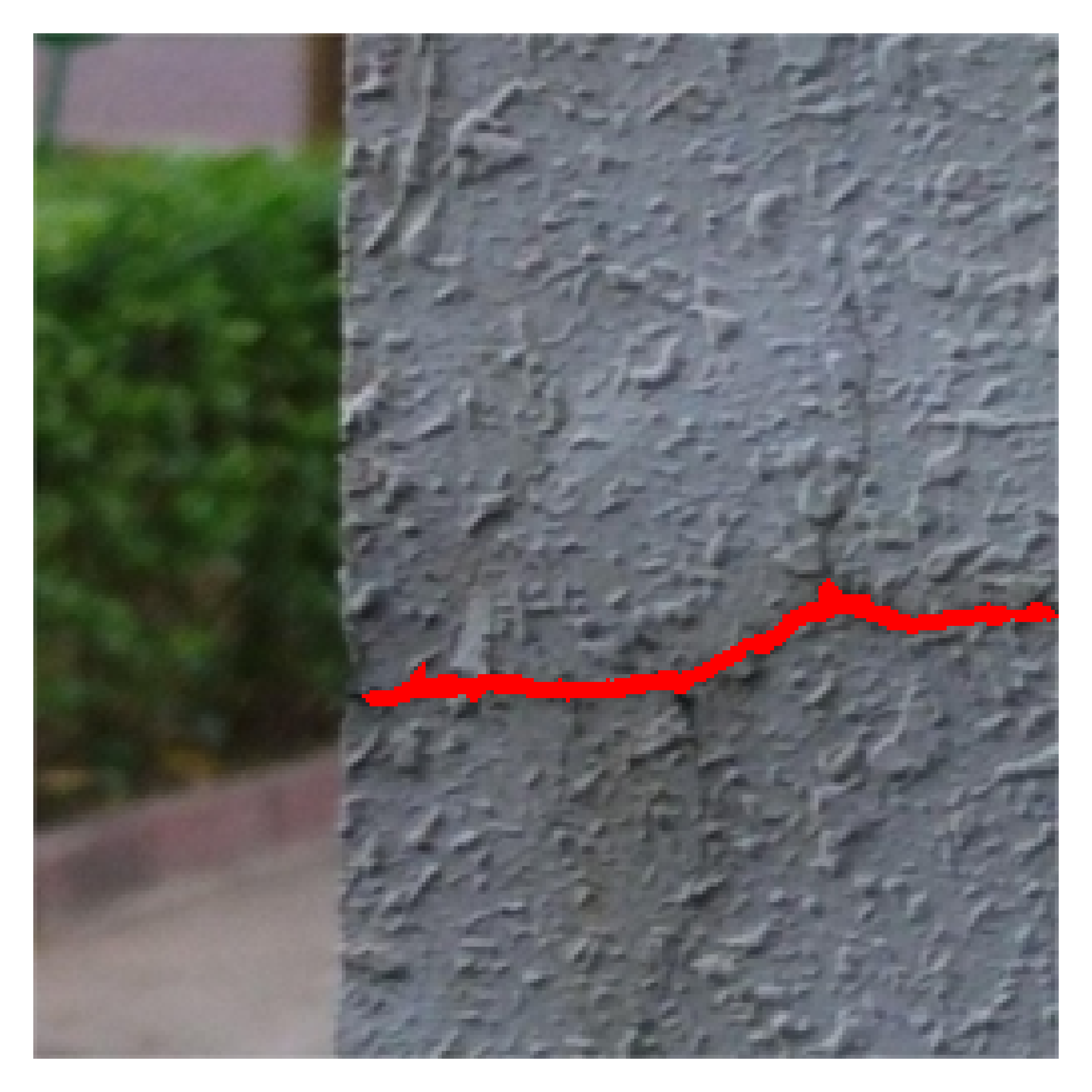}}
    \end{subfigure}
    \begin{subfigure}[b]{0.1\textwidth}
        \adjustbox{trim=10 10 10 10,clip,width=1.6cm,height=1.6cm}{\includegraphics{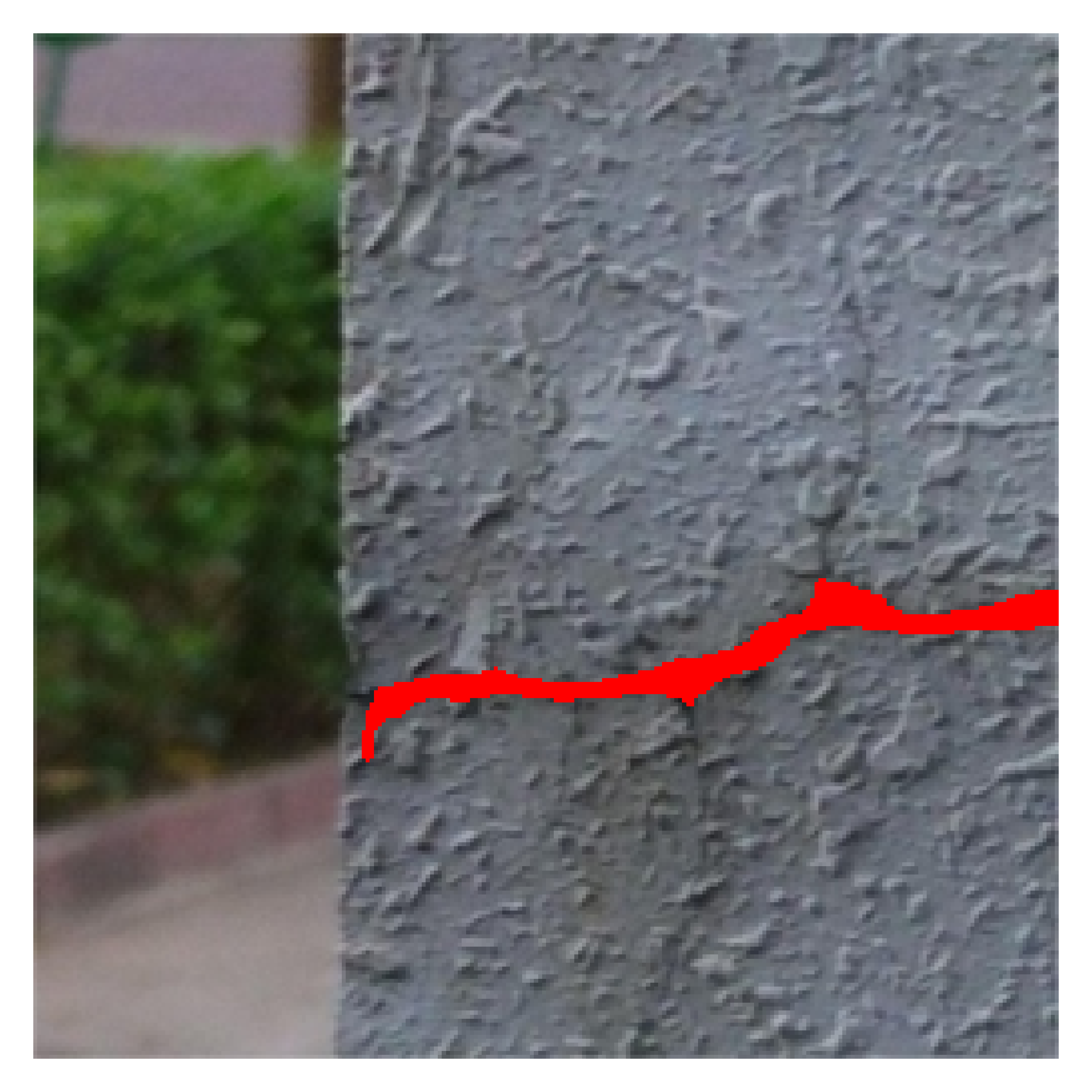}}
    \end{subfigure}

    \vspace{0.5cm}

    \begin{subfigure}[b]{0.1\textwidth}
        \adjustbox{trim=10 10 10 10,clip,width=1.6cm,height=1.6cm}{\includegraphics{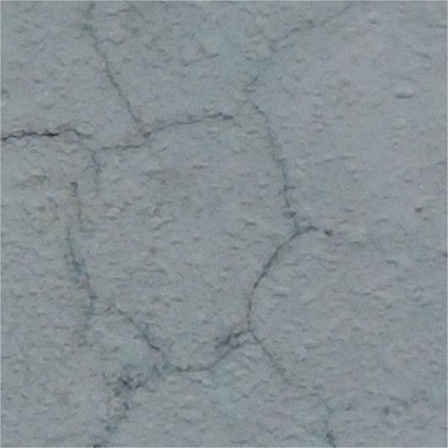}}
    \end{subfigure}
    \begin{subfigure}[b]{0.1\textwidth}
        \adjustbox{trim=10 10 10 10,clip,width=1.6cm,height=1.6cm}{\includegraphics{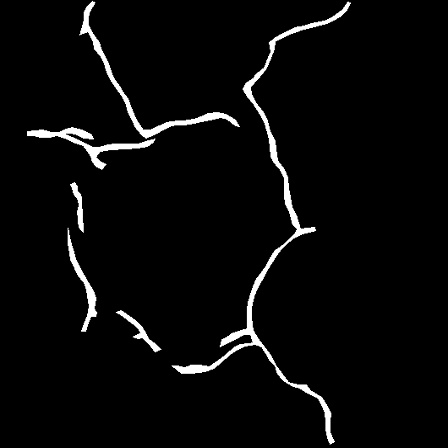}}
    \end{subfigure}  
    \begin{subfigure}[b]{0.1\textwidth}
        \adjustbox{trim=10 10 10 10,clip,width=1.6cm,height=1.6cm}{\includegraphics{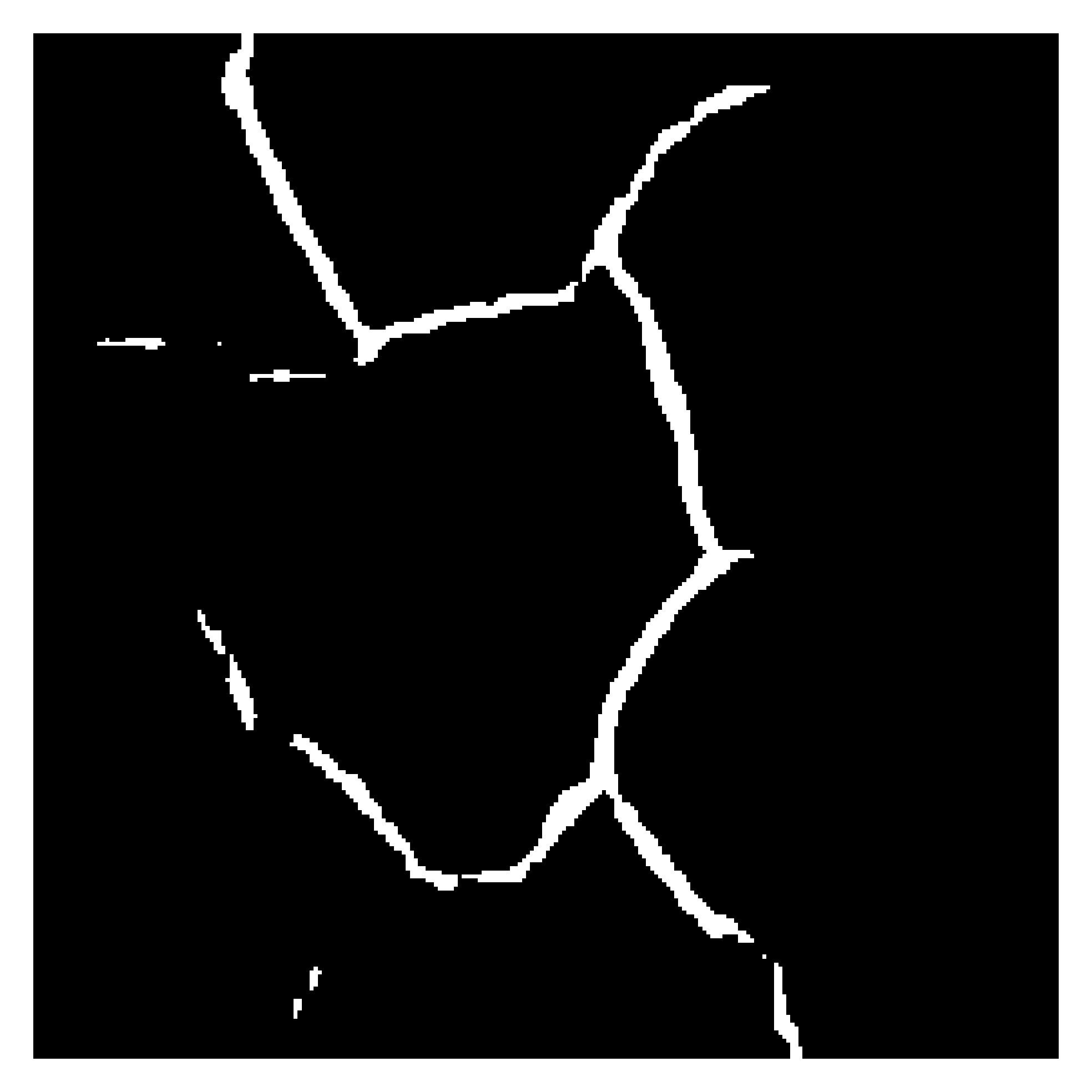}}
    \end{subfigure}
    \begin{subfigure}[b]{0.1\textwidth}
        \adjustbox{trim=10 10 10 10,clip,width=1.6cm,height=1.6cm}{\includegraphics{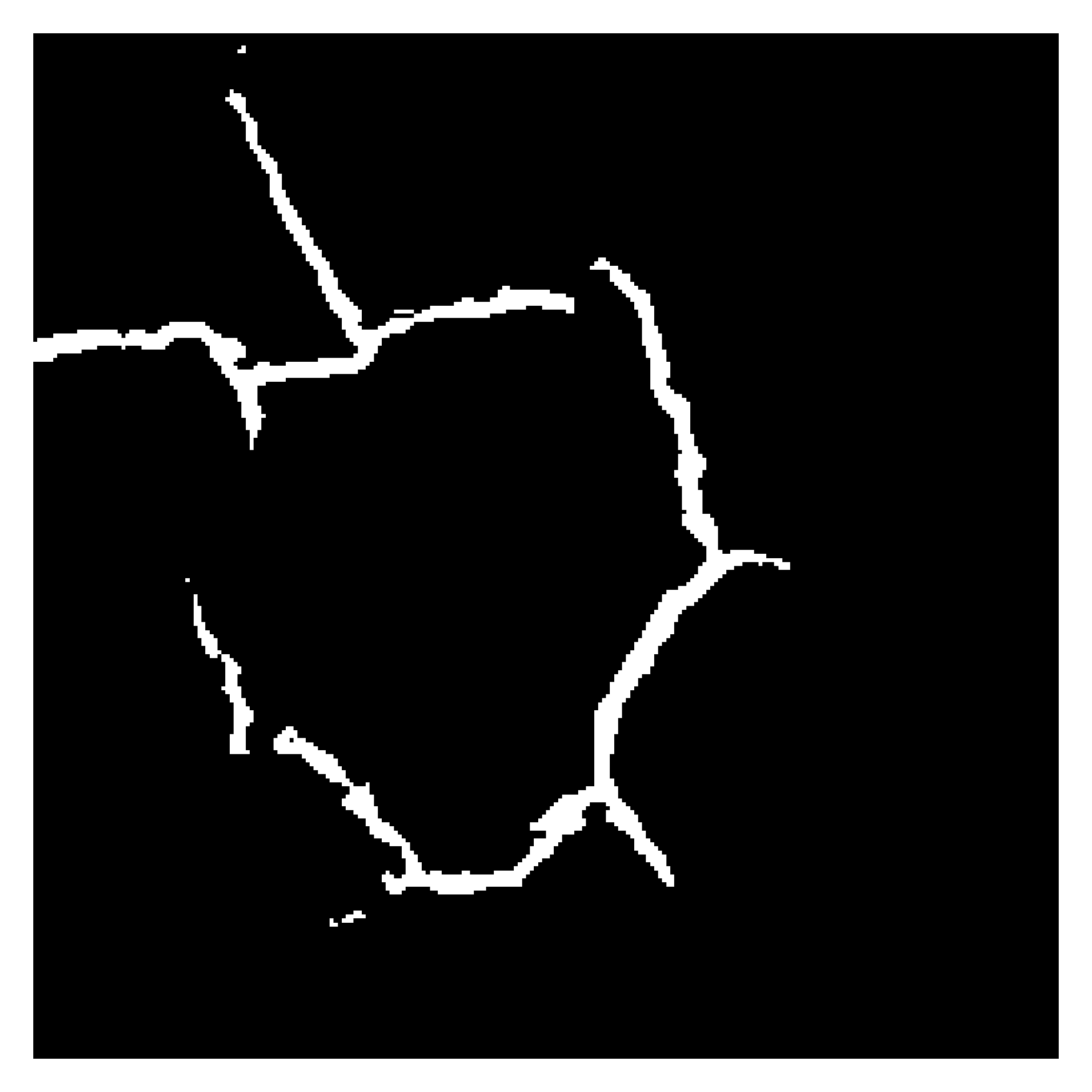}}
    \end{subfigure}
    \begin{subfigure}[b]{0.1\textwidth}
        \adjustbox{trim=10 10 10 10,clip,width=1.6cm,height=1.6cm}{\includegraphics{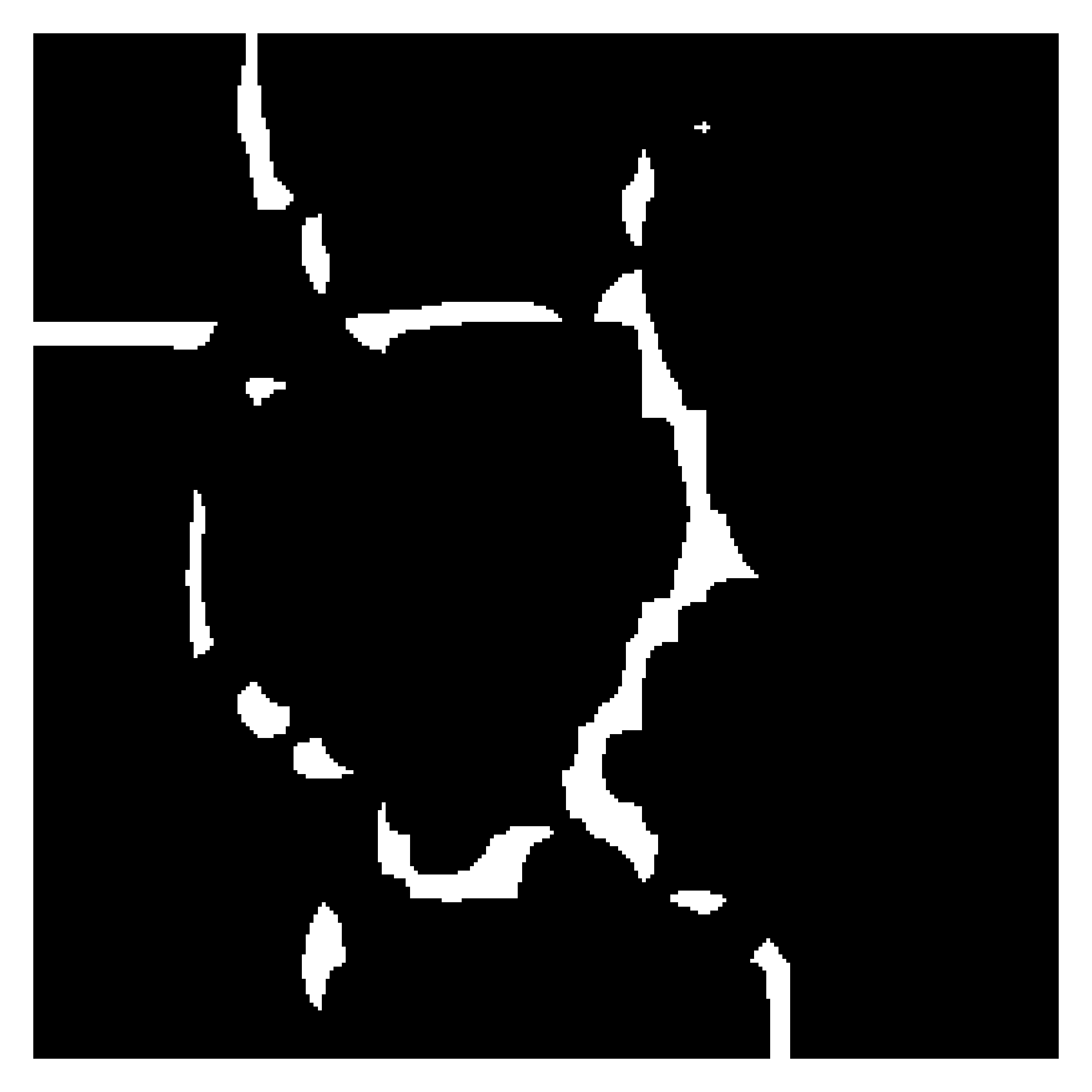}}
    \end{subfigure}
    \begin{subfigure}[b]{0.1\textwidth}
        \adjustbox{trim=10 10 10 10,clip,width=1.6cm,height=1.6cm}{\includegraphics{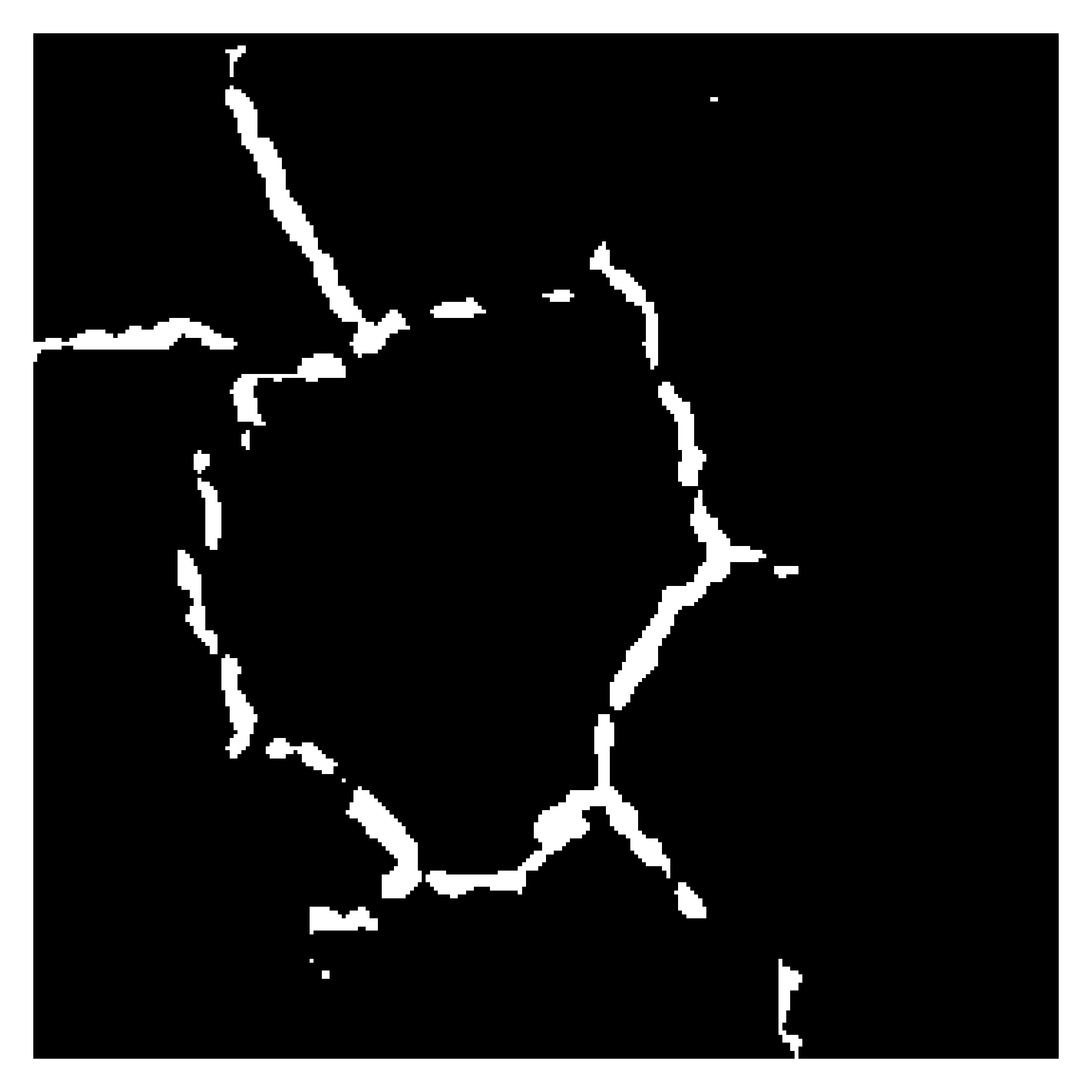}}
    \end{subfigure}
    \begin{subfigure}[b]{0.1\textwidth}
        \adjustbox{trim=10 10 10 10,clip,width=1.6cm,height=1.6cm}{\includegraphics{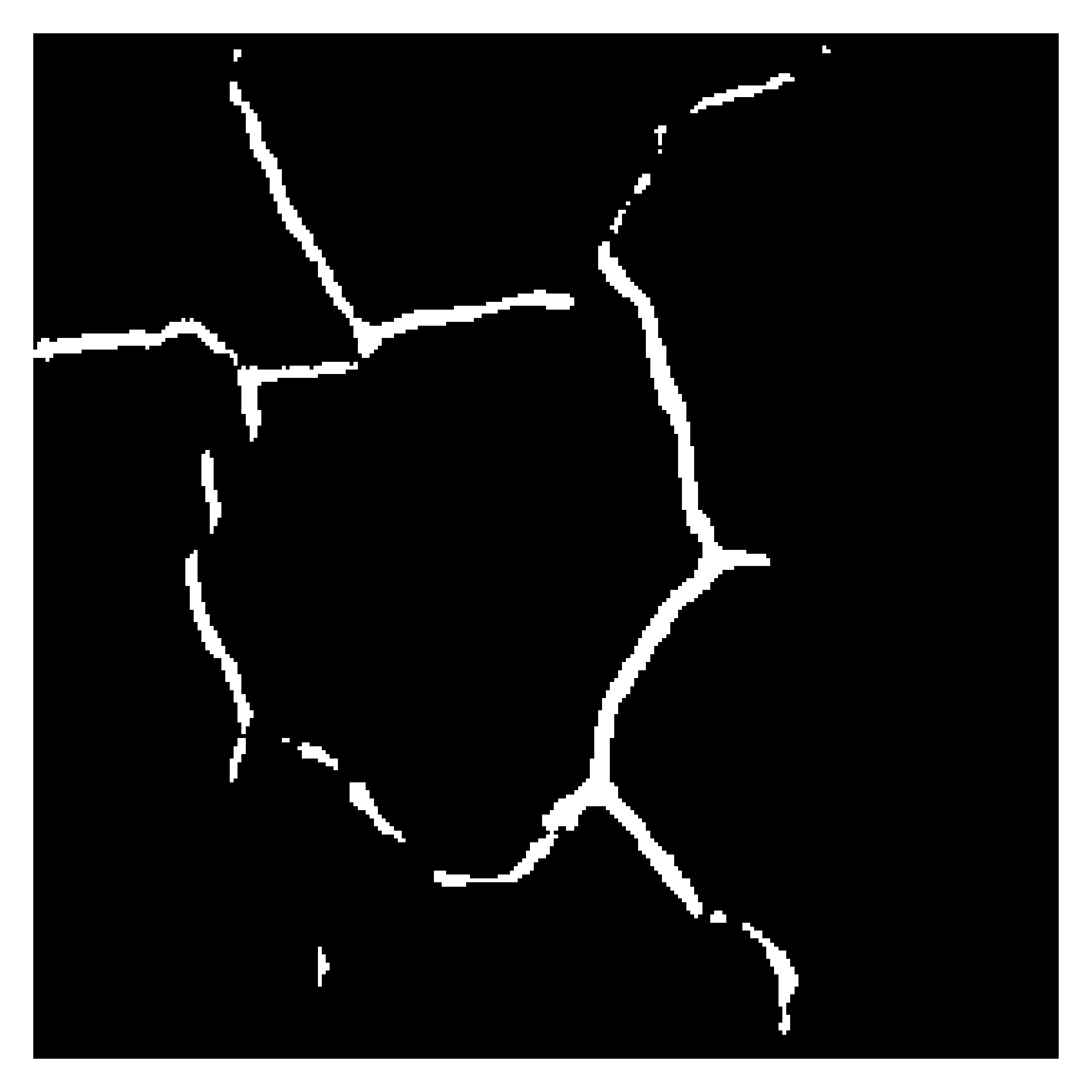}}
    \end{subfigure}
    \begin{subfigure}[b]{0.1\textwidth}
        \adjustbox{trim=10 10 10 10,clip,width=1.6cm,height=1.6cm}{\includegraphics{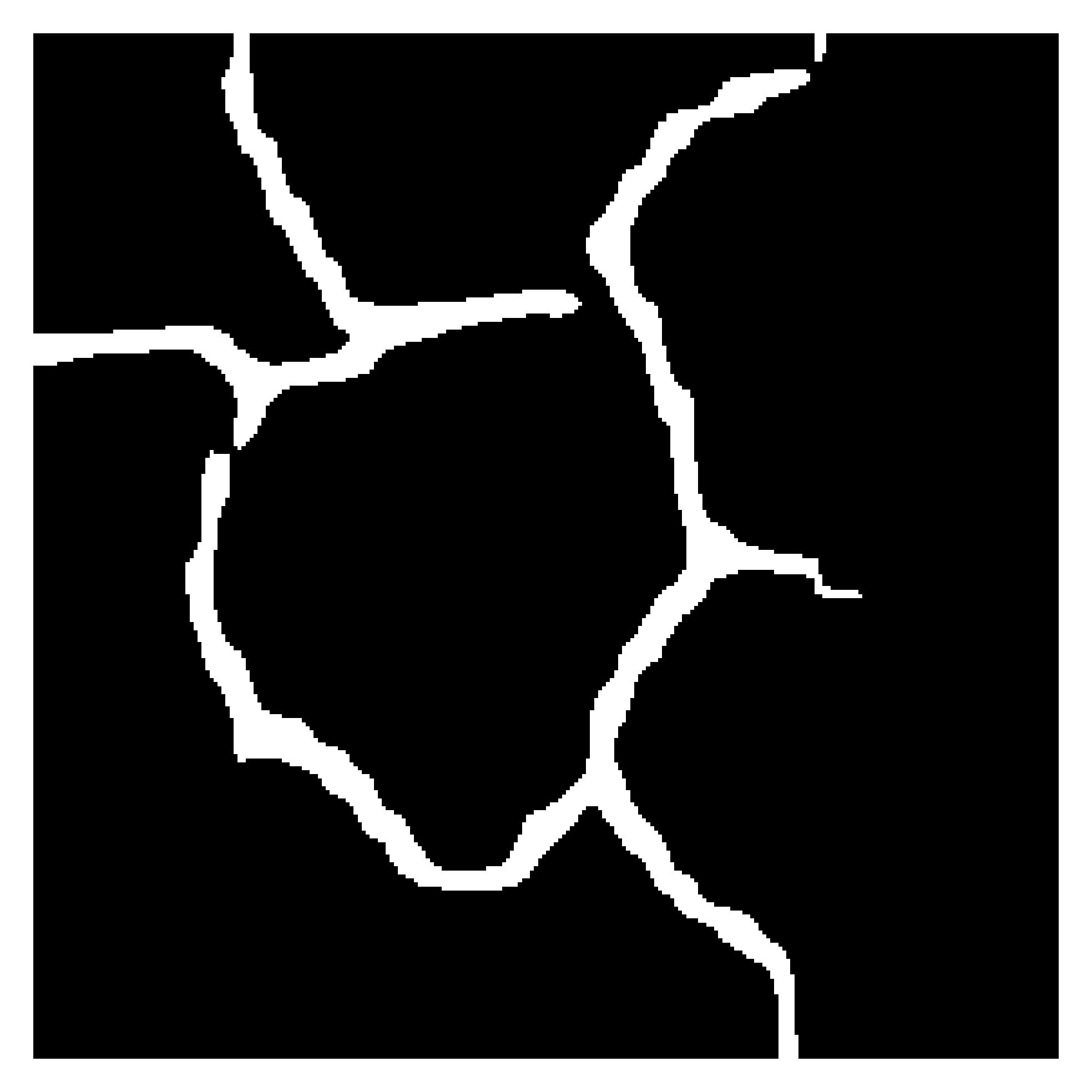}}
    \end{subfigure}

    \vspace{0.2cm}
    
    \begin{subfigure}[b]{0.1\textwidth}
        \adjustbox{trim=10 10 10 10,clip,width=1.6cm,height=1.6cm}{\includegraphics{figures/facade390_zero_shot/3/DJ_Wall_218.jpg}}
    \end{subfigure}
    \begin{subfigure}[b]{0.1\textwidth}
        \adjustbox{trim=10 10 10 10,clip,width=1.6cm,height=1.6cm}{\includegraphics{figures/facade390_zero_shot/3/DJ_Wall_218_mask.jpg}}
    \end{subfigure}  
    \begin{subfigure}[b]{0.1\textwidth}
        \adjustbox{trim=10 10 10 10,clip,width=1.6cm,height=1.6cm}{\includegraphics{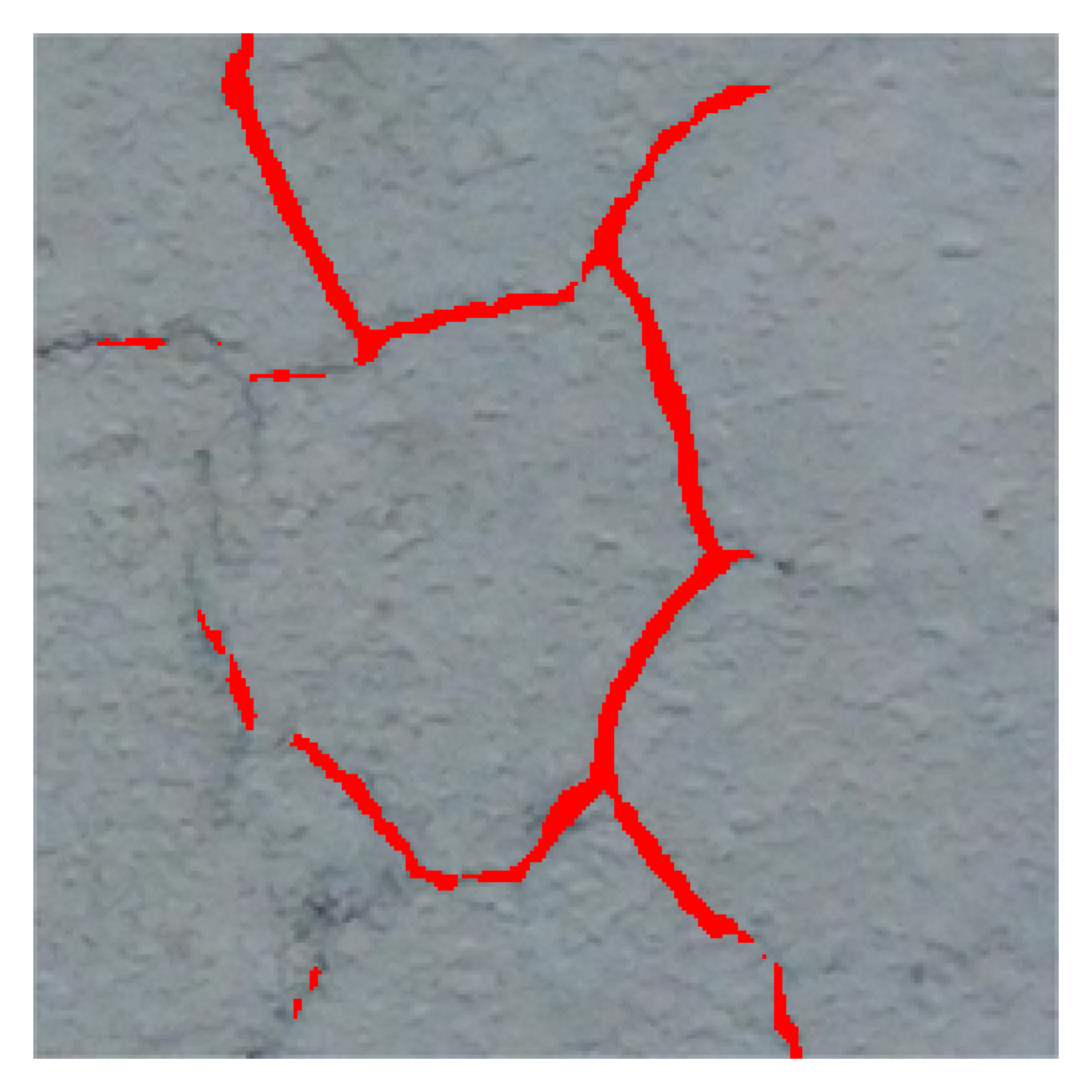}}
    \end{subfigure}
    \begin{subfigure}[b]{0.1\textwidth}
        \adjustbox{trim=10 10 10 10,clip,width=1.6cm,height=1.6cm}{\includegraphics{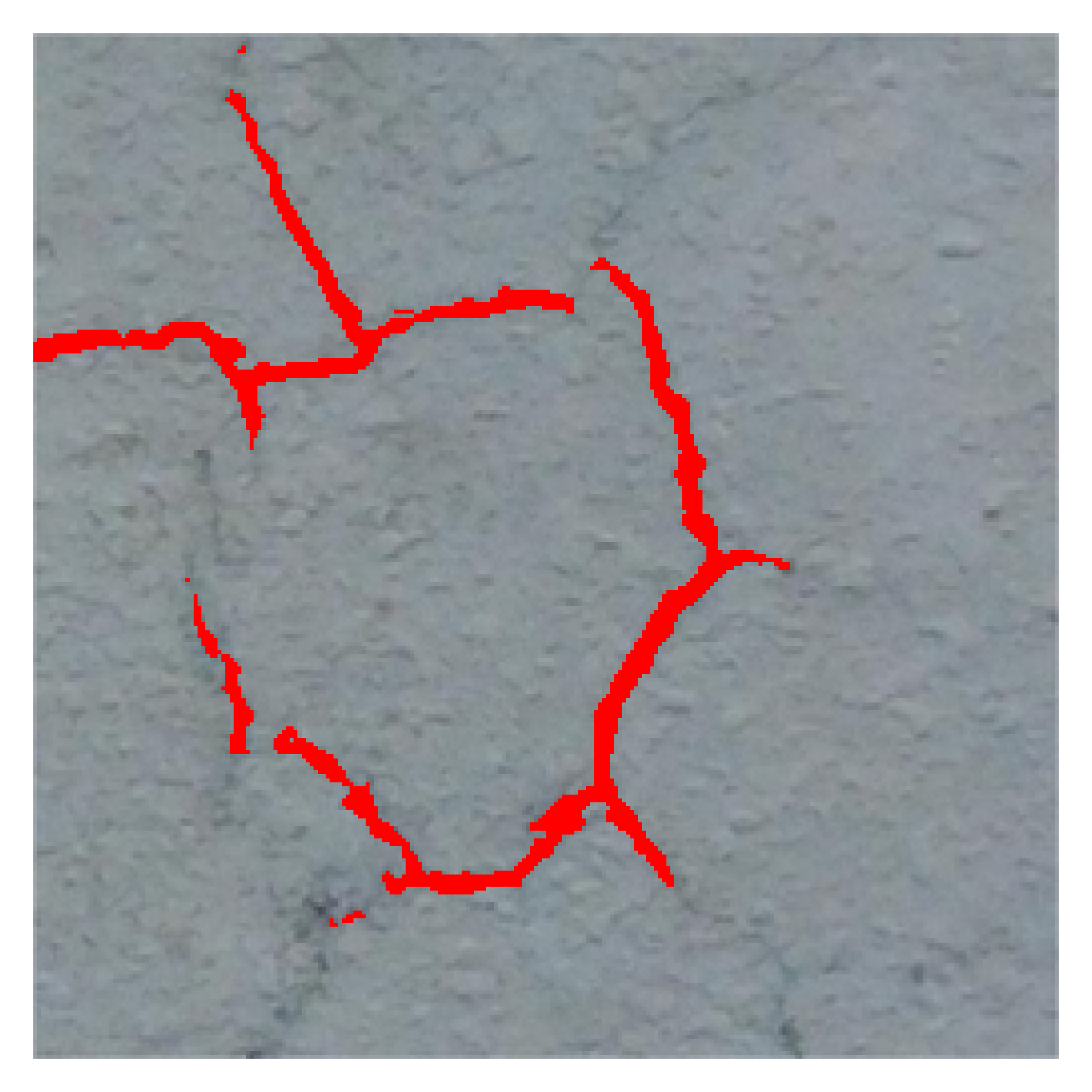}}
    \end{subfigure}
    \begin{subfigure}[b]{0.1\textwidth}
        \adjustbox{trim=10 10 10 10,clip,width=1.6cm,height=1.6cm}{\includegraphics{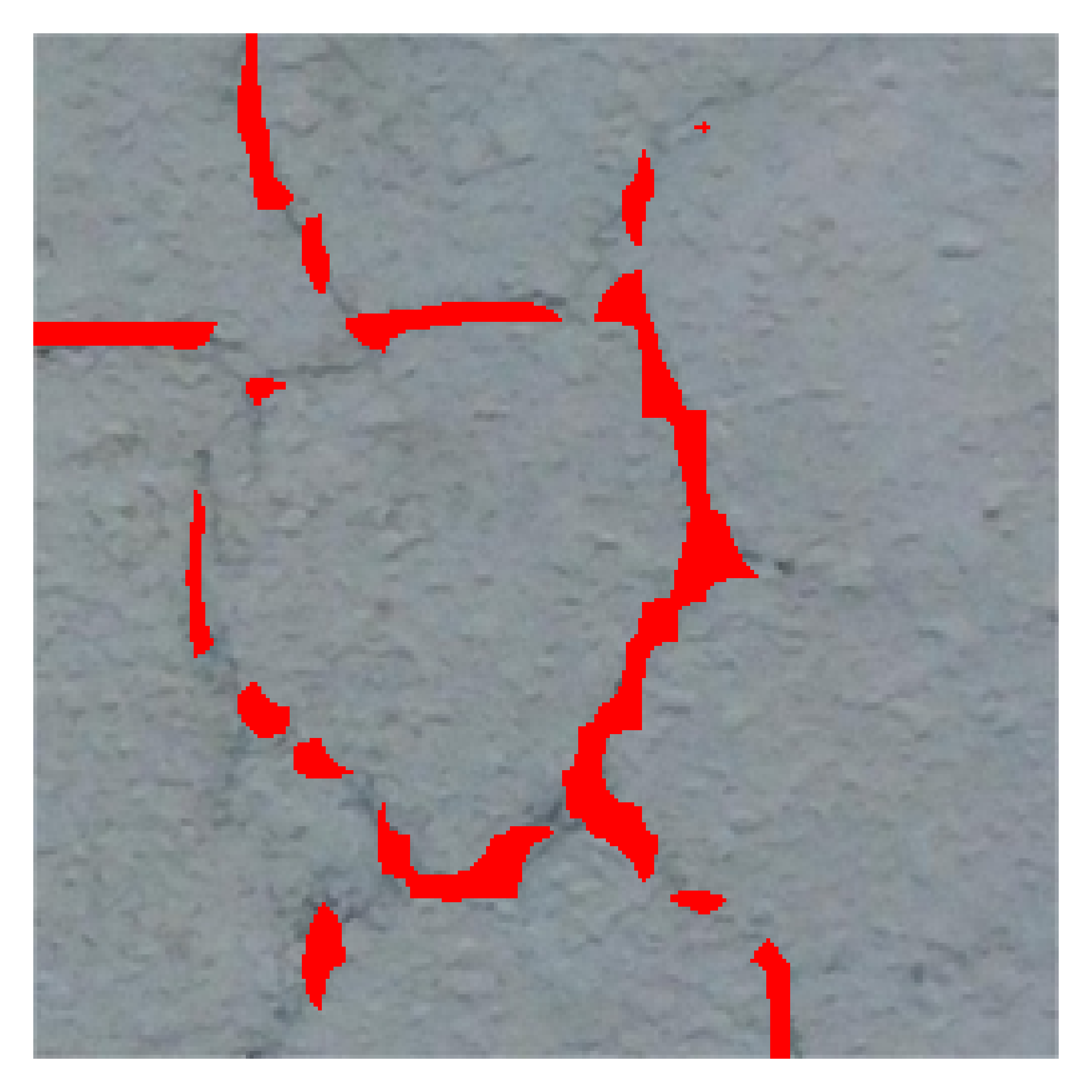}}
    \end{subfigure}
    \begin{subfigure}[b]{0.1\textwidth}
        \adjustbox{trim=10 10 10 10,clip,width=1.6cm,height=1.6cm}{\includegraphics{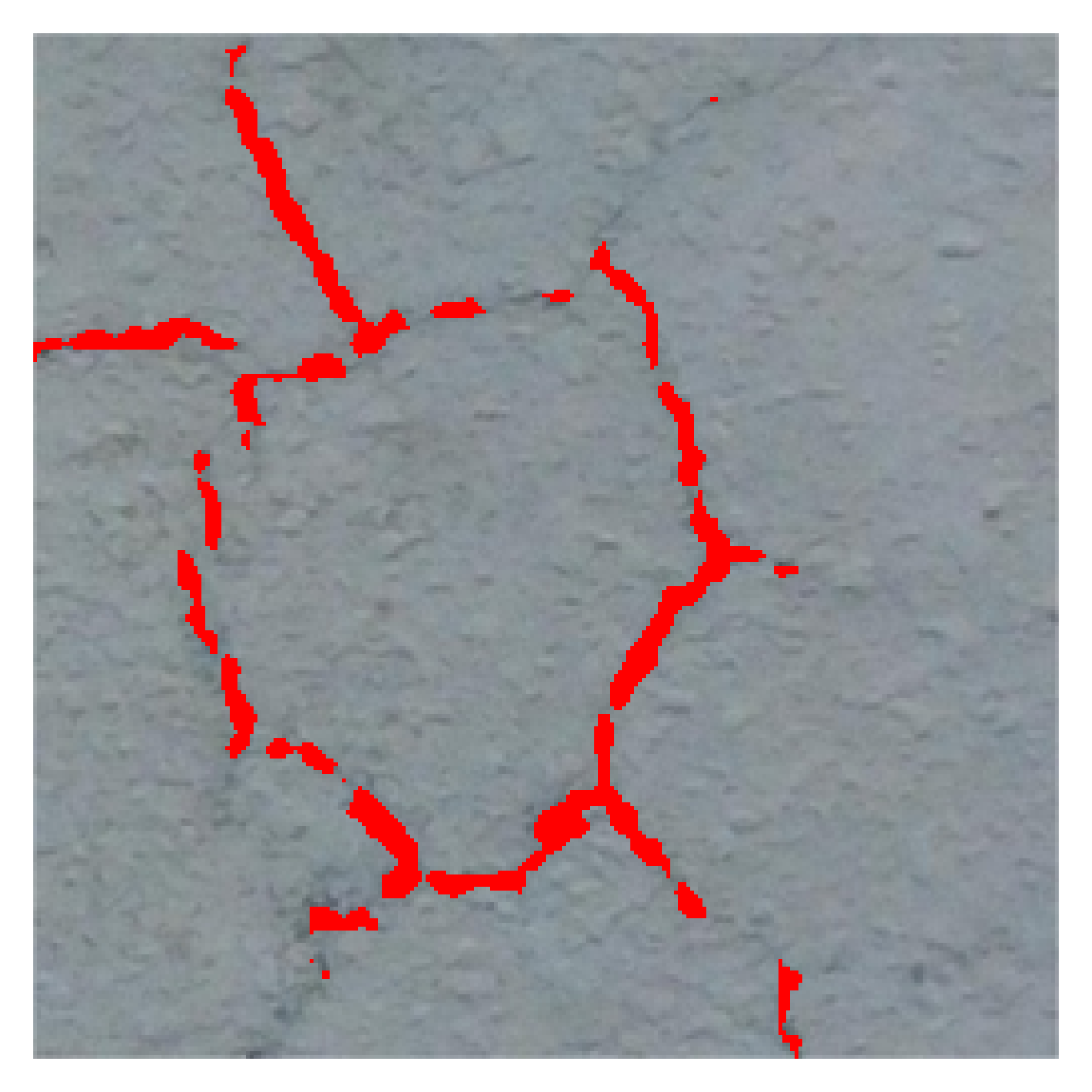}}
    \end{subfigure}
    \begin{subfigure}[b]{0.1\textwidth}
        \adjustbox{trim=10 10 10 10,clip,width=1.6cm,height=1.6cm}{\includegraphics{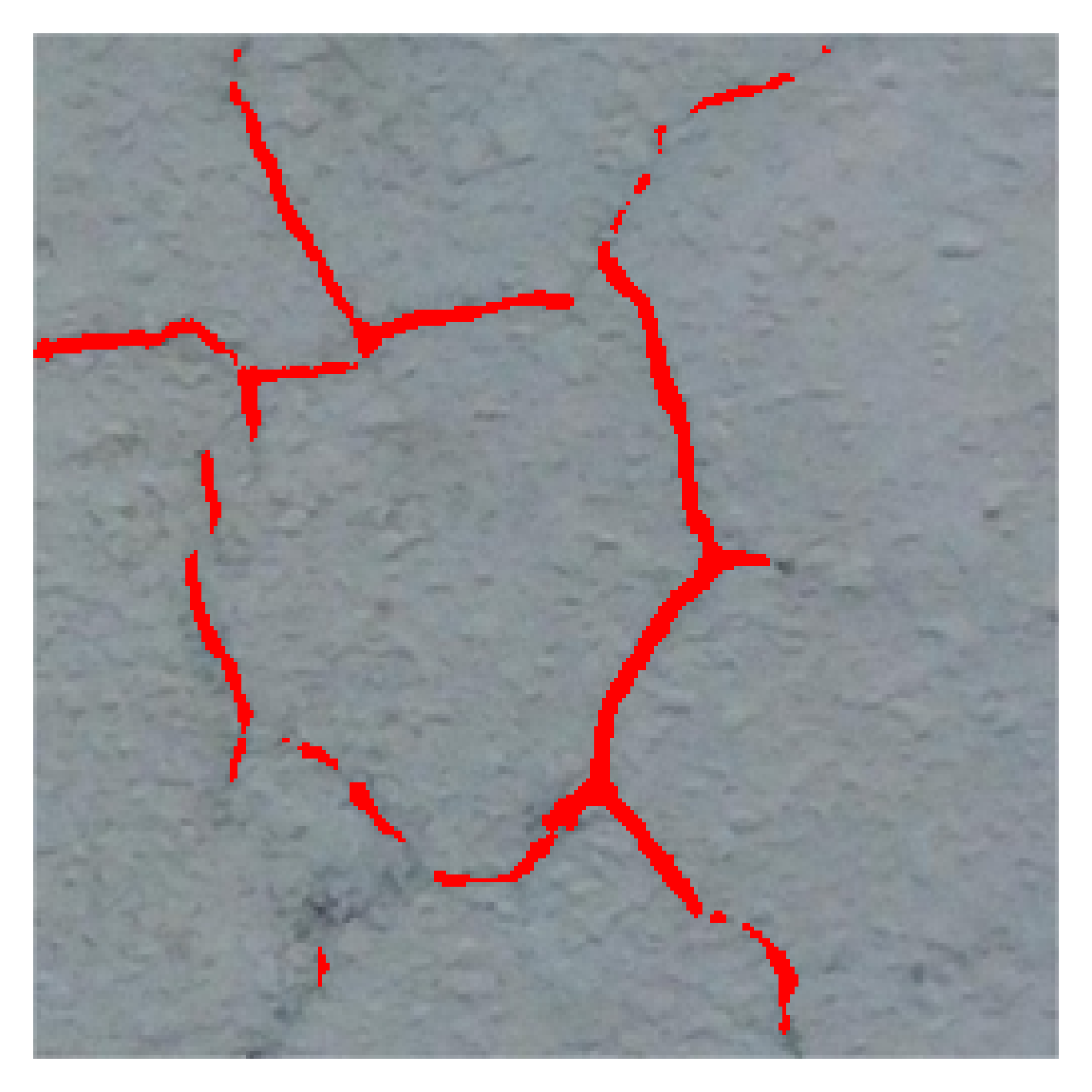}}
    \end{subfigure}
    \begin{subfigure}[b]{0.1\textwidth}
        \adjustbox{trim=10 10 10 10,clip,width=1.6cm,height=1.6cm}{\includegraphics{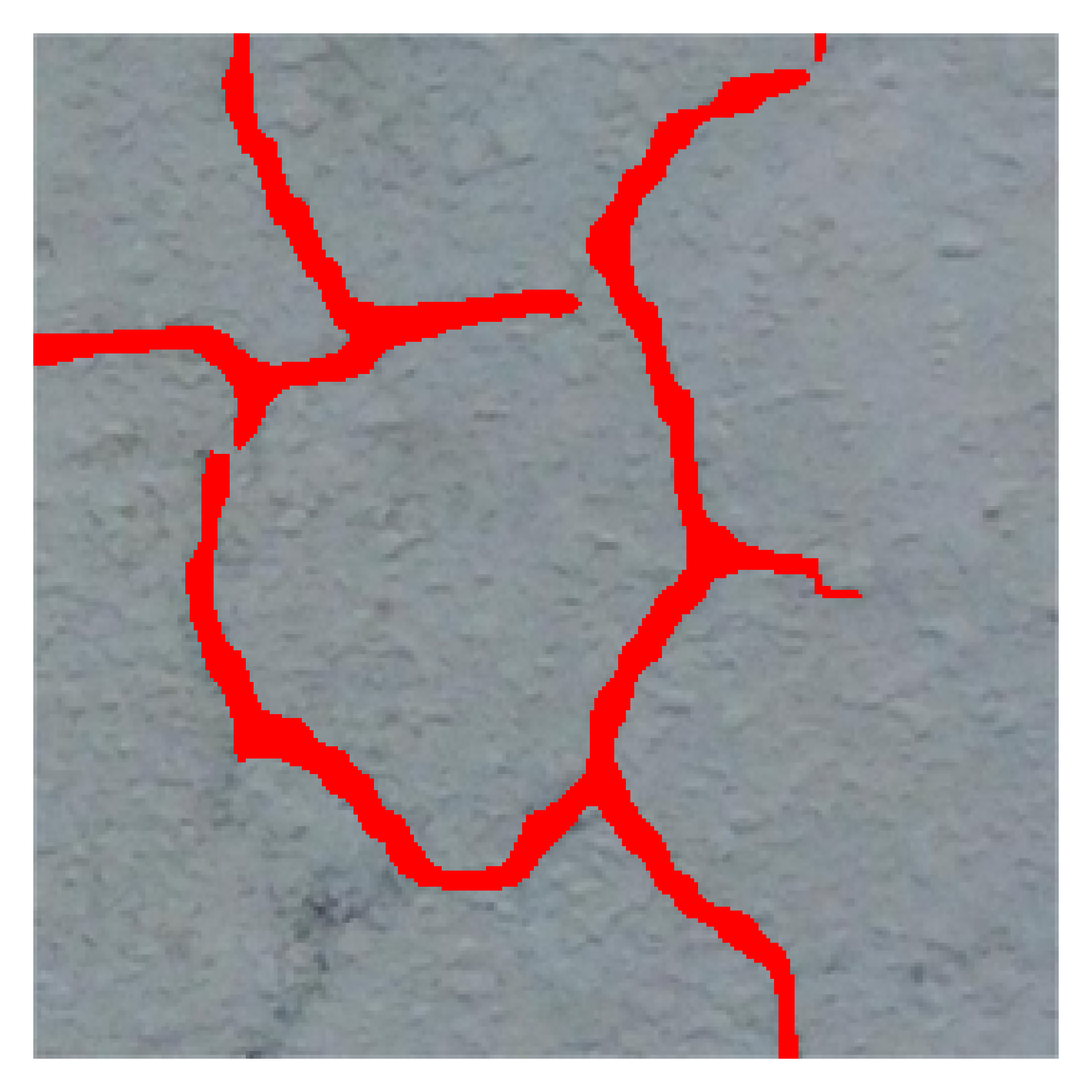}}
    \end{subfigure}
    \vspace{0.05cm}
    \vspace{0.2cm}
    \caption{Zero-shot performance of the fine-tuned models on Facade390 Dataset}
    \label{fig:zero_shot_facade390_qualitative}
\end{figure}

\begin{figure}[h]
    \captionsetup[subfigure]{labelformat=empty, font=footnotesize, justification=centering, position=top} 
    \centering
    \begin{subfigure}[b]{0.1\textwidth}
        \subcaption*{Input Image} 
        \adjustbox{trim=10 10 10 10,clip,width=1.6cm,height=1.6cm}{\includegraphics{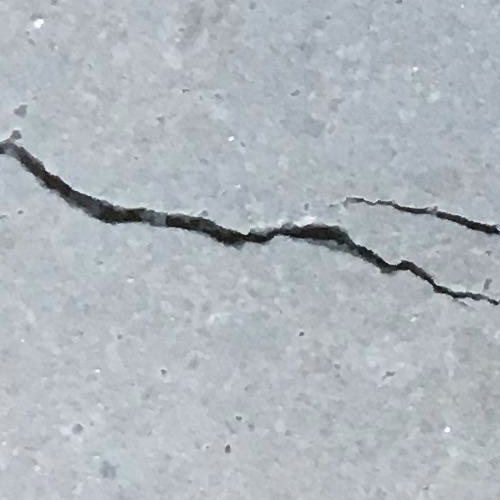}}
    \end{subfigure}
    \begin{subfigure}[b]{0.1\textwidth}
        \subcaption*{GroundTruth} 
        \adjustbox{trim=10 10 10 10,clip,width=1.6cm,height=1.6cm}{\includegraphics{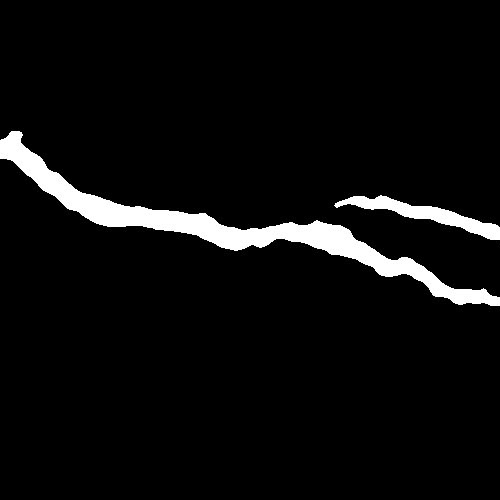}}
    \end{subfigure}  
    \begin{subfigure}[b]{0.1\textwidth}
        \subcaption*{SAC} 
        \adjustbox{trim=10 10 10 10,clip,width=1.6cm,height=1.6cm}{\includegraphics{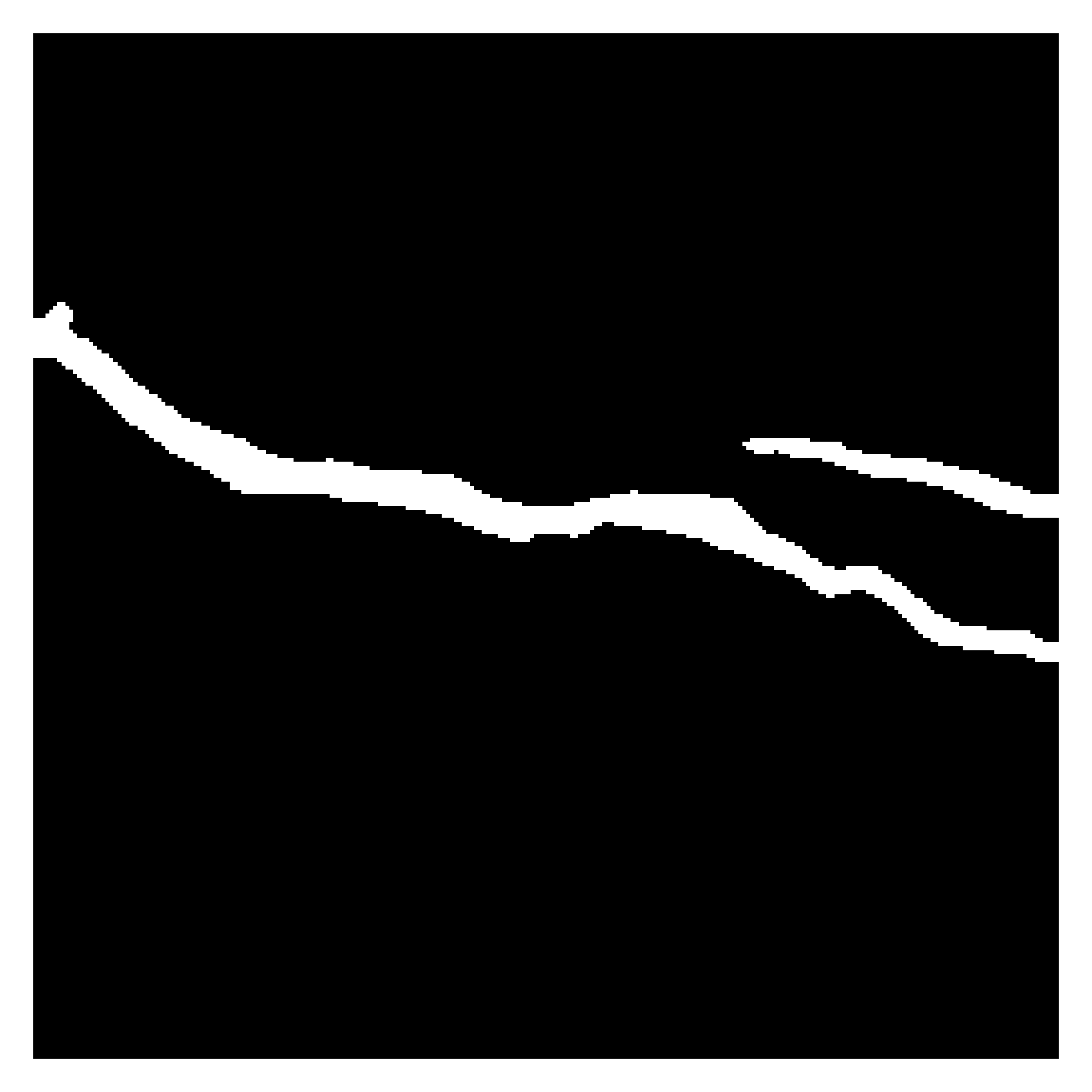}}
    \end{subfigure}
    \begin{subfigure}[b]{0.1\textwidth}
        \subcaption*{DeepCrackL} 
        \adjustbox{trim=10 10 10 10,clip,width=1.6cm,height=1.6cm}{\includegraphics{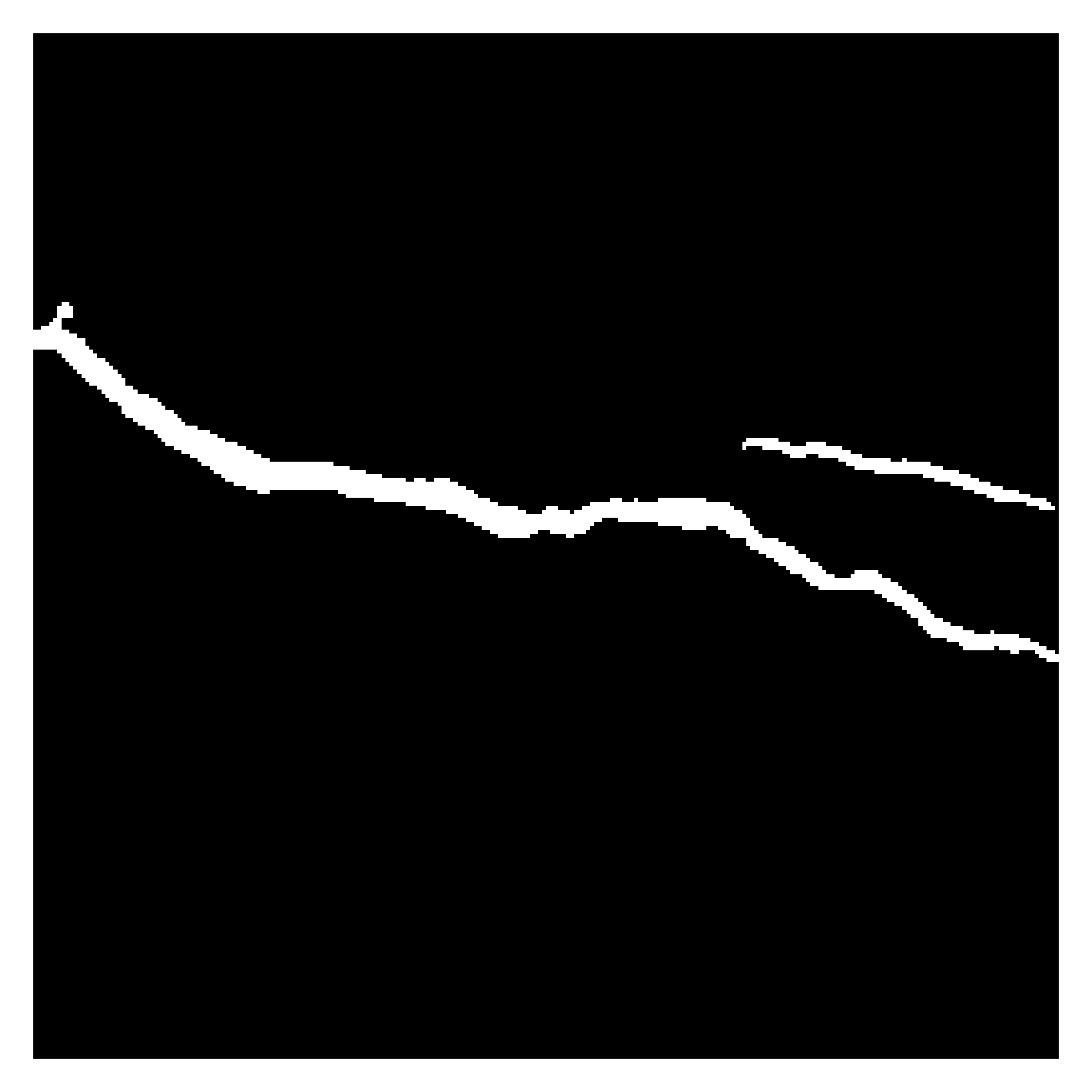}}
    \end{subfigure}
    \begin{subfigure}[b]{0.1\textwidth}
        \subcaption*{CrackFormer} 
        \adjustbox{trim=10 10 10 10,clip,width=1.6cm,height=1.6cm}{\includegraphics{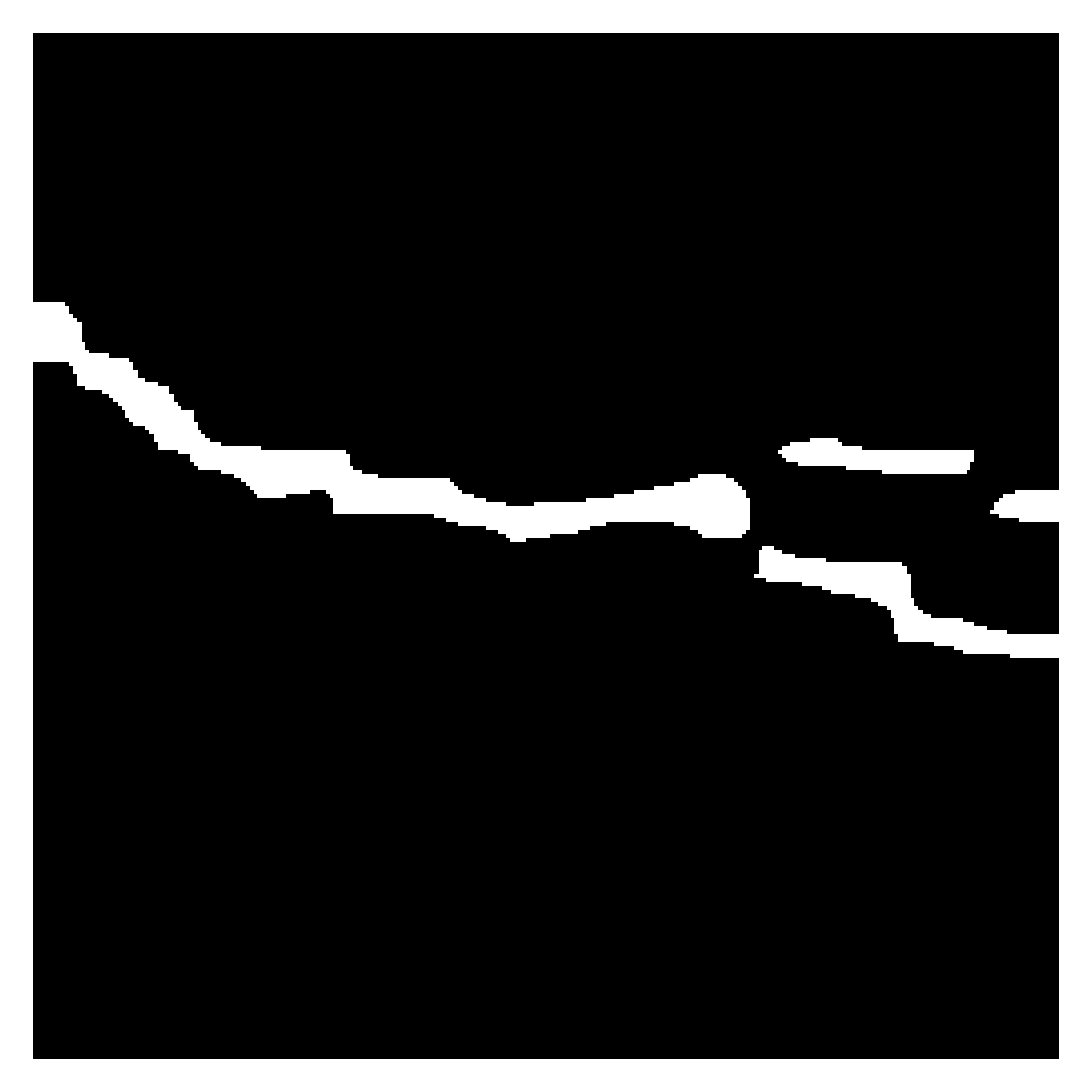}}
    \end{subfigure}
    \begin{subfigure}[b]{0.1\textwidth}
        \subcaption*{SegFormer} 
        \adjustbox{trim=10 10 10 10,clip,width=1.6cm,height=1.6cm}{\includegraphics{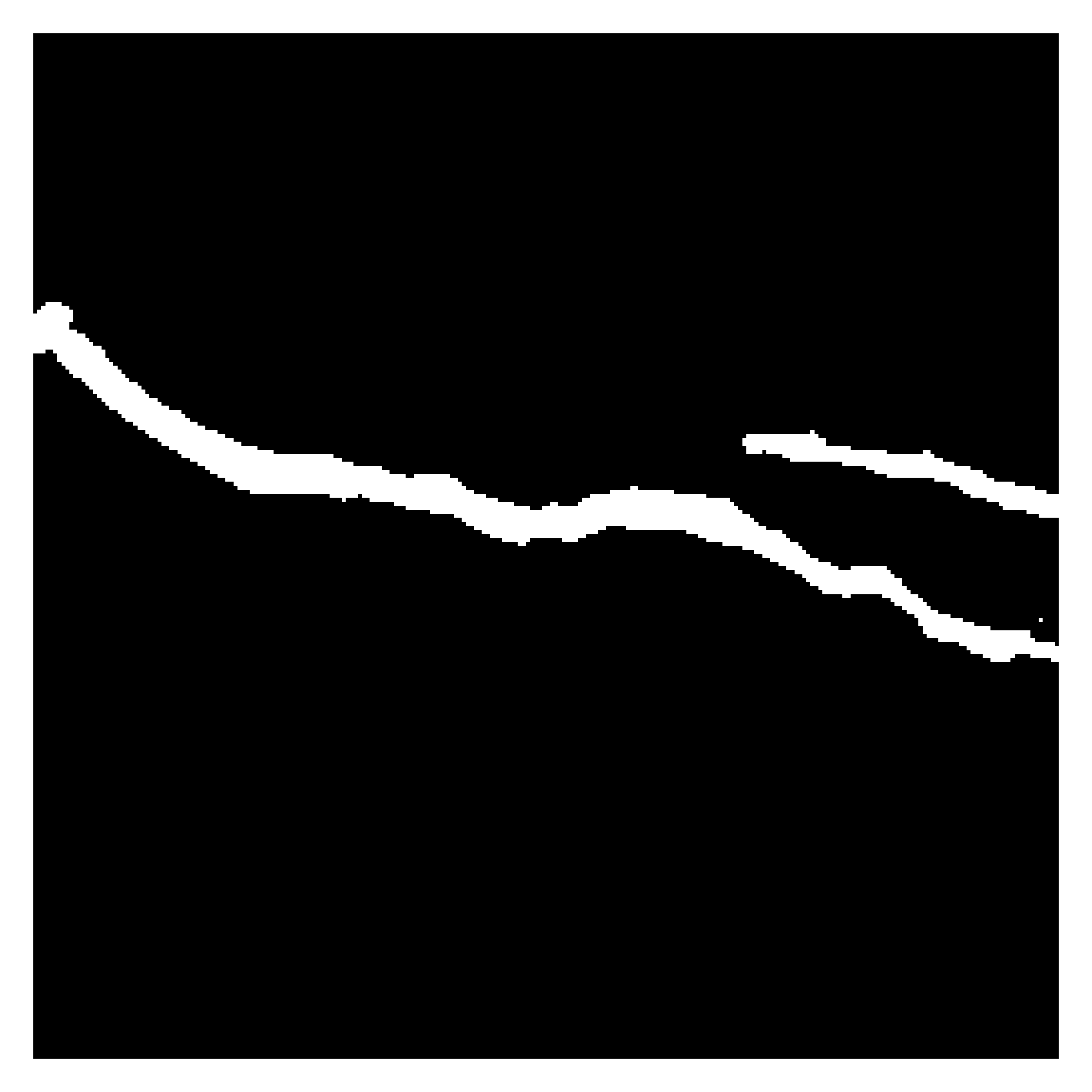}}
    \end{subfigure}
    \begin{subfigure}[b]{0.1\textwidth}
        \subcaption*{U-Net} 
        \adjustbox{trim=10 10 10 10,clip,width=1.6cm,height=1.6cm}{\includegraphics{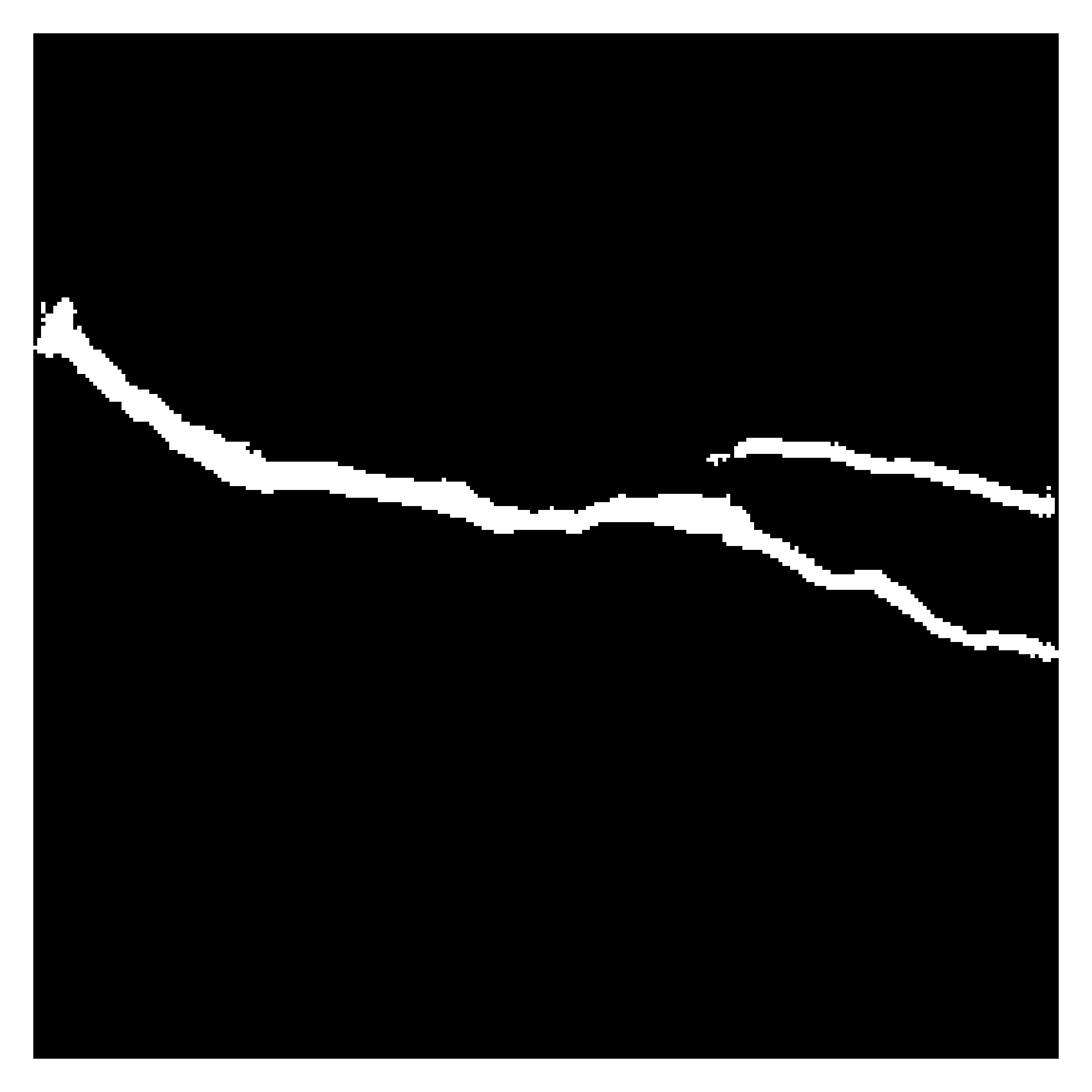}}
    \end{subfigure}
    \begin{subfigure}[b]{0.1\textwidth}
        \subcaption*{DeepLabv3+} 
        \adjustbox{trim=10 10 10 10,clip,width=1.6cm,height=1.6cm}{\includegraphics{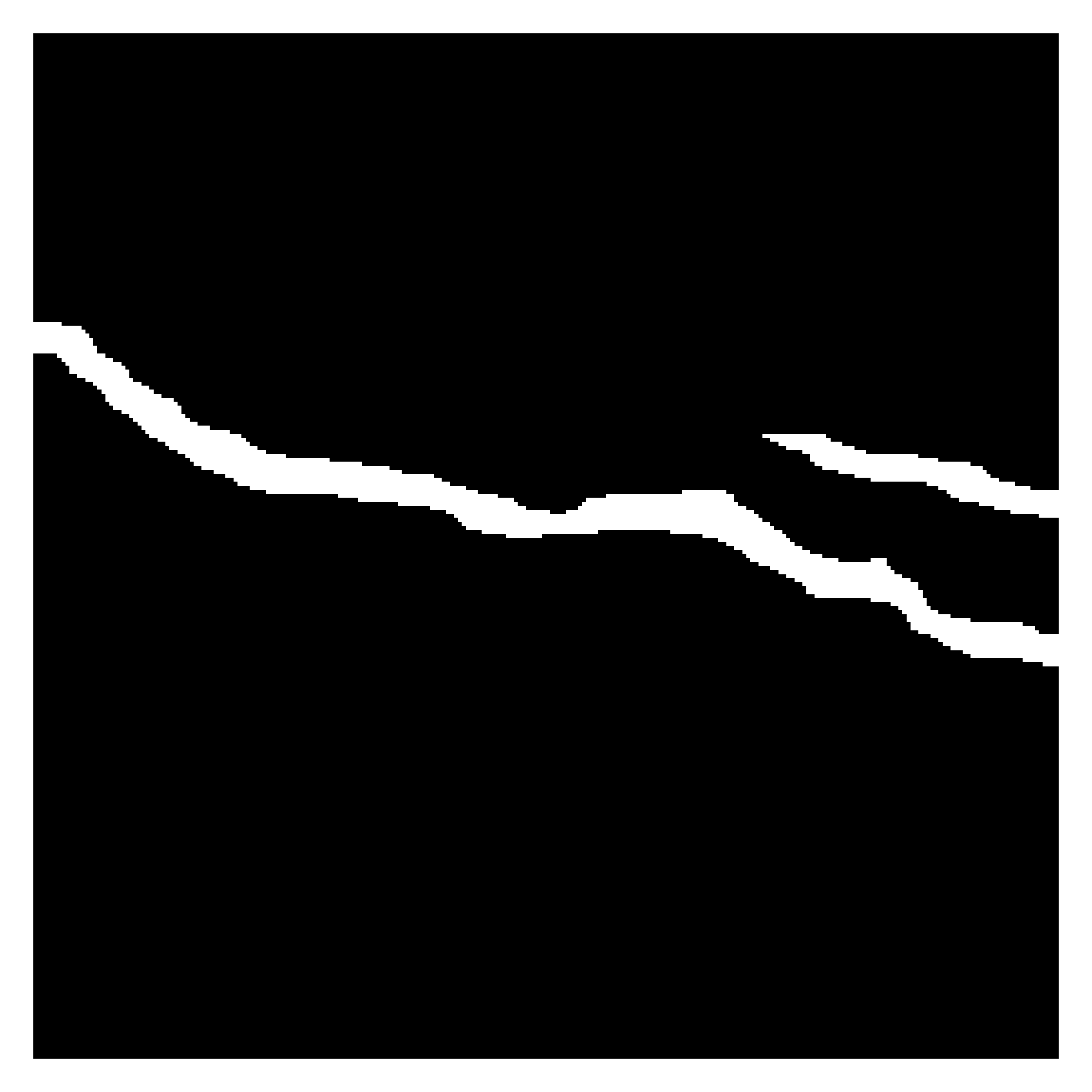}}
    \end{subfigure}

    \vspace{0.2cm}
    
    \begin{subfigure}[b]{0.1\textwidth}
        \adjustbox{trim=10 10 10 10,clip,width=1.6cm,height=1.6cm}{\includegraphics{figures/concrete3k_zero_shot/1/174_18.jpg}}
    \end{subfigure}
    \begin{subfigure}[b]{0.1\textwidth}
        \adjustbox{trim=10 10 10 10,clip,width=1.6cm,height=1.6cm}{\includegraphics{figures/concrete3k_zero_shot/1/174_18_mask.jpg}}  
    \end{subfigure}
    \begin{subfigure}[b]{0.1\textwidth}
        \adjustbox{trim=10 10 10 10,clip,width=1.6cm,height=1.6cm}{\includegraphics{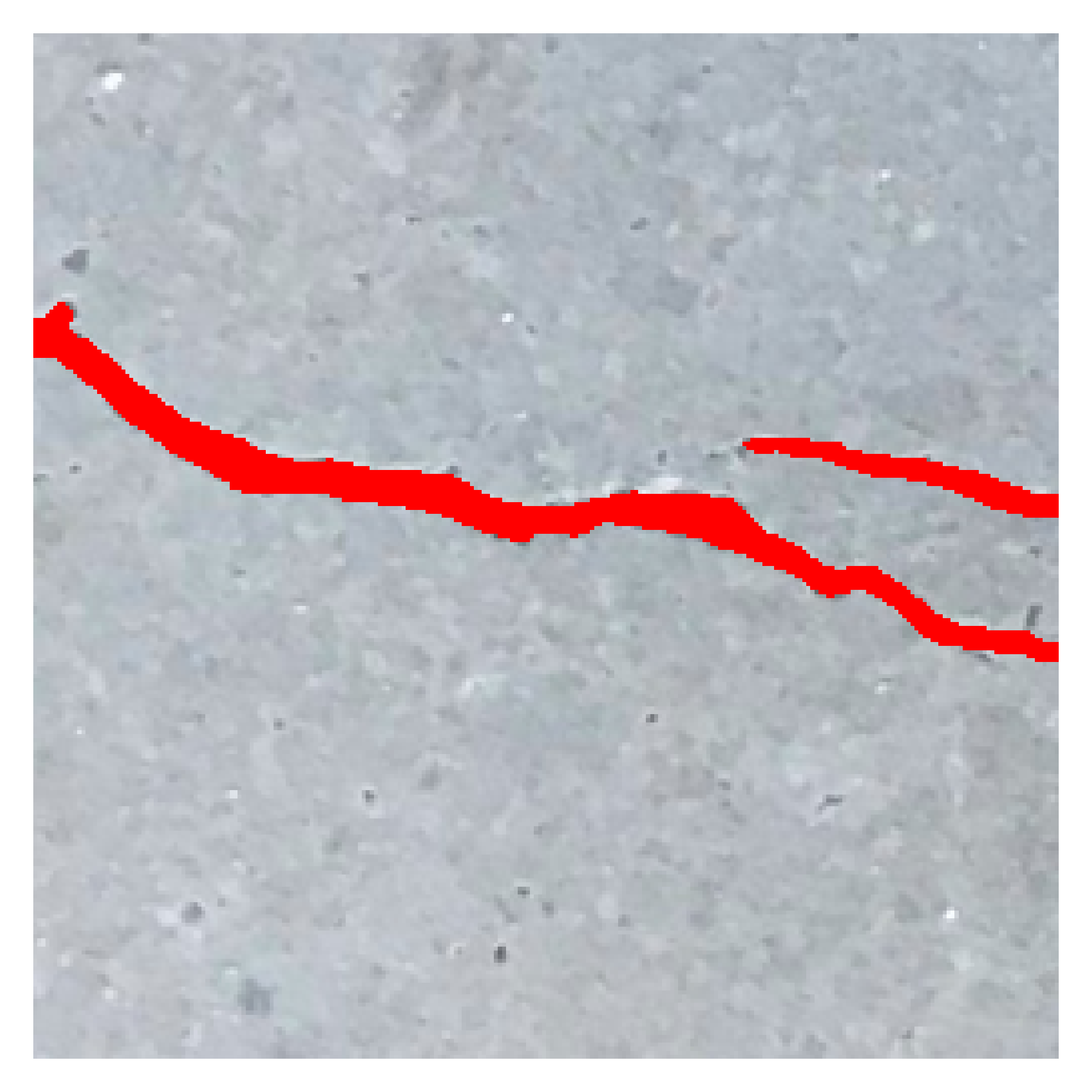}}
    \end{subfigure}
    \begin{subfigure}[b]{0.1\textwidth}
        \adjustbox{trim=10 10 10 10,clip,width=1.6cm,height=1.6cm}{\includegraphics{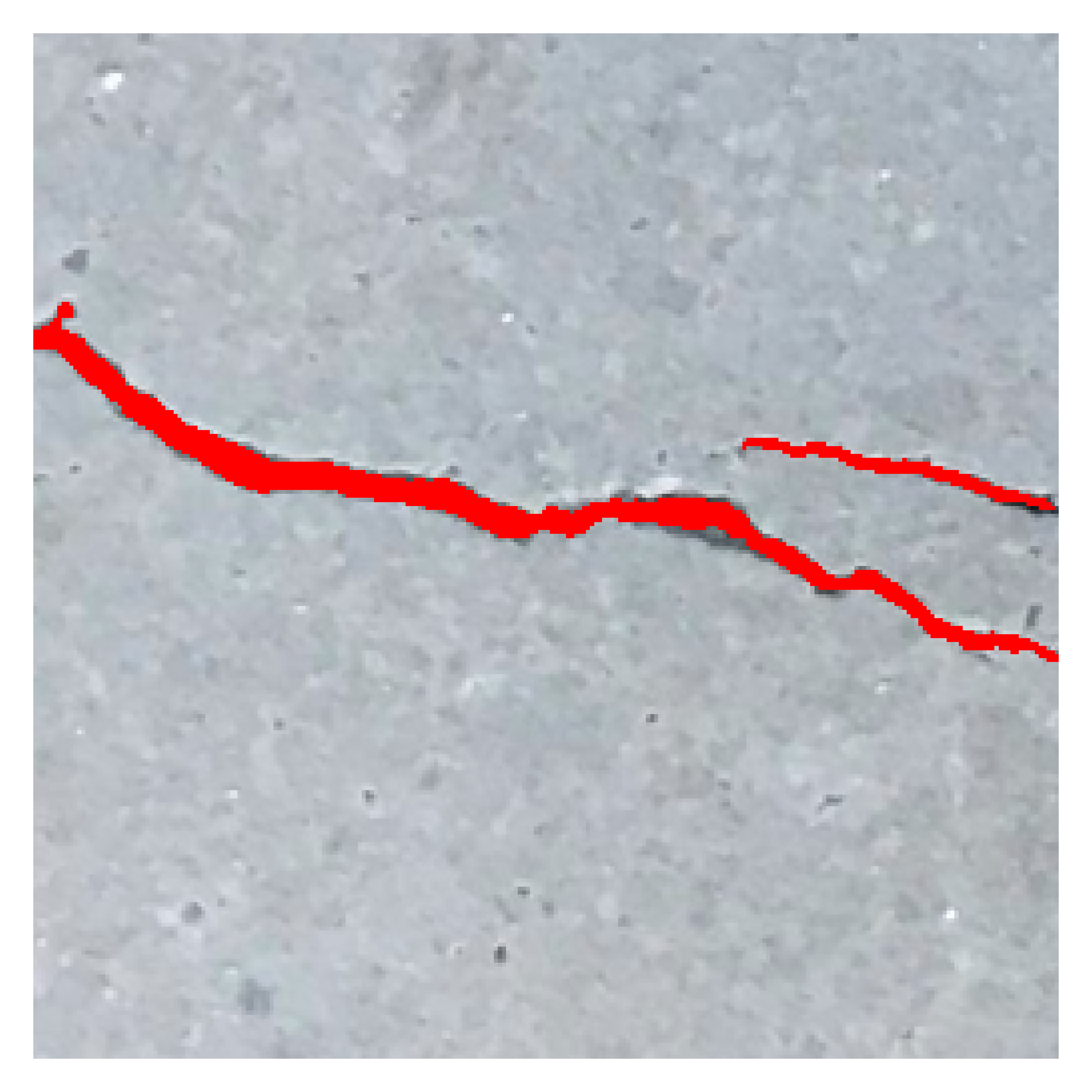}}
    \end{subfigure}
    \begin{subfigure}[b]{0.1\textwidth}
        \adjustbox{trim=10 10 10 10,clip,width=1.6cm,height=1.6cm}{\includegraphics{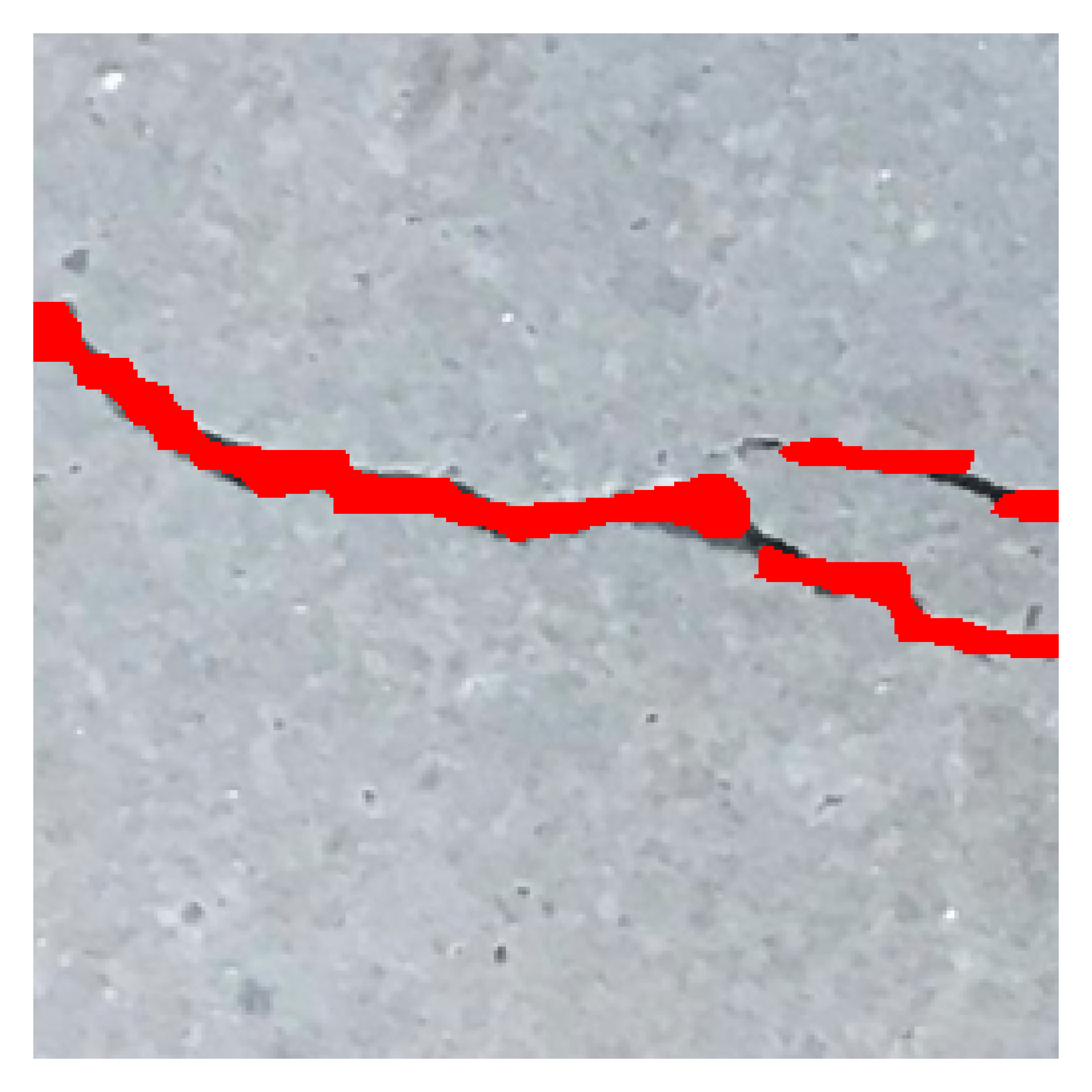}}
    \end{subfigure}
    \begin{subfigure}[b]{0.1\textwidth}
        \adjustbox{trim=10 10 10 10,clip,width=1.6cm,height=1.6cm}{\includegraphics{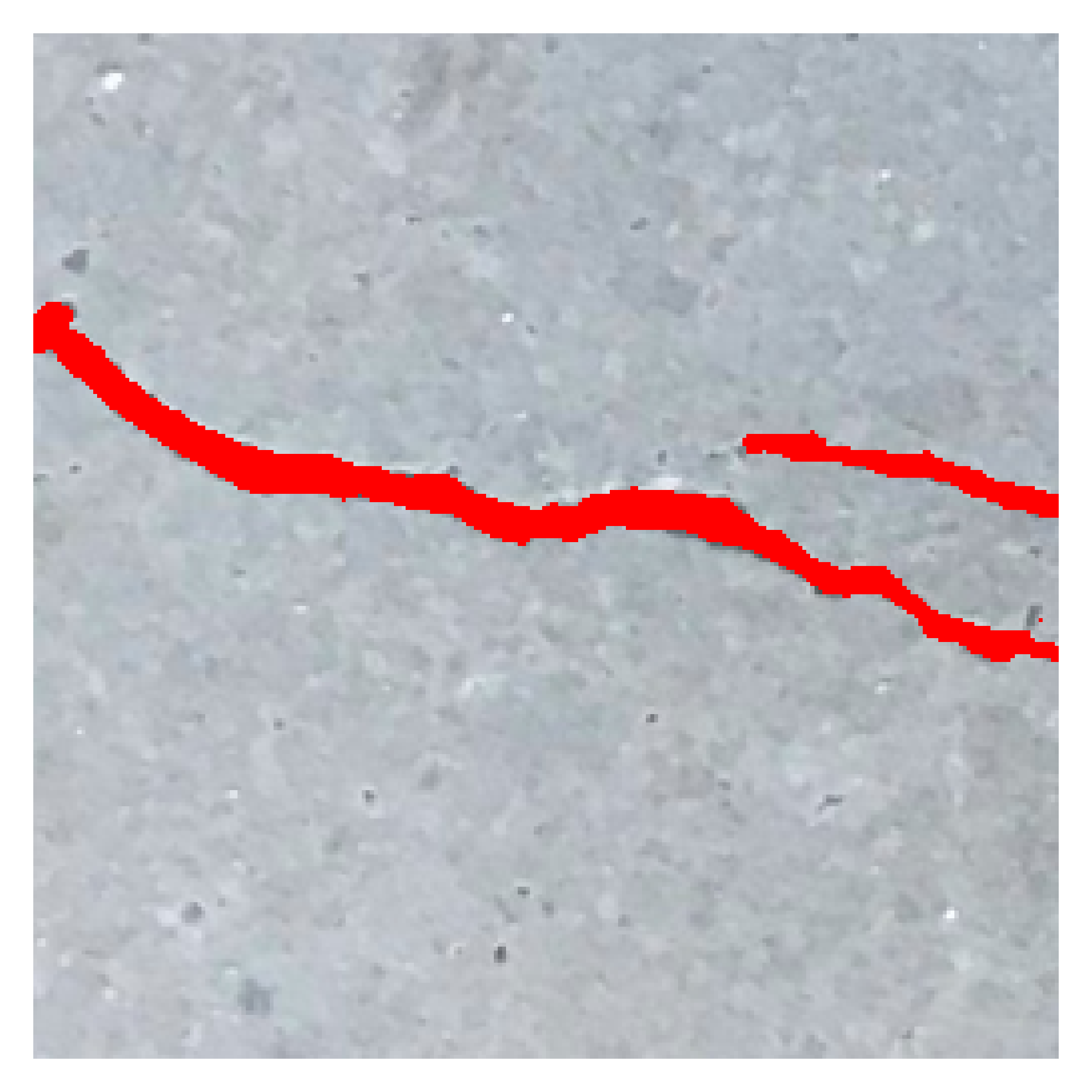}}
    \end{subfigure}
    \begin{subfigure}[b]{0.1\textwidth}
        \adjustbox{trim=10 10 10 10,clip,width=1.6cm,height=1.6cm}{\includegraphics{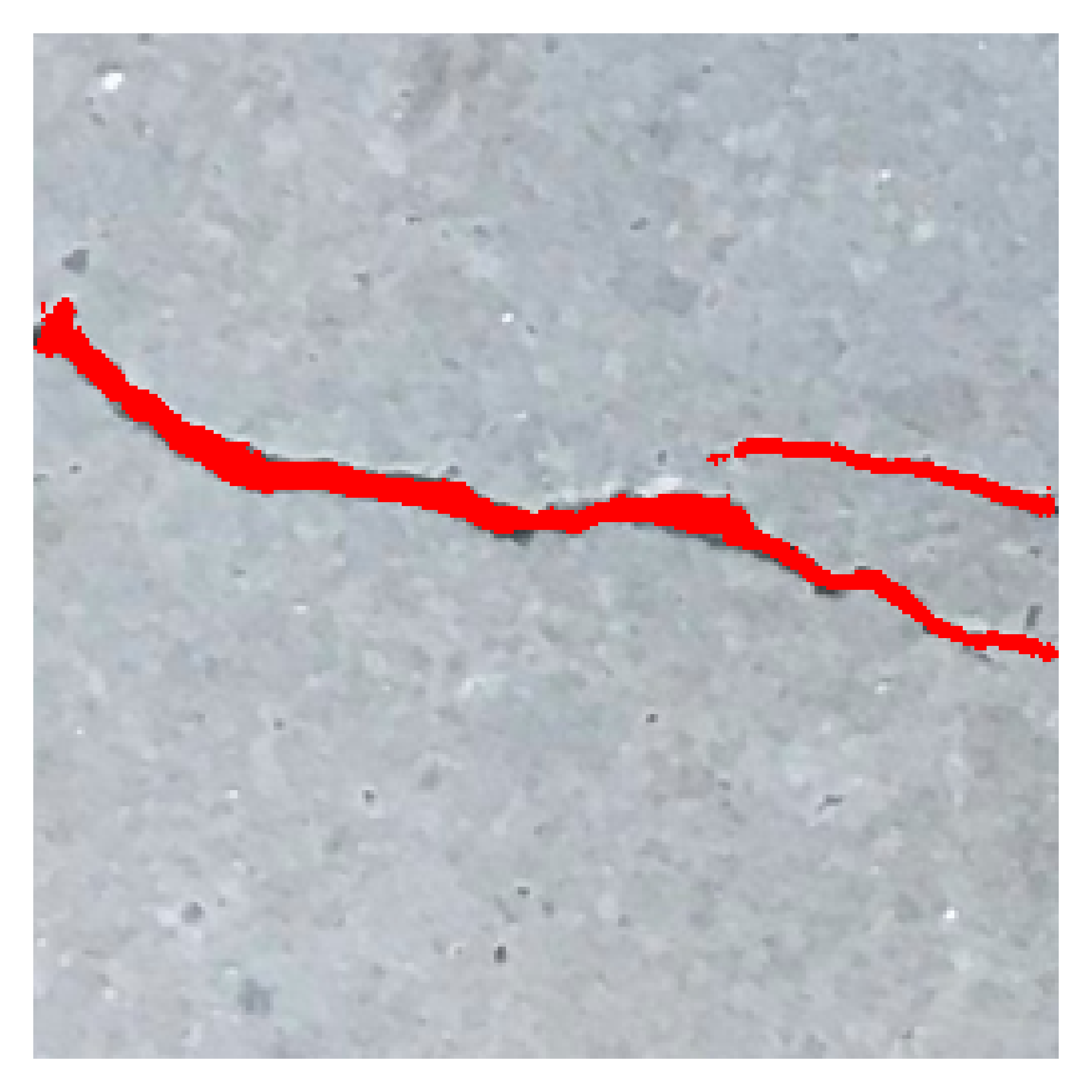}}
    \end{subfigure}
    \begin{subfigure}[b]{0.1\textwidth}
        \adjustbox{trim=10 10 10 10,clip,width=1.6cm,height=1.6cm}{\includegraphics{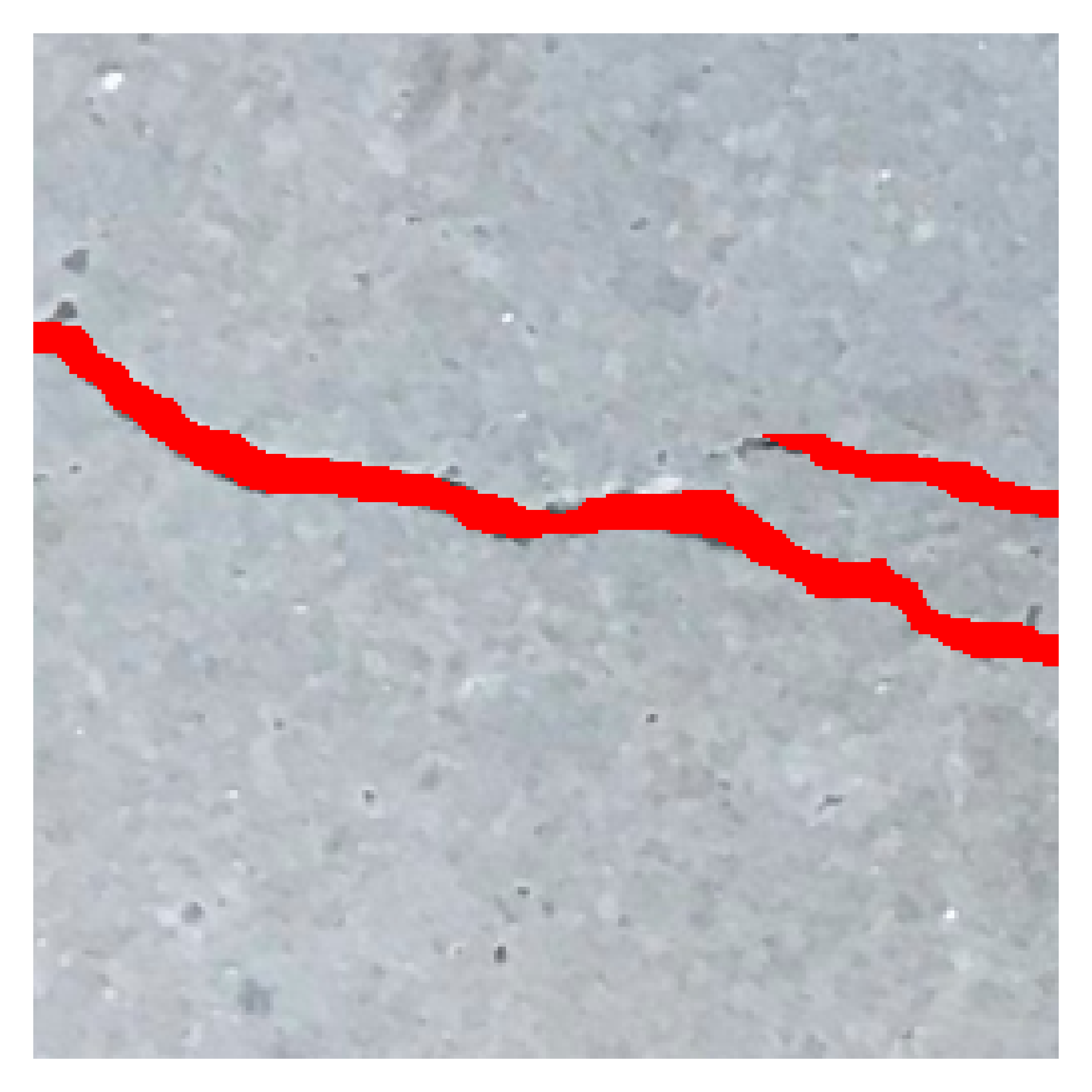}}
    \end{subfigure}
    \vspace{0.5cm}

    \begin{subfigure}[b]{0.1\textwidth}
        \adjustbox{trim=10 10 10 10,clip,width=1.6cm,height=1.6cm}{\includegraphics{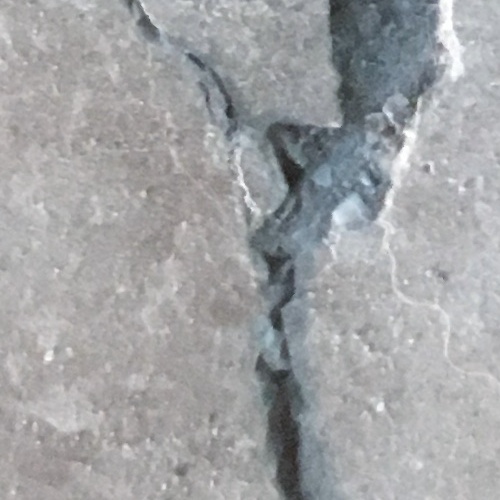}}
    \end{subfigure}
    \begin{subfigure}[b]{0.1\textwidth}
        \adjustbox{trim=10 10 10 10,clip,width=1.6cm,height=1.6cm}{\includegraphics{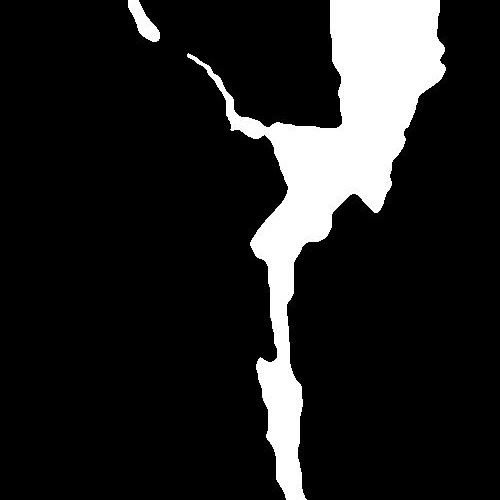}}
    \end{subfigure}  
    \begin{subfigure}[b]{0.1\textwidth}
        \adjustbox{trim=10 10 10 10,clip,width=1.6cm,height=1.6cm}{\includegraphics{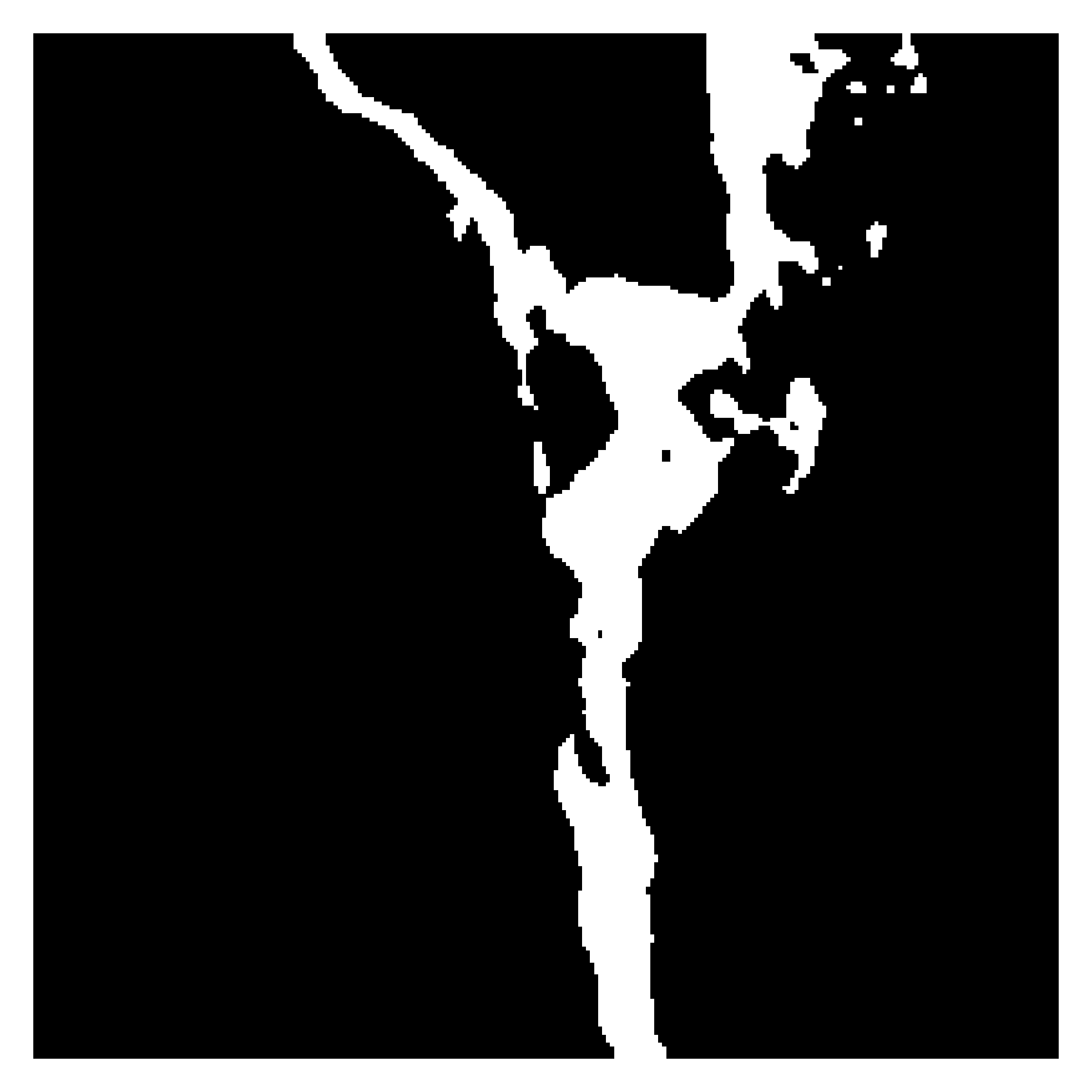}}
    \end{subfigure}
    \begin{subfigure}[b]{0.1\textwidth}
        \adjustbox{trim=10 10 10 10,clip,width=1.6cm,height=1.6cm}{\includegraphics{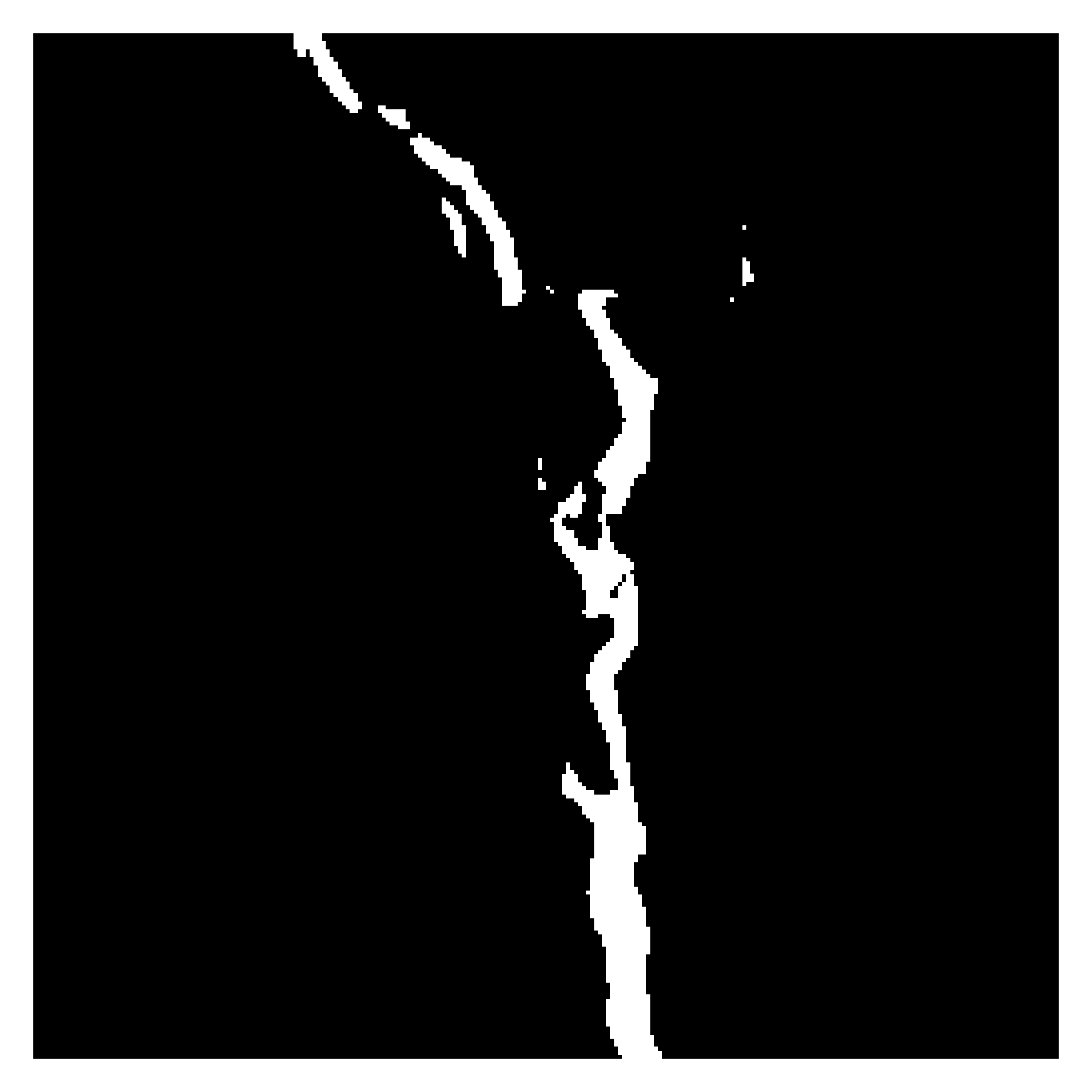}}
    \end{subfigure}
    \begin{subfigure}[b]{0.1\textwidth}
        \adjustbox{trim=10 10 10 10,clip,width=1.6cm,height=1.6cm}{\includegraphics{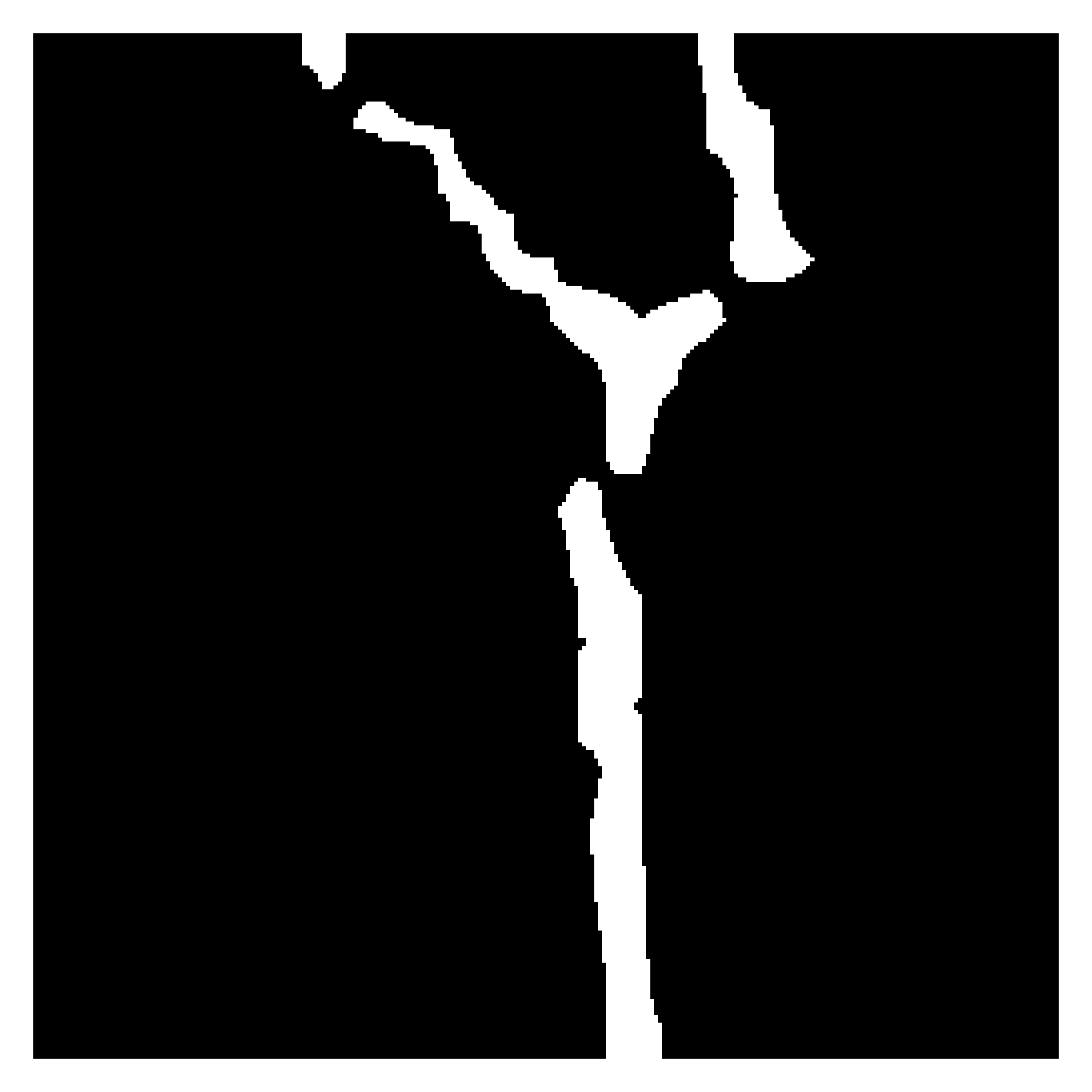}}
    \end{subfigure}
    \begin{subfigure}[b]{0.1\textwidth}
        \adjustbox{trim=10 10 10 10,clip,width=1.6cm,height=1.6cm}{\includegraphics{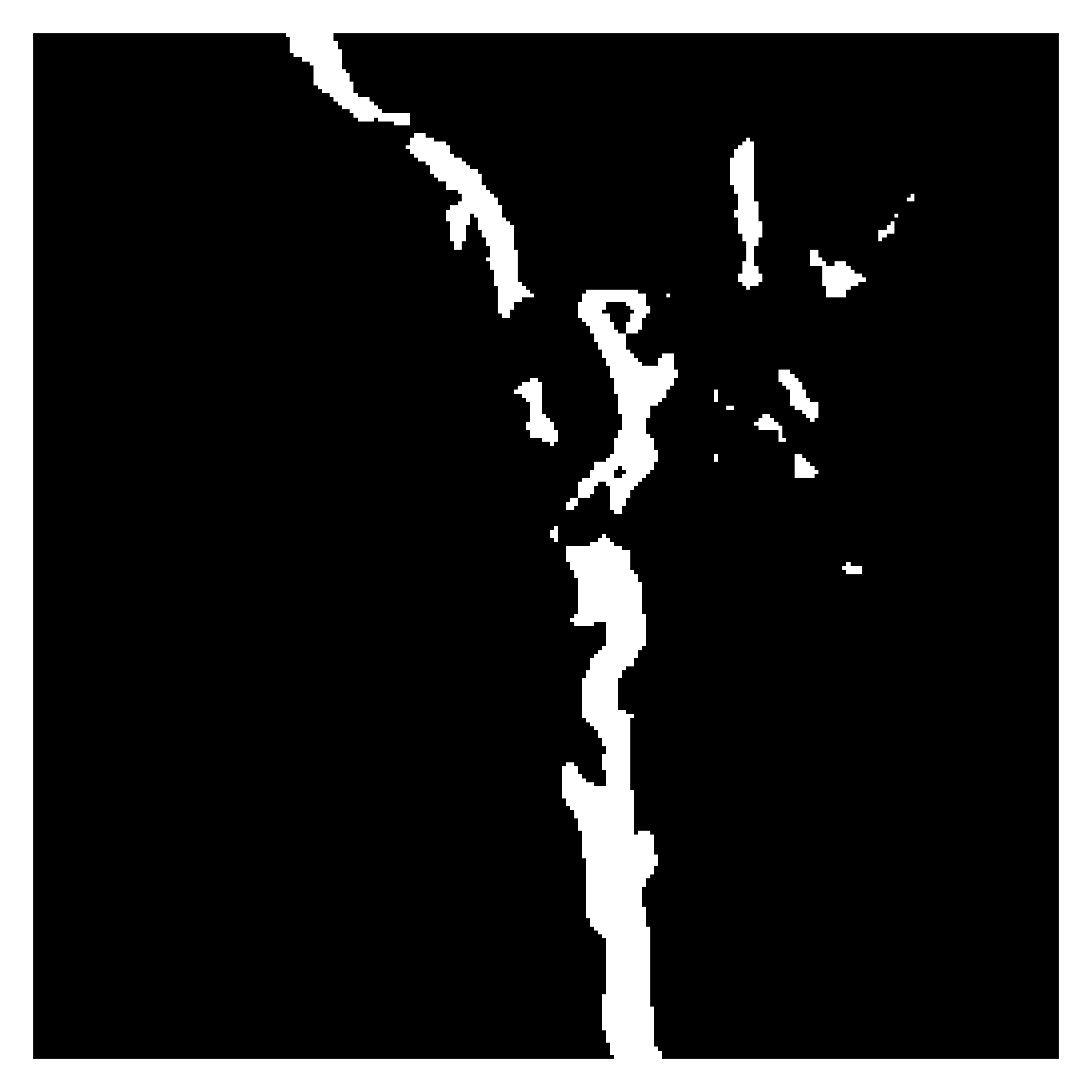}}
    \end{subfigure}
    \begin{subfigure}[b]{0.1\textwidth}
        \adjustbox{trim=10 10 10 10,clip,width=1.6cm,height=1.6cm}{\includegraphics{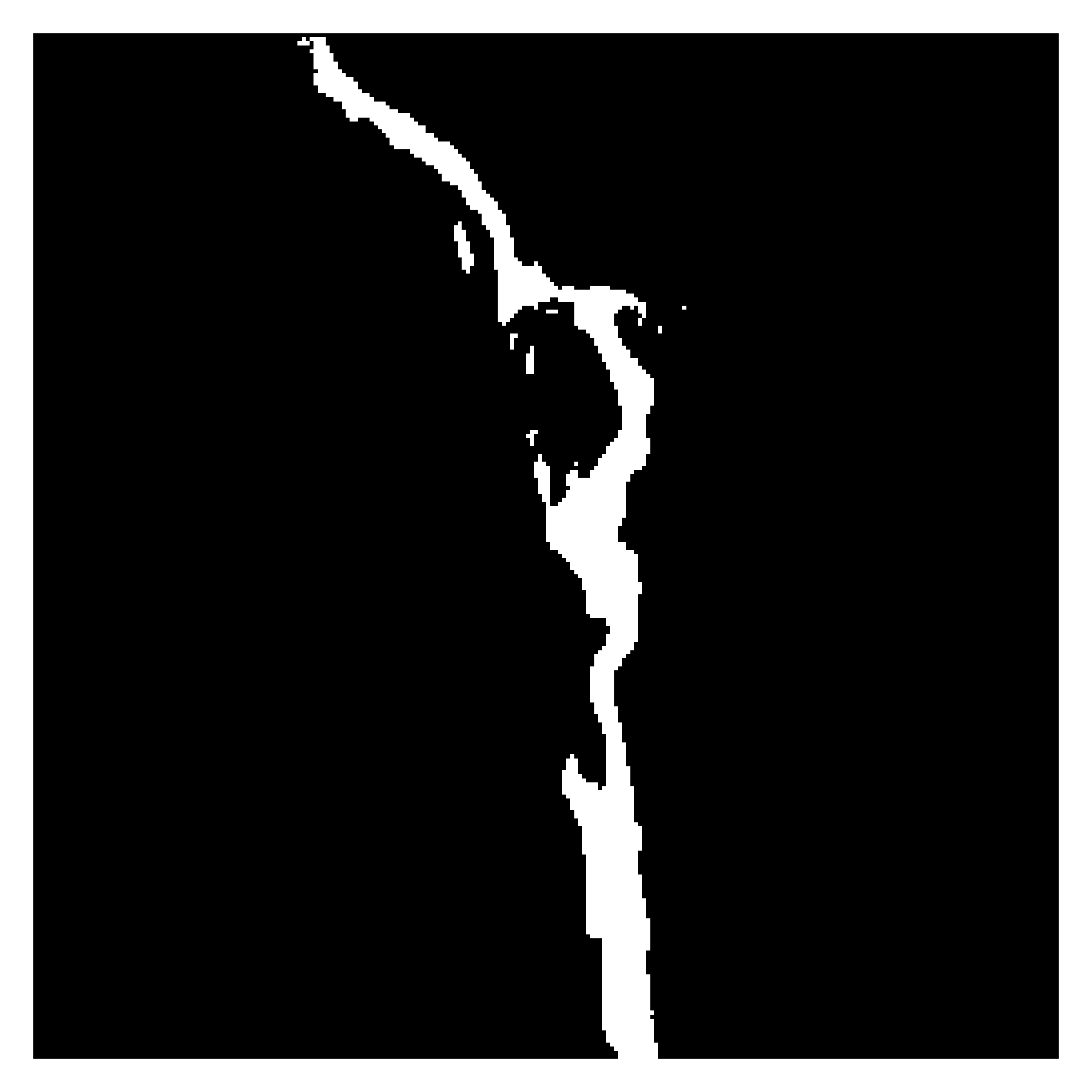}}
    \end{subfigure}
    \begin{subfigure}[b]{0.1\textwidth}
        \adjustbox{trim=10 10 10 10,clip,width=1.6cm,height=1.6cm}{\includegraphics{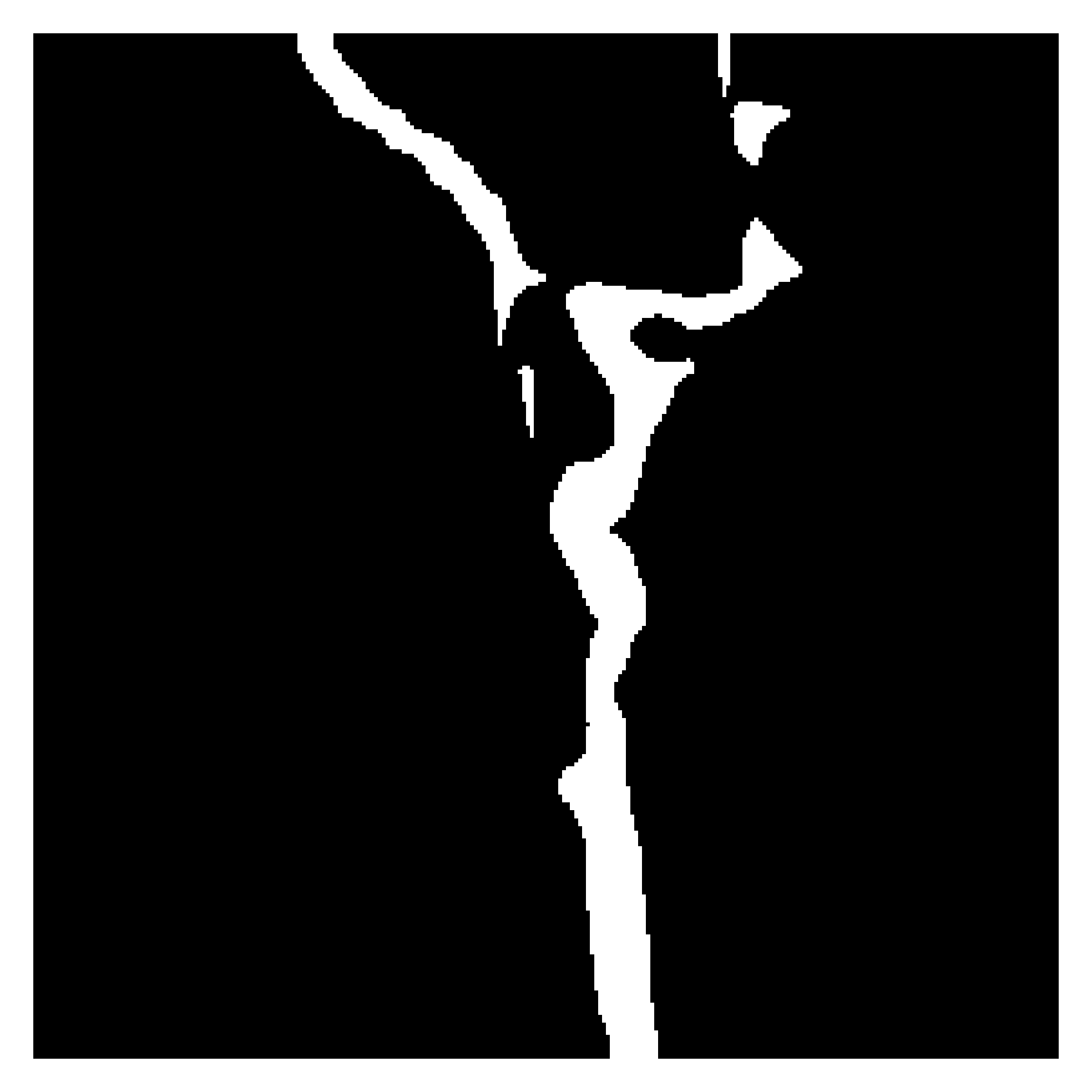}}
    \end{subfigure}

    \vspace{0.2cm}
    
    \begin{subfigure}[b]{0.1\textwidth}
        \adjustbox{trim=10 10 10 10,clip,width=1.6cm,height=1.6cm}{\includegraphics{figures/concrete3k_zero_shot/2/418_44.jpg}}
    \end{subfigure}
    \begin{subfigure}[b]{0.1\textwidth}
        \adjustbox{trim=10 10 10 10,clip,width=1.6cm,height=1.6cm}{\includegraphics{figures/concrete3k_zero_shot/2/418_44_mask.jpg}}
    \end{subfigure}  
    \begin{subfigure}[b]{0.1\textwidth}
        \adjustbox{trim=10 10 10 10,clip,width=1.6cm,height=1.6cm}{\includegraphics{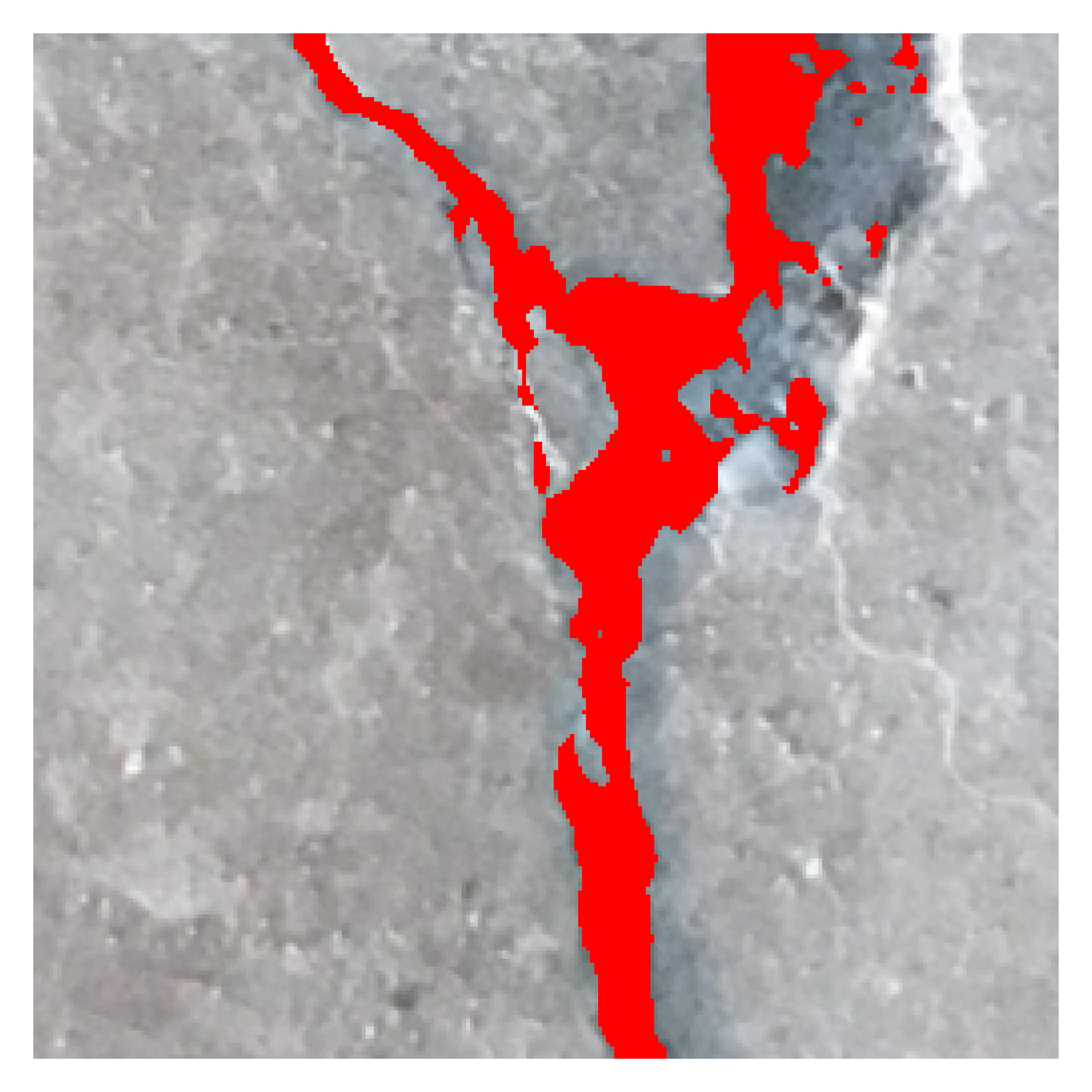}}
    \end{subfigure}
    \begin{subfigure}[b]{0.1\textwidth}
        \adjustbox{trim=10 10 10 10,clip,width=1.6cm,height=1.6cm}{\includegraphics{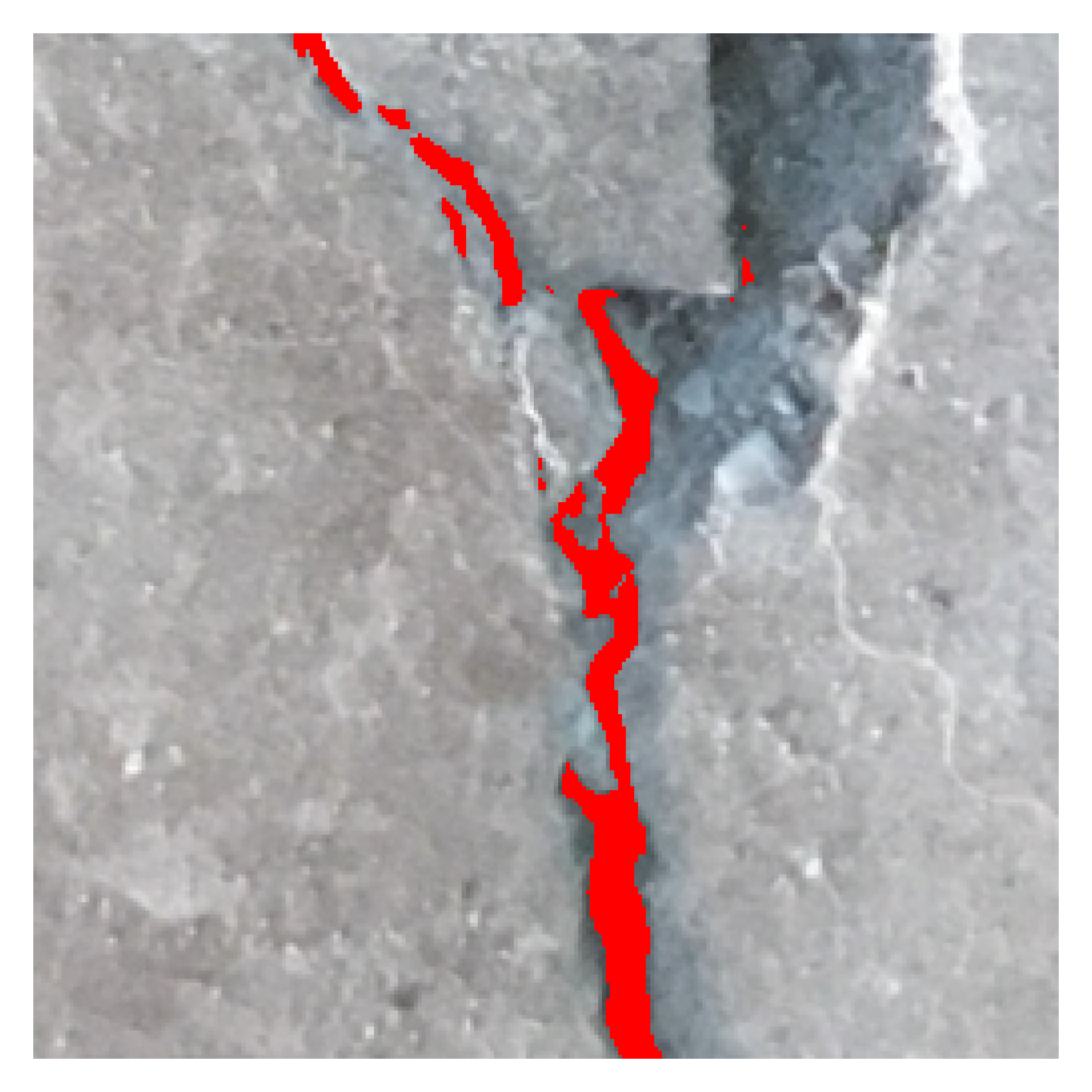}}
    \end{subfigure}
    \begin{subfigure}[b]{0.1\textwidth}
        \adjustbox{trim=10 10 10 10,clip,width=1.6cm,height=1.6cm}{\includegraphics{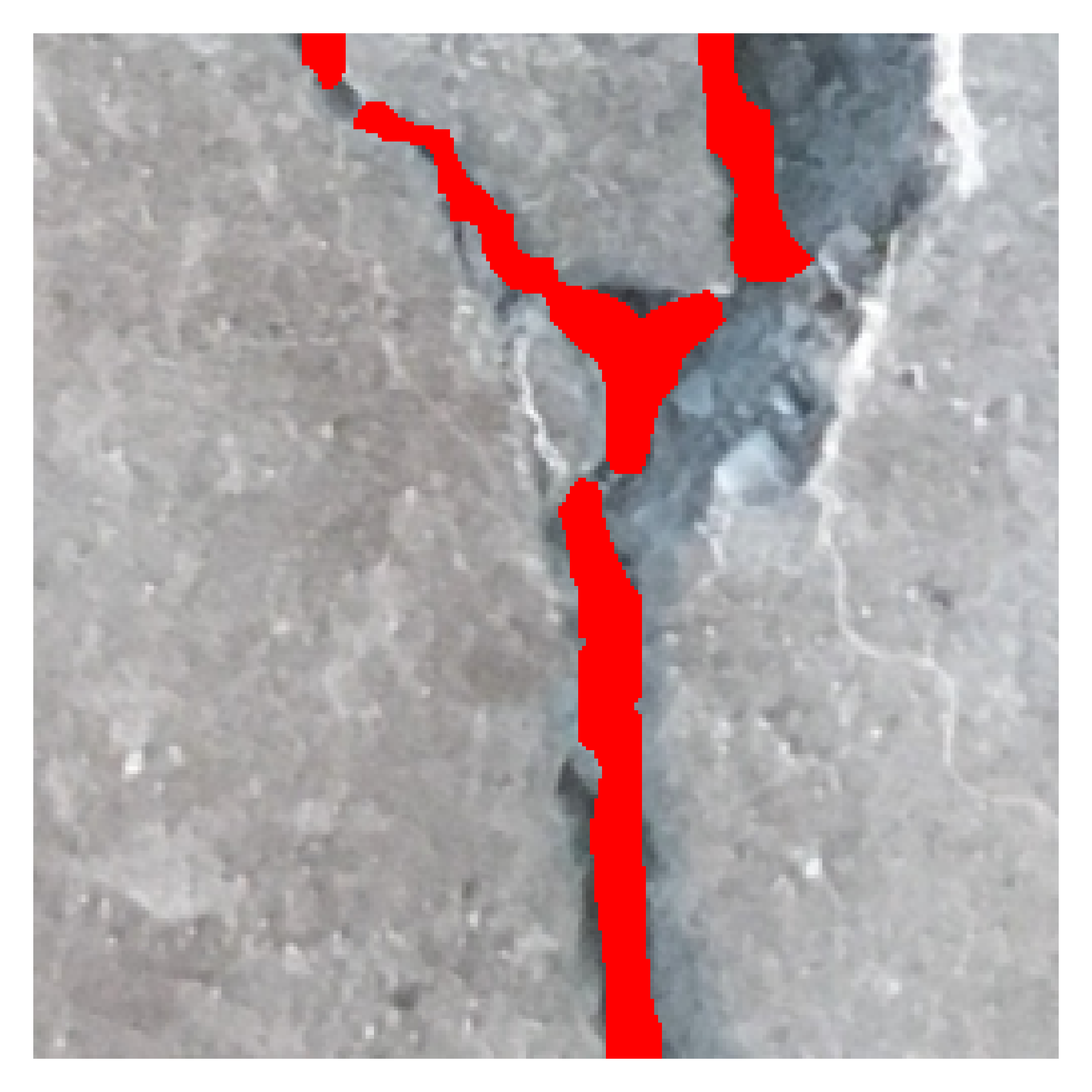}}
    \end{subfigure}
    \begin{subfigure}[b]{0.1\textwidth}
        \adjustbox{trim=10 10 10 10,clip,width=1.6cm,height=1.6cm}{\includegraphics{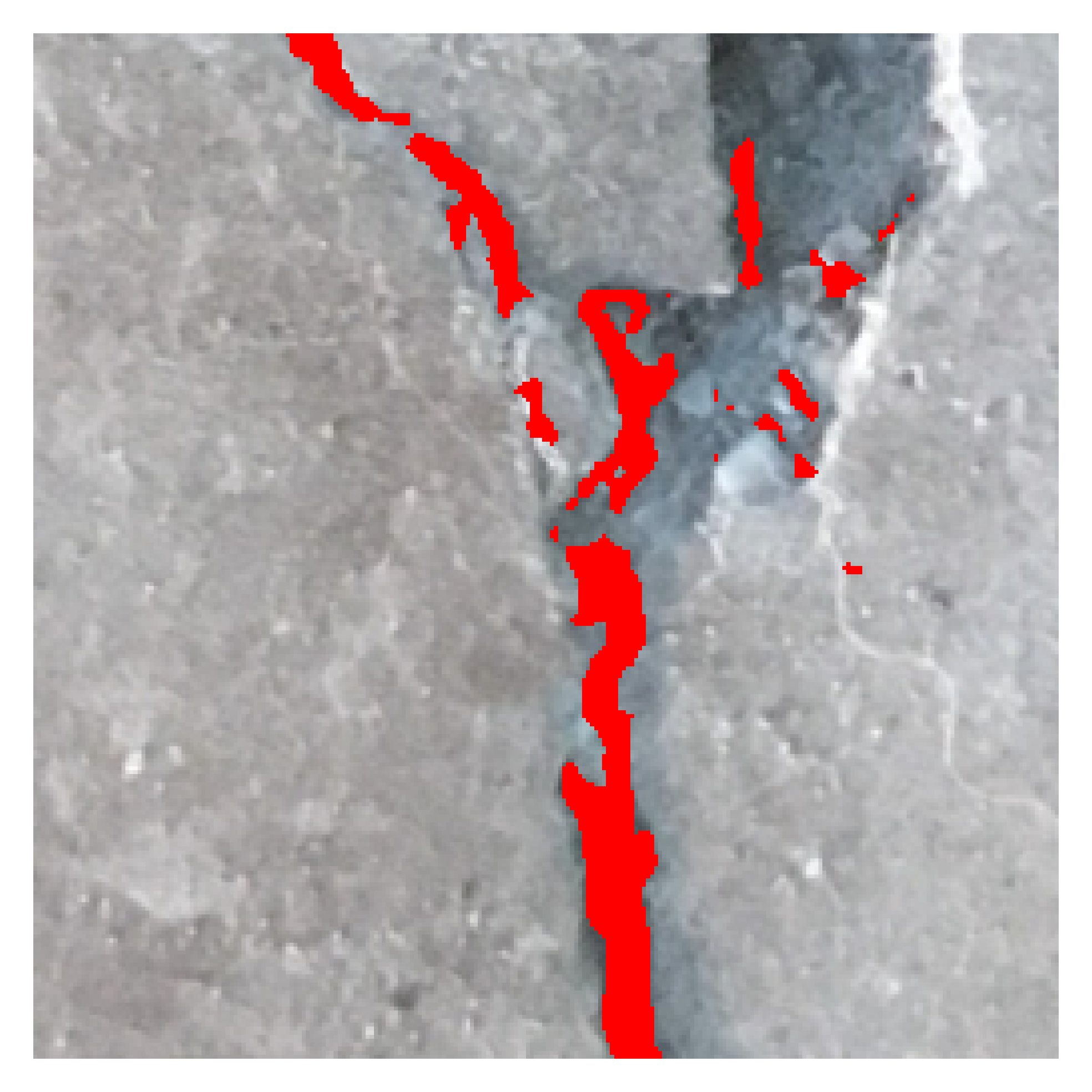}}
    \end{subfigure}
    \begin{subfigure}[b]{0.1\textwidth}
        \adjustbox{trim=10 10 10 10,clip,width=1.6cm,height=1.6cm}{\includegraphics{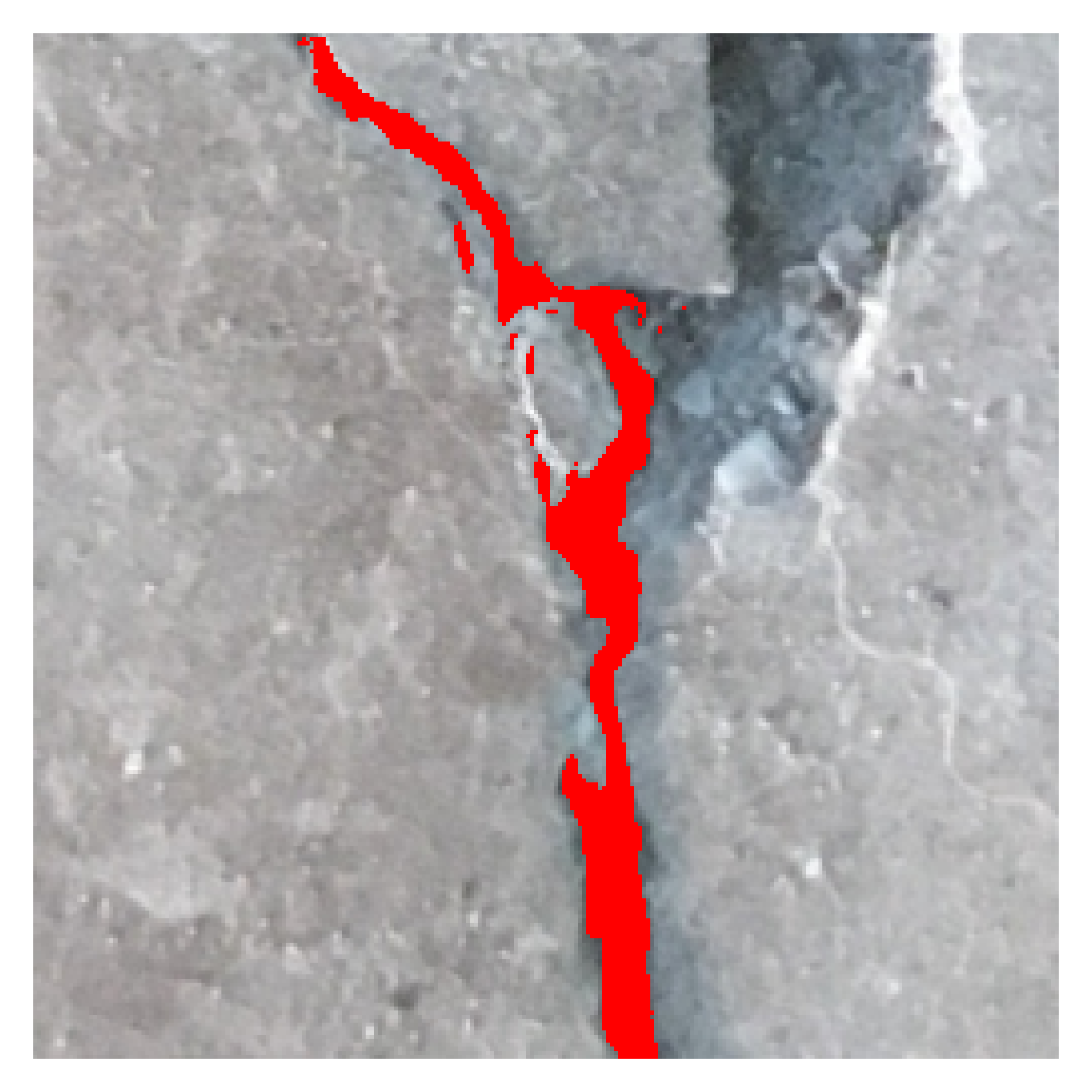}}
    \end{subfigure}
    \begin{subfigure}[b]{0.1\textwidth}
        \adjustbox{trim=10 10 10 10,clip,width=1.6cm,height=1.6cm}{\includegraphics{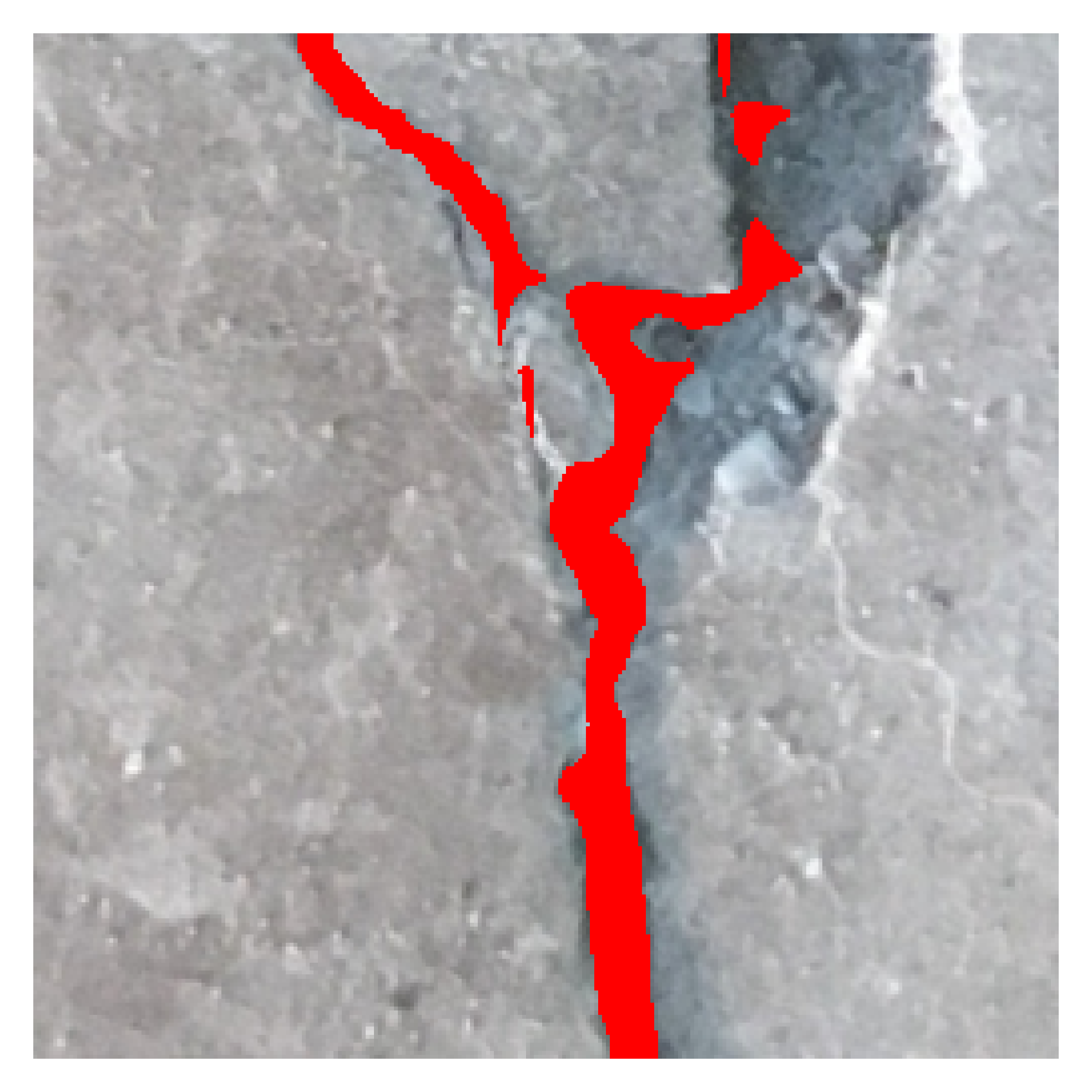}}
    \end{subfigure}
    \vspace{0.5cm}

    \begin{subfigure}[b]{0.1\textwidth}
        \adjustbox{trim=10 10 10 10,clip,width=1.6cm,height=1.6cm}{\includegraphics{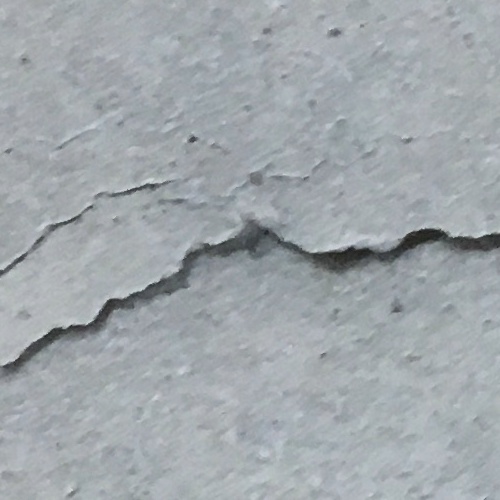}}
    \end{subfigure}
    \begin{subfigure}[b]{0.1\textwidth}
        \adjustbox{trim=10 10 10 10,clip,width=1.6cm,height=1.6cm}{\includegraphics{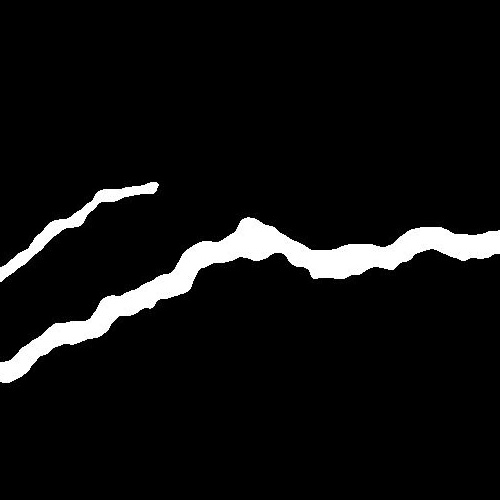}}
    \end{subfigure}  
    \begin{subfigure}[b]{0.1\textwidth}
        \adjustbox{trim=10 10 10 10,clip,width=1.6cm,height=1.6cm}{\includegraphics{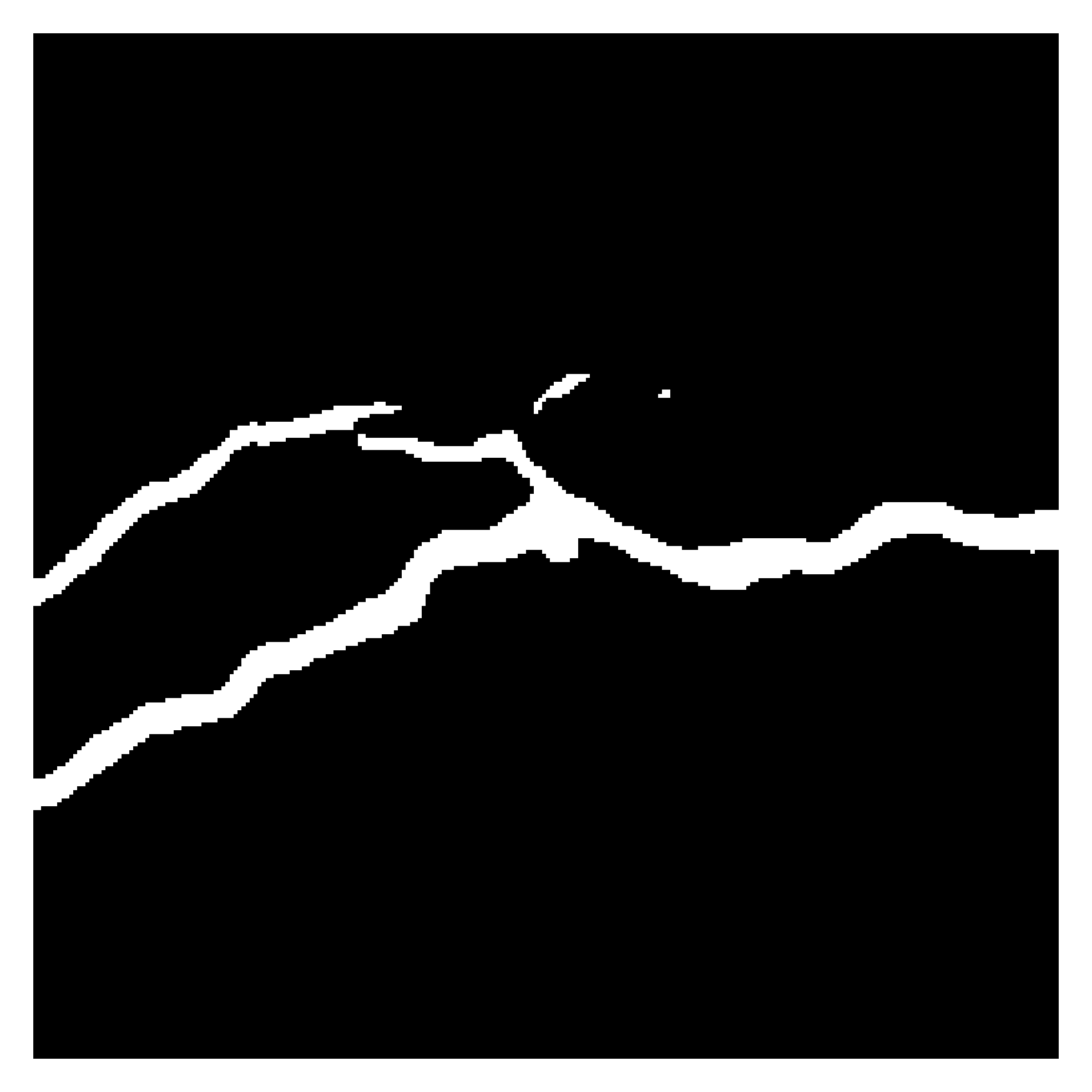}}
    \end{subfigure}
    \begin{subfigure}[b]{0.1\textwidth}
        \adjustbox{trim=10 10 10 10,clip,width=1.6cm,height=1.6cm}{\includegraphics{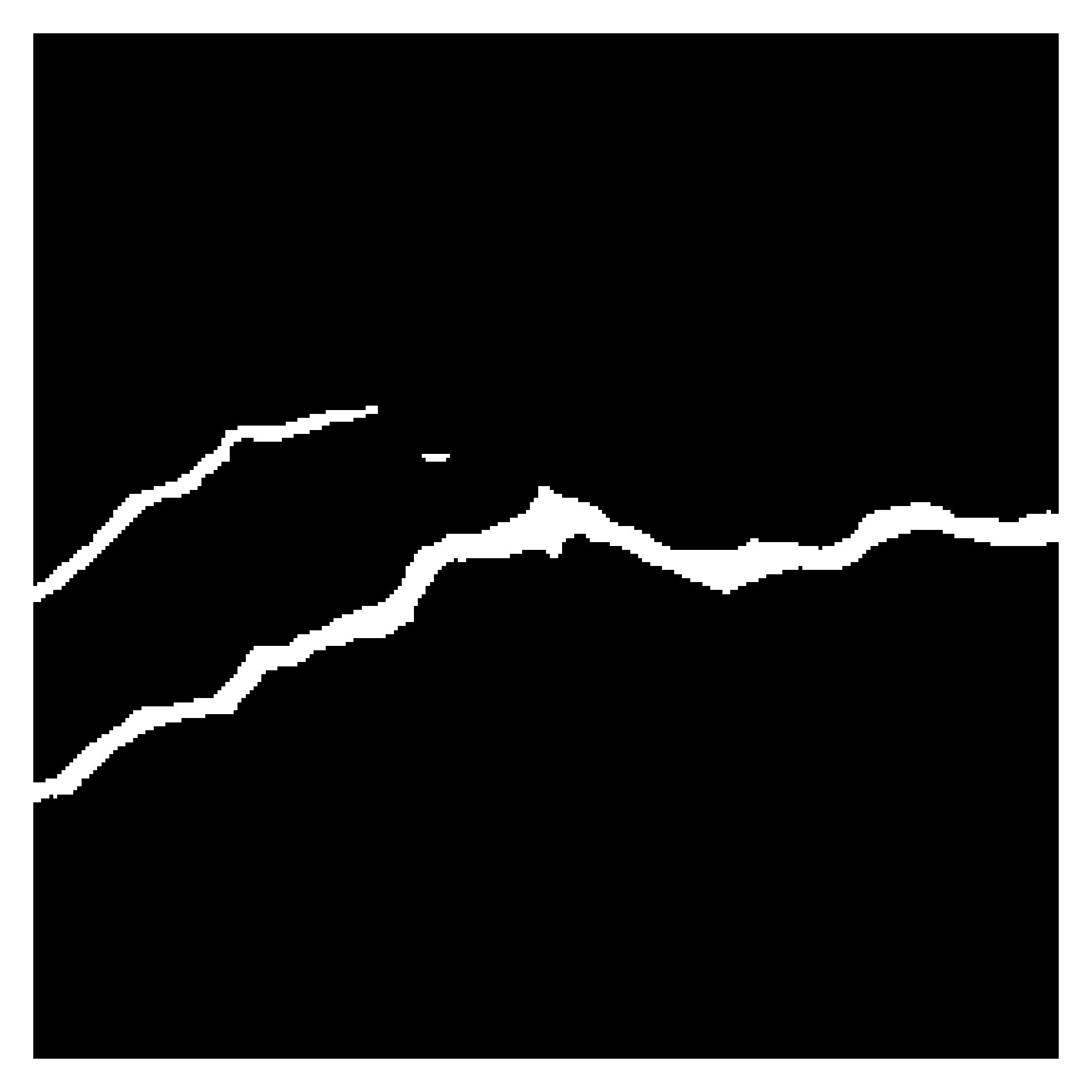}}
    \end{subfigure}
    \begin{subfigure}[b]{0.1\textwidth}
        \adjustbox{trim=10 10 10 10,clip,width=1.6cm,height=1.6cm}{\includegraphics{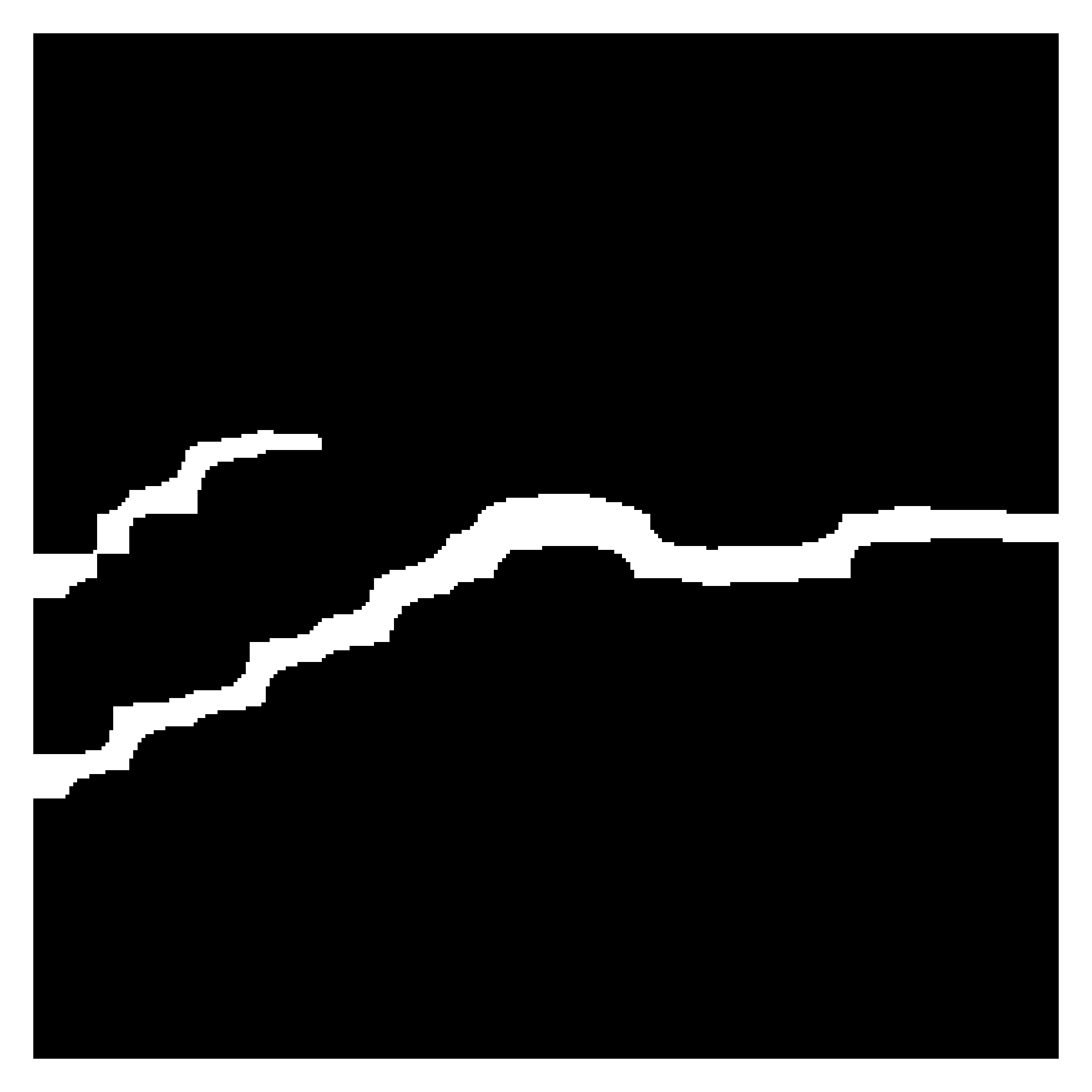}}
    \end{subfigure}
    \begin{subfigure}[b]{0.1\textwidth}
        \adjustbox{trim=10 10 10 10,clip,width=1.6cm,height=1.6cm}{\includegraphics{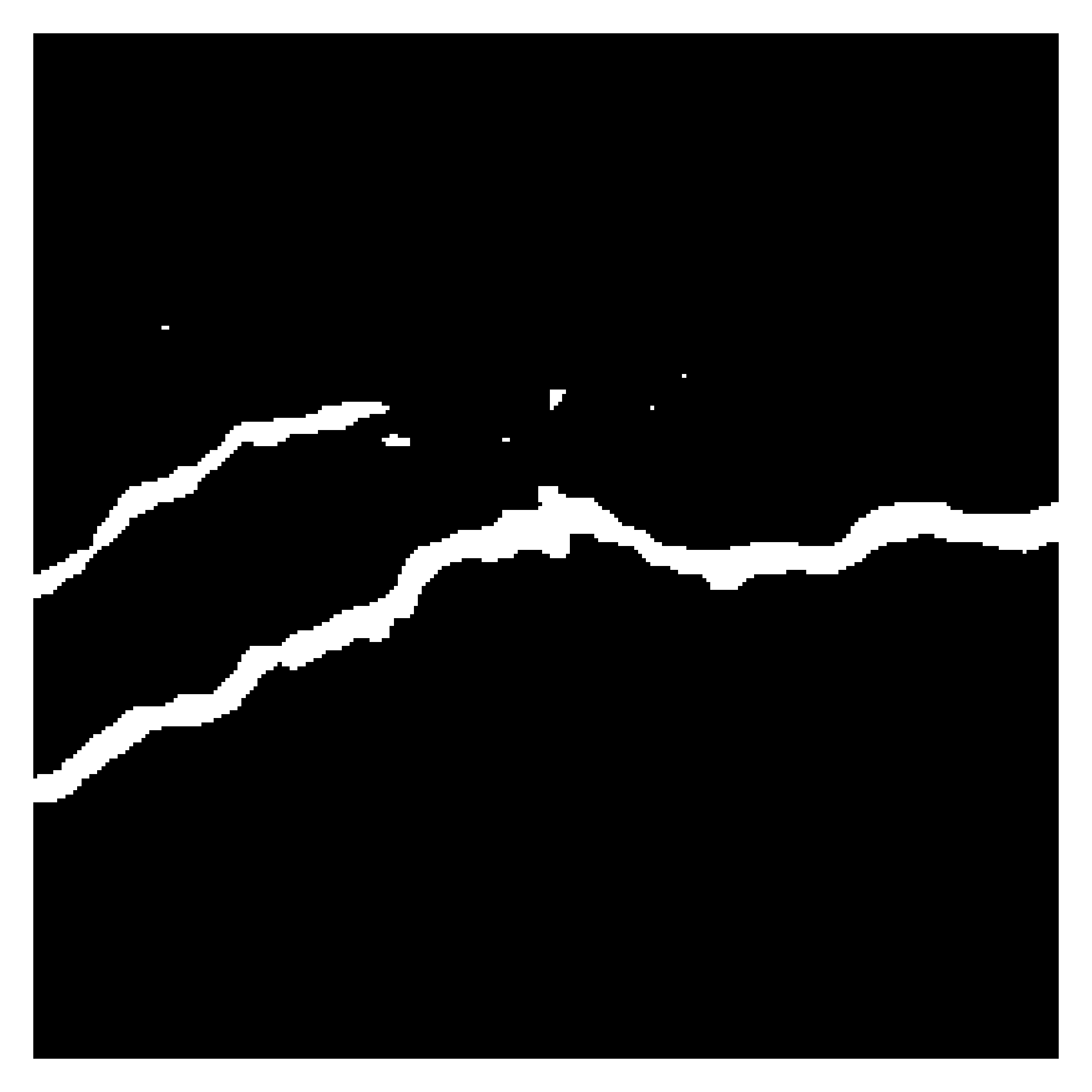}}
    \end{subfigure}
    \begin{subfigure}[b]{0.1\textwidth}
        \adjustbox{trim=10 10 10 10,clip,width=1.6cm,height=1.6cm}{\includegraphics{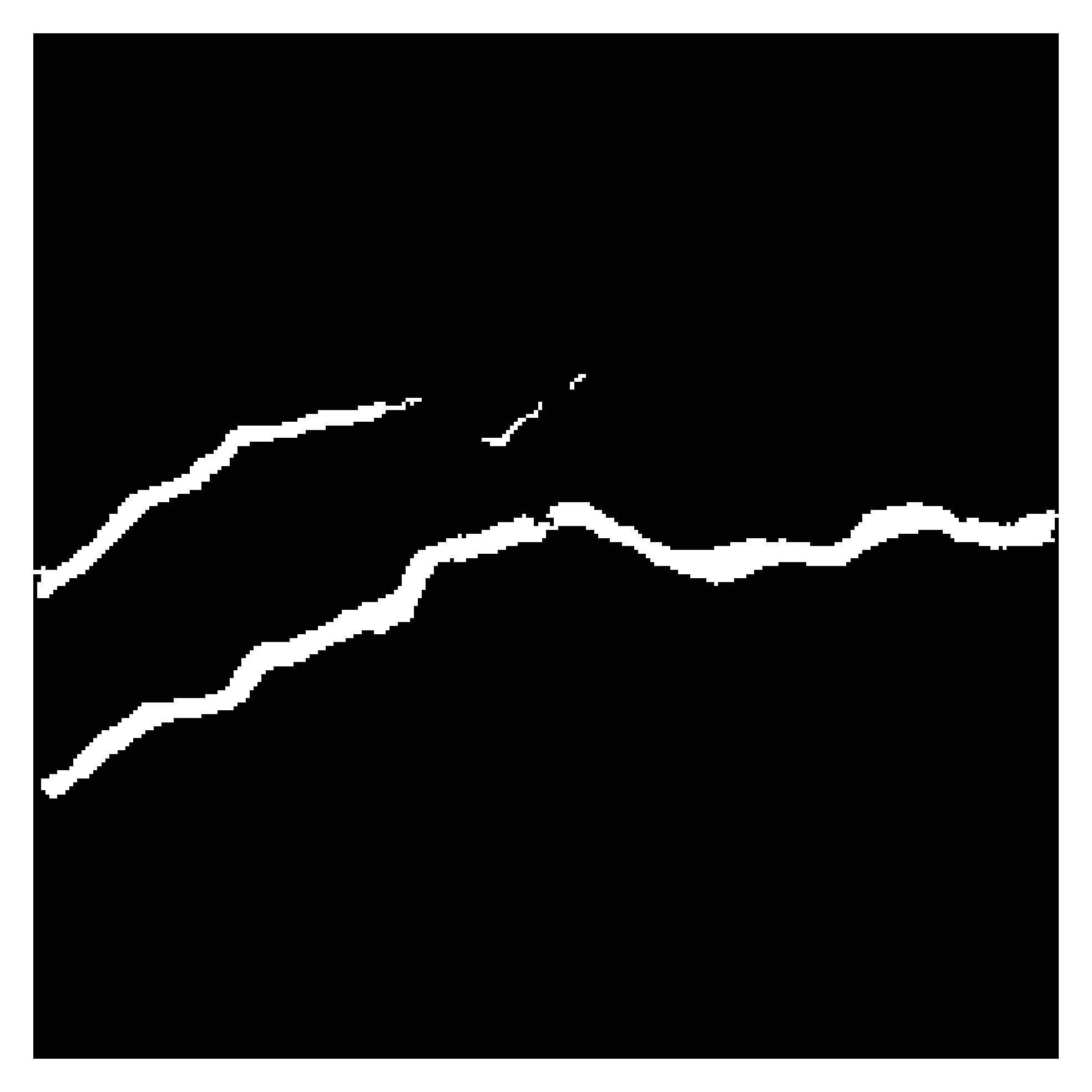}}
    \end{subfigure}
    \begin{subfigure}[b]{0.1\textwidth}
        \adjustbox{trim=10 10 10 10,clip,width=1.6cm,height=1.6cm}{\includegraphics{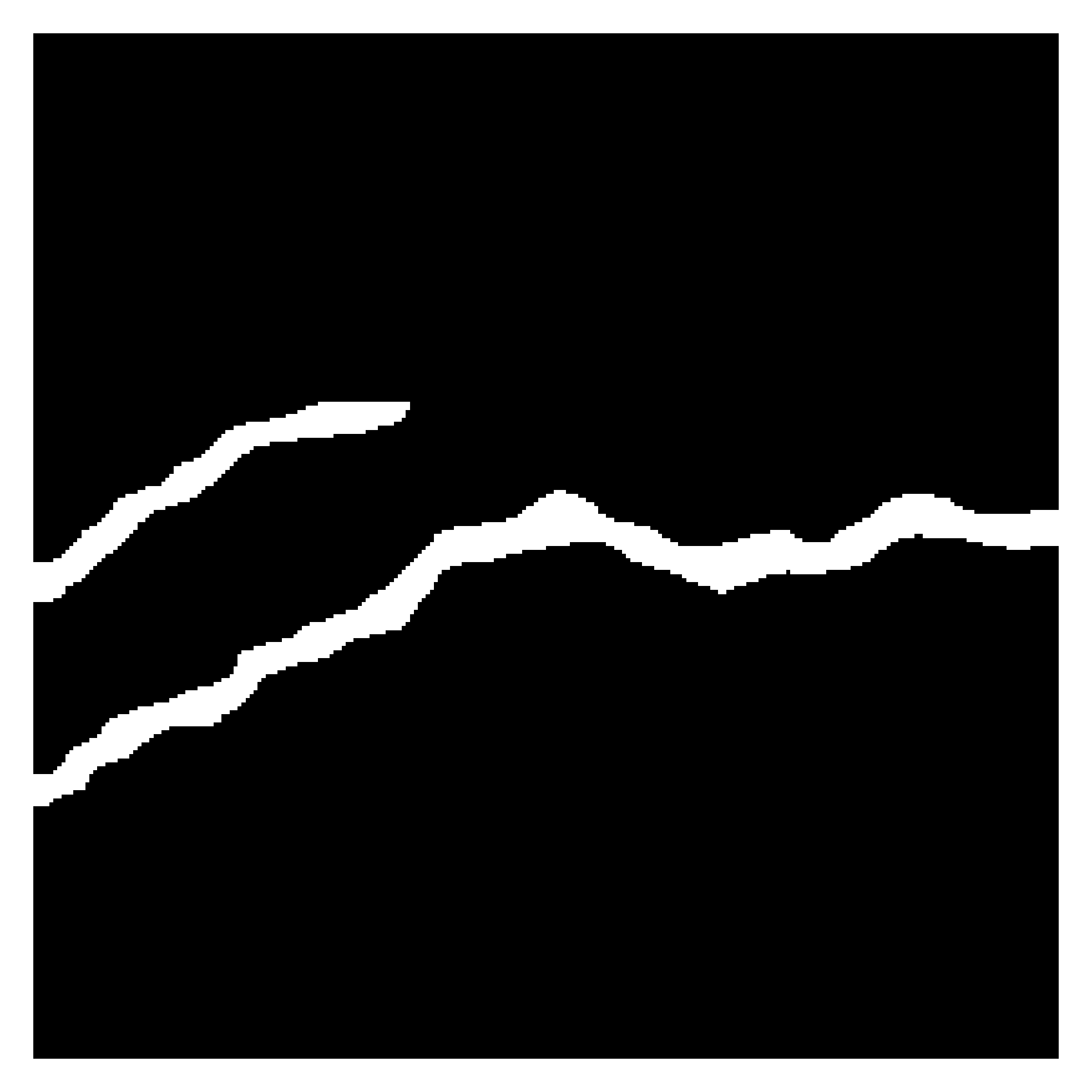}}
    \end{subfigure}

    \vspace{0.2cm}
    
    \begin{subfigure}[b]{0.1\textwidth}
        \adjustbox{trim=10 10 10 10,clip,width=1.6cm,height=1.6cm}{\includegraphics{figures/concrete3k_zero_shot/3/501_25.jpg}}
    \end{subfigure}
    \begin{subfigure}[b]{0.1\textwidth}
        \adjustbox{trim=10 10 10 10,clip,width=1.6cm,height=1.6cm}{\includegraphics{figures/concrete3k_zero_shot/3/501_25_mask.jpg}}
    \end{subfigure}  
    \begin{subfigure}[b]{0.1\textwidth}
        \adjustbox{trim=10 10 10 10,clip,width=1.6cm,height=1.6cm}{\includegraphics{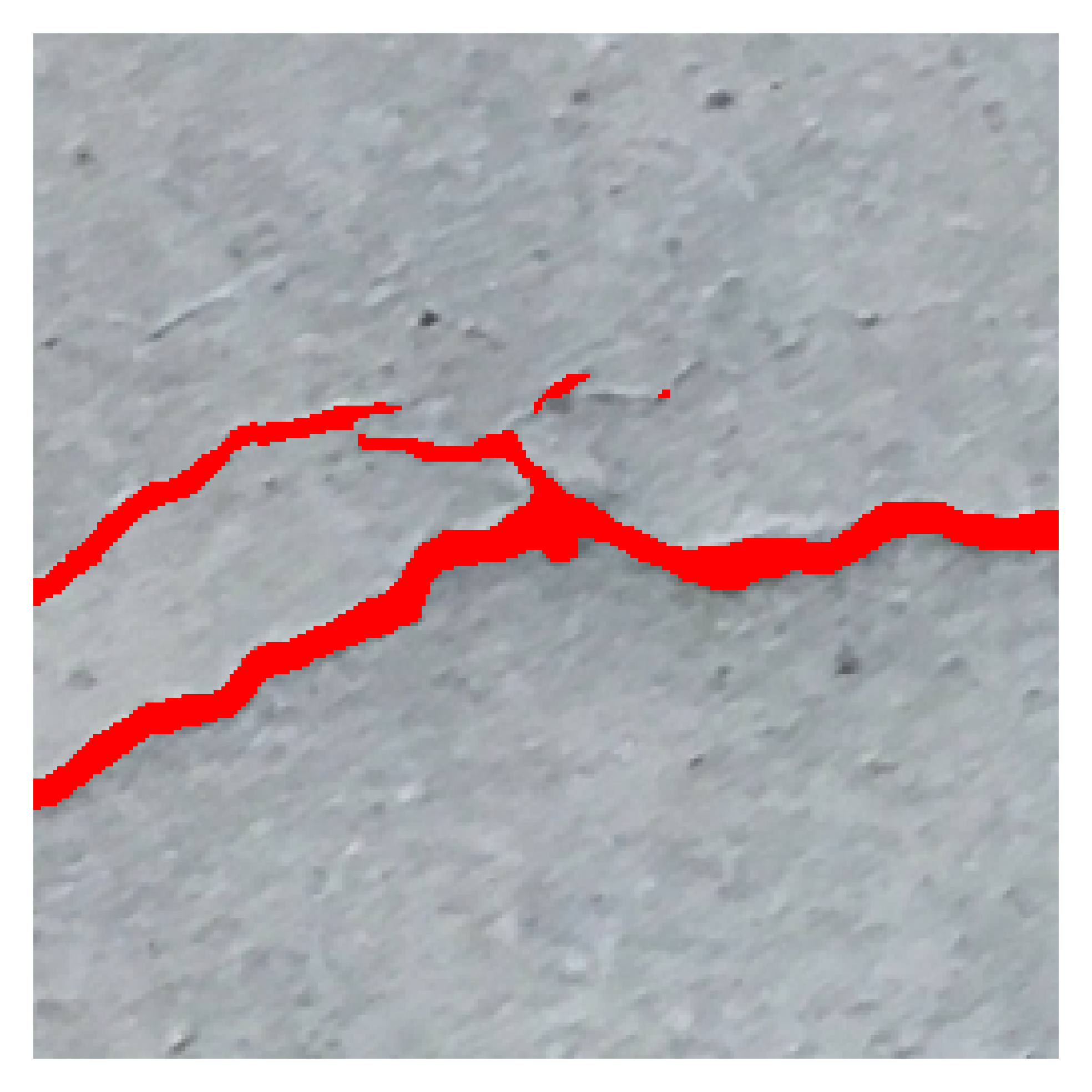}}
    \end{subfigure}
    \begin{subfigure}[b]{0.1\textwidth}
        \adjustbox{trim=10 10 10 10,clip,width=1.6cm,height=1.6cm}{\includegraphics{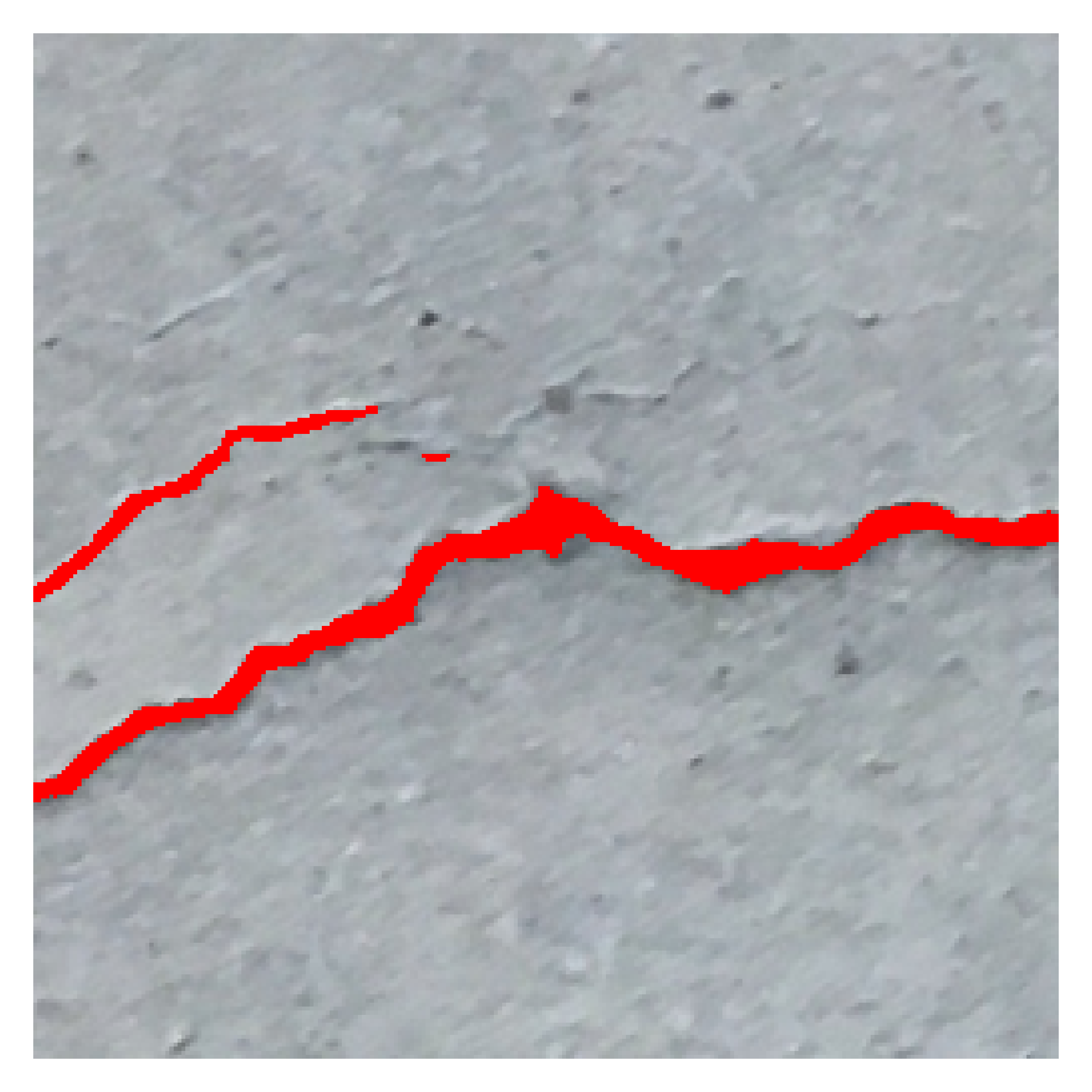}}
    \end{subfigure}
    \begin{subfigure}[b]{0.1\textwidth}
        \adjustbox{trim=10 10 10 10,clip,width=1.6cm,height=1.6cm}{\includegraphics{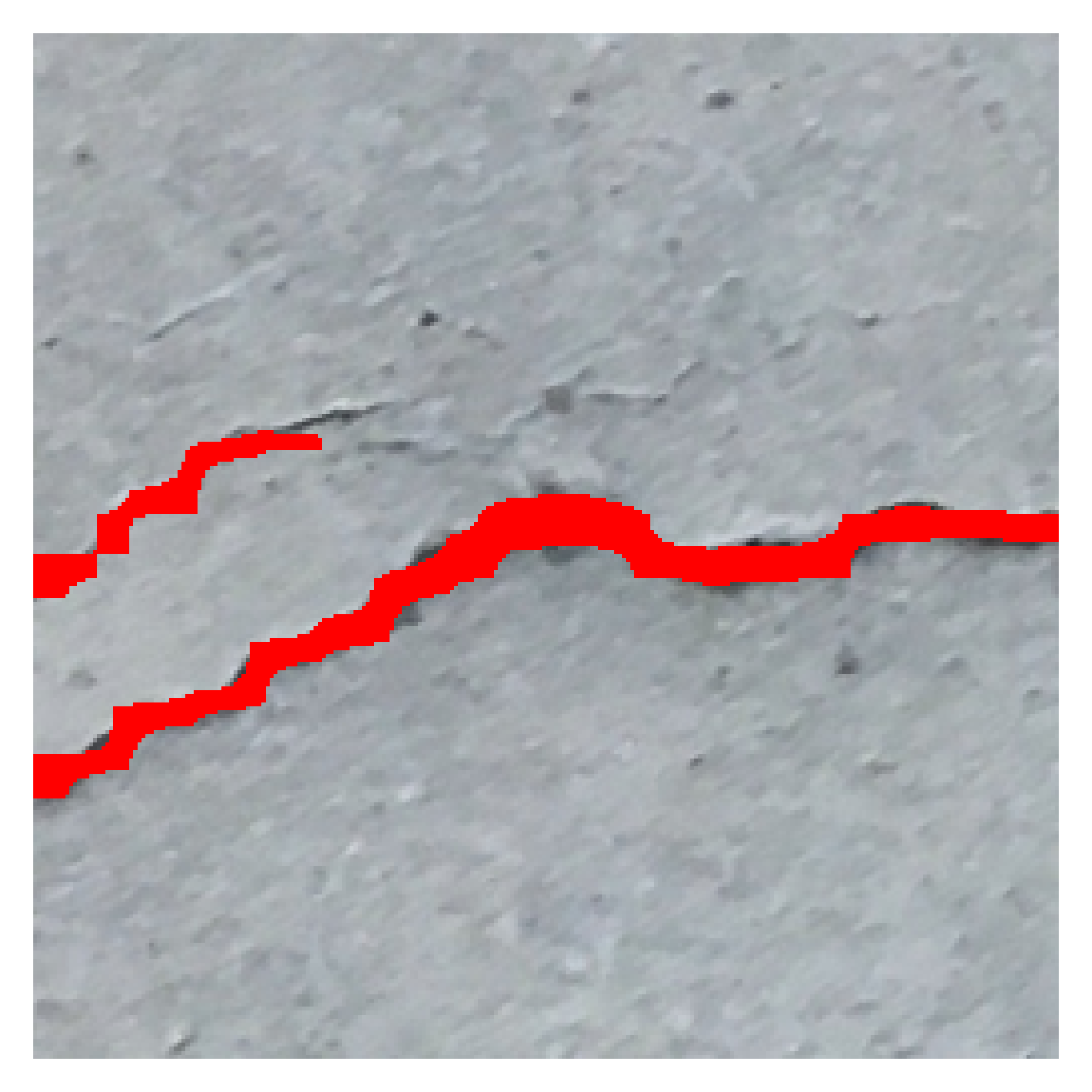}}
    \end{subfigure}
    \begin{subfigure}[b]{0.1\textwidth}
        \adjustbox{trim=10 10 10 10,clip,width=1.6cm,height=1.6cm}{\includegraphics{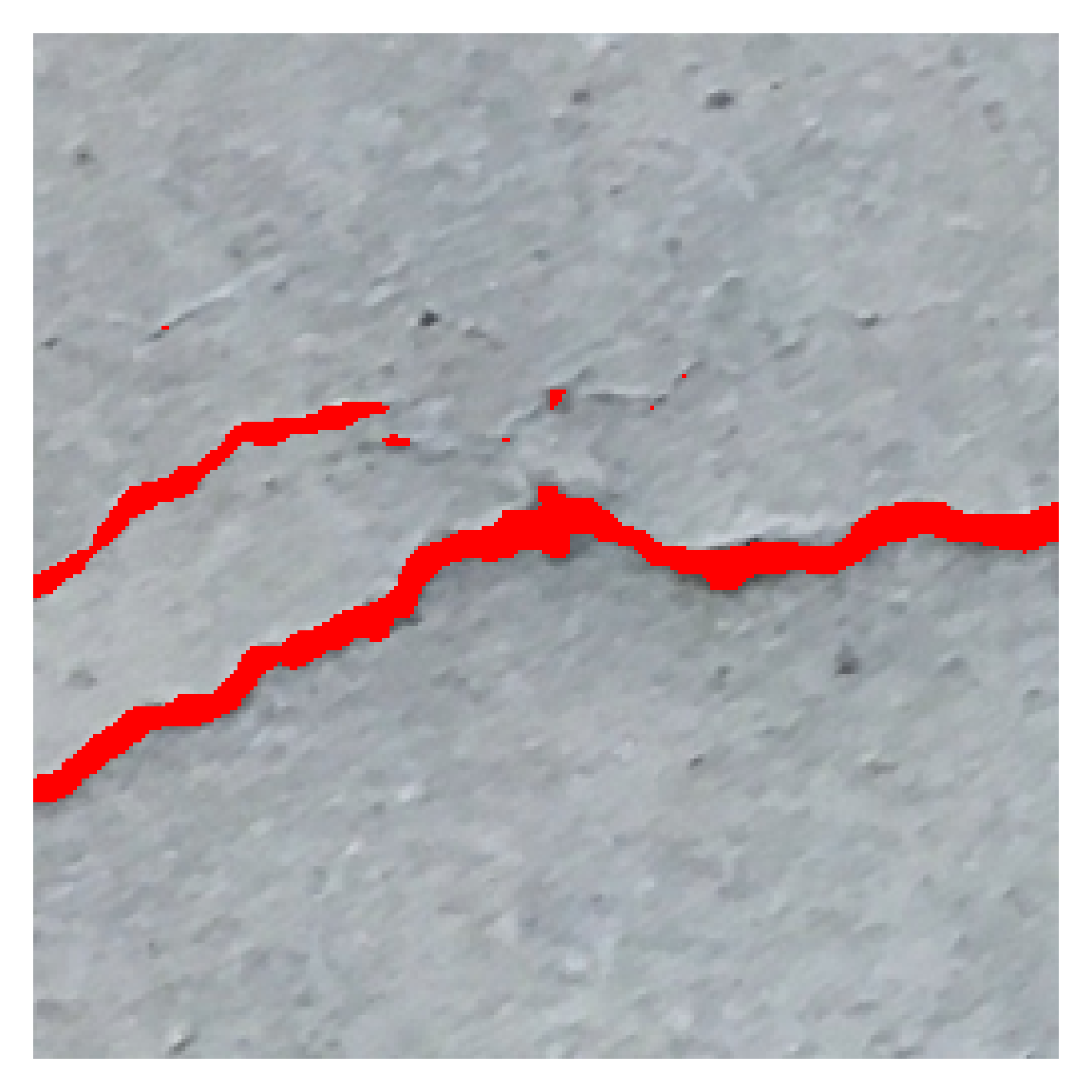}}
    \end{subfigure}
    \begin{subfigure}[b]{0.1\textwidth}
        \adjustbox{trim=10 10 10 10,clip,width=1.6cm,height=1.6cm}{\includegraphics{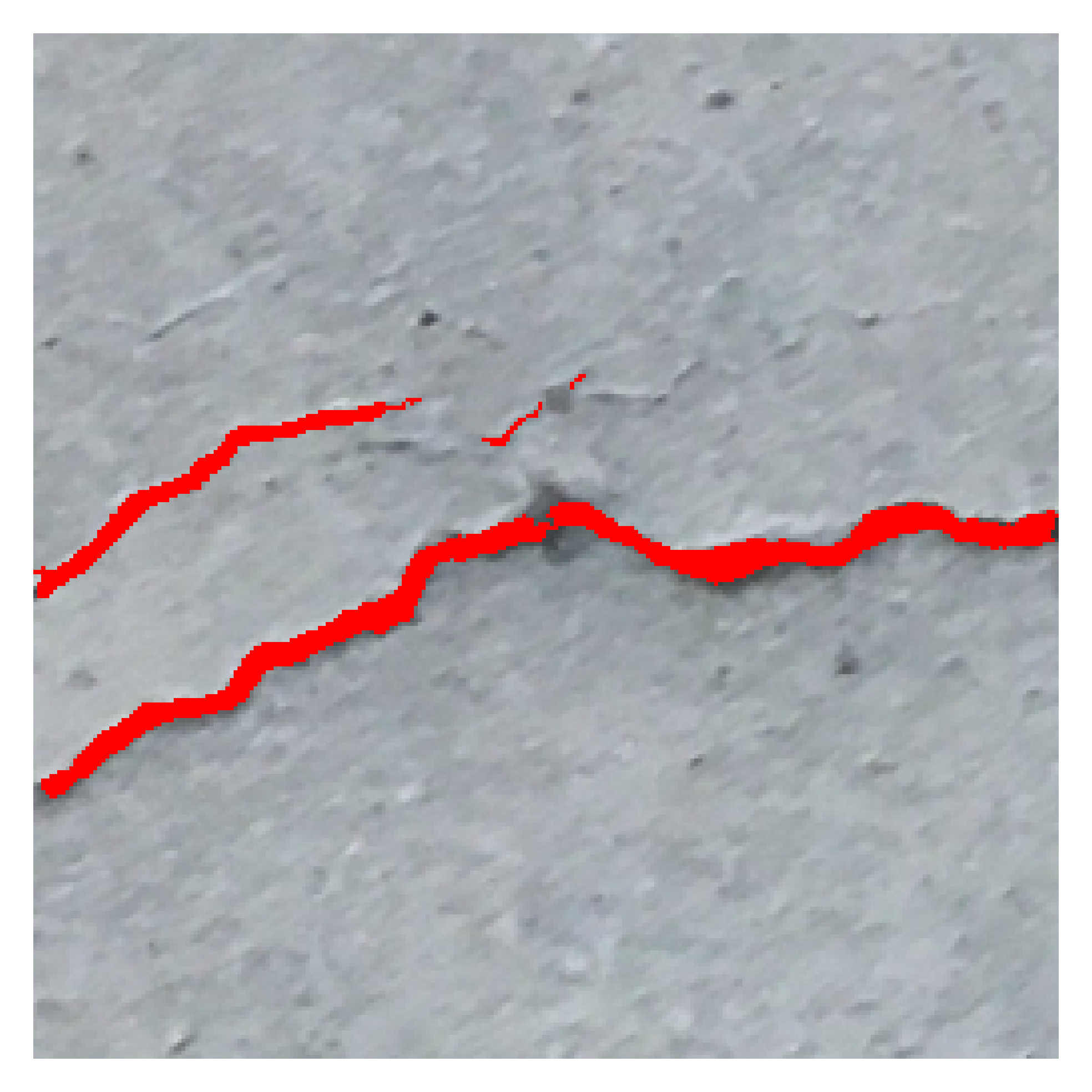}}
    \end{subfigure}
    \begin{subfigure}[b]{0.1\textwidth}
        \adjustbox{trim=10 10 10 10,clip,width=1.6cm,height=1.6cm}{\includegraphics{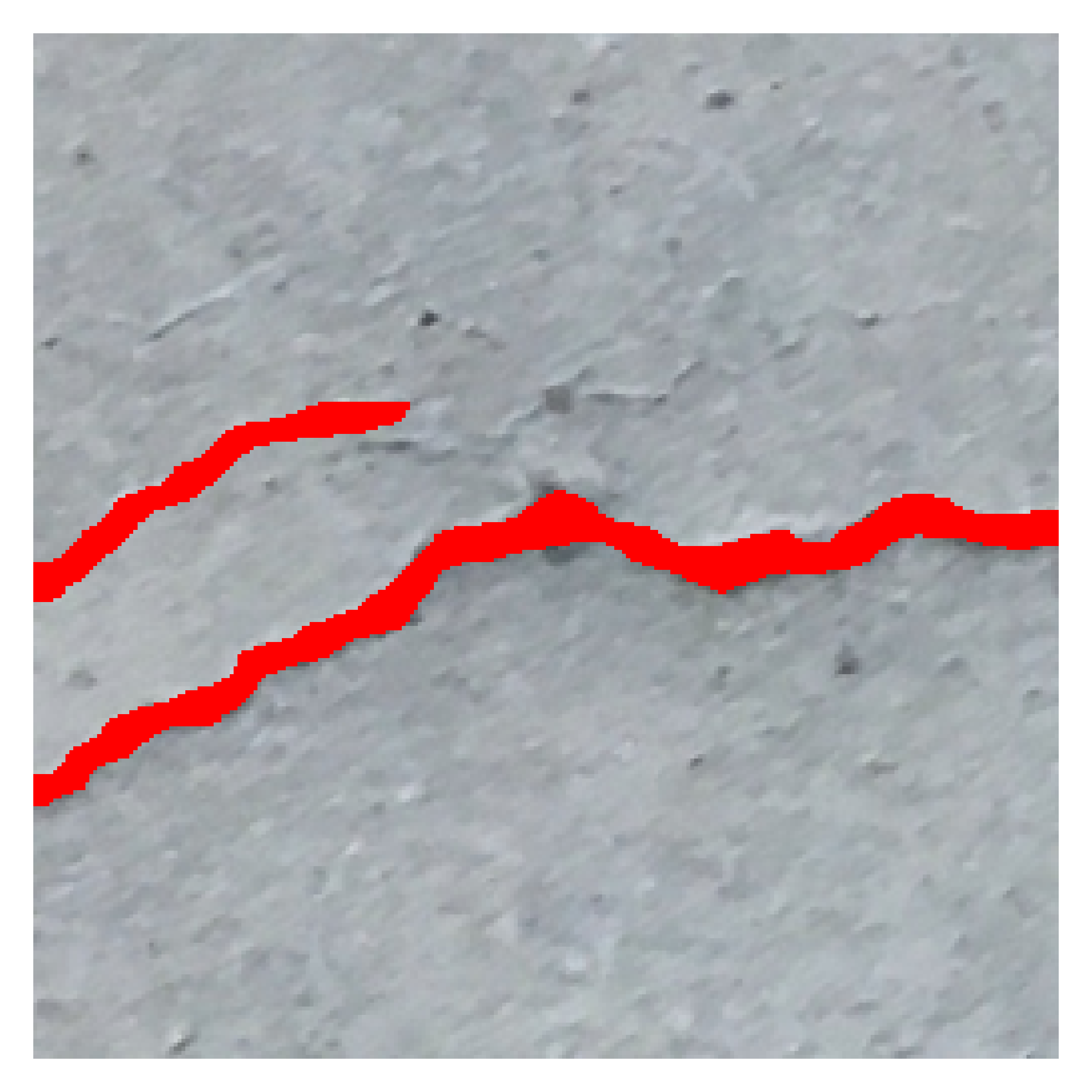}}
    \end{subfigure}
    \vspace{0.05cm}
    \vspace{0.2cm}
    \caption{Zero-shot performance of the fine-tuned models on Concrete3k Dataset}
    \label{fig:zero_shot_concrete3k_qualitative}
\end{figure}

Samples of qualitative segmentation results are presented in Figures \ref{fig:zero_shot_road420_qualitative}, \ref{fig:zero_shot_facade390_qualitative}, and \ref{fig:zero_shot_concrete3k_qualitative}.
It can be observed that the SAC model predicts crack pixels with better continuity and effectively preserves the structural integrity of detected crack paths. Additionally, the model has high precision in detecting crack boundaries, capturing fine details even in complex surface conditions.

\clearpage
\begin{landscape}
\thispagestyle{empty} 
    \begin{table}[h]
        \centering
        \caption{Zero-Shot evaluation results of the full-tuned segmentation methods.}
        \label{table:Fulltuning ZeroShot Results}
        \small
        \renewcommand{\arraystretch}{1.25}
        \begin{tabular}{l c | c c | c c | c c | c c}
            \hline\hline
            \multicolumn{2}{c|}{} & 
            \multicolumn{2}{c|}{Road420} &
            \multicolumn{2}{c|}{Facade390} & 
            \multicolumn{2}{c|}{Concrete3k} & 
            \multicolumn{2}{c}{Avg. Performance} \\
            \multicolumn{1}{c}{Model} &  
            \multicolumn{1}{c|}{\parbox[t]{2cm}{\# Tunable \\Parameters}} &
            \multicolumn{1}{c}{\parbox[t]{1.5cm}{F1-Score\\ (\%)}} & 
            \multicolumn{1}{c|}{\parbox[t]{1.5cm}{IoU\\ (\%)}} & 
            \multicolumn{1}{c}{\parbox[t]{1.5cm}{F1-Score\\ (\%)}} & 
            \multicolumn{1}{c|}{\parbox[t]{1.5cm}{IoU\\ (\%)}} & 
            \multicolumn{1}{c}{\parbox[t]{1.5cm}{F1-Score \\(\%)}} & 
            \multicolumn{1}{c|}{\parbox[t]{1.5cm}{IoU\\ (\%)}} &
            \multicolumn{1}{c}{\parbox[t]{2cm}{F1-Score \\(avg., std.)}} & 
            \multicolumn{1}{c}{\parbox[t]{2cm}{IoU\\ (avg., std.)}} \\
            \hline\hline
            DeepCrackZ & 31M & 56.73 & 39.6 & 64.47 & 47.62 & 68.85 & 52.5 & $63.35_{\pm 5.01}$ & $46.57_{\pm5.82}$ \\
            DeepCrackL & 15M & 34.71 & 21.02 & 68.76 & 52.40 & 64.37 & 47.47 & $55.95_{\pm15.12}$ & $40.3_{\pm13.78}$ \\
            CrackFormer & 5M & 46.83 & 30.57 & 43.34 & 27.68 & 68.32 & 51.89 & $52.83_{\pm11.05}$ & $36.71_{\pm10.80}$ \\
            \hline
            SegFormer & 3.7M & \textbf{58.36} & \textbf{41.21} & 53.4 & 36.46 & 69.29 & 53.02 & $60.35_{\pm6.64}$ & $43.56_{\pm6.96}$ \\
            U-Net & 31M & 30.63 & 18.11 & \textbf{69.00} & \textbf{52.69} & 63.85 & 46.91 & $54.49_{\pm17.00}$ & $39.24_{\pm15.12}$ \\
            DeepLabv3+-ResNet50 & 42M & 52.87 & 35.94 & 56.71 & 39.62 & 73.86 & 58.56 & $61.15_{\pm9.13}$ & $44.71_{\pm9.91}$ \\
            \textbf{DeepLabv3+-ResNet101} & 61M & 51.68 & 34.68 & 58.88 & 41.75 & \textbf{74.53} & \textbf{59.41} & $\mathbf{61.70}_{\pm\mathbf{9.54}}$ & $\mathbf{45.28}_{\pm\mathbf{10.40}}$ \\
            \hline
\end{tabular}%
\normalsize
\end{table}
\end{landscape}
\clearpage

\clearpage
\begin{landscape}
\thispagestyle{empty}
\begin{table}[h]
\centering
\caption{Zero-Shot evaluation results of the fine-tuned segmentation methods.}
\label{table:Finetuning ZeroShot Results}
\small
\renewcommand{\arraystretch}{1.25}
\begin{tabular}{l c | c c | c c | c c | c c}
    \hline\hline
    \multicolumn{2}{c|}{} & 
            \multicolumn{2}{c|}{Road420} &
            \multicolumn{2}{c|}{Facade390} & 
            \multicolumn{2}{c|}{Concrete3k} & 
            \multicolumn{2}{c}{Avg. Performance} \\
            \multicolumn{1}{c}{Model} &  
            \multicolumn{1}{c|}{\parbox[t]{2cm}{\# Tunable \\Parameters}} &
            \multicolumn{1}{c}{\parbox[t]{1.5cm}{F1-Score\\ (\%)}} & 
            \multicolumn{1}{c|}{\parbox[t]{1.5cm}{IoU\\ (\%)}} & 
            \multicolumn{1}{c}{\parbox[t]{1.5cm}{F1-Score\\ (\%)}} & 
            \multicolumn{1}{c|}{\parbox[t]{1.5cm}{IoU\\ (\%)}} & 
            \multicolumn{1}{c}{\parbox[t]{1.5cm}{F1-Score \\(\%)}} & 
            \multicolumn{1}{c|}{\parbox[t]{1.5cm}{IoU\\ (\%)}} &
            \multicolumn{1}{c}{\parbox[t]{2cm}{F1-Score \\(avg., std.)}} & 
            \multicolumn{1}{c}{\parbox[t]{2cm}{IoU\\ (avg., std.)}} \\
    \hline\hline
    DeepCrackZ & - & - & - & - & - & - & - & - & - \\
    DeepCrackL & 8K & 40.69 & 25.57 & 61.17 & 44.09 & 55.15 & 38.08 & $52.34_{\pm8.60}$ & $35.91_{\pm7.71}$\\
    CrackFormer & 39K & 50.93 & 34.17 & 43.83 & 28.08 & 66.76 & 50.11 & $53.84_{\pm9.59}$ & $37.45_{\pm9.29}$ \\
    \hline
    SegFormer & 7.6K & 48.77 & 32.27 & 56.7 & 39.58 & 70.57 & 54.53 & $58.68_{\pm9.01}$ & $42.13_{\pm9.26}$\\
    U-Net & 55K & 52.72 & 35.83 & \textbf{68.4} & \textbf{51.99} & 69.03 & 52.71 & $63.38_{\pm7.55}$ & $46.84_{\pm7.79}$ \\
    DeepLabv3+-ResNet50 & 57K & 49.93 & 33.28 & 53.85 & 36.86 & 75.03 & 60.05 & $59.59_{\pm11.03}$ & $43.40_{\pm11.87}$ \\
    DeepLabv3+-ResNet101 & 110K & 56.8 & 39.68 & 54.79 & 37.75 & \textbf{76.09} & \textbf{61.42} & $62.56_{\pm9.60}$ & $46.28_{\pm10.73}$\\
    \textbf{SAC} & \textbf{41K} & \textbf{64.22} & \textbf{47.3} & 61.74 & 44.68 & 75.63 & 60.82 & $\mathbf{67.20}_{\pm\mathbf{6.05}}$ & $\mathbf{50.93}_{\pm\mathbf{7.07}}$  \\ 
    \hline
\end{tabular}%
\normalsize
\end{table}
\end{landscape}
\clearpage

\section{Conclusion and Future Work}
This study introduced an efficient selective fine-tuning strategy for crack segmentation by focusing on tuning only the normalization parameters of pre-trained segmentation models. the Segment Any Crack (SAC) model was introduced based on the Segment Anything Model (SAM) and the adaptation strategy proposed in this study. Through extensive experiments on the OmniCrack30k benchmark dataset, the proposed method achieved 61.22\% F1-score and 44.13\% IoU, outperforming full fine-tuning and other common fine-tuning techniques in both performance and computational efficiency. Furthermore, the SAC model achieved the highest performance across three zero-shot datasets with the lowest standard deviation, confirming its robustness and generalization capabilities. The findings highlight the effectiveness of the proposed parameter-efficient fine-tuning in addressing domain shift and improving crack segmentation accuracy without incurring the high computational costs associated with full model adaptation. Also, the results of this study demonstrate the potential of vision foundation models in infrastructure health monitoring, specifically, for automated crack detection. 

By demonstrating that tuning only normalization components is sufficient for domain adaptation, this work provides a practical solution for deploying deep segmentation models in real-world crack detection scenarios. Future research could further explore hybrid fine-tuning approaches that combine normalization tuning with lightweight architectural modifications to enhance adaptability across diverse structural conditions. Additionally, extending this method to other structural defect detection tasks could further validate its broader applicability in automated infrastructure health assessment and inspection systems.

\section{Data Availability Statement}
Some or all data, models, or code that support the findings of this study are available from the corresponding author upon reasonable request.
\section{Acknowledgment}
This research was funded by the Discovery Grant (Grant No.: RGPIN-2022-05151) of the Natural Sciences and Engineering Research Council of Canada (NSERC).

\clearpage
\begingroup
\setstretch{1.0}
\bibliography{ascexmpl-new}
\endgroup
\end{document}